\documentclass{SCIS2024}
\usepackage{graphicx}
\usepackage{amsmath}
\usepackage{amssymb}
\usepackage{booktabs}
\usepackage{algorithm}
\usepackage{algorithmic}
\usepackage{multirow}
\usepackage{amsthm}
\usepackage{url}
\usepackage{rotate}
\usepackage{bm}
\usepackage{setspace}
\usepackage{color}
\usepackage{enumerate}
\usepackage{cases}
\usepackage{makecell}
\usepackage{threeparttable}

\usepackage[colorlinks]{hyperref}
\usepackage{cite}

\newtheorem{thm}{Theorem}[section]
\begin{document}

	\ArticleType{RESEARCH PAPER}
	\Year{2024}
	\Month{}
	\Vol{}
	\No{}
	\DOI{}
	\ArtNo{}
	\ReceiveDate{}
	\ReviseDate{}
	\AcceptDate{}
	\OnlineDate{}

	\title{Practical exposure correction via compensation\\}{Practical exposure corrector}

	\author[1]{Long MA}{}
	\author[1]{Nan AN}{}
	\author[2]{Jinyuan LIU}{}
	\author[1]{Xin FAN}{}
	\author[1]{\\Zhongxuan LUO}{}
	\author[3,4]{Deyu MENG}{}
	\author[1]{Risheng LIU}{rsliu@dlut.edu.cn}

	\AuthorMark{Ma L}
	
	\address[1]{School of Software Technology, Dalian University of Technology, Dalian {\rm 116024}, China}
	\address[2]{School of Mechanical Engineering, Dalian University of Technology, Dalian {\rm 116024}, China}
	\address[3]{School of Mathematics and Statistics, Xi'an Jiaotong University, Xi'an {\rm 710049}, China}
	\address[4]{Macao Institute of Systems Engineering, Macau University of Science and Technology, Macao {\rm 999078}, China}

	\abstract{In computer vision, correcting the exposure level is a fundamental task for enhancing the visual quality of observations with inappropriate lightness. However, existing methodologies tend to be impractical because they lack adaptability to unknown scenes due to restricted modeling patterns and struggle to achieve satisfactory efficiency due to complex computational flows. To tackle these challenges, we establish a new practical exposure corrector (PEC) that excels in both quality and efficiency. Specifically, to overcome the limited expressive power of existing modeling patterns, we build a general model with exposure-sensitive compensation to provide an intuitive modeling perspective. We also design a simple but effective exposure adversarial function to catalyze scene-adaptive compensation. Building on the aforementioned key concepts, we develop a stable and robust iterative shrinkage scheme, avoiding the complex inferences encountered in existing studies. Extensive experimental evaluations across eight challenging datasets showcase the strong adaptability of the developed model to unknown environments. The model offers impressive processing speed, requiring only 0.0009 s to handle a 2K image on a device equipped with a GeForce RTX 2080Ti GPU. Experimental analysis of different downstream vision tasks further verifies the flexibility of the model. The code is available at \url{https://github.com/vis-opt-group/PEC}.}

	
	\maketitle

	\section{Introduction}
	\label{sec:intro}
	
	With the increasing demand for high-quality images with correct exposure in various domains~\cite{wang2022unsupervised,onzon2021neural, lee2023human,  yan2022optical}, the imaging capabilities, such as the photosensitivity and aperture size, of terminal devices have gained significant importance. However, improving the device-level capabilities may not be sufficient for addressing imaging challenges in severe scenarios involving back-lighting or non-uniform illumination, among other issues.
	Exposure correction techniques~\cite{xu2022snr, liu2022learning, ma2022toward, afifi2021learning, huang2022cvpr, ma2023bilevel} have emerged as a necessary solution to these challenges. Such techniques focus on two typical scenes: underexposure and overexposure. Underexposure has received more attention, leading to faster development for addressing this issue compared to overexposure. 

	Traditional optimization approaches~\cite{fu2016weighted, guo2017lime, zhang2018high, ma2022low, xu2020star, zhang2019dual}, which involve modeling the physical principles derived from Retinex theory~\cite{land1971lightness} under statistical priors, are the core methods for correcting underexposure. However, these methods often result in inconsistent exposure, particularly in challenging scenes with non-uniform brightness.
	Owing to the availability of large-scale data, data-driven learning schemes~\cite{liu2023coconet, su2023global, ma2021learning, liu2022attention, liu2023holoco} have gained dominance in various fields, including exposure correction. A series of deep networks~\cite{zheng2022learn, fu2023learning, xu2023low, wu2023learning, fu2023you, zhou2023fourmer} featuring well-designed architectures and training strategies has been proposed for correcting underexposure. Although most of these deep networks achieve the desired visual quality in specific scenarios, high computational cost is often an attendant drawback, making these methods time-consuming.
	Recently, certain studies~\cite{huang2022acmmm, wang2023decoupling, huang2023learning, ma2020joint} introduced unified models that can simultaneously handle underexposure and overexposure. However, these methods are limited by the supervised learning paradigm, making it challenging for them to effectively handle real-world, unknown scenes with incorrect exposure. The reliance on labeled data restricts the adaptability of these methods to various and unpredictable scenarios, which remain a significant challenge for exposure correction.

	In summary, the limitations of existing techniques, including their low adaptability to unseen degraded scenes and the time-consuming testing burden, heavily impact their practicality. These challenges hinder the seamless application of conventional approaches in real-world scenarios, highlighting the necessity for more efficient and versatile solutions for this task. Developing techniques that can handle diverse exposure conditions and achieve faster processing times is crucial for advancing the practicality of various applications.
	
\begin{figure*}
		\footnotesize
		\centering
		\begin{tabular}{c@{\extracolsep{0.3em}}c@{\extracolsep{0.3em}}c} 
			\includegraphics[width=0.299\linewidth]{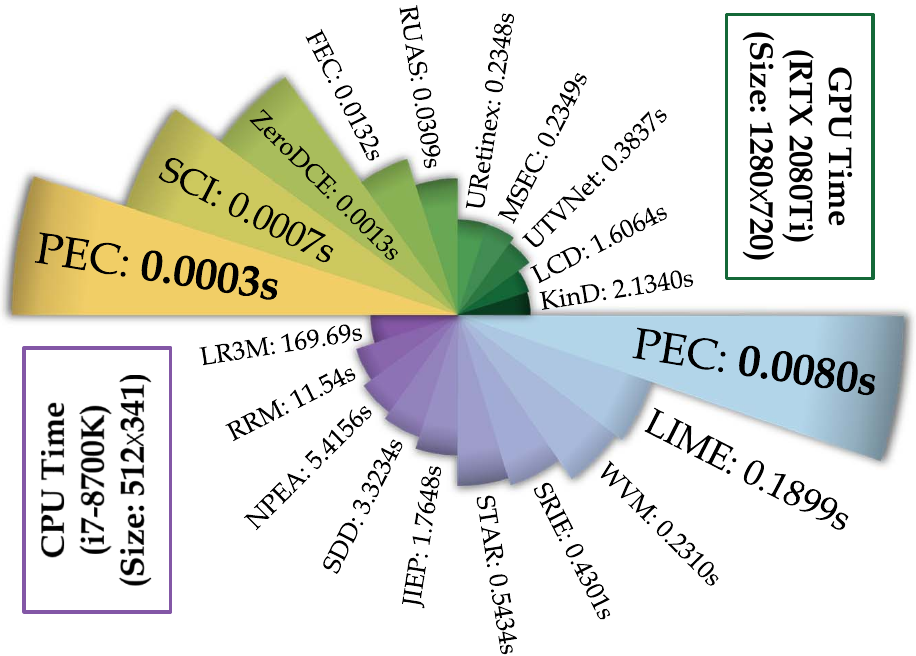}&
			\includegraphics[width=0.335\linewidth]{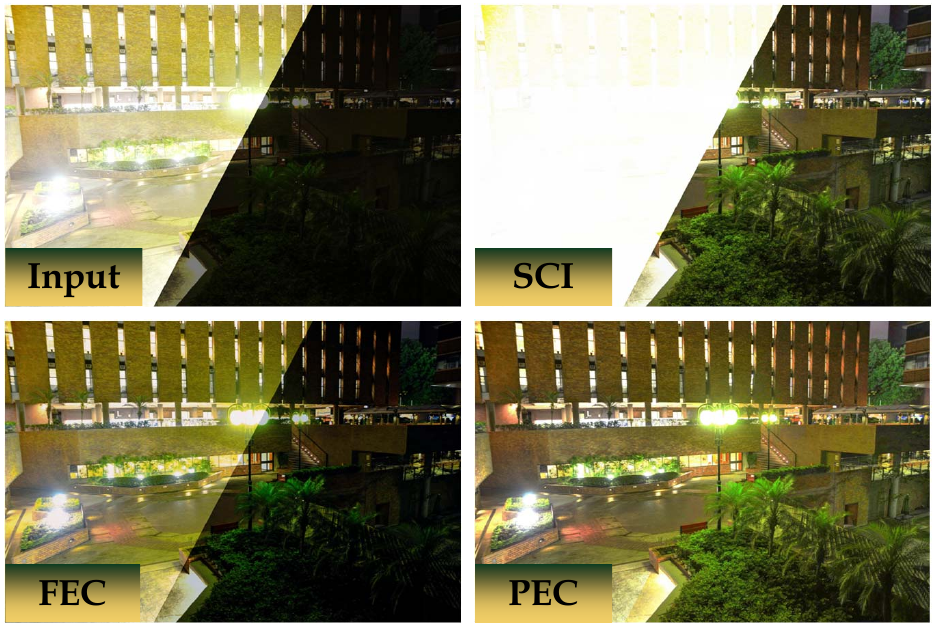}&
			\vspace{-0.2cm}
			\includegraphics[width=0.335\linewidth]{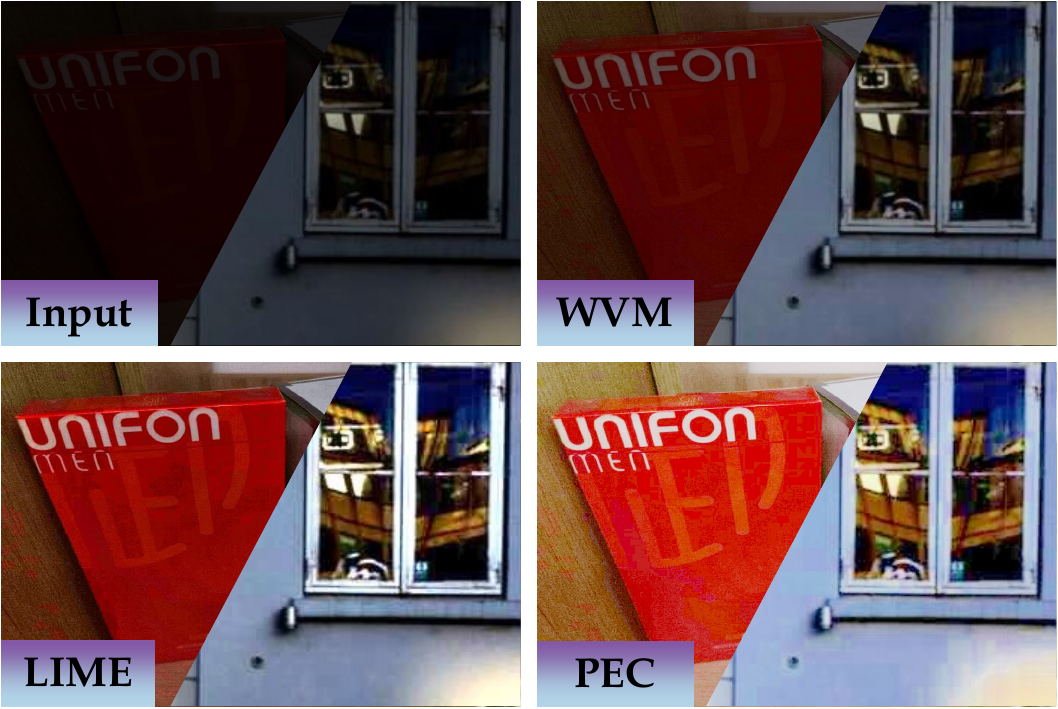}\\
			(a) Comparison of running time &(b) Comparison with deep networks & (c) Comparison with traditional methods\\
		\end{tabular}
		\caption{Practicality evaluation: Panel (a) compares the performance of nine advanced deep networks and nine traditional methods using various computational resources (comparisons using more platforms are presented in Table~\ref{table: Mobile Time}). Panel (b) presents visual comparisons among two deep networks and PEC for the same scene~\cite{cai2018Learning} but with different exposure conditions (the upper left and lower right images depict overexposure and underexposure, respectively). Panel (c) presents visual comparisons of two traditional methods with the PEC for different scenes~\cite{hai2023r2rnet,nada2018pushing} with varying degrees of underexposure. The developed PEC achieves the best scene adaptability and simultaneously requires the shortest running time, confirming its high practicality.}
		\label{fig:FirstFigure}
	\end{figure*}

	To tackle the abovementioned challenges, we introduce a new method completely devised from knowledge of the task, which breaks through the restrictions of existing modeling patterns and complex computational flows, affording a practical exposure corrector (PEC) that is both highly efficient and remarkably effective.
	Figure~\ref{fig:FirstFigure} presents a comprehensive evaluation of the practicality of the proposed PEC compared to existing state-of-the-art approaches, focusing on two vital aspects: the adaptability to scenes with incorrect exposure levels from the three datasets in~\cite{cai2018Learning, hai2023r2rnet, nada2018pushing}, and computational efficiency on different platforms. Compared to other competitive approaches, the developed PEC exhibits superior visual quality across a range of challenging scenes, demonstrating its high adaptability. Additionally, this method achieves the fastest inference speed, underscoring its exceptional efficiency. Simply, the practicality of the PEC surpasses that of existing methods. 
	The main contributions of this study are threefold:
	\begin{itemize}
		\item By rethinking the task demand, we introduce an exposure-sensitive compensation to develop a general model for exposure correction. This approach provides an intuitive but effective modeling perspective for overcoming existing restricted modeling patterns. 
		\item An exposure adversarial function is developed to efficiently leverage latent crucial structures from the observation itself, thereby facilitating scene-adaptive compensation. This function circumvents the introduction of complex operators, ensuring highly efficient inference.
		\item We present an iterative shrinkage scheme driven by the exposure adversarial function, yielding a PEC. The corrector requires only a few manually adjusted parameters to reduce the testing cost, and its shrinkage property guarantees stability and robustness in diverse scenarios.
	\end{itemize}

	\section{{Modeling exposure correction}}\label{sec:REC}
	
	\subsection{Modeling with compensation}
	Underexposure correction in images aims to enhance the visibility of low pixel values in the underexposed areas, bringing out more details and information from darker regions. Conversely, overexposure correction aims to reduce the intensity of large pixel values to restore the overexposed areas to a more natural and visually pleasing appearance.
	
	\textit{From this straightforward perspective, introducing a map into the underexposure observation is a simple approach for achieving the magnification effect. Similarly, we can remove a map from the overexposure observation to obtain the shrunken normal image. These maps can act as guides to adjust the pixel values and achieve the desired exposure correction effect.}
	
	Based on the above, we define the following formula:
	\begin{equation}\label{eq:motivation}
		\mathbf{x}=\mathbf{y}+\Gamma_{\mathbf{y}}\bm{\epsilon}(\mathbf{y}), 
	\end{equation}
	where $\mathbf{y}$ and $\mathbf{x}$ denote the observation and output, respectively. The function $\bm{\epsilon}(\mathbf{y})$ is the defined exposure-sensitive compensation developed herein, which is closely related to the given observation. $\Gamma_{\mathbf{y}}$ denotes the indicator discriminating the exposure level covering underexposure and overexposure.

	\begin{figure}[t]
		\centering
		\footnotesize
		\begin{tabular}{c@{\extracolsep{0.8em}}c}			
			\vspace{-0.2cm}
			\includegraphics[height=0.175\linewidth]{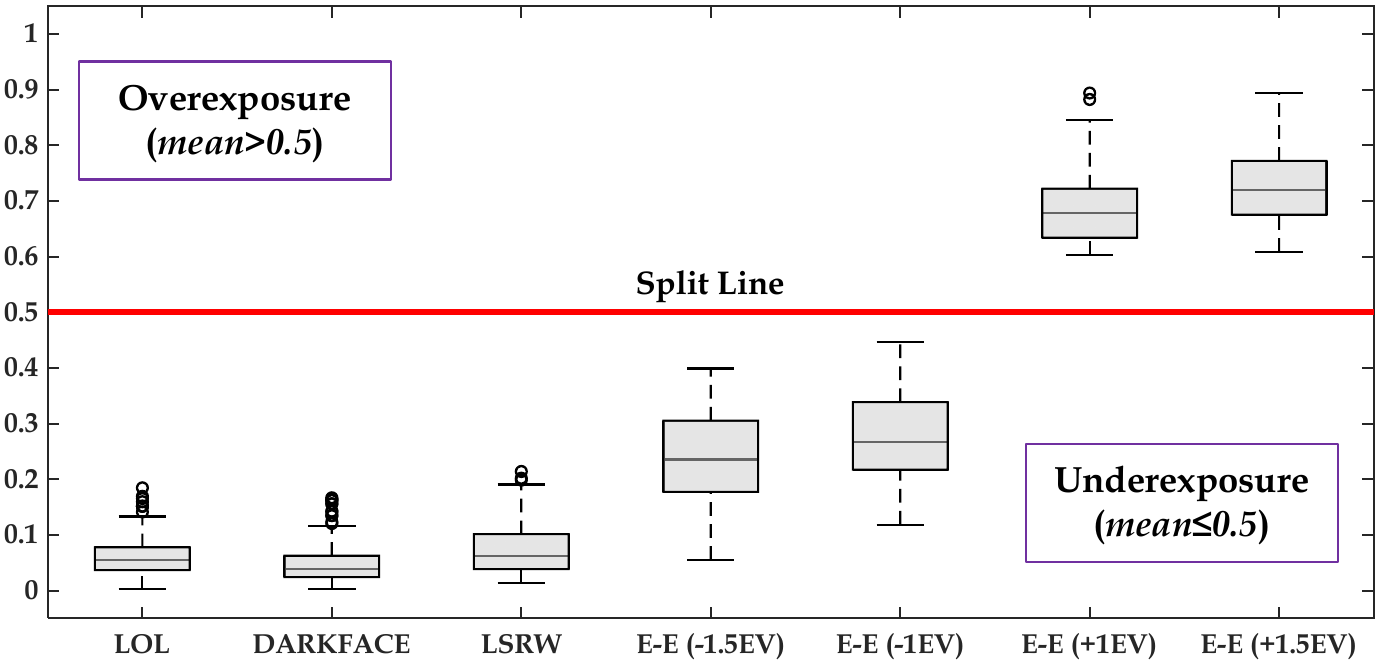}&
			\includegraphics[height=0.175\linewidth]{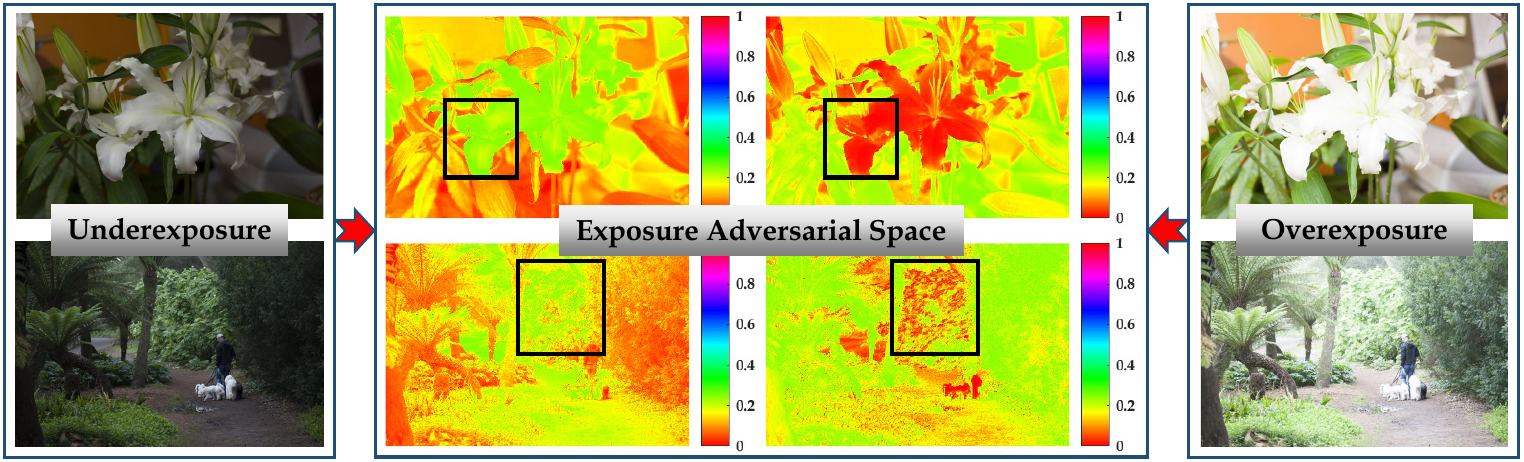}\\
			(a)&(b)\\
		\end{tabular}
		\caption{(a) Statistical analysis of four datasets (each with 100 randomly sampled images) with different exposure levels. (b) Exploration of the exposure adversarial function. We define the exposure adversarial space in the middle part to show the results after applying the function. To illustrate the differences, we adopt the Hue, Saturation, Value (HSV) colormap in the channel-averaged results.}
		\label{fig: exposurelevel}
	\end{figure}

	\subsection{Discriminating exposure level}
	Figure~\ref{fig: exposurelevel} (a) presents the statistical metrics for different datasets, including Exposure-Errors~\cite{afifi2021learning} (abbreviated as “E-E”; “EV” represents the exposure value), LSRW~\cite{hai2023r2rnet}, LOL~\cite{Chen2018Retinex}, and DARKFACE~\cite{yang2020advancing} datasets using a box plot. It is evident that a clear boundary is apparent, \textit{i.e.,} if the mean value is lower than 0.5, the observation is classified as an underexposure case; otherwise, it is categorized as an overexposure case. Thus, we obtain:
	\begin{equation}
		\Gamma_{\mathbf{y}}=\left\{
		\begin{aligned}
			&1, \mathrm{if}\;\mathtt{mean}(\mathbf{y}) \leq 0.5,\\
			&-1, \mathrm{if}\;\mathtt{mean}(\mathbf{y}) > 0.5,\\
		\end{aligned}
		\right.
	\end{equation}
	where $\mathtt{mean}(\cdot)$ represents the mean function.

	By identifying the critical component of the exposure correction task, we can then concentrate on estimating the exposure-sensitive compensation during the exposure correction process. The ensuing section elaborates on the design of an effective algorithm for accurately achieving this compensation.

	\section{Practical exposure corrector}
	
	\subsection{Exposure adversarial function}
	Before delving into the construction of the exposure-sensitive compensation $\bm{\epsilon}$, we design an exposure adversarial function to serve as the core support for generating the compensation. For a given variable $\mathbf{z}\in[0,1]^{m\times n}$ (representing either an underexposed or overexposed image), the exposure adversarial function is formulated as
	\begin{equation}\label{eq:SEC}
		f(\mathbf{z}) =c\mathbf{z}\otimes(\mathbf{1}-\mathbf{z}), \\
	\end{equation}
where $\otimes$ denotes element-wise multiplication, and $0<c\leq1$ is a coefficient controlling the exposure level.

The constructed function combines the input $\mathbf{z}$ and its complement $\mathbf{1-z}$ to generate a result within a closed-form range, specifically $f(\mathbf{z})\in[0,0.25c]^{m\times n}$. The relationship between $\mathbf{z}$ and $\mathbf{1-z}$ exhibits a “checks and balances” dynamic, and is thus referred to as "adversarial." Importantly, when confronted with different observations having varying exposure levels, this function consistently produces a stable solution because of its axisymmetric property (where $c$ remains constant), meaning that $f(\mathbf{1-z})=f(\mathbf{z})$. This key property ensures the effectiveness of the function in handling diverse exposure conditions and consistently providing reliable solutions.
	
	\begin{figure}[t]
		\centering
		\footnotesize
		\begin{tabular}{c@{\extracolsep{0.6em}}c}			
			\vspace{-0.2cm}
			\includegraphics[width=0.47\linewidth]{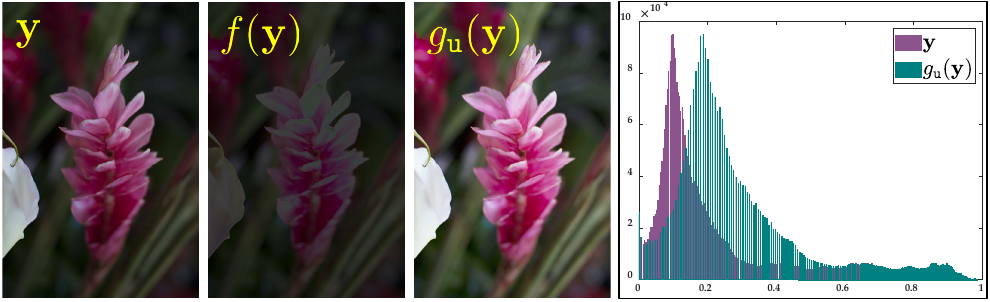}&
			\includegraphics[width=0.47\linewidth]{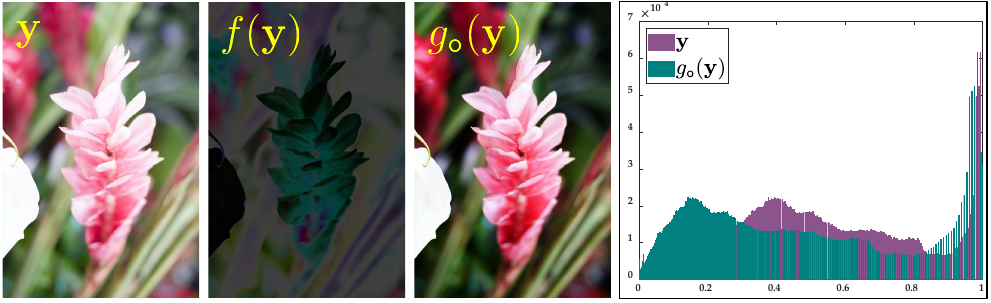}\\
			(a) Underexposure image correction&(b) Overexposure image correction \\
		\end{tabular}
		\caption{Effects of the warm start. The left three columns display the visual results of each component in the warm start, and the last column presents a comparison of the histograms for the results before and after correction. \textit{It is noticeable that the corrected outputs largely maintain the same distribution trend as the inputs}.}
		\label{fig: warmstart}
	\end{figure}

	Figure~\ref{fig: exposurelevel} (b) shows different observations with incorrect exposure, which are mapped to a similar distribution space after applying the exposure adversarial function ($c=1$). Importantly, the generated results effectively reflect the levels of exposure associated with the observations. As illustrated in the black rectangles of the first row, the girl’s face in the underexposed case is not bright enough, whereas in the overexposed case, it is acceptable. Thus, the results in the exposure adversarial space accurately capture the corresponding levels, where values on the left are greater than those on the right.

	\subsection{Warm start with compensation}
	We construct a warm start with compensation to provide an initialization $g(\mathbf{y})$, formulated as
	\begin{equation}
		g(\mathbf{y})=\mathbf{y}+\Gamma_{\mathbf{y}}f(\mathbf{y}).
	\end{equation}
	
	Figure~\ref{fig: warmstart} shows the effects of the warm start operation. The warm start operation successfully improves the exposure of the input. Notably, from the histogram distribution, we observe that the corrected outputs do not disrupt the intrinsic distribution of the observation, meaning that the original pixel-level correspondences in the observation are preserved.
	However, it is essential to acknowledge that the warm start method is suitable for correcting exposure in some simple cases only. In more complex scenarios, an additional computational process is essential for achieving accurate exposure correction.
	
	\begin{lemma}\label{lemma2} If $c$ is identical, the warm start satisfies the symmetric property; that is, $g_{\mathtt{u}}(\mathbf{1-y})=1-g_{\mathtt{o}}(\mathbf{y})$, where $g_{\mathtt{u}}$ and $g_{\mathtt{o}}$ are the corrected outputs for underexposure and overexposure cases, respectively. 
	\end{lemma}
	
	\begin{proof} 
		For any $\mathbf{y}$ $\in$ $\left [ 0,1 \right ]^{m\times n}$, the warm start process shows that
		$g_{\mathtt{u}}(\mathbf{1-y})=\mathbf{1}-\mathbf{y}+f(\mathbf{1}-\mathbf{y})$. Based on $f(\mathbf{1}-\mathbf{y}) = f(\mathbf{y})$, assuming the exposure level is consistent, we get $g_{\mathtt{u}}(\mathbf{1-y})=\mathbf{1}-g_{\mathtt{o}}(\mathbf{y})$.
	\end{proof}

	\begin{algorithm}[t]
				\caption{{Practical exposure corrector}}\label{alg:PEC}
				\footnotesize
				\begin{algorithmic}[1]
					\REQUIRE 
					The observation $\mathbf{y}$, the iteration numbers $T$ and $K$, the coefficient $c$. 
					\STATE $g(\mathbf{y})=\mathbf{y}+\Gamma_{\mathbf{y}}f(\mathbf{y})$, $\mathbf{s}^{1}=g(\mathbf{y})$, $\mathbf{x}^{1}=\mathbf{0}$,
					\FOR{{$t=1$ to $T$}}
					\FOR{$k=1$ to $K$}
					\STATE $\mathbf{x}^{k+1}=\mathbf{s}^{t}+\Gamma_{\mathbf{y}}f(\mathbf{x}^{k})$,
					\ENDFOR
					\STATE $\mathbf{s}^{t+1}=\mathbf{x}^{K}$. 
					\ENDFOR\\
					\RETURN The corrected output $\mathbf{x}^{K}$.
				\end{algorithmic}	
	\end{algorithm}
	
	It is evident that underexposed images often consist of numerous pixels with small values, some even close to zero. Conversely, overexposed images typically have many pixels with significant values, nearly equal to 1. \textit{Based on this observation, a mutual connection or symmetry plausibly exists between underexposure and overexposure correction}. As depicted in Lemma~\ref{lemma2}, the developed warm start process adheres to a symmetric property among various exposure cases, further validating the rationality of the method.
	
	Compared to the task modeling in Eq.~\eqref{eq:motivation}, during the warm start process, the desired exposure-sensitive compensation $\bm{\epsilon}$ is precisely equal to the exposure adversarial function $f$, meaning that $\bm{\epsilon}(\mathbf{y})=f(\mathbf{y})$. 
	
	\subsection{Iterative shrinkage scheme}
	We successfully achieved an acceptable level of correction quality, as demonstrated in Figure~\ref{fig: warmstart}. However, there is still a need for further research to address more common and challenging scenes.
	By combining the exposure adversarial function and the solution presented in Eq.~\eqref{eq:motivation}, we define an iterative shrinkage scheme that is composed of the shrinkage built-in block. Each built-in block belongs to an iterative process for which the basic iterative step is
		\begin{equation}\label{eq: PEC}
			h^{k}(\mathbf{y}): \mathbf{x}^{k+1}=\mathbf{s}^{t}+\Gamma_{\mathbf{y}}f(\mathbf{x}^{k}),\\
		\end{equation}
		where $1\leq k\leq K (1\leq K\leq 3)$ represents the iterative number in the $t$-th ($1\leq t \leq T$, $1\leq T\leq 3$) shrinkage built-in block. In the initial stage, we set $\mathbf{s}^{1}=g(\mathbf{y}), \mathbf{x}^{1}=\mathbf{0}$. 
		
	\begin{figure*}[t]
		\centering
		\footnotesize
		\begin{tabular}{c}			
			\includegraphics[width=0.98\linewidth]{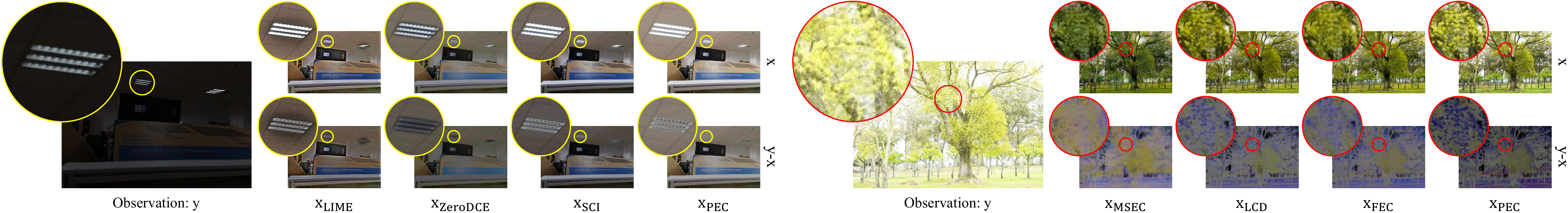}\\
		\end{tabular}
		\caption{Comparison of derived compensations for application of different methods to underexposure (left) and overexposure (right) cases. We compare three representative underexposure correction methods: LIME~\cite{guo2017lime}, ZeroDCE~\cite{guo2020zero}, and SCI~\cite{ma2022toward}, along with three recent overexposure correction methods: MSEC~\cite{afifi2021learning}, LCD~\cite{wang2022local}, and FEC~\cite{huang2022eccv}. \textit{Best viewed full screen to see details.}}
		\label{fig: Compensation}
	\end{figure*}
	
	\begin{thm}\label{thm:sym} According to Lemma~\ref{lemma2}, a symmetric property emerges between underexposure and overexposure cases within each iterative step, \textit{i.e.,} $h_{\mathtt{u}}^{k}(\mathbf{1-y})=\mathbf{1}-h_{\mathtt{o}}^{k}(\mathbf{y})$, where $h_{\mathtt{u}}$ and $h_{\mathtt{o}}$ correspond to underexposure and overexposure cases, respectively.
	\end{thm}
	\begin{proof} 
		When $k = 1 $, the function $h^k(\cdot)$ satisfies 
		\begin{equation}
			\begin{aligned}
				h_u^1(\mathbf{1-y})&=g_{\mathtt{u}}(\mathbf{1-y}),\\
				&=\mathbf{1}-g_{\mathtt{o}}(\mathbf{y})=\mathbf{1}-h_o^1(\mathbf{y}).
			\end{aligned}
		\end{equation}
		
		When $k = 2 $, based on Lemma 1, we have
		$h_u^{2}(\mathbf{1-y})=\mathbf{1}-h_o^{2}(\mathbf{y}).$
		According to the inductive proof, we assume that this property still holds when $k-1$; then we get 
		\begin{equation}
			\begin{aligned}
				h_u^{k}(\mathbf{1-y})&=g_\mathtt{u}(\mathbf{1-y})+f\big(h_u^{k-1}\big(\mathbf{1-y})\big),\\
				&=\mathbf{1}-g_\mathtt{o}(\mathbf{y})+f\big(\mathbf{1}-h_o^{k-1}(\mathbf{y})\big),\\
				&=\mathbf{1}-h_o^{k}(\mathbf{y}).
			\end{aligned}
		\end{equation}
		
		Overall, we have $h_u^{k}(\mathbf{1-y}) = \mathbf{1}-h_o^{k}(\mathbf{y}).$
	\end{proof}

	Furthermore, we can deduce Theorem~\ref{thm:sym} from the property presented in Lemma~\ref{lemma2}.
	Theorem~\ref{thm:sym} establishes a clear and explicit connection between underexposure and overexposure correction for each basic iterative step.
	In fact, the PEC demonstrates a remarkable shrinkage effect, indicating a reduction in the magnitude as the number of iterations increases.
	In conclusion, the devised iterative shrinkage scheme preserves the inherent relationship among various exposure correction tasks. Moreover, it implicitly imposes a constraint that narrows the solution space, ensuring stability during the correction process.
	
	Finally, in each iterative step, we have $\bm{\epsilon}(\mathbf{y})=f(\mathbf{y})+f(\mathbf{x}^{k-1})$, where $\mathbf{x}^{k-1}$ is related to $\mathbf{y}$. Because of the nested iteration, we refrain from providing a general, final form for the compensation, as it is contingent on the specific iterative process.
	Algorithm~\ref{alg:PEC} outlines the computational flow.

	\begin{table*}[t]
		\footnotesize
		\renewcommand\arraystretch{1.2} 
		\setlength{\tabcolsep}{0.8mm}
		\centering
		\caption{Quantitative comparison on the Exposure-Errors dataset. Three full-reference metrics and three no-reference metrics are reported. The top two are respectively marked \textbf{bold} and \underline{underline}; the same representation is used below.}
		\begin{tabular}{|c|c|c|c|c|c|c|c|c|c|c|c|}
			\hline 
			\multicolumn{2}{|c|}{{Metrics}}&{{KinD\cite{zhang2019kindling}}}& {{ZeroDCE\cite{guo2020zero}}}& {{MSEC\cite{afifi2021learning}}}&{{RUAS\cite{liu2021retinex}}}& {{UTVNet\cite{zheng2021adaptive}}}&{{LCD\cite{wang2022local}}}& {{FEC\cite{huang2022eccv}}}& {{URetinex\cite{wu2022uretinex}}}& {{SCI\cite{ma2022toward}}}&{PEC}\\
			\hline
			\multirow{6}{*}{\rotatebox{90}{Underexposure}}&PSNR$\uparrow$ &15.6897&14.8025&20.7083&13.8939&18.5640&{\underline{22.6598}}&{\textbf{22.9983}}&13.4200&20.5109&18.4411\\
			&SSIM$\uparrow$ &0.6845&0.6047&0.7758&0.1943&0.5893&{\underline{0.8113}}&{\textbf{0.8203}}&0.6378&0.7129&0.7382\\
			&LPIPS$\downarrow$ &0.2478&0.2668&0.1808&0.2745&0.2533&{\textbf{0.1471}}&{\underline{0.1782}}&0.2537&0.1843&0.1841\\
			\cline{2-12}
			&DE$\uparrow$ &7.0382&6.7807&\underline{7.4151}&6.7804&{7.1458}&7.4114&7.3600&6.9806&7.3594&{\textbf{7.4450}}\\
			&LOE$\downarrow$ &377.69&350.96&260.45&496.93&227.48&120.03&147.49&417.99&{\underline{95.65}}&{\textbf{22.29}}\\
			&NIQE$\downarrow$ &2.8416&3.3212&2.9387&3.1368&4.2256&2.9458&3.0721&2.7346&{\underline{2.6997}}&{\textbf{2.6843}}\\
			\hline
			\multirow{6}{*}{\rotatebox{90}{Overexposure}}&PSNR$\uparrow$ &9.7730&8.5916&19.3523&4.8380&13.5211&{\underline{22.2832}}&{\textbf{22.8995}}&8.5030&6.9746&19.1298\\
			&SSIM$\uparrow$ &0.5856&0.4764&0.7674&0.1943&0.2907&{\underline{0.8235}}&{\textbf{0.8253}}&0.5106&0.3612&{0.7803}\\
			&LPIPS$\downarrow$ &0.2945&0.3405&0.1834&0.7410&0.4923&{\textbf{0.1341}}&{\underline{0.1609}}&0.3611&0.5055&0.1677\\
			\cline{2-12}
			&DE$\uparrow$ &6.9735&6.7789&{\underline{7.5965}}&2.1515&6.6062&7.4810&7.4671&6.8123&6.1999&{\textbf{7.6040}}\\
			&LOE$\downarrow$ &{\underline{102.57}}&338.90&354.95&1663.92&762.05&195.01&257.92&389.22&231.17&{\textbf{9.10}}\\
			&NIQE$\downarrow$ &2.6464&2.7607&2.6890&6.5999&5.0407&2.6631&2.9606&{\underline{2.6201}}&2.6709&{\textbf{2.5588}}\\
			\hline
		\end{tabular}
		\label{table: ExposureErrors}
	\end{table*}

	\begin{figure*}[t]
		\centering
		\begin{tabular}{c@{\extracolsep{0.2em}}c@{\extracolsep{0.2em}}c@{\extracolsep{0.2em}}c@{\extracolsep{0.2em}}c@{\extracolsep{0.2em}}c}			
			\includegraphics[width=0.16\linewidth]{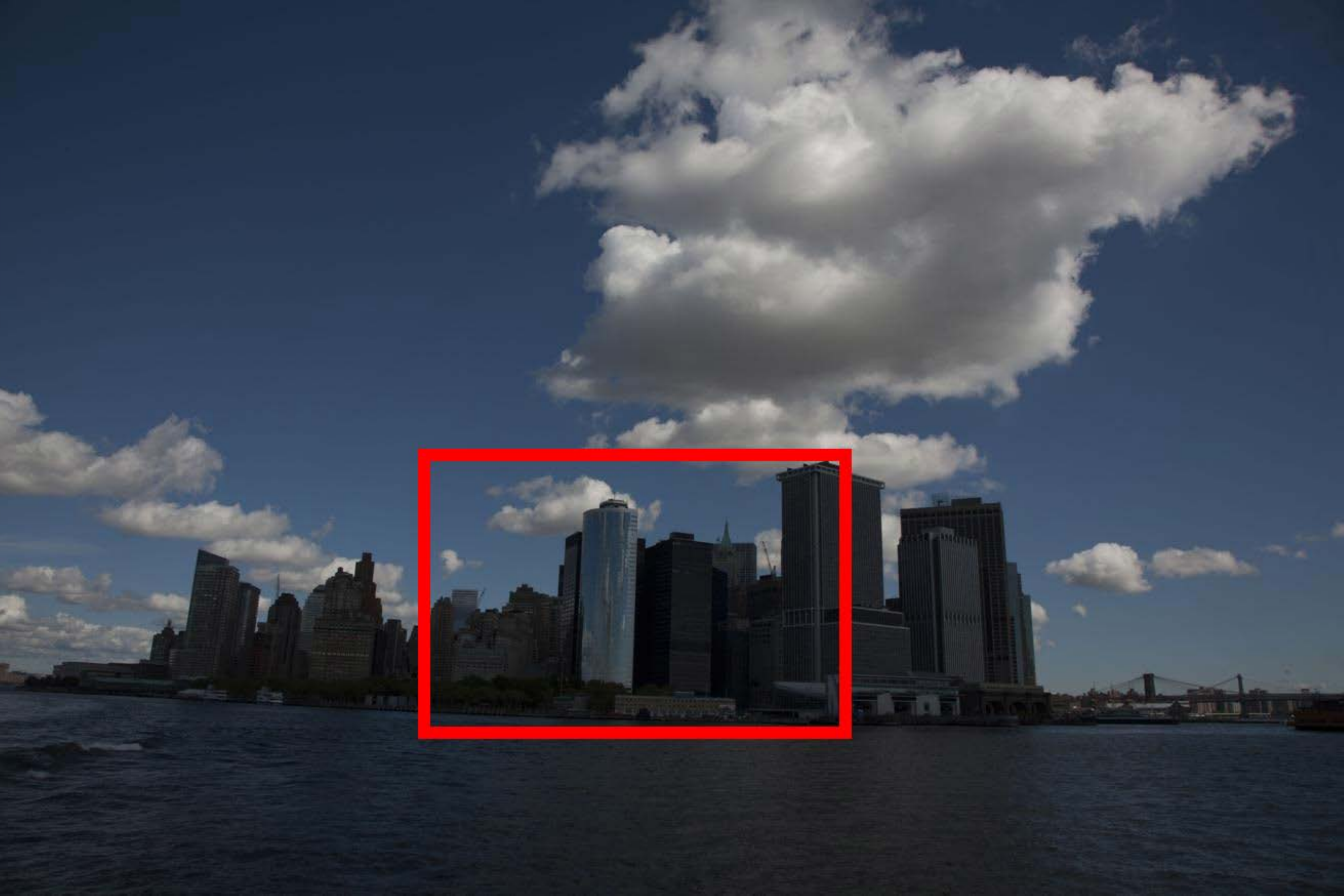}&
			\includegraphics[width=0.16\linewidth]{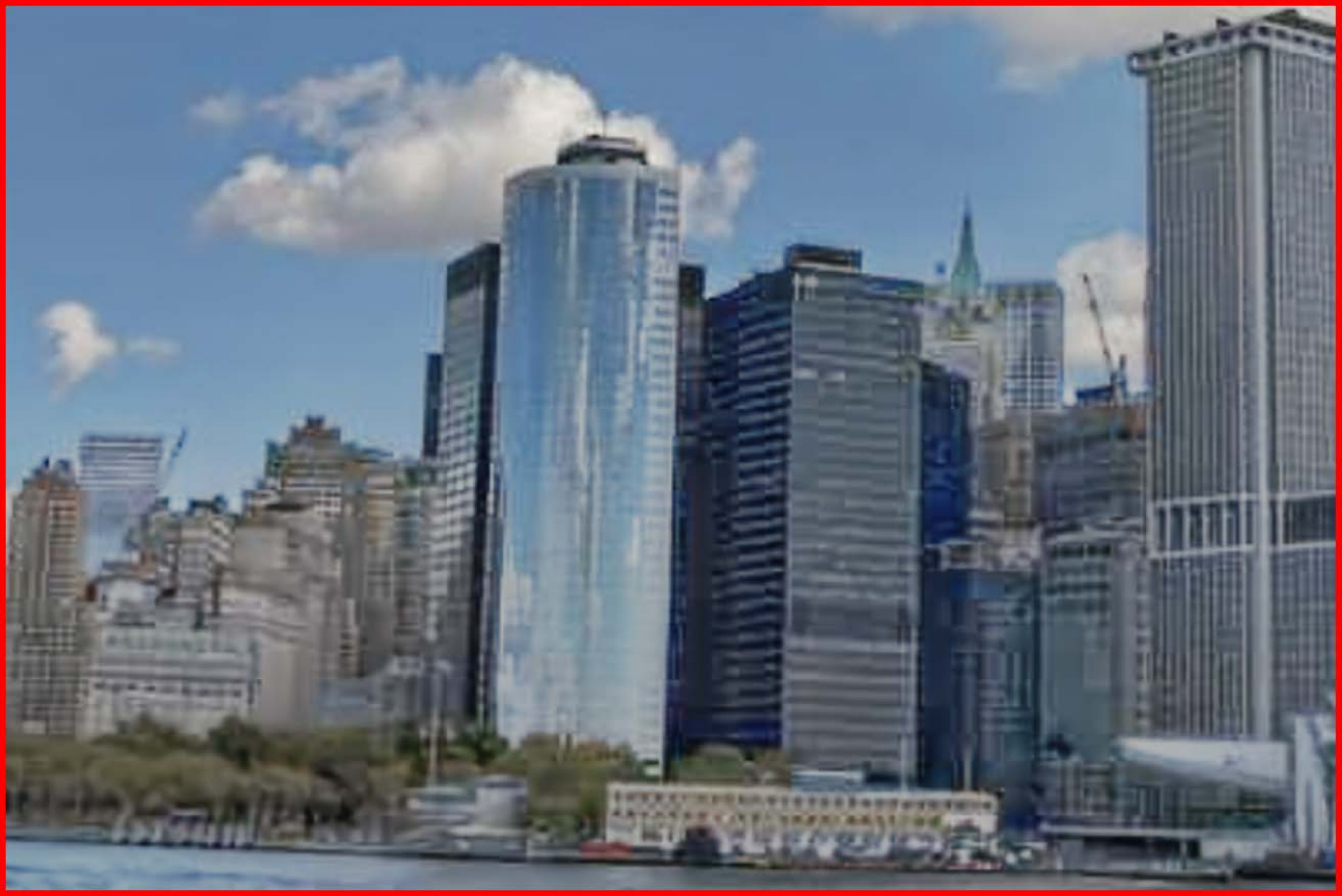}&
			\includegraphics[width=0.16\linewidth]{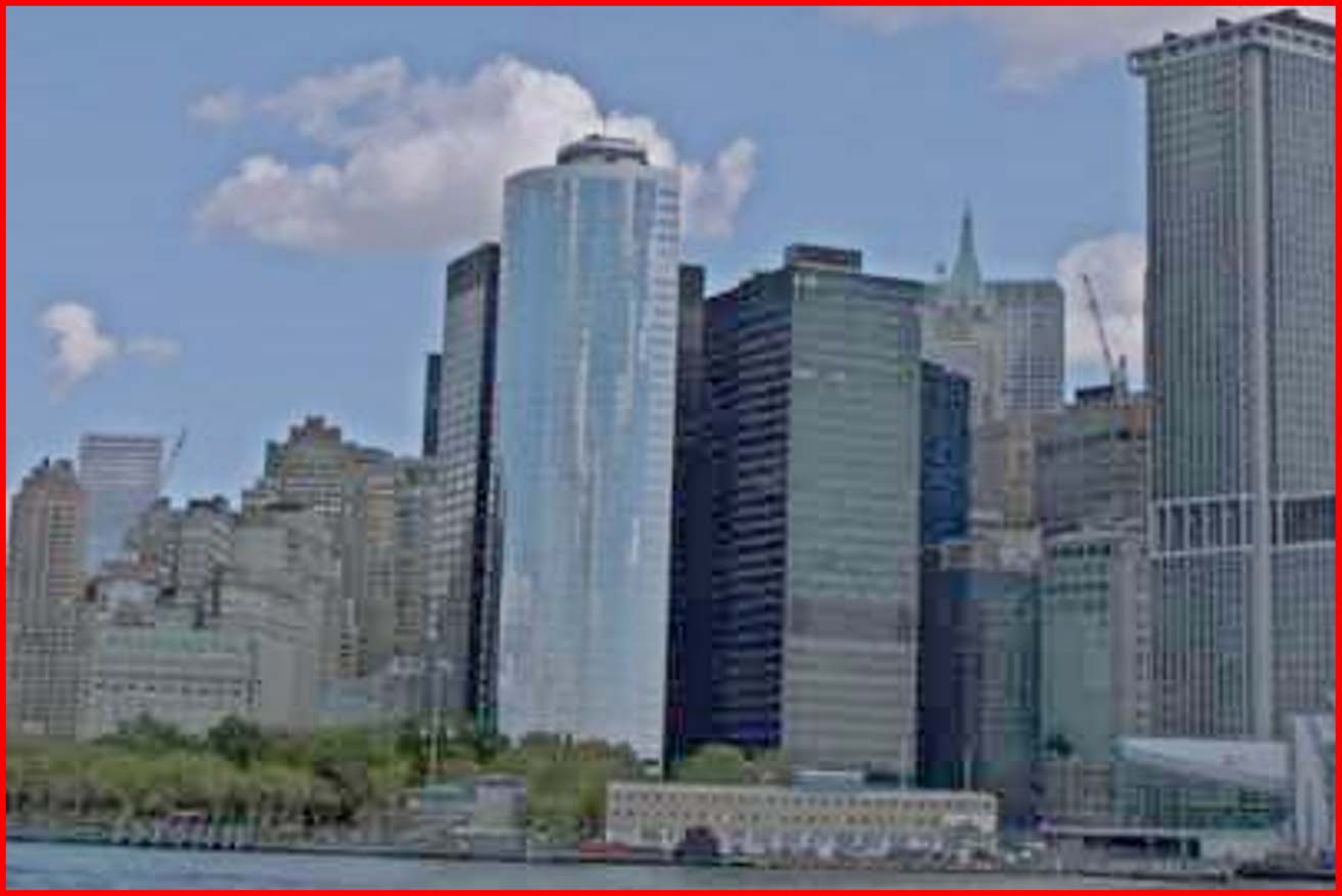}&
			\includegraphics[width=0.16\linewidth]{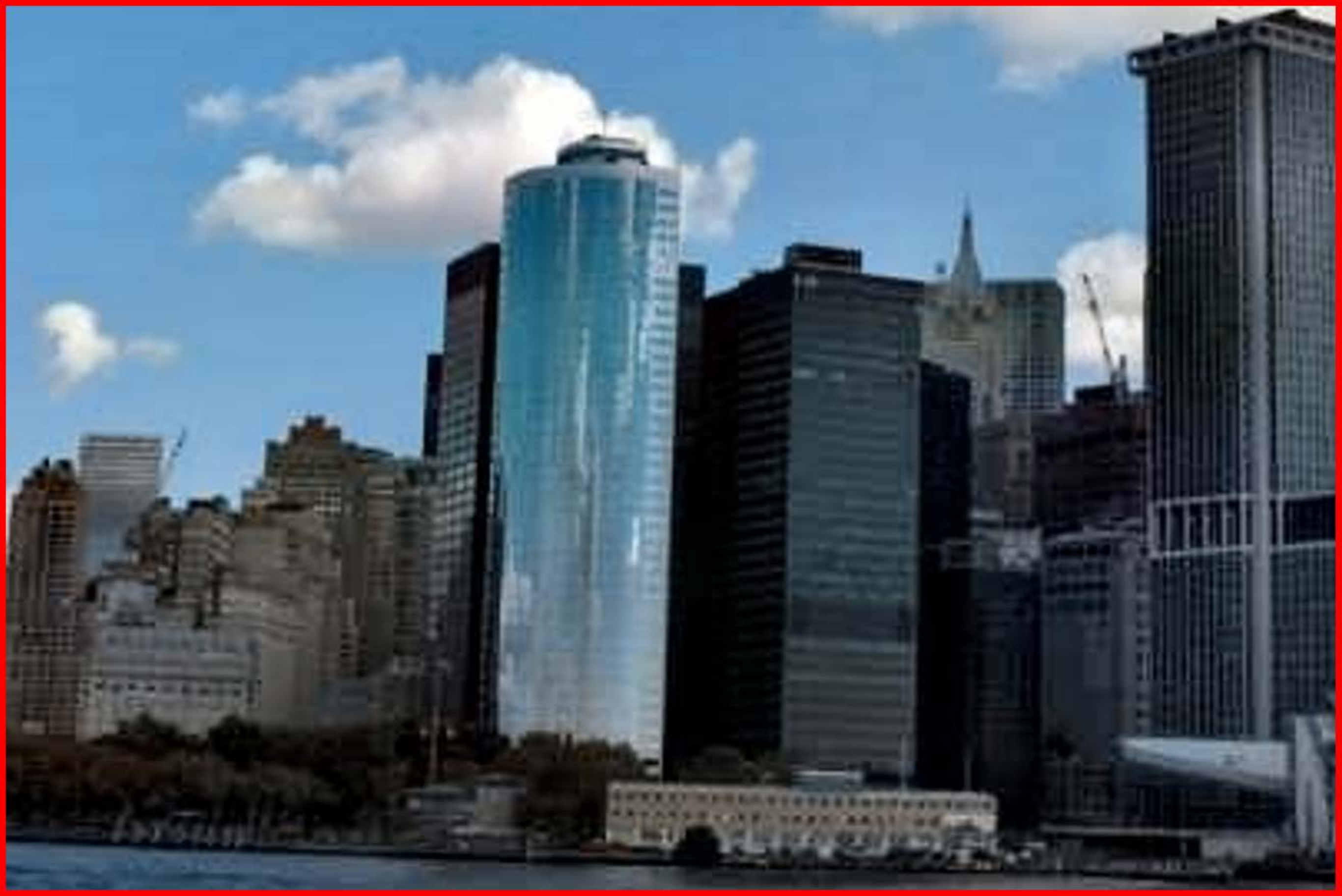}&
			\includegraphics[width=0.16\linewidth]{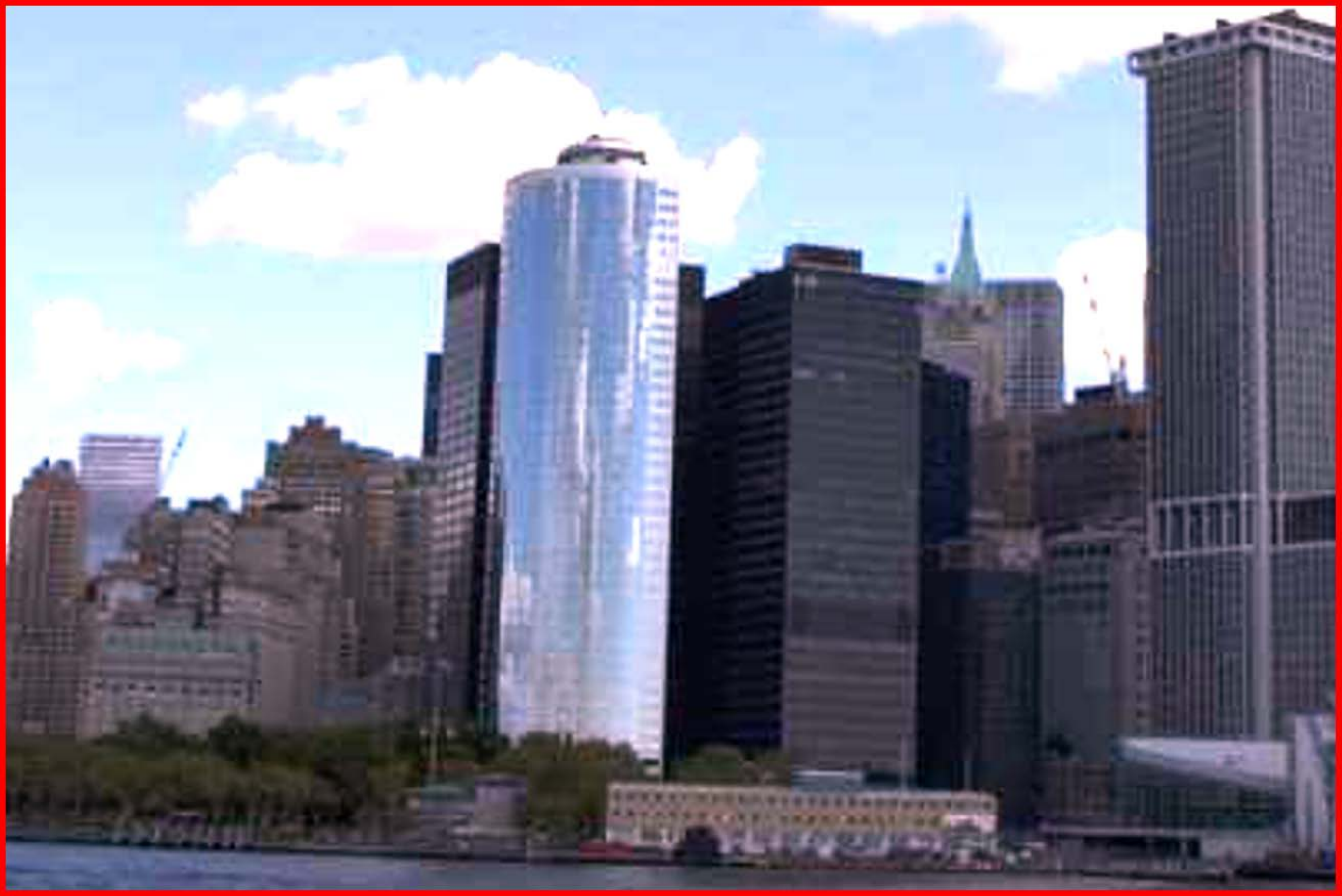}&
			\includegraphics[width=0.16\linewidth]{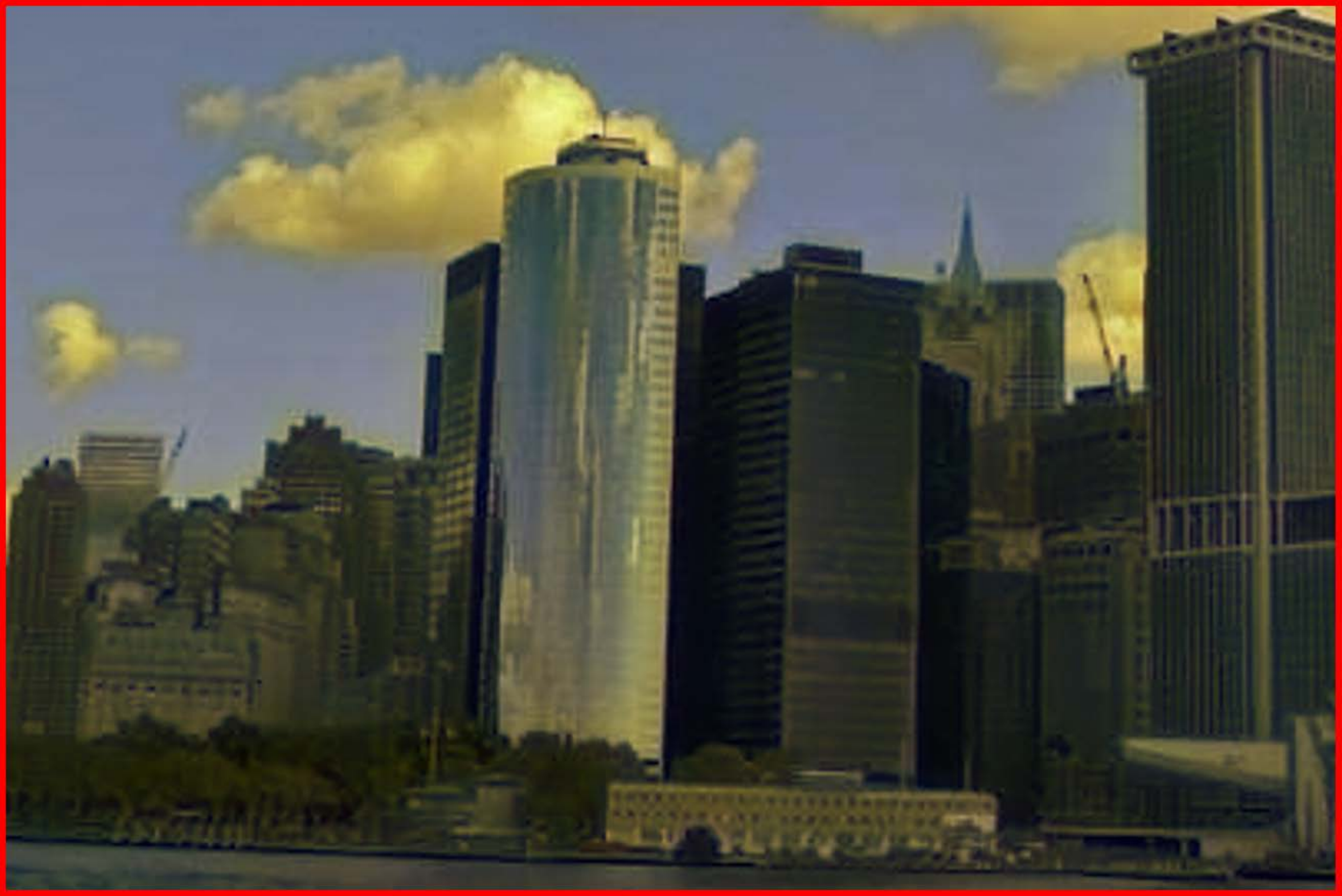}\\\vspace{-0.2cm}
			\includegraphics[width=0.16\linewidth]{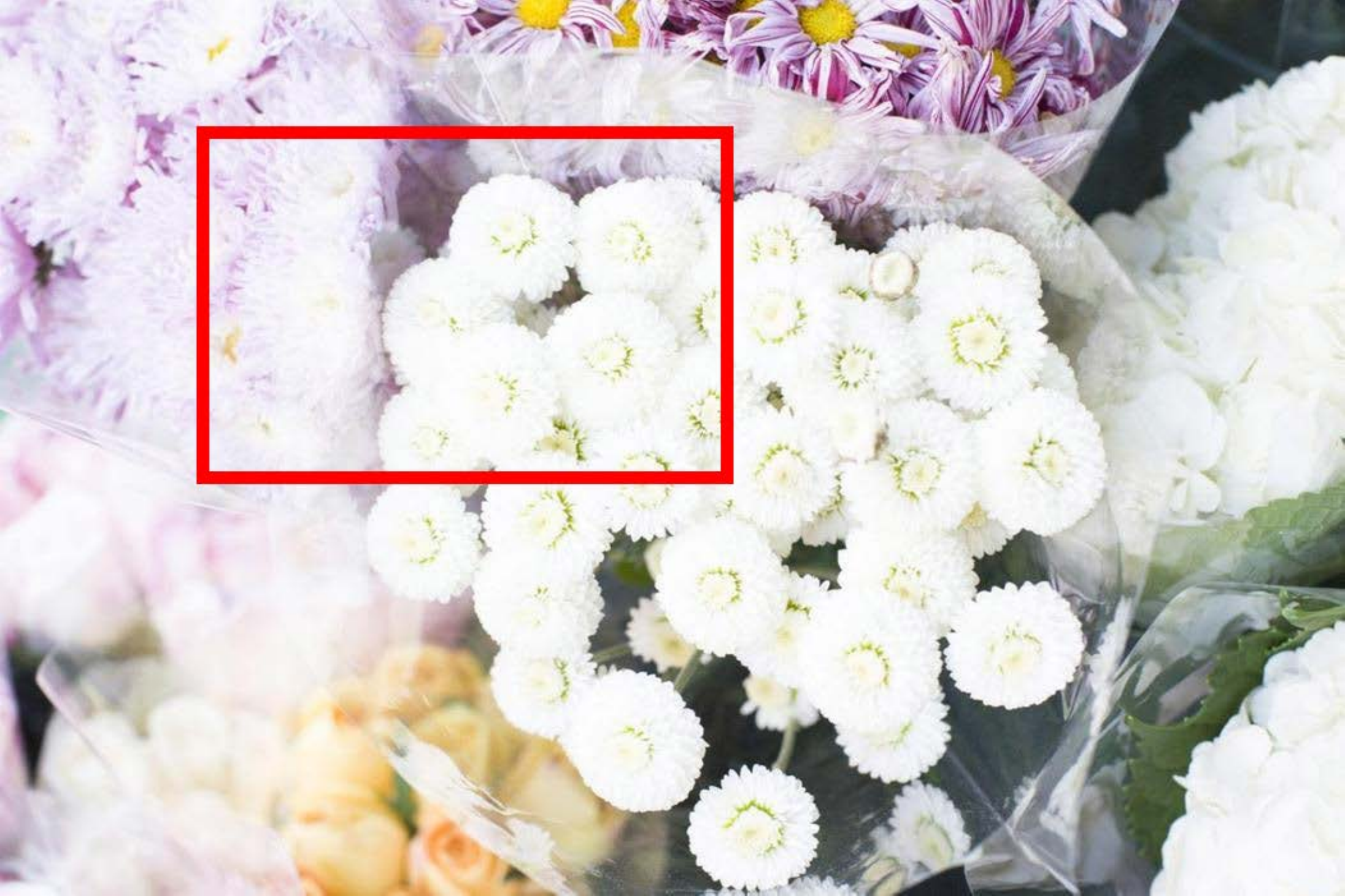}&
			\includegraphics[width=0.16\linewidth]{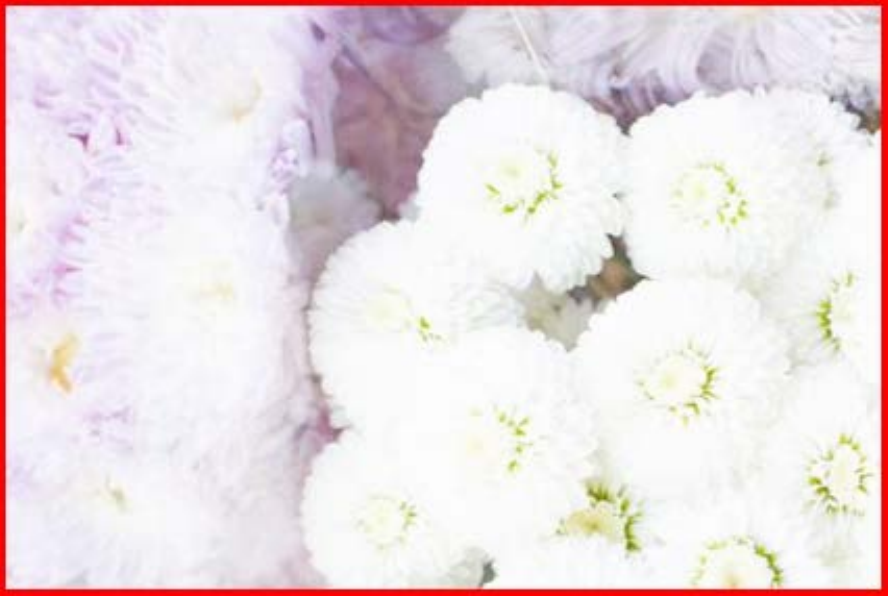}&
			\includegraphics[width=0.16\linewidth]{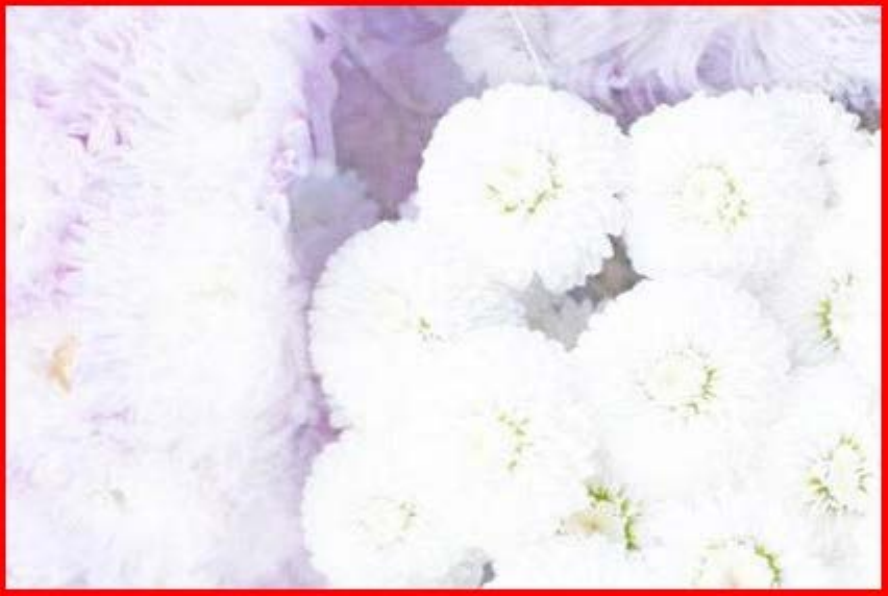}&
			\includegraphics[width=0.16\linewidth]{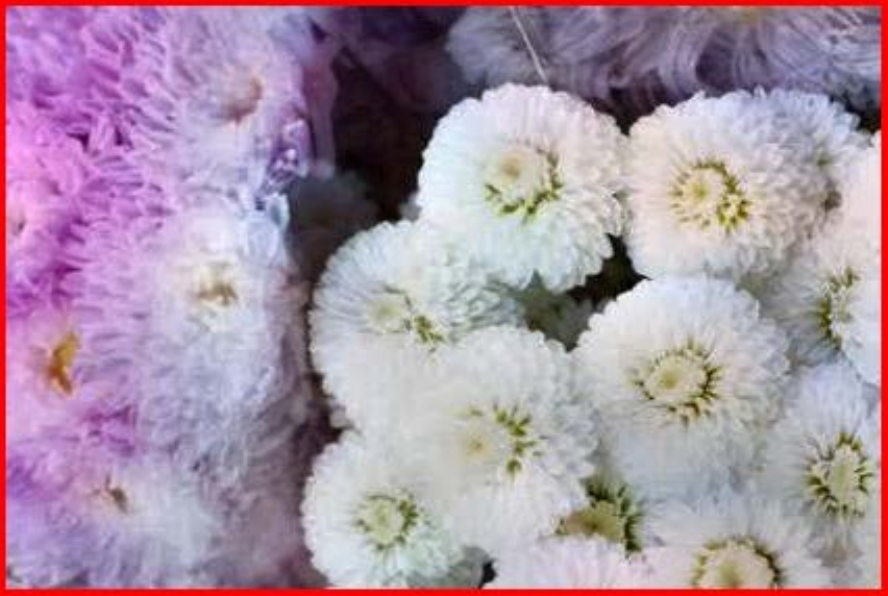}&
			\includegraphics[width=0.16\linewidth]{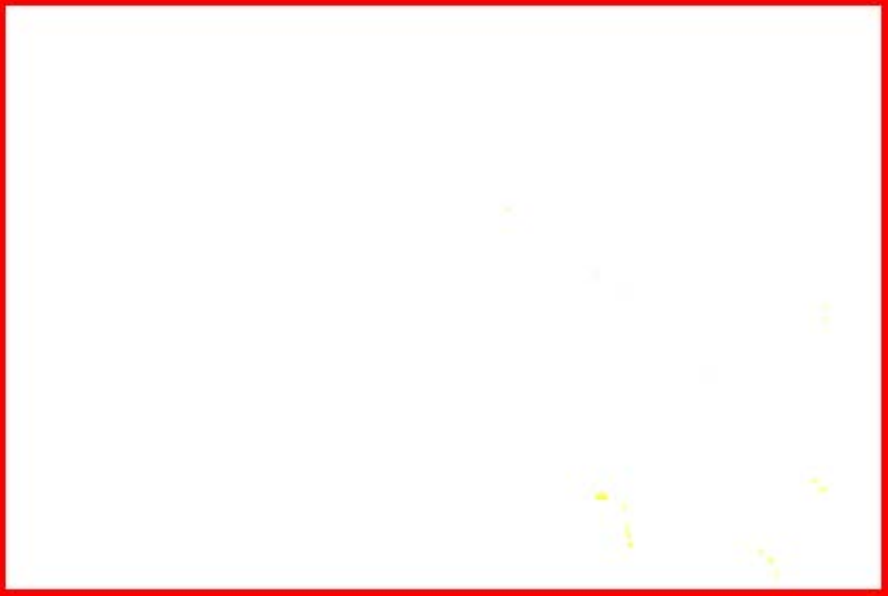}&
			\includegraphics[width=0.16\linewidth]{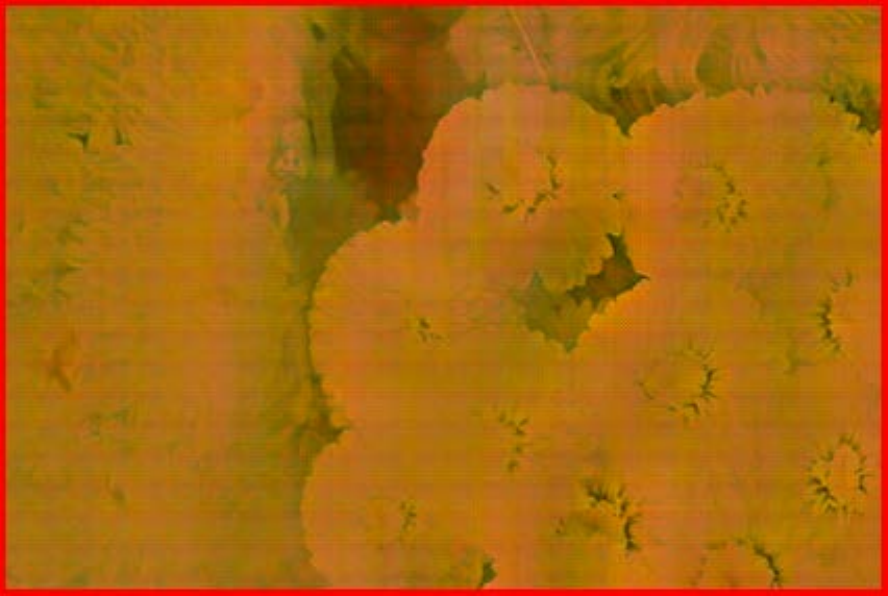}\\
			\footnotesize Input (Full Size)&\footnotesize KinD~\cite{zhang2019kindling}&\footnotesize ZeroDCE~\cite{guo2020zero}&\footnotesize MSEC~\cite{afifi2021learning}&\footnotesize RUAS~\cite{liu2021retinex}&\footnotesize UTVNet~\cite{zheng2021adaptive}\\
			\includegraphics[width=0.16\linewidth]{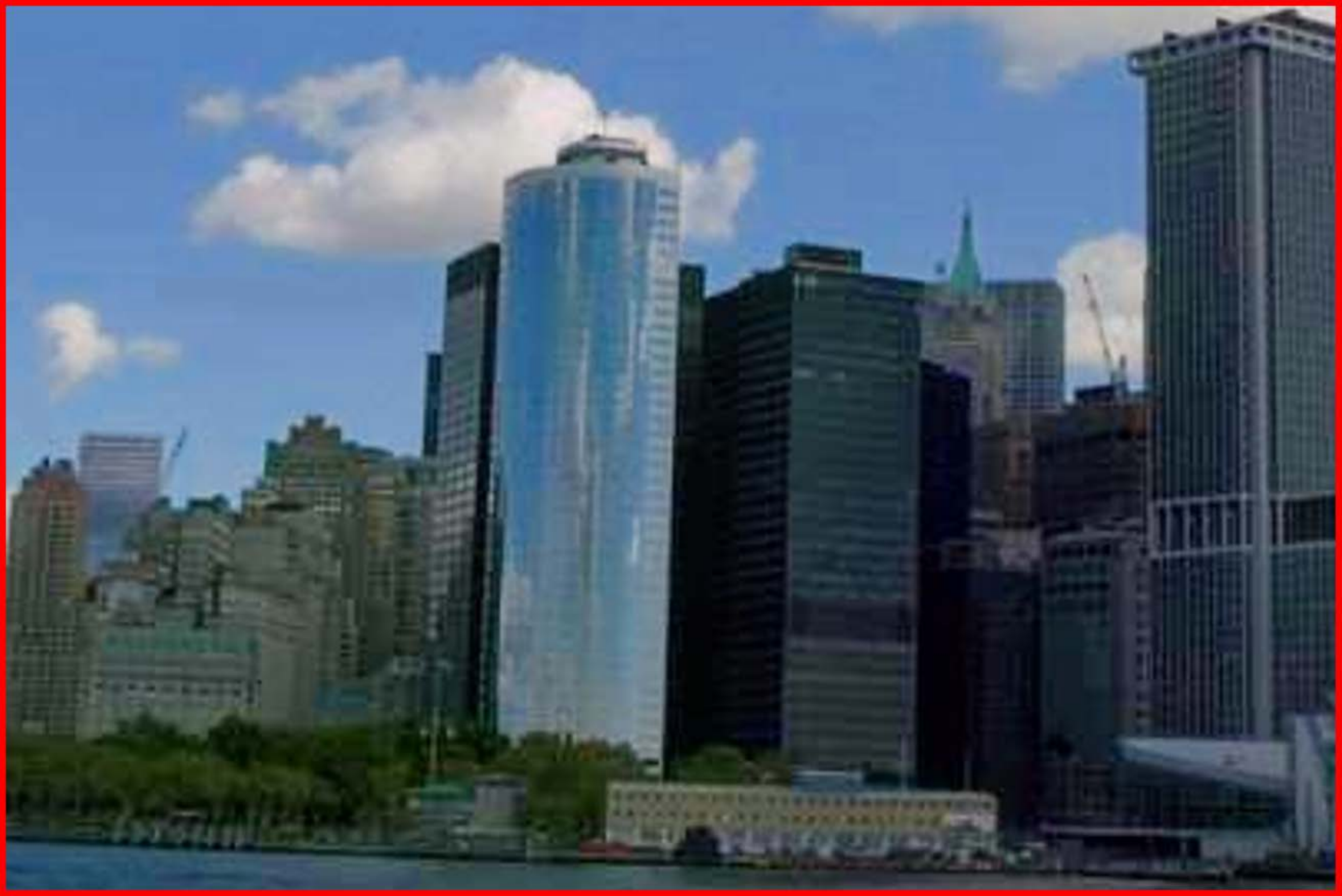}&
			\includegraphics[width=0.16\linewidth]{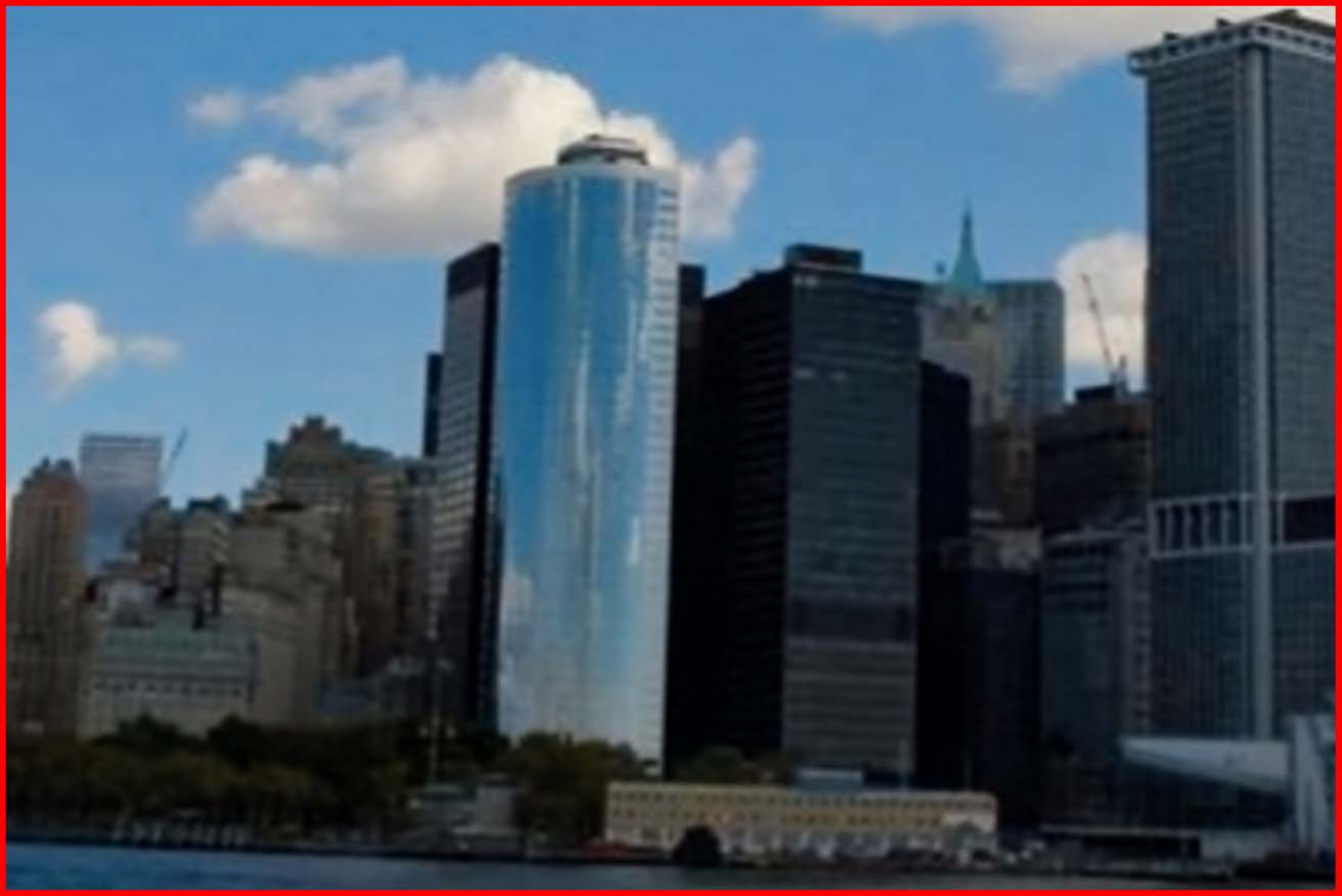}&
			\includegraphics[width=0.16\linewidth]{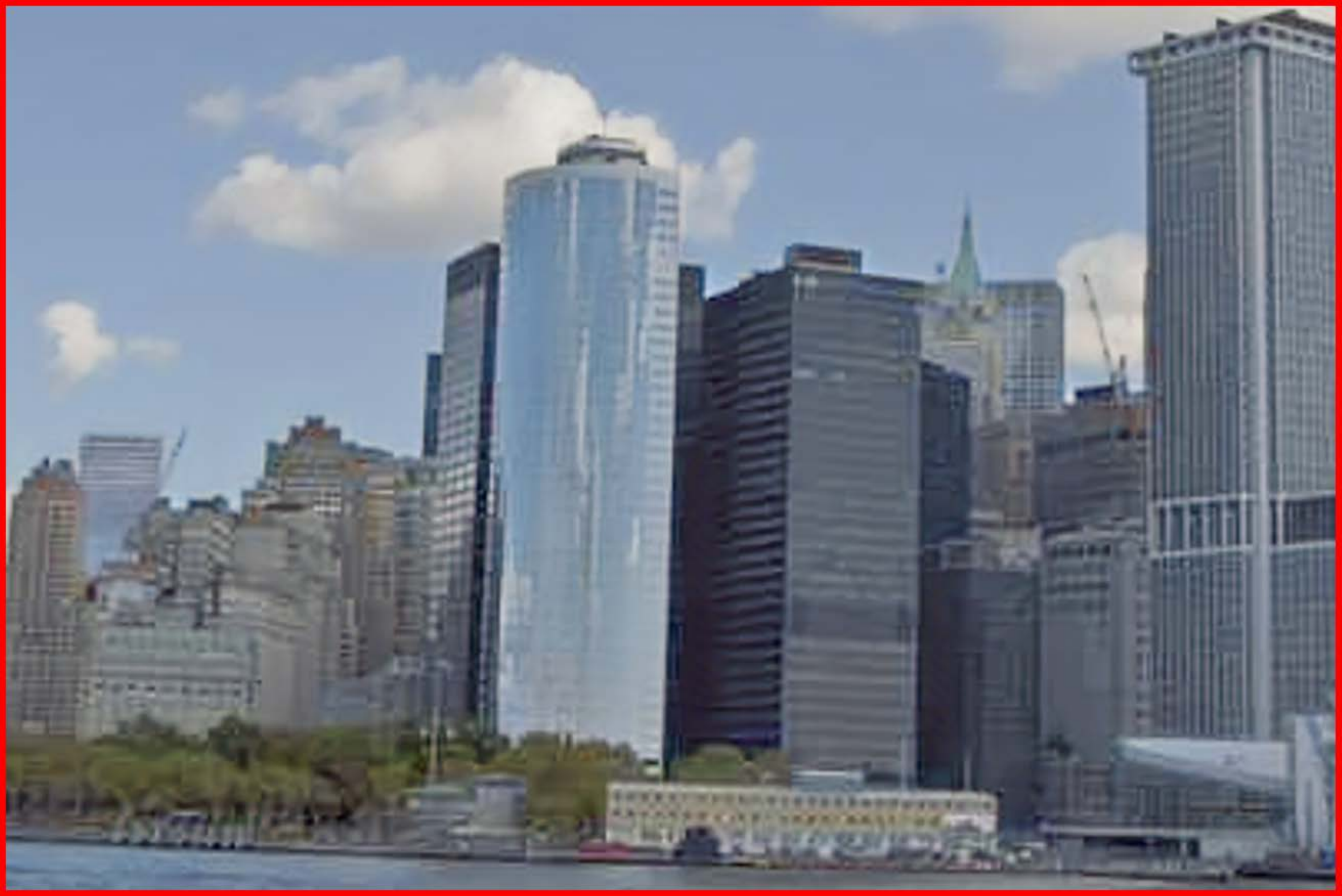}&
			\includegraphics[width=0.16\linewidth]{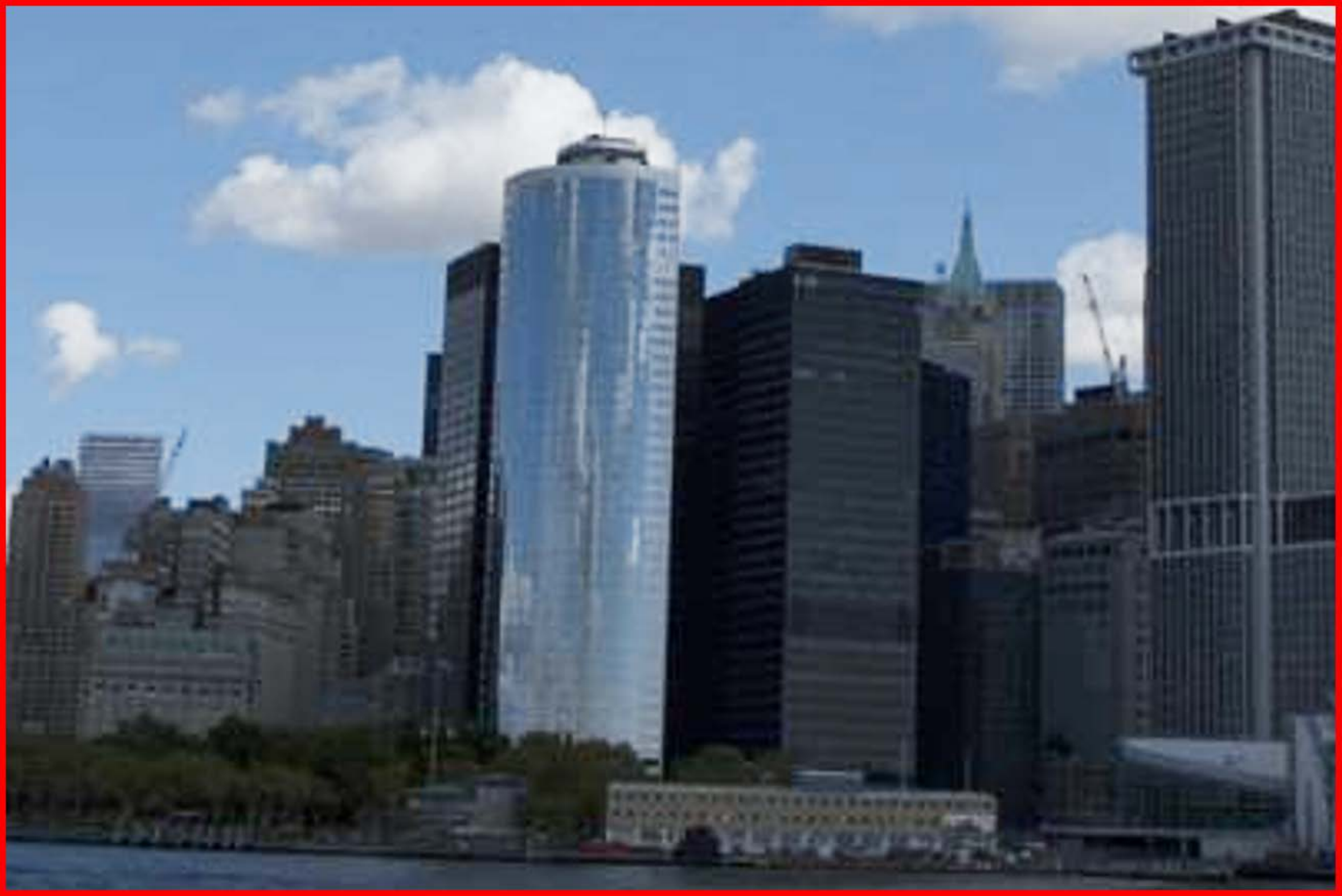}&
			\includegraphics[width=0.16\linewidth]{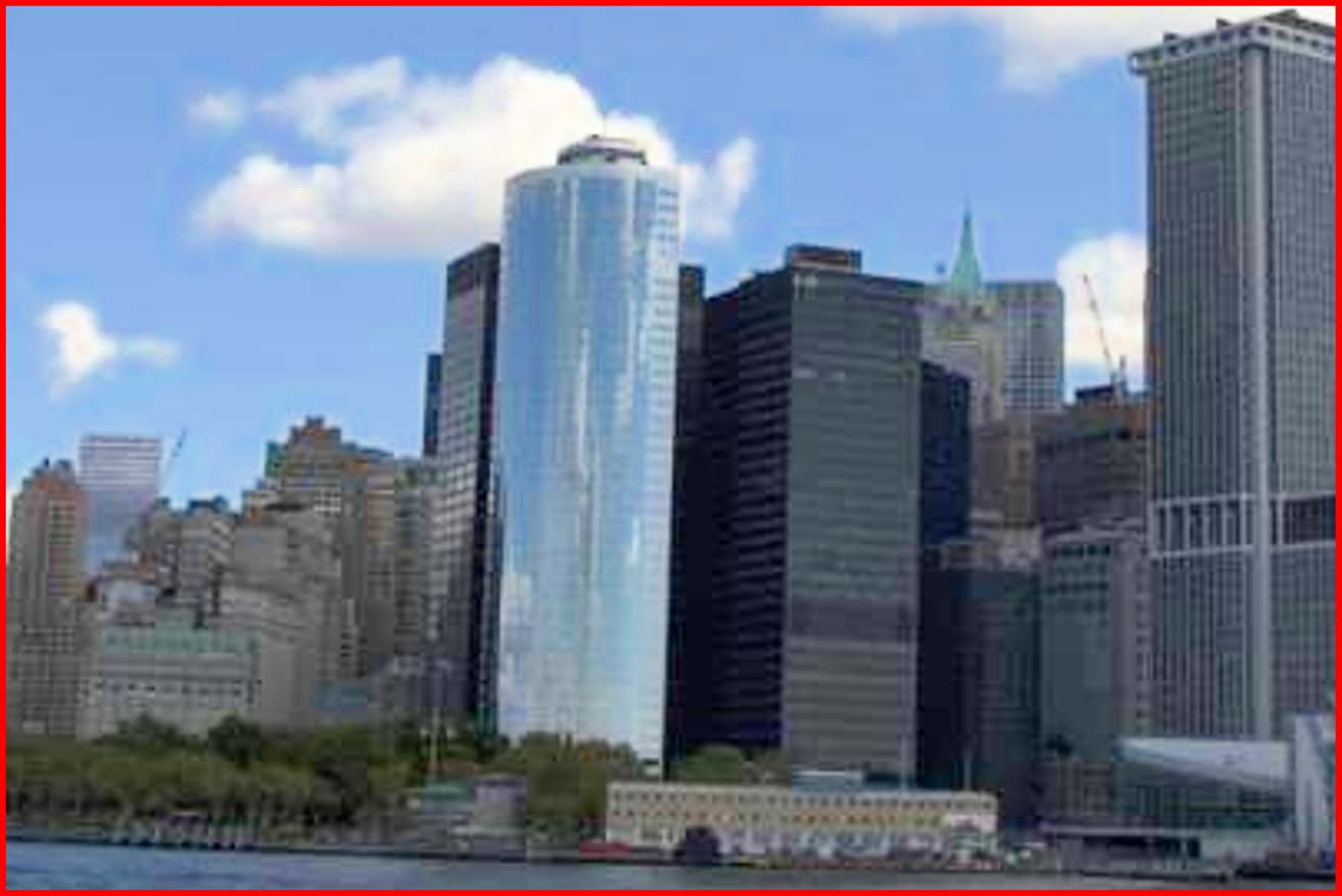}&
			\includegraphics[width=0.16\linewidth]{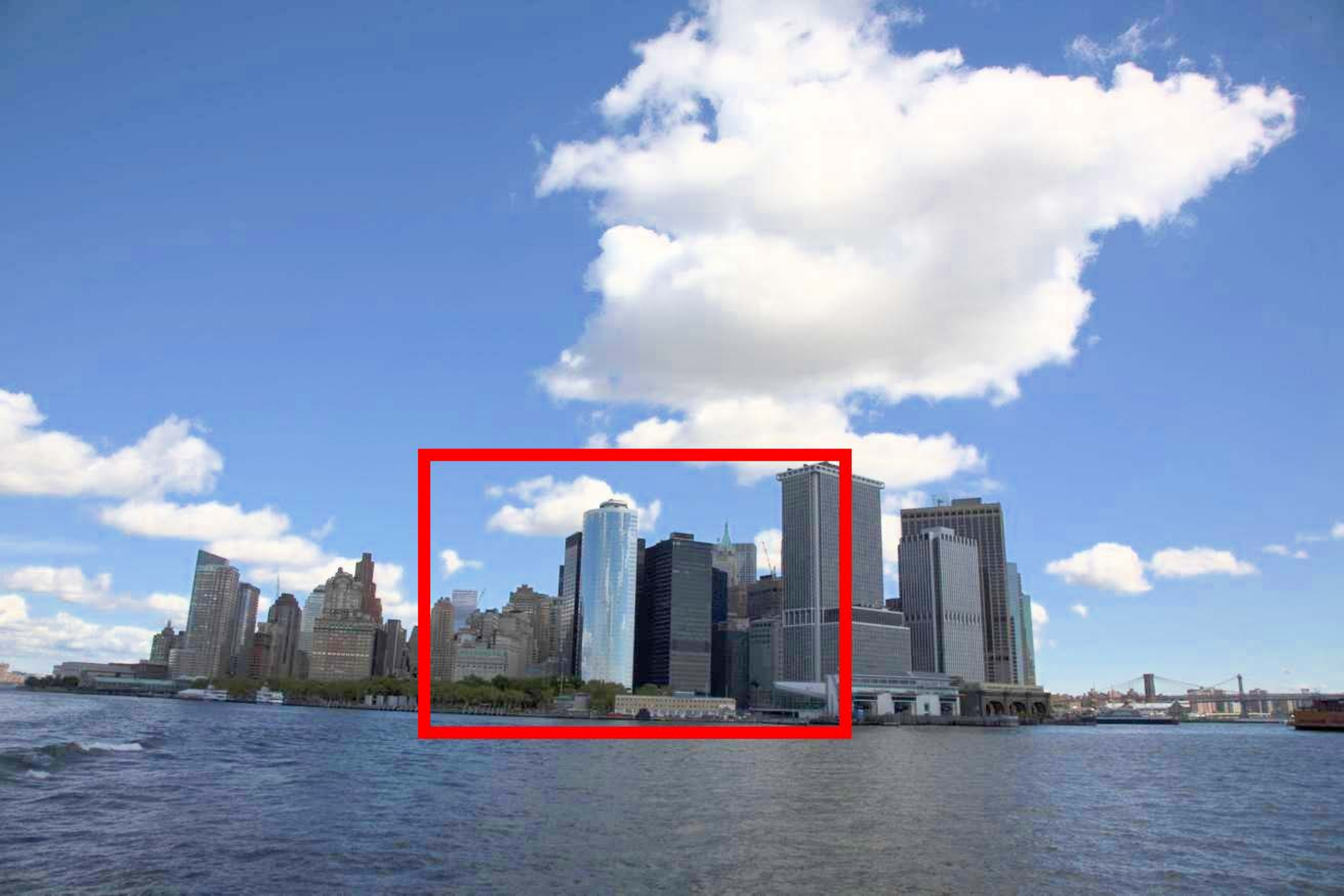}\\\vspace{-0.2cm}
			\includegraphics[width=0.16\linewidth]{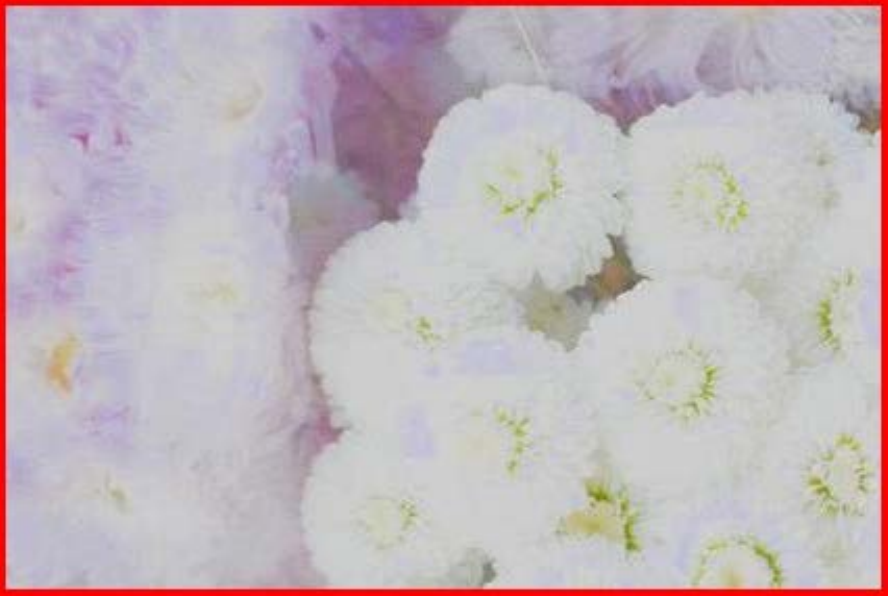}&
			\includegraphics[width=0.16\linewidth]{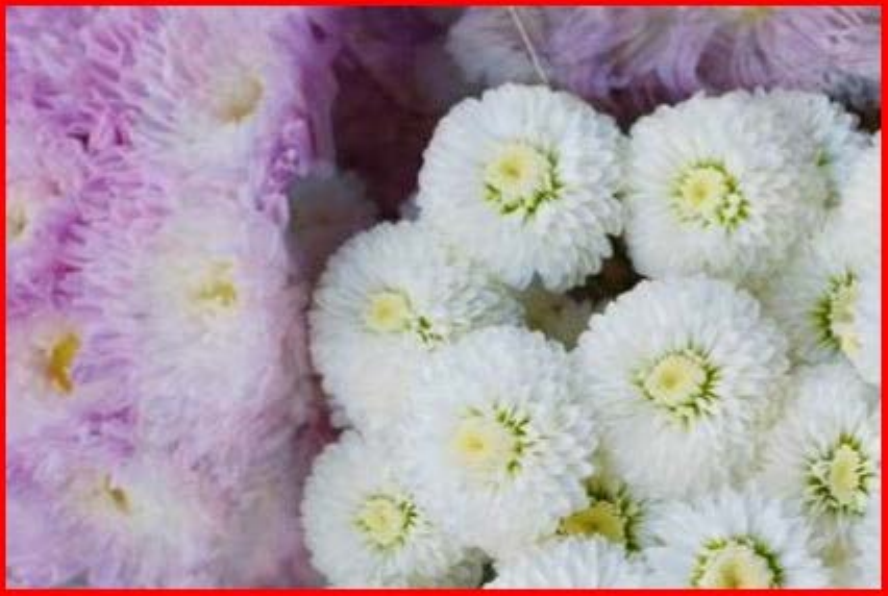}&
			\includegraphics[width=0.16\linewidth]{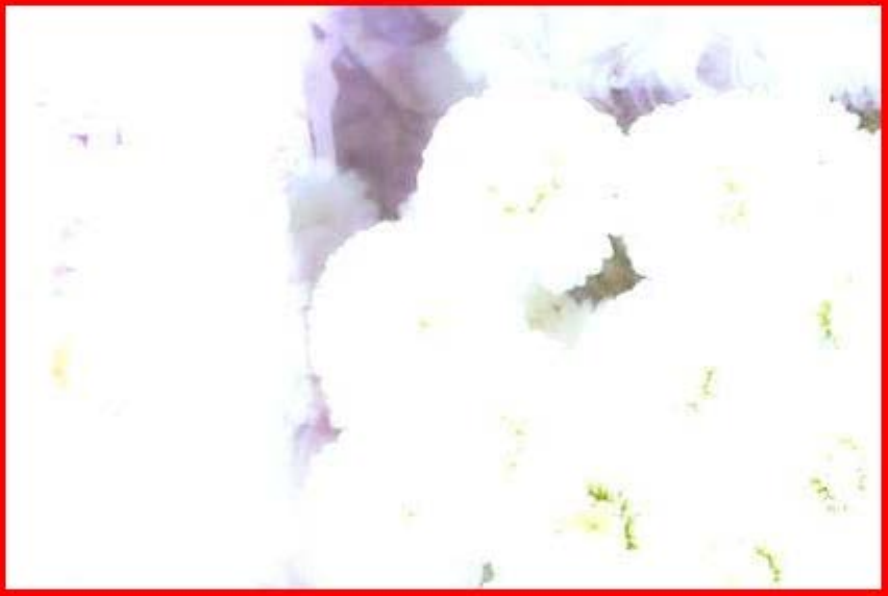}&
			\includegraphics[width=0.16\linewidth]{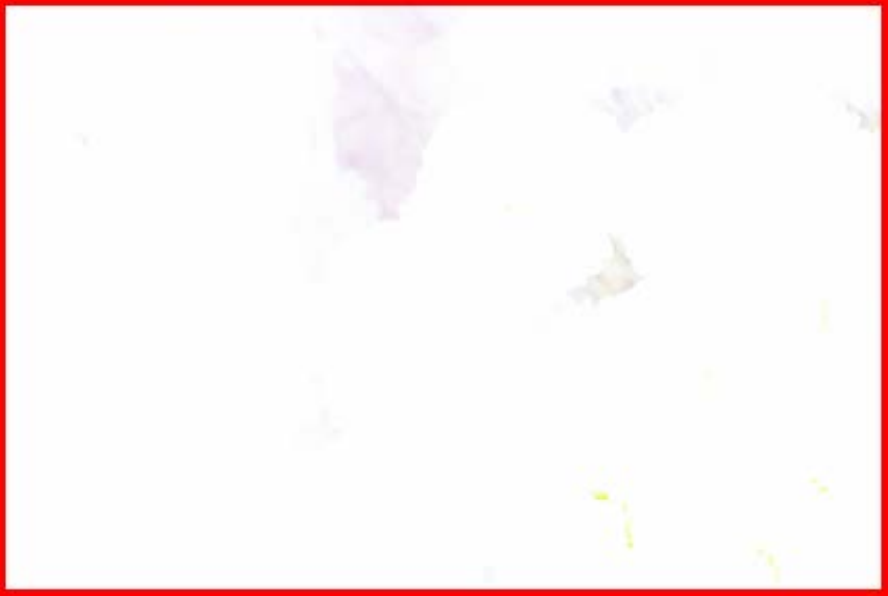}&
			\includegraphics[width=0.16\linewidth]{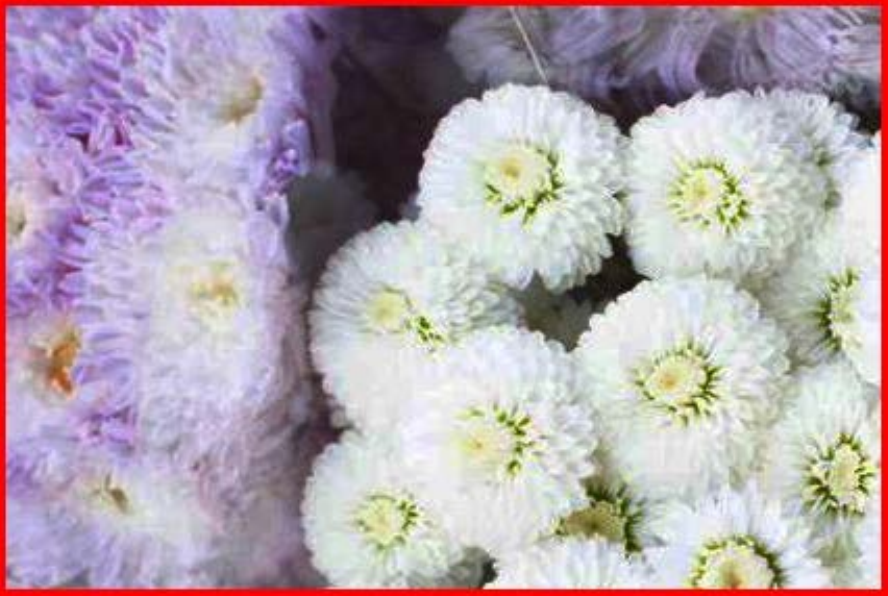}&
			\includegraphics[width=0.16\linewidth]{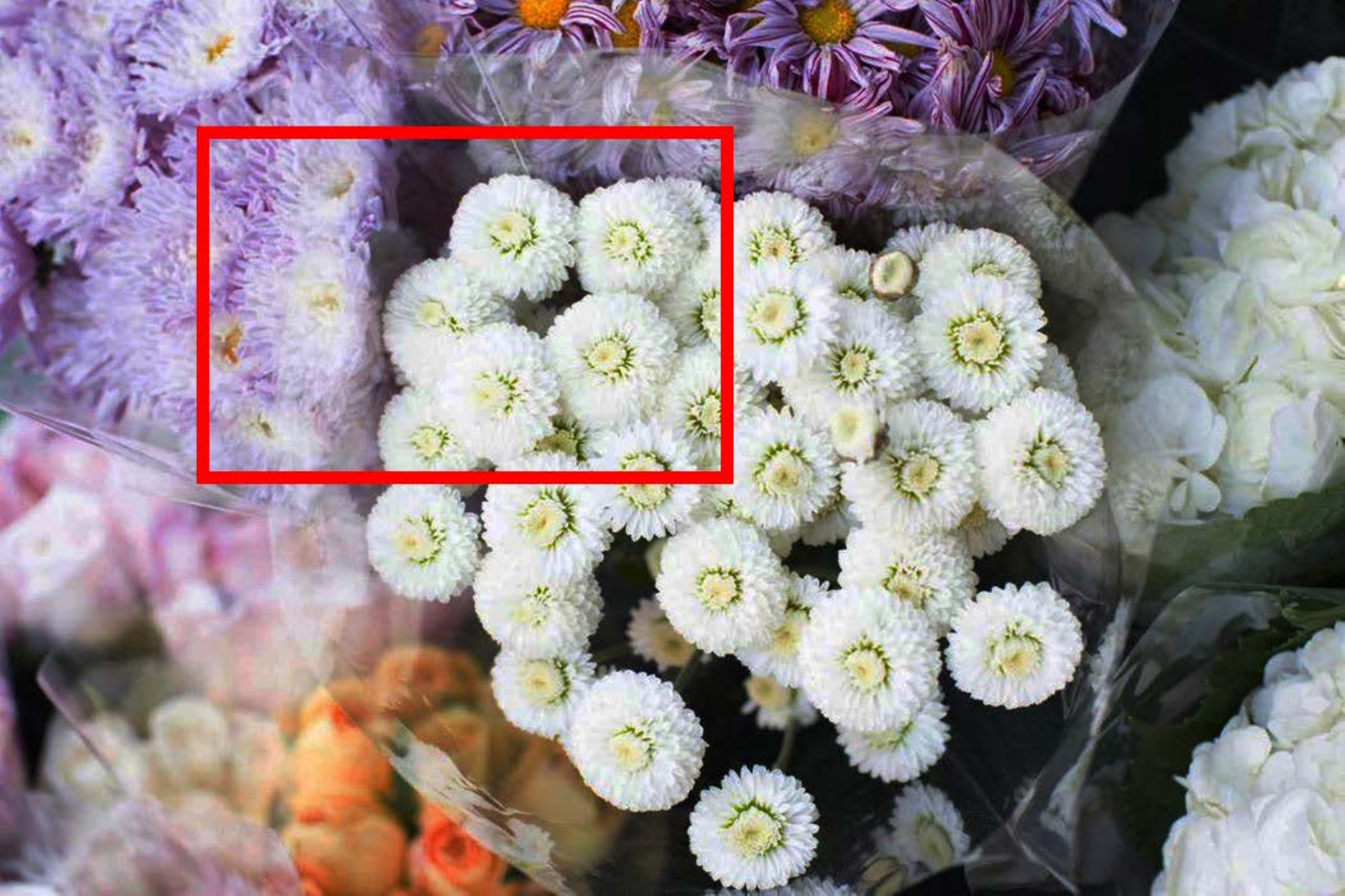}\\
			\footnotesize LCD~\cite{wang2022local}&\footnotesize FEC~\cite{huang2022eccv}&\footnotesize URetinex~\cite{wu2022uretinex}&\footnotesize SCI~\cite{ma2022toward}&\footnotesize PEC&\footnotesize PEC (Full Size)\\
		\end{tabular}	
		\caption{Visual comparison on the Exposure-Errors dataset~\cite{afifi2021learning}.}
		\label{fig: MSEC}
	\end{figure*}

	\begin{table*}[t]
		\footnotesize
		\renewcommand\arraystretch{1.2} 
		\setlength{\tabcolsep}{0.7mm}
		\centering
		\caption{Numerical scores among different methods using MEF and LIME datasets.}
		\begin{tabular}{|c|c|c|c|c|c|c|c|c|c|c|c|}
			\hline 
			Datasets&Metrics&SRIE~\cite{fu2015probabilistic}&NPEA~\cite{wang2013naturalness}&WVM~\cite{fu2016weighted}&JIEP~\cite{cai2017joint}&LIME~\cite{guo2017lime}&RRM~\cite{li2018structure}&LR3M~\cite{ren2020lr3m}&STAR~\cite{xu2020star}&SDD~\cite{hao2020low}&PEC\\
			\hline
			\multirow{3}{*}{MEF}&DE$\uparrow$&6.9282&7.0714&6.9436&7.0119&{\textbf{7.3263}}&7.0530&6.9869&6.9377&{\underline{7.0835}}&7.0390\\
			~&LOE$\downarrow$&261.70&422.32&210.26&469.86&802.33&311.39&359.26&{\underline{59.33}}&436.54&{\textbf{30.76}}\\  
			~&NIQE$\downarrow$&4.6592&3.5469&3.4742&{\underline{3.4314}}&3.7662&3.9385&4.4172&3.5251&4.3136&{\textbf{3.0053}}\\
			\hline
			\multirow{3}{*}{LIME}&DE$\uparrow$&6.8967&6.8963&6.8516&6.8924&{\textbf{7.5492}}&6.9878&6.8873&6.7729&7.0355&{\underline{7.0752}}\\
			~&LOE$\downarrow$&212.61&436.45&140.12&226.84&621.42&283.46&297.97&{\textbf{51.39}}&309.93&{\underline{76.63}}\\  
			~&NIQE$\downarrow$&4.9814&5.2273&4.0273&{\underline{3.9927}}&4.1357&4.2415&4.4396&4.4369&4.5146&{\textbf{3.8963}}\\
			\hline
		\end{tabular}
		\label{table: Score of traditional}
	\end{table*}
	
	\begin{figure*}[t]
		\centering
		\begin{tabular}{c@{\extracolsep{0.3em}}c@{\extracolsep{0.3em}}c@{\extracolsep{0.3em}}c@{\extracolsep{0.3em}}c@{\extracolsep{0.3em}}c@{\extracolsep{0.3em}}c@{\extracolsep{0.3em}}c@{\extracolsep{0.3em}}c}			
			\vspace{-0.2cm}
			\includegraphics[height=0.143\linewidth]{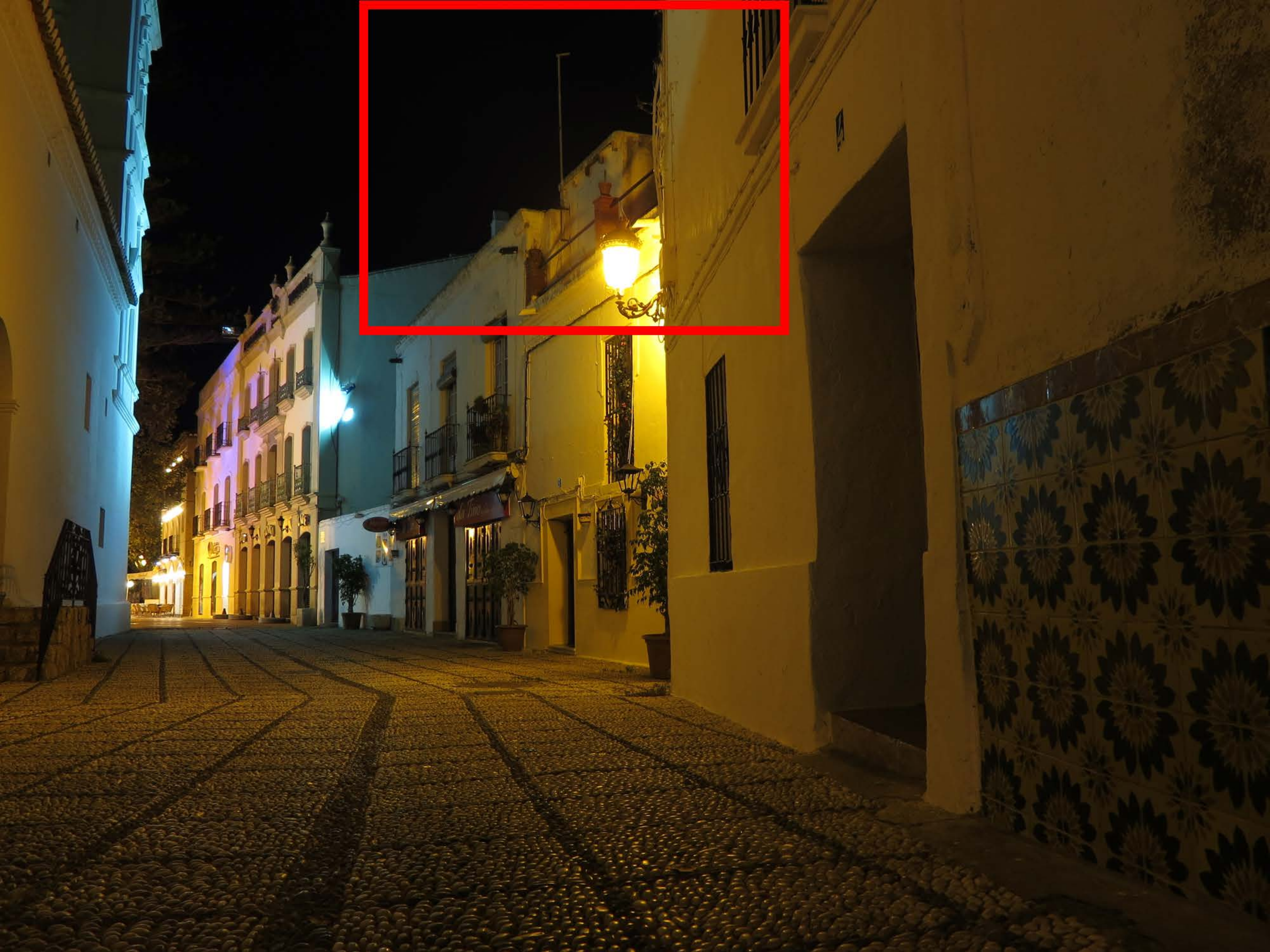}&
			\includegraphics[height=0.143\linewidth]{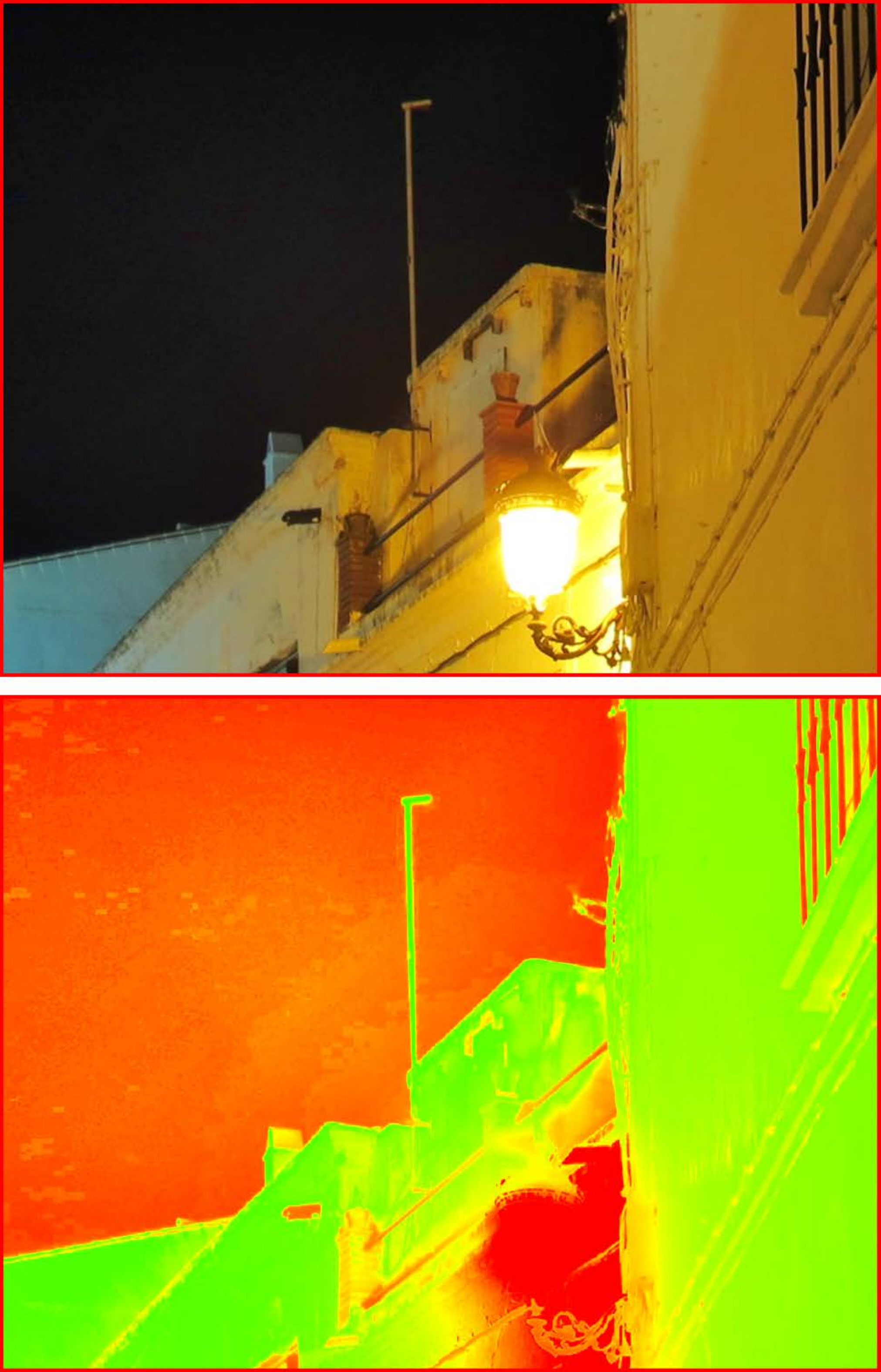}&
			\includegraphics[height=0.143\linewidth]{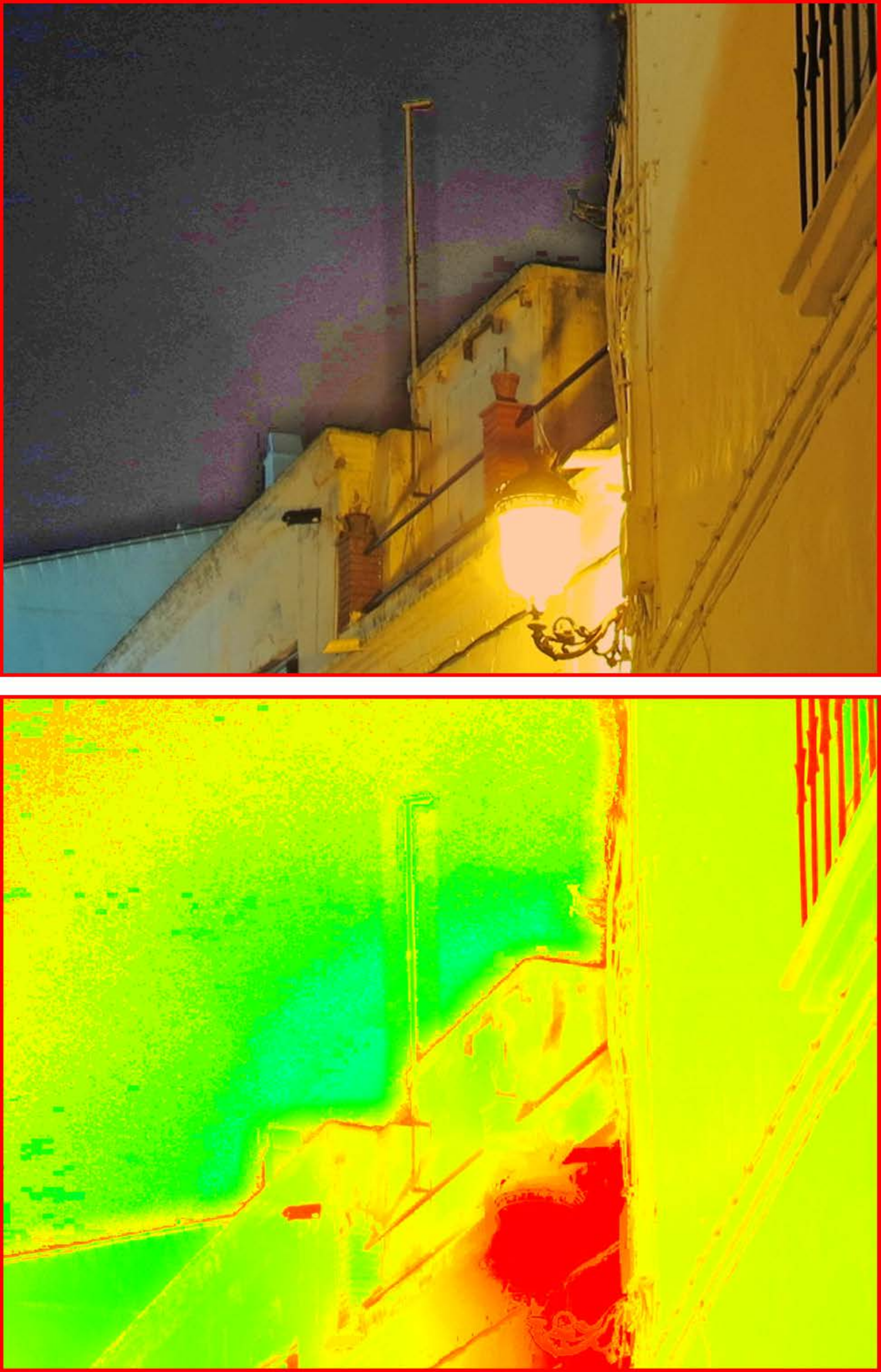}&
			\includegraphics[height=0.143\linewidth]{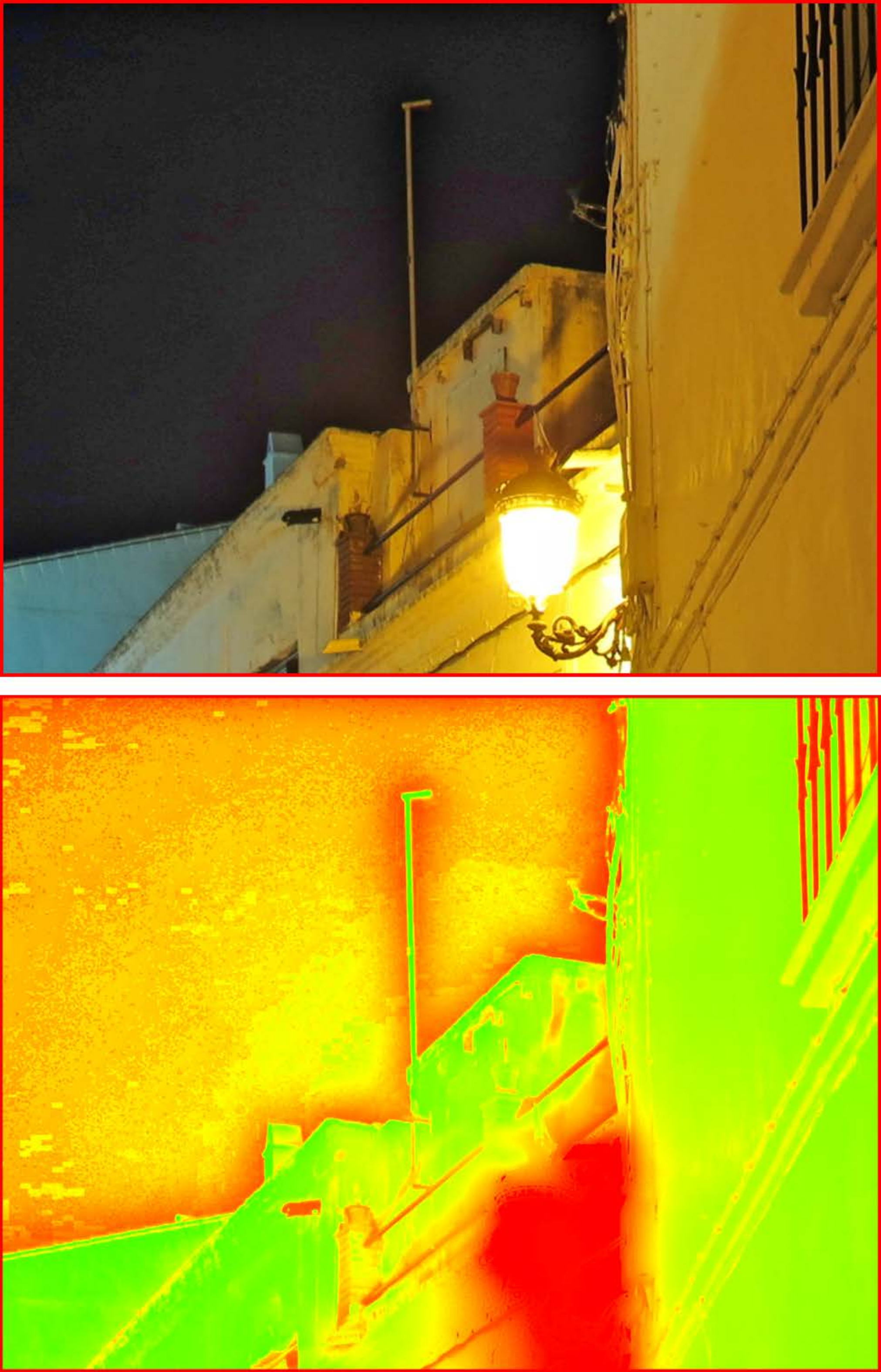}&
			\includegraphics[height=0.143\linewidth]{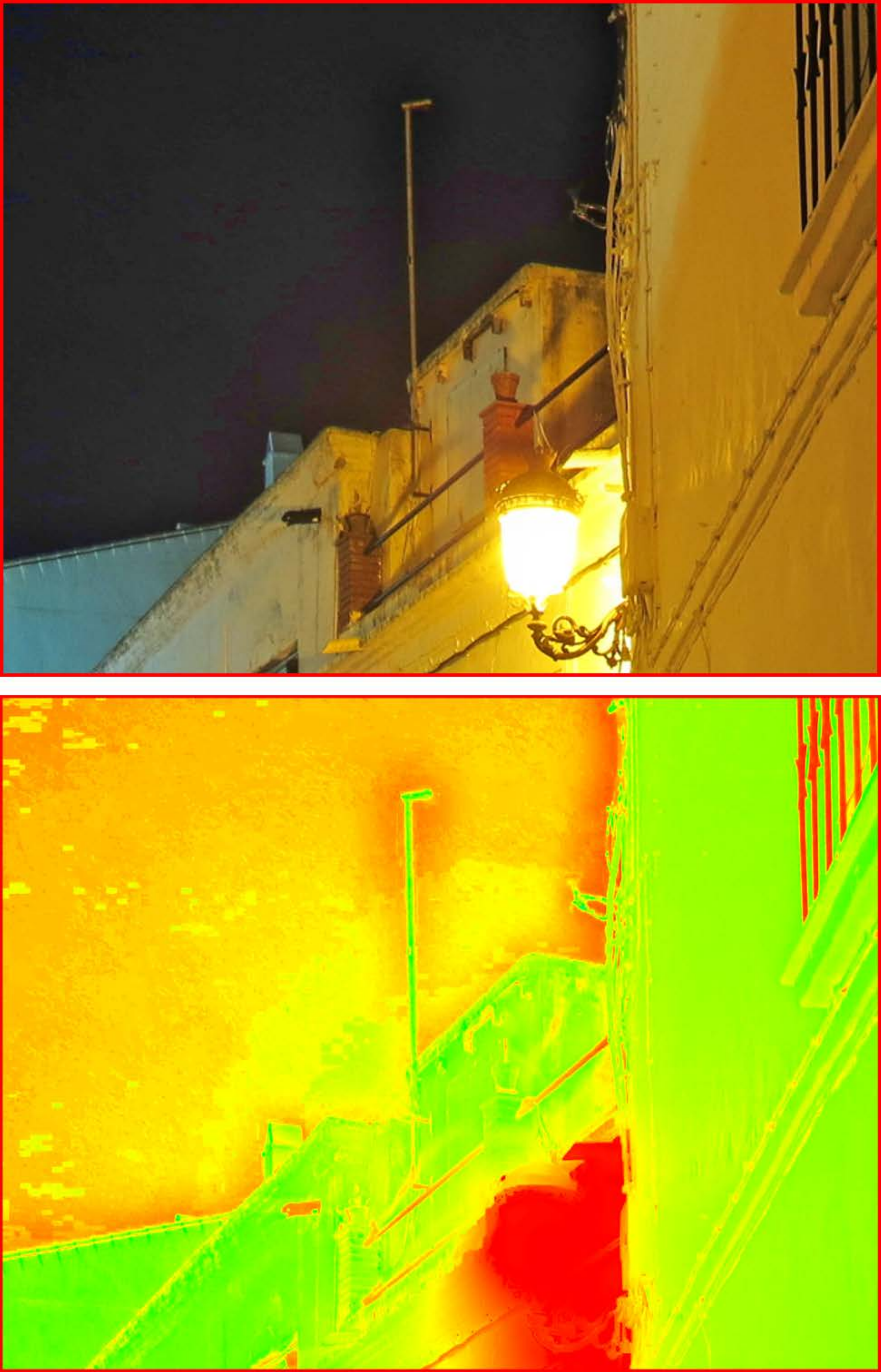}&
			\includegraphics[height=0.143\linewidth]{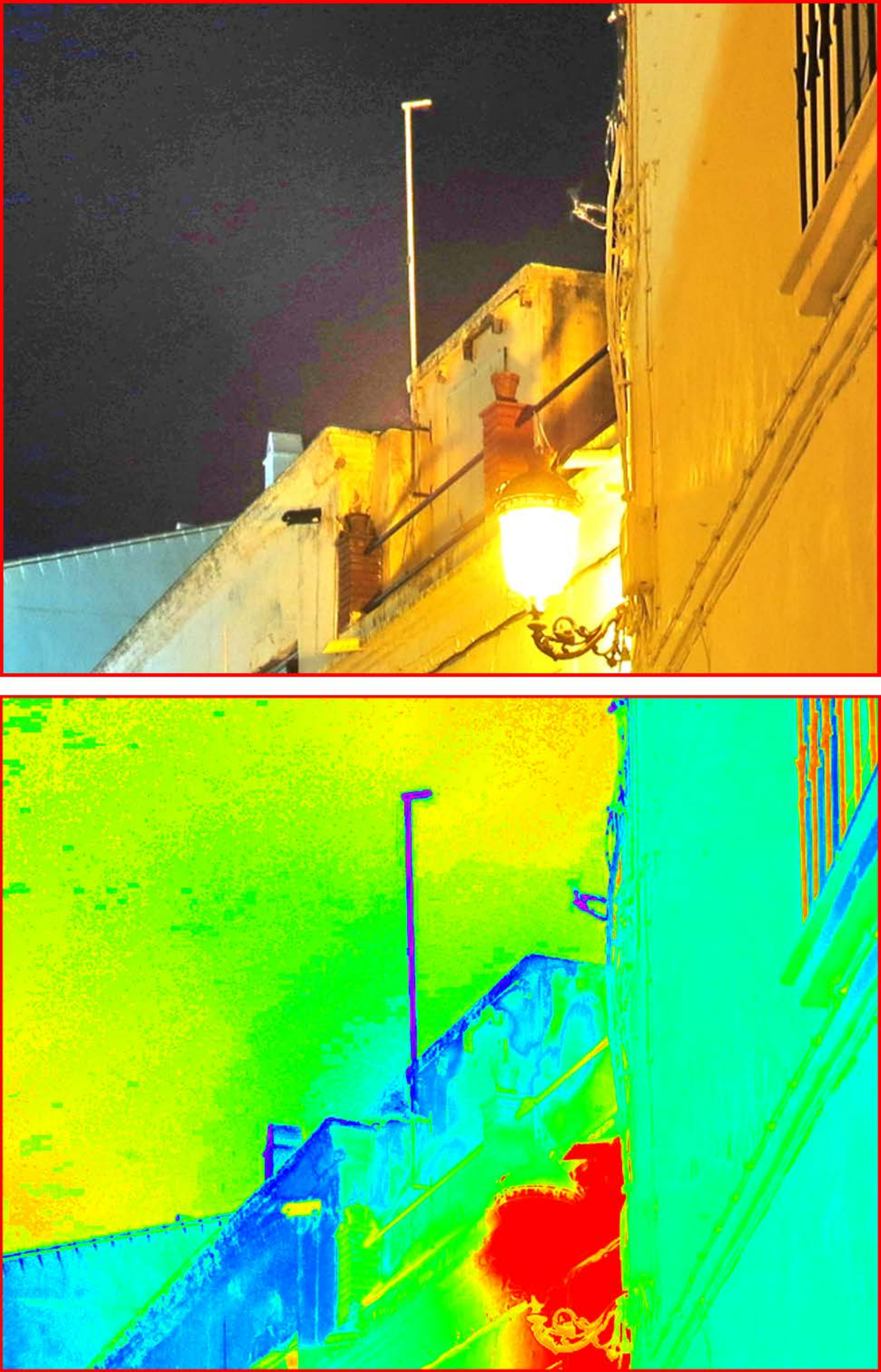}&
			\includegraphics[height=0.143\linewidth]{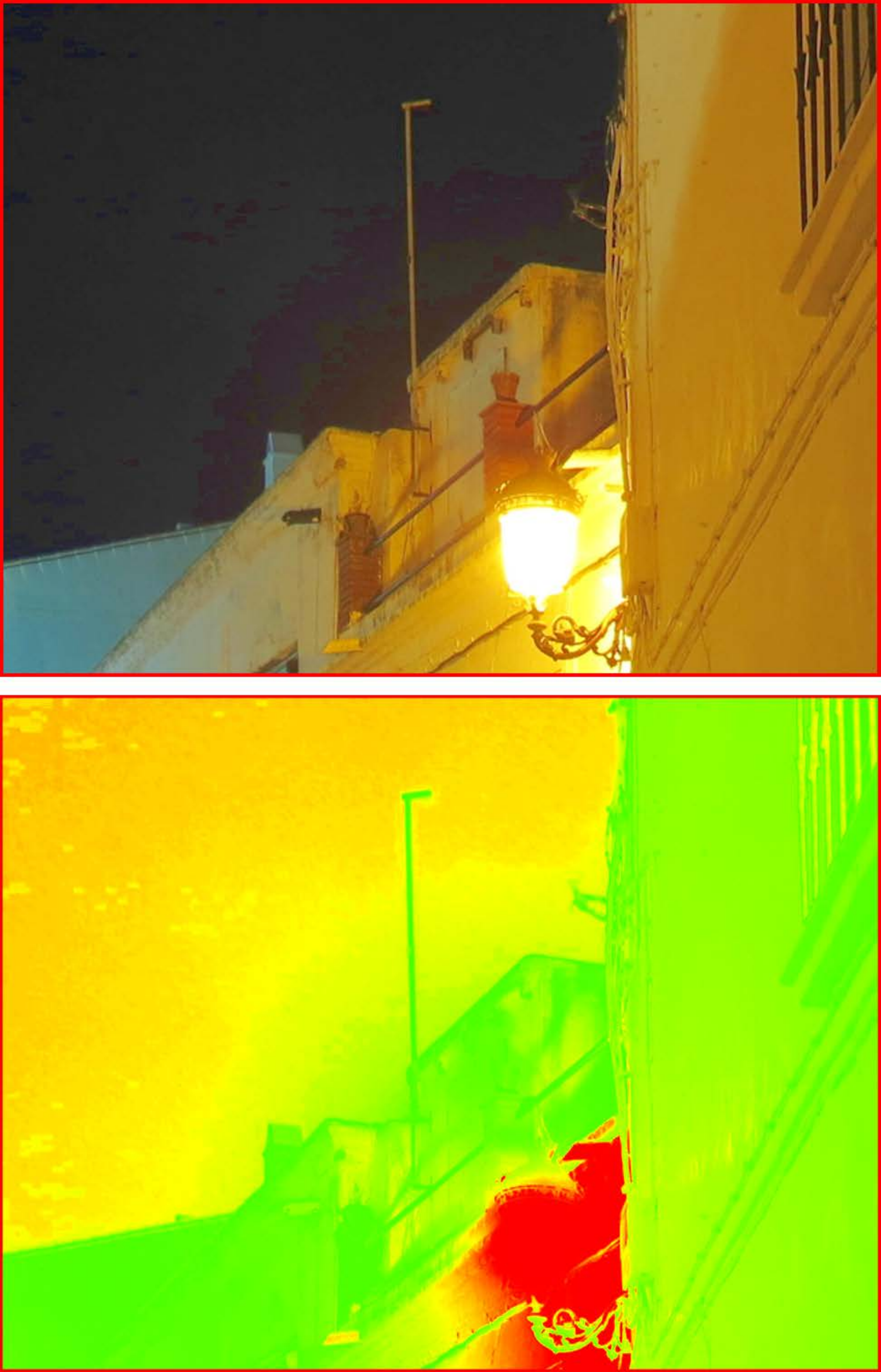}&
			\includegraphics[height=0.143\linewidth]{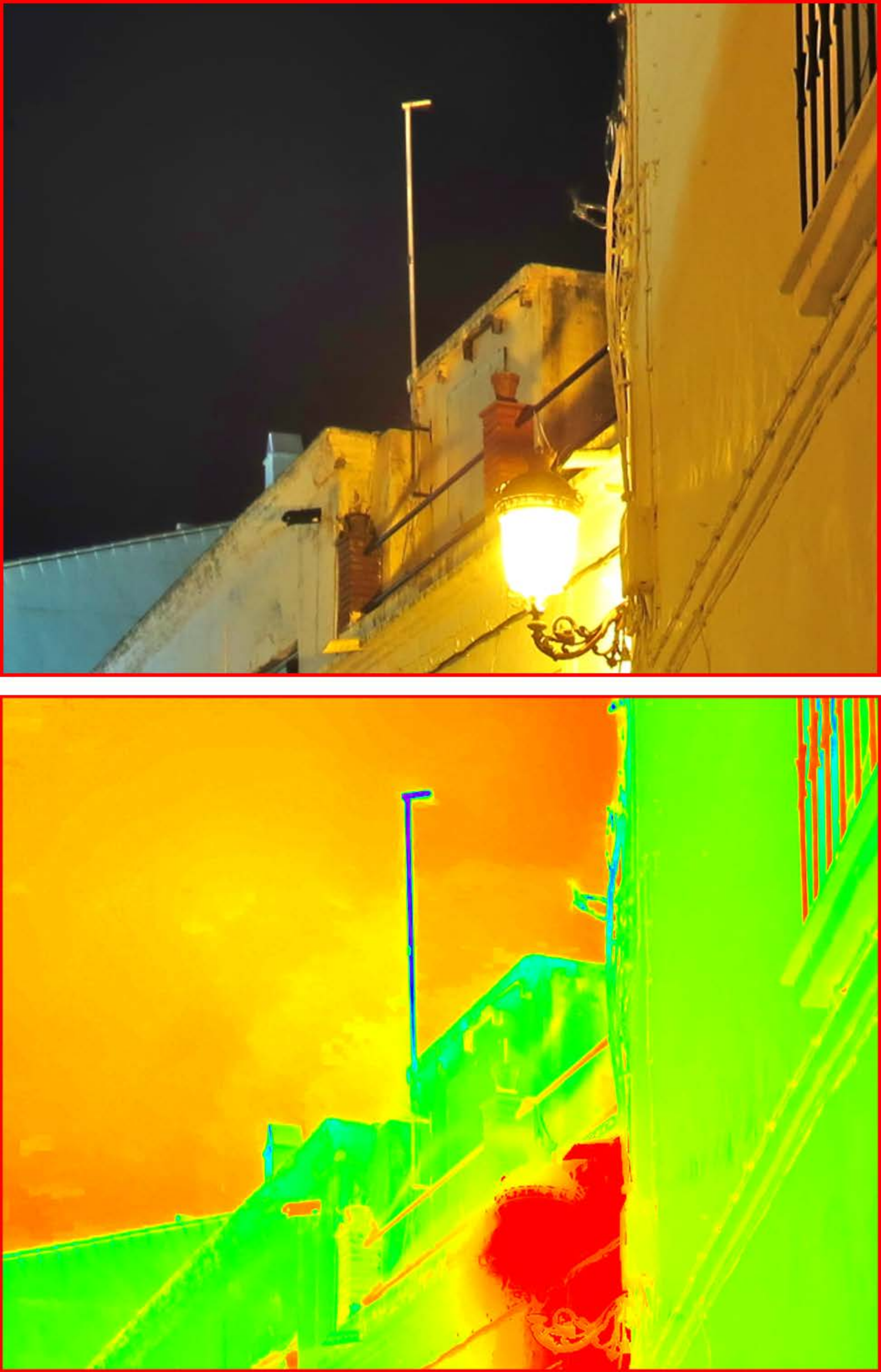}&
			\includegraphics[height=0.143\linewidth]{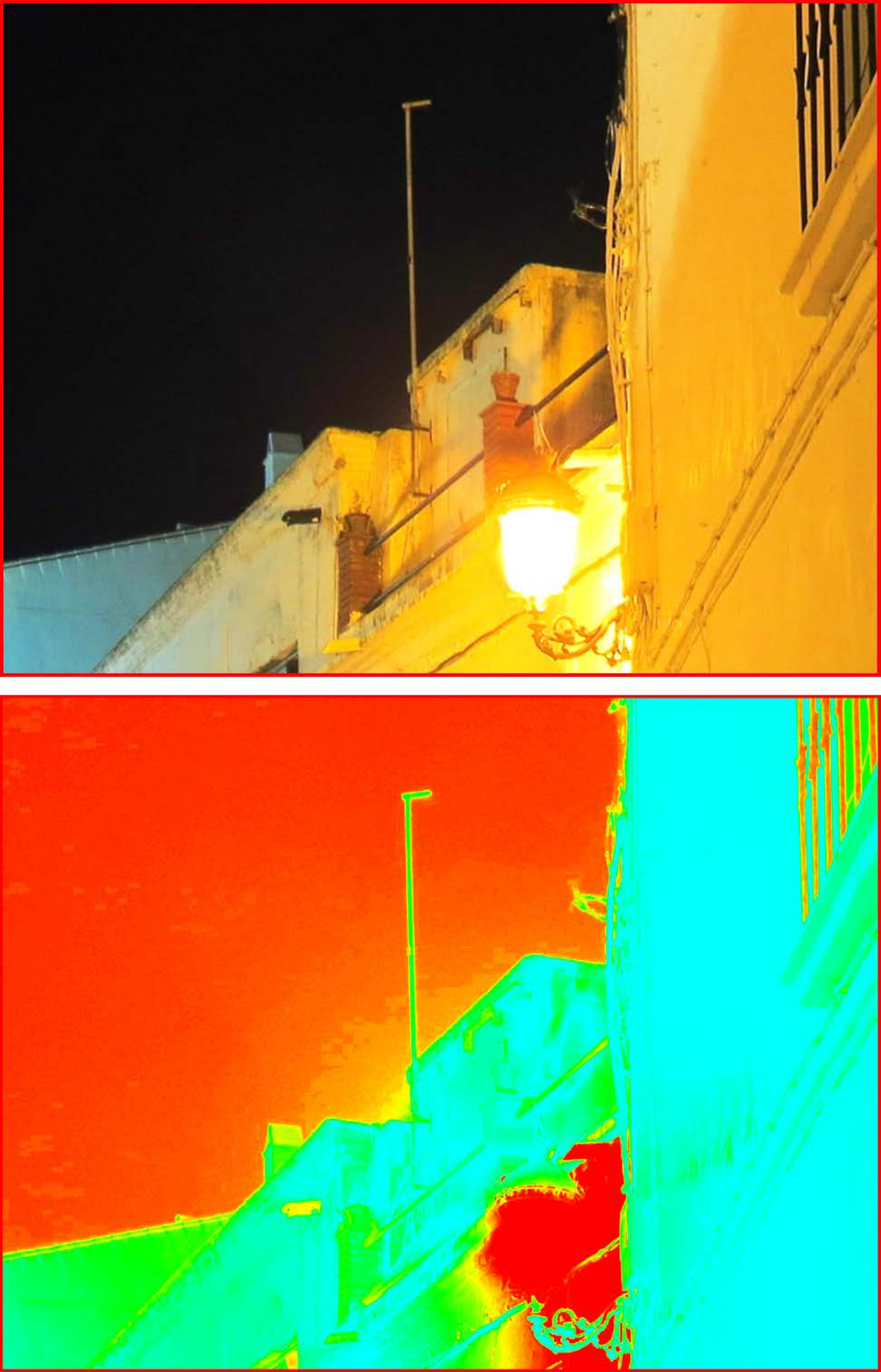}\\
			\footnotesize Input&\footnotesize SRIE~\cite{fu2015probabilistic}&\footnotesize NPEA~\cite{wang2013naturalness}&\footnotesize WVM~\cite{fu2016weighted}&\footnotesize JIEP~\cite{cai2017joint}&\footnotesize LIME~\cite{guo2017lime}&\footnotesize STAR~\cite{xu2020star}&\footnotesize SDD~\cite{hao2020low}&\footnotesize {PEC}\\
		\end{tabular}
		\caption{Visual comparison on the LIME dataset. From the second to last column, we plot the zoomed-in regions and the corresponding difference map with the original input in the gray-scale space. The HSV colormap is applied to highlight the results.} 
		\label{fig: Others}
	\end{figure*}
	
	\begin{figure*}[t]
		\centering
		\footnotesize
		\begin{tabular}{c}			
			\includegraphics[width=0.98\linewidth]{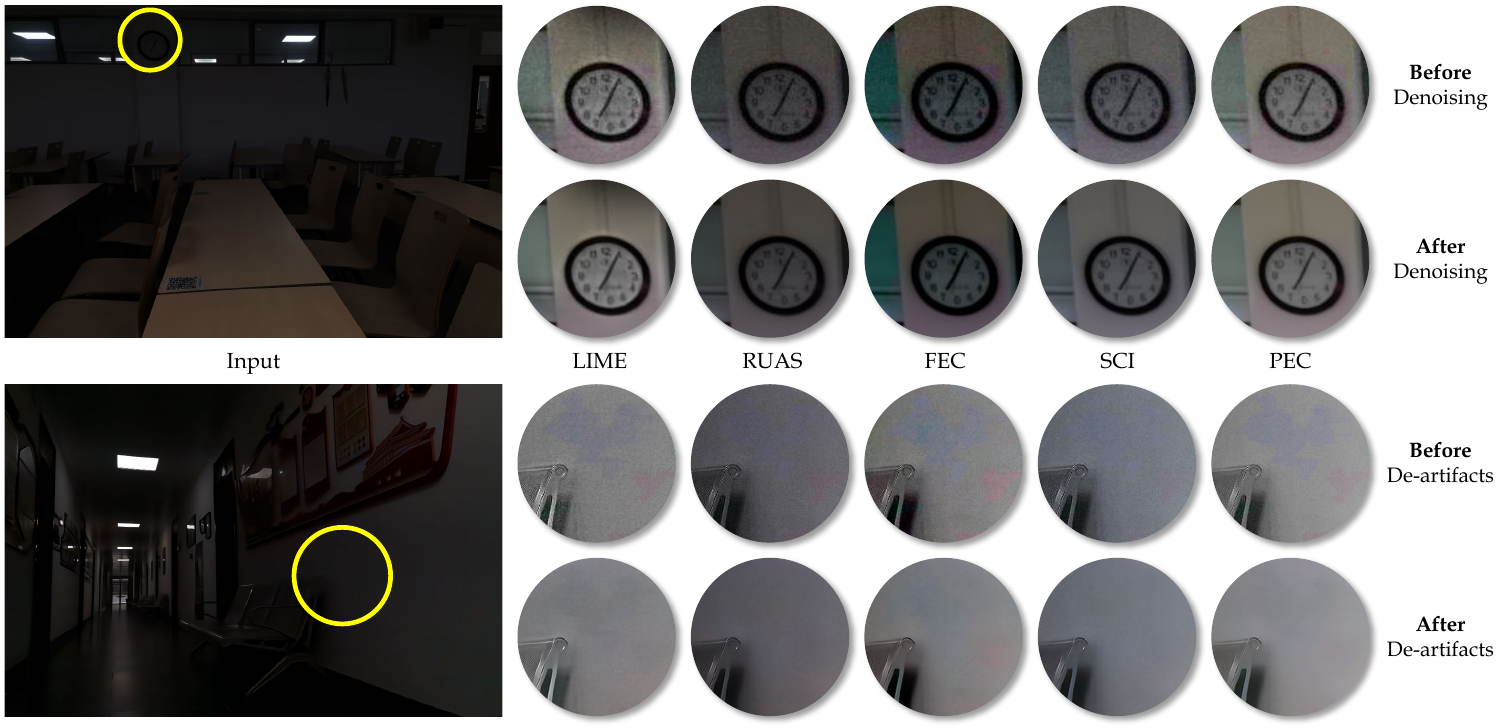}\\
		\end{tabular}
		\caption{Analysis of the necessity of post-processing operations for challenging underexposure scenarios with noise/artifacts.}
		\label{fig: Analysis}
	\end{figure*}

	\begin{table*}[t]
		\footnotesize
		\renewcommand\arraystretch{1.2} 
		\setlength{\tabcolsep}{3mm}
		\centering
		\caption{Comparison of running time (seconds) for samples with different resolutions from the Exposure-Errors dataset. The running time is computed on a PC with a GeForce RTX 2080Ti GPU and an Intel Xeon W-2135 processor.}
		\begin{threeparttable}[b]
			\begin{tabular}{|c|c|c|c|c|c|c|}
				\hline 
				\multirow{2}{*}{Method}&\multicolumn{2}{c|}{1280$\times$720}&\multicolumn{2}{c|}{1920$\times$1080}&\multicolumn{2}{c|}{2560$\times$1440}\\
				\cline{2-7}
				~&CPU&GPU&CPU&GPU&CPU&GPU\\
				\hline
				KinD~\cite{zhang2019kindling}&2.3844&2.1340&5.1322&4.6483&8.0632&7.9920\\
				\hline
				ZeroDCE~\cite{guo2020zero}&17.7944&0.0013&40.4684&0.0014&69.7165&0.0016\\
				\hline
				MSEC\tnote{$\dagger$}~\cite{afifi2021learning}&0.3007&0.2349&0.4899&0.4730&0.9081&0.8838\\
				\hline
				RUAS~\cite{liu2021retinex}&2.8807&0.0309&4.8915&0.0463&5.6561&0.0821\\
				\hline
				UTVNet\tnote{$\ddagger$}~\cite{zheng2021adaptive}&8.6610&0.3837&13.4690&-&22.5862&-\\
				\hline
				LCD~\cite{wang2022local}&4.1292&1.6064&8.2461&2.8085&14.7200&4.7721\\
				\hline
				FEC~\cite{huang2022eccv}&4.8920&0.0132&12.1834&0.0143&23.0975&0.0164\\
				\hline
				URetinex~\cite{wu2022uretinex}&8.9093&0.2348&23.3995&0.4876&33.4536&0.8318\\
				\hline
				SCI~\cite{ma2022toward}&{\underline{0.0823}}&{\underline{0.0007}}&{\underline{0.1550}}&{\underline{0.0009}}&{\underline{0.2702}}&{\underline{0.0020}}\\
				\hline
				{PEC}&{\textbf{0.0269}}{\tiny  \textbf{$\uparrow${67\%}}}&{\textbf{0.0003}}{\tiny  \textbf{$\uparrow${57\%}}}&{\textbf{0.0515}}{\tiny  \textbf{$\uparrow${66\%}}}&{\textbf{0.0006}}{\tiny  \textbf{$\uparrow${33\%}}}&{\textbf{0.0927}}{\tiny  \textbf{$\uparrow${65\%}}}&{\textbf{0.0009}}{\tiny  \textbf{$\uparrow${55\%}}}\\
				\hline
			\end{tabular} 
			\begin{tablenotes}
				\tiny \item[$\dagger$]The running platform in this study is different from other methods and was evaluated using a PC with a NVIDIA TITAN Xp GPU.
				\tiny \item[$\ddagger$]The developed approach requires too much video memory for high-resolution images, leading to a failure in producing results. 
			\end{tablenotes}
		\end{threeparttable}
		\label{table: PC Time}
	\end{table*}
    
	\begin{figure*}[t]
		\centering
		\begin{tabular}{c@{\extracolsep{0.2em}}c@{\extracolsep{0.2em}}c@{\extracolsep{0.2em}}c@{\extracolsep{0.2em}}c@{\extracolsep{0.2em}}c}	
			\vspace{-0.1cm}	
			\includegraphics[width=0.16\linewidth]{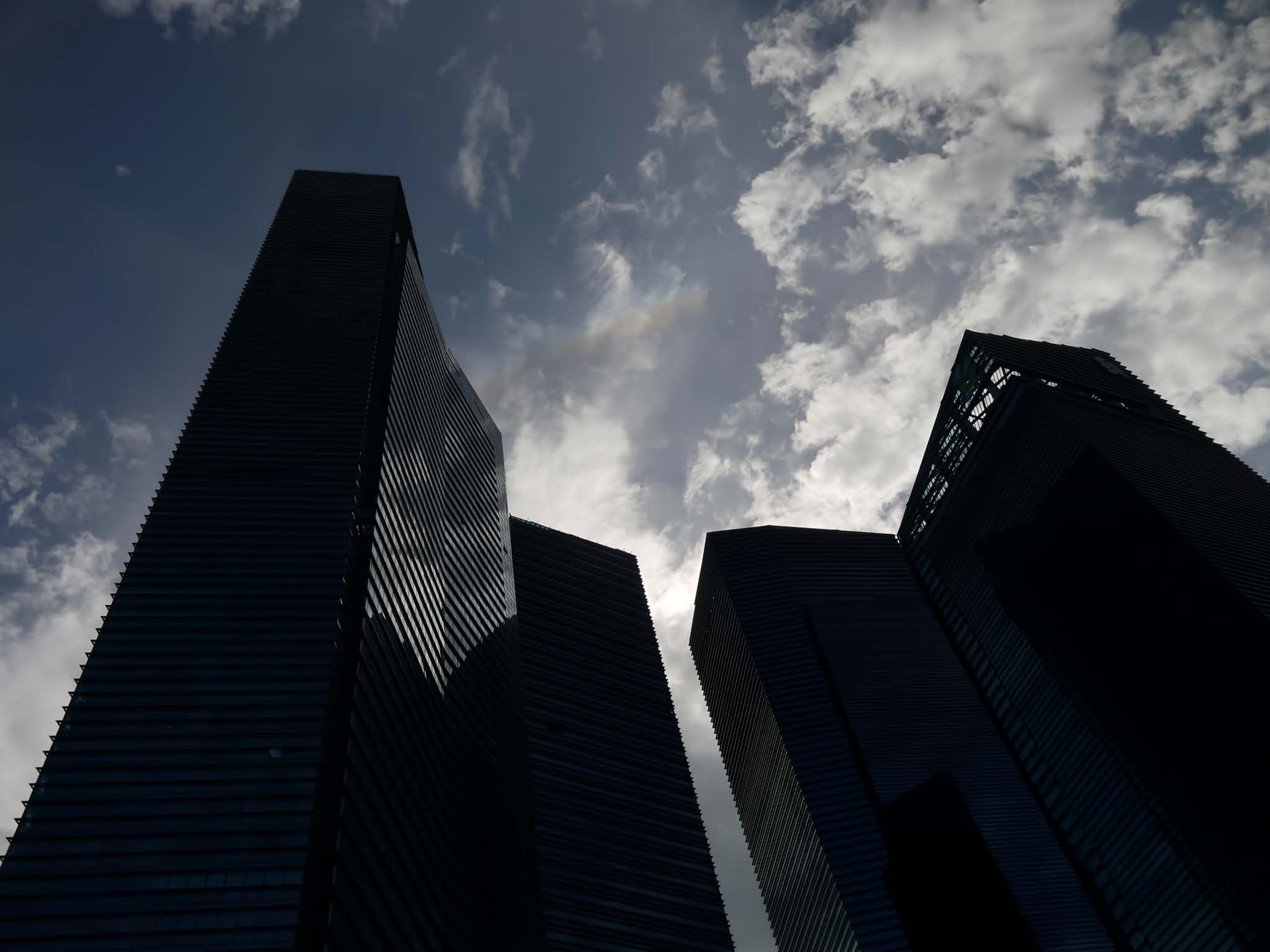}&	
			\includegraphics[width=0.16\linewidth]{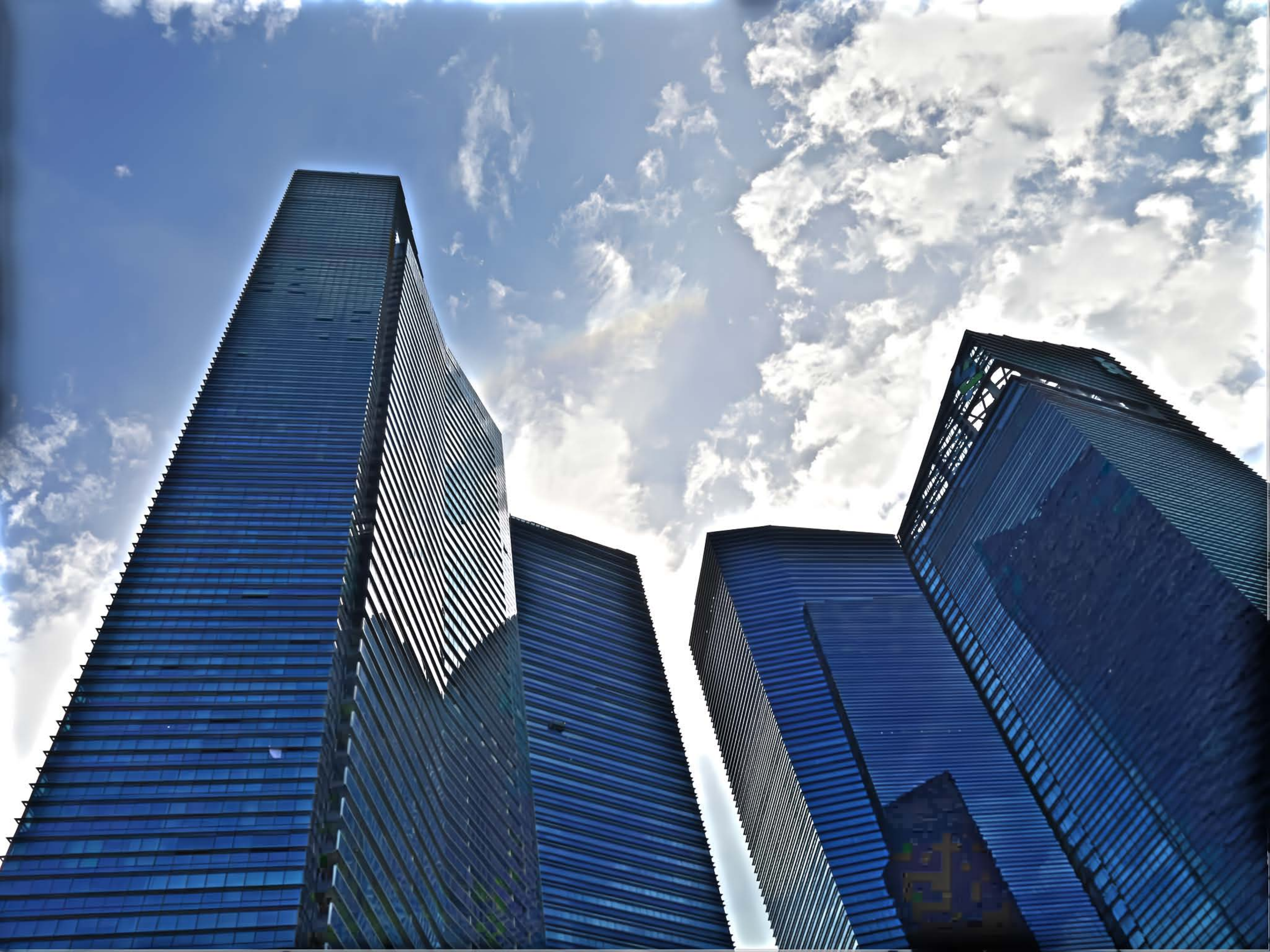}&	
			\includegraphics[width=0.16\linewidth]{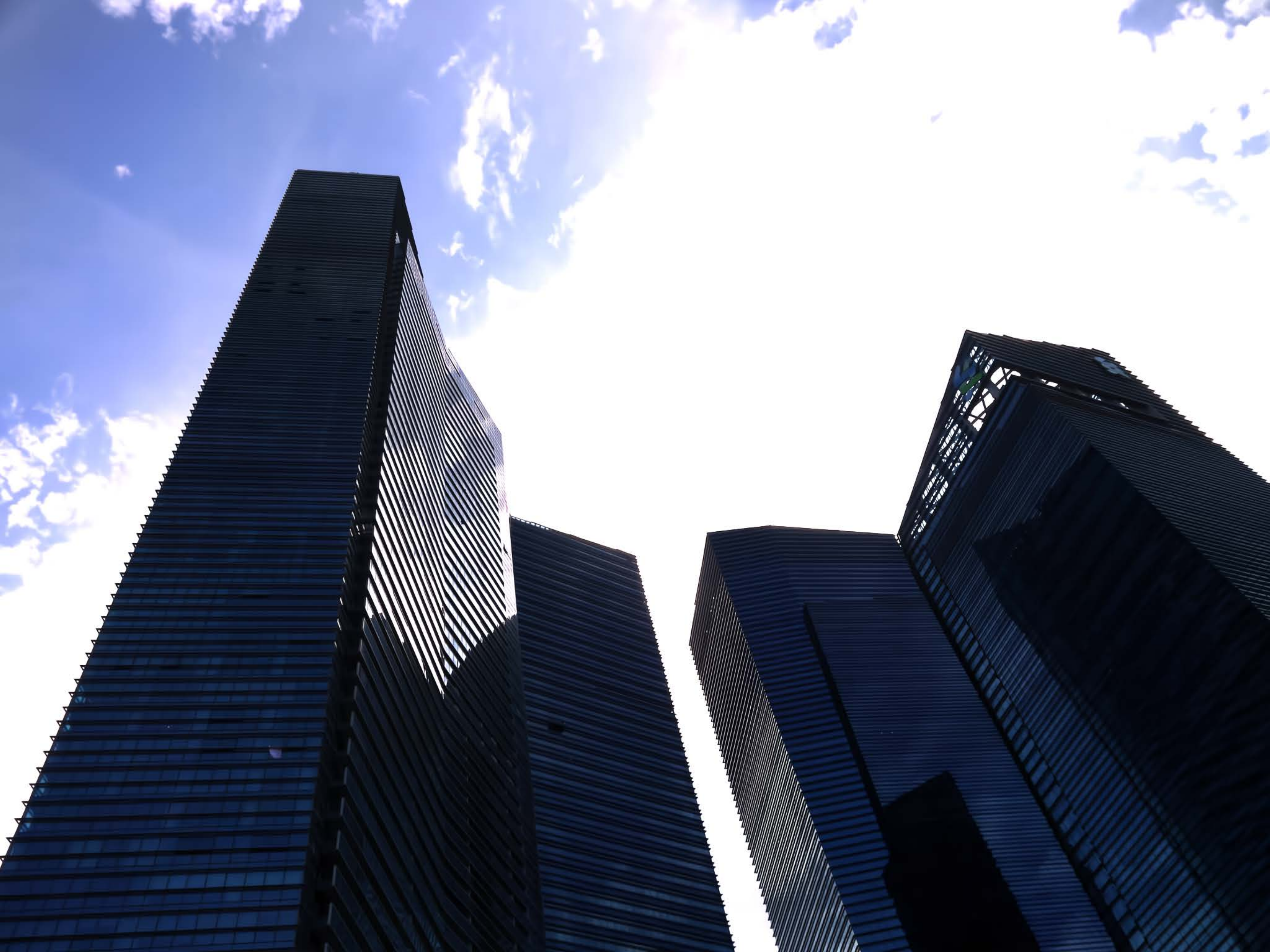}&	
			\includegraphics[width=0.16\linewidth]{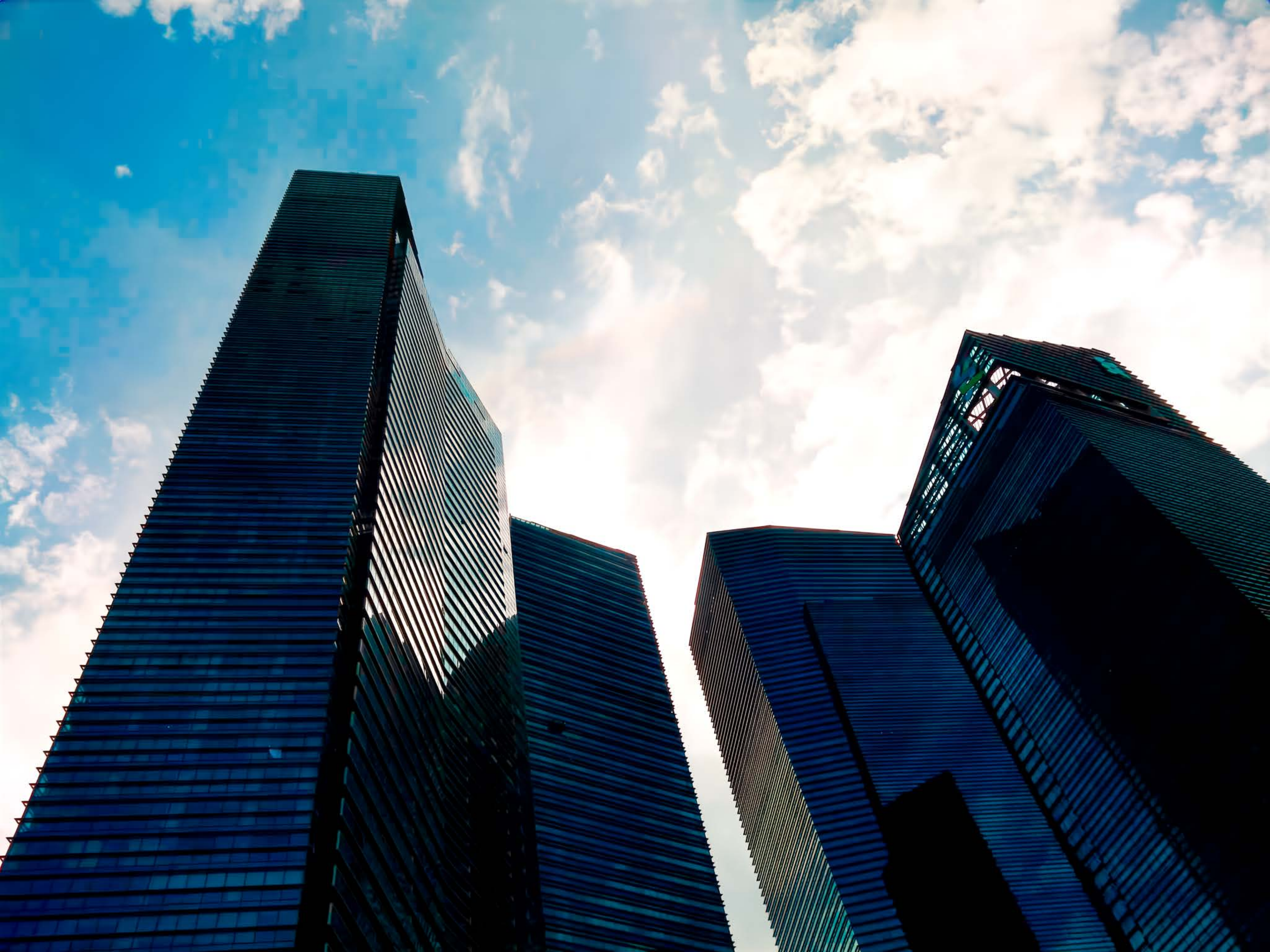}&	
			\includegraphics[width=0.16\linewidth]{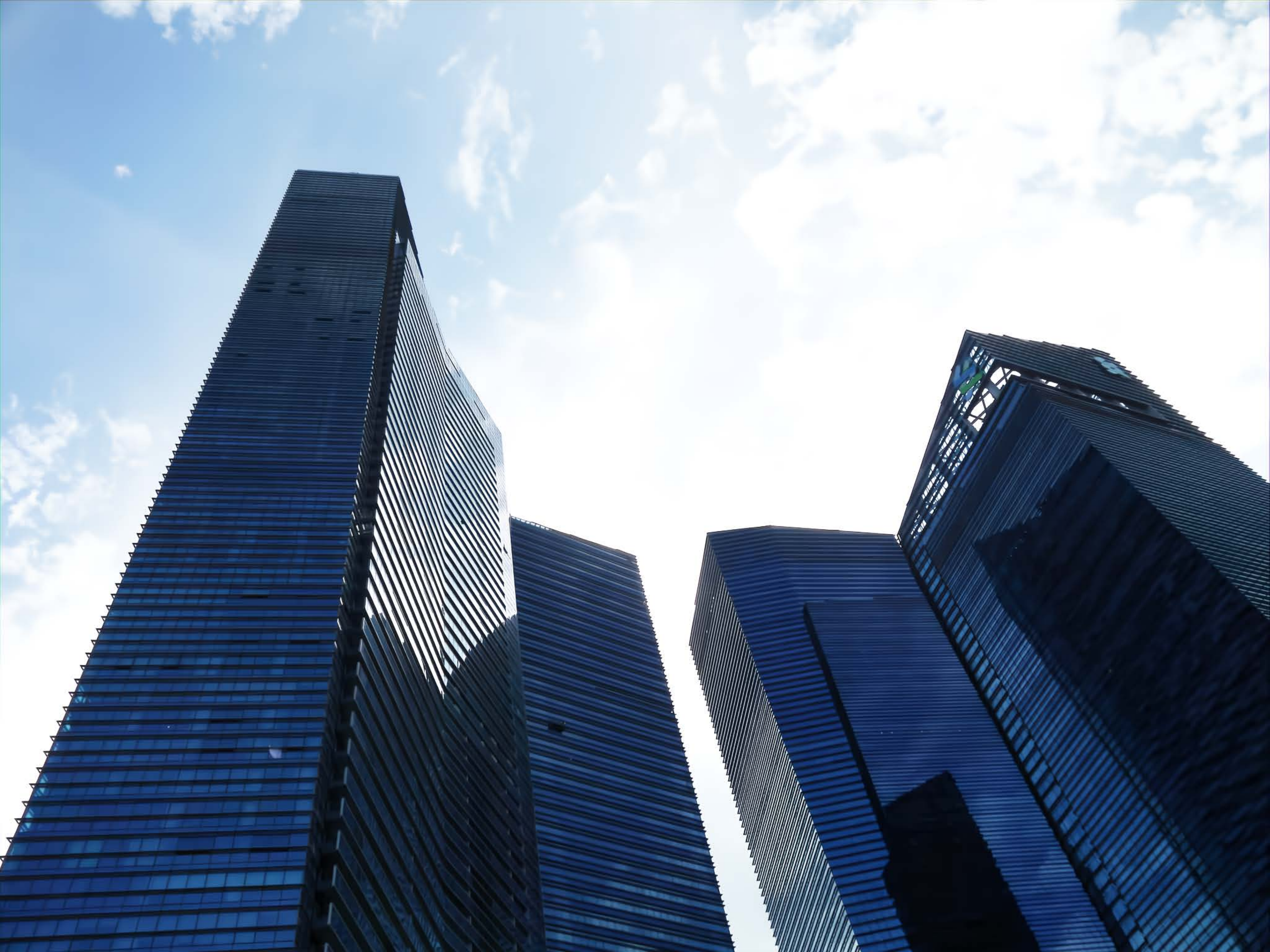}&	
			\includegraphics[width=0.16\linewidth]{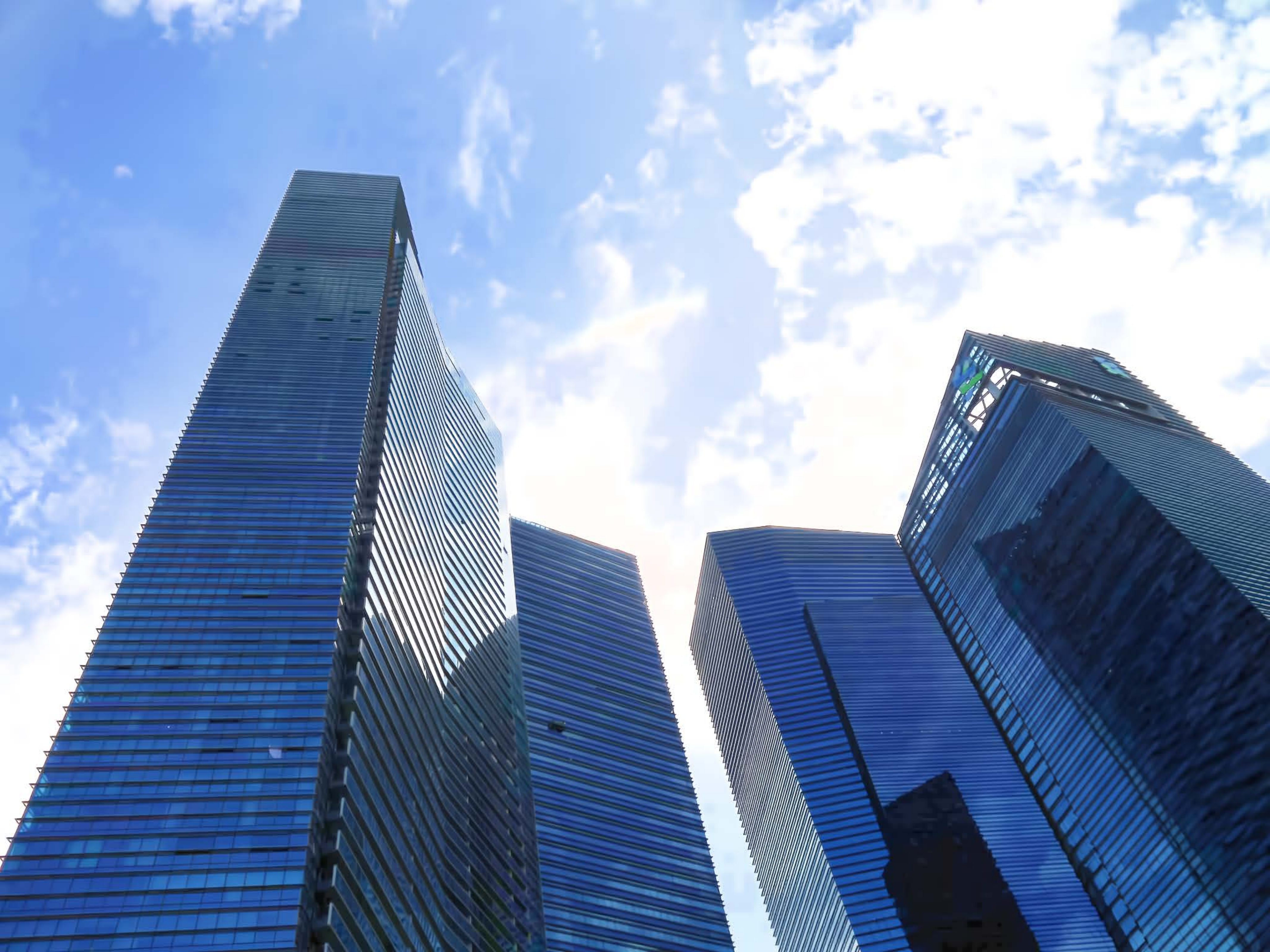}\\
			\vspace{-0.1cm}	
			\includegraphics[width=0.16\linewidth]{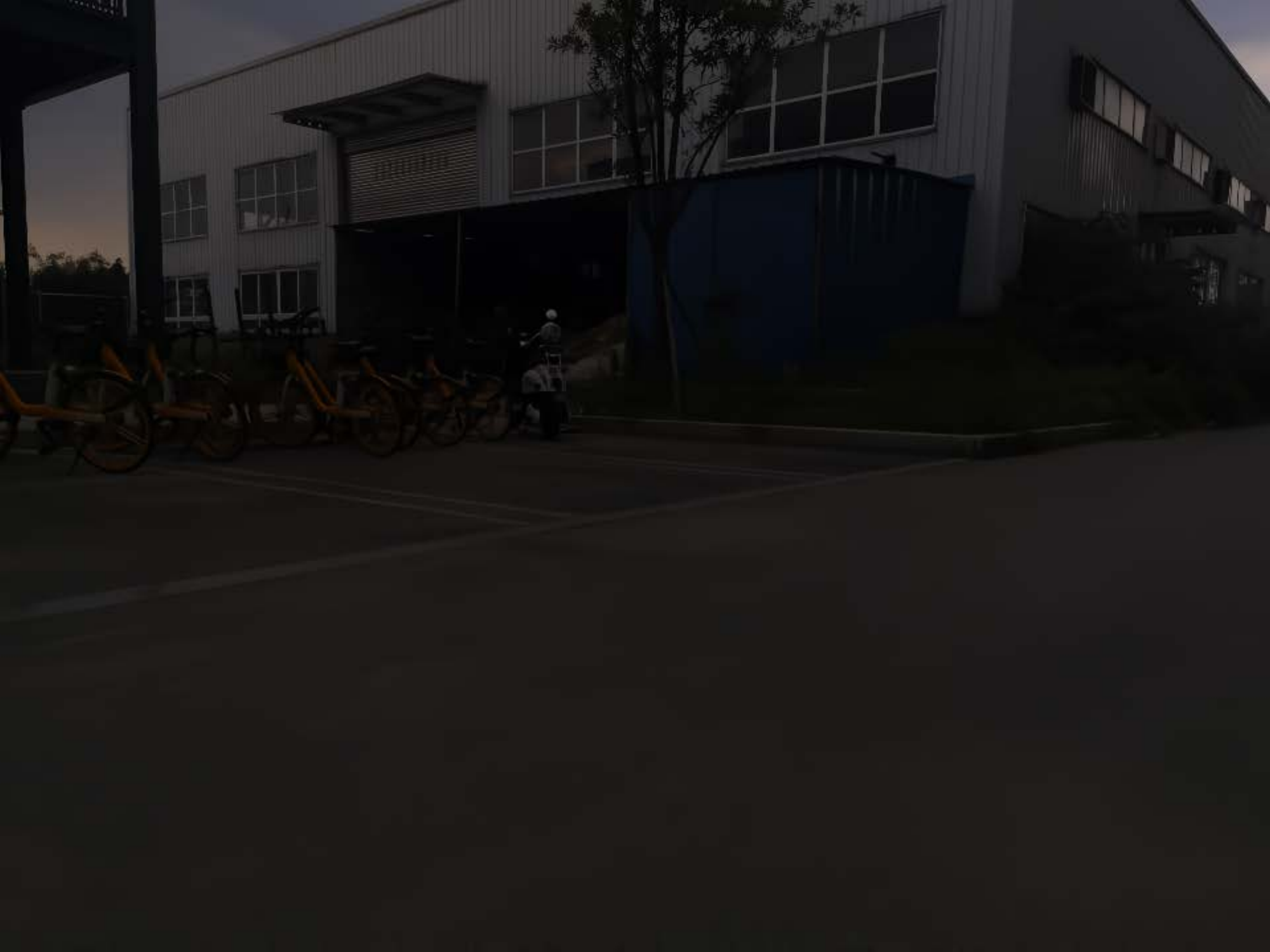}&	
			\includegraphics[width=0.16\linewidth]{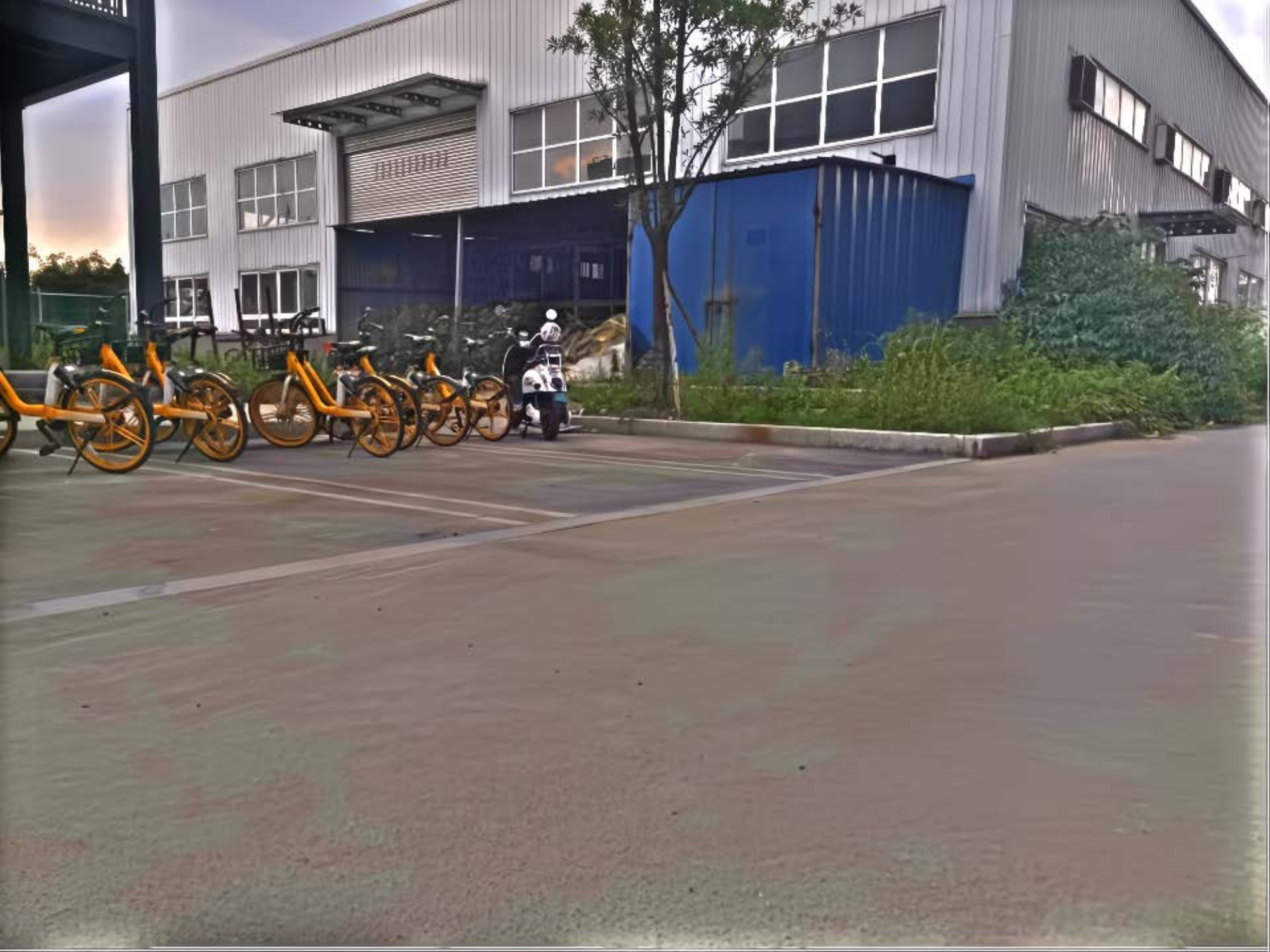}&	
			\includegraphics[width=0.16\linewidth]{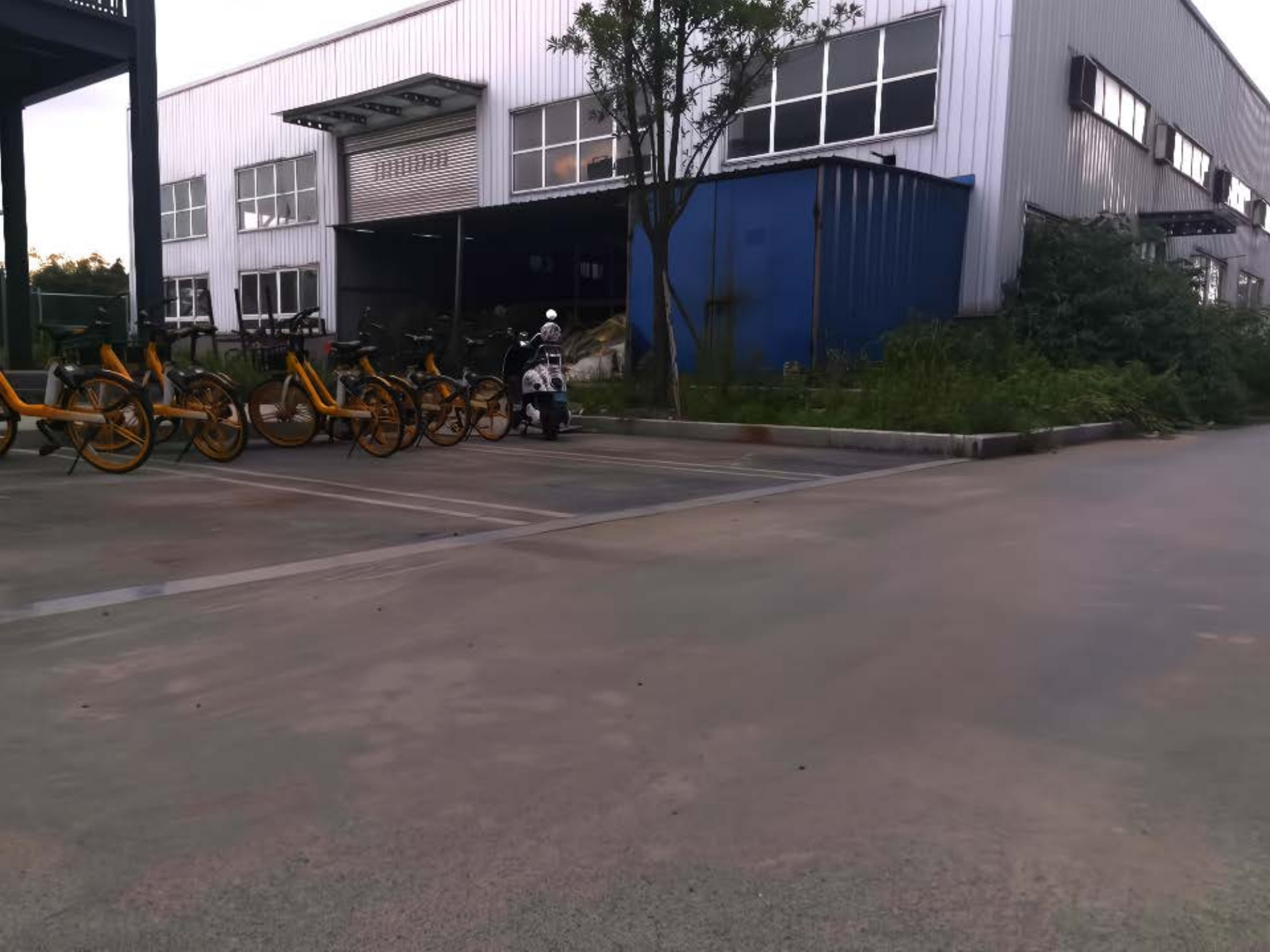}&	
			\includegraphics[width=0.16\linewidth]{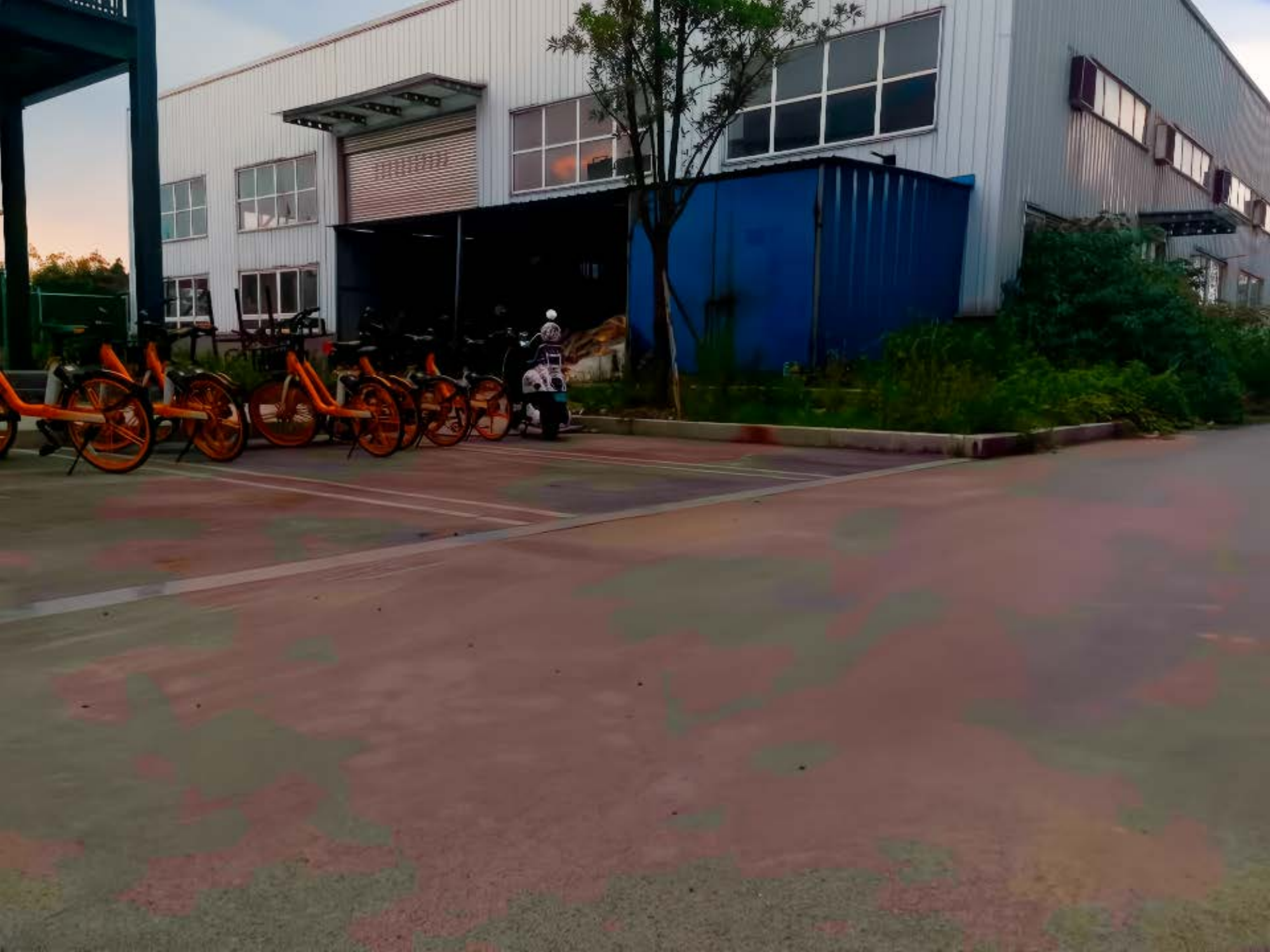}&	
			\includegraphics[width=0.16\linewidth]{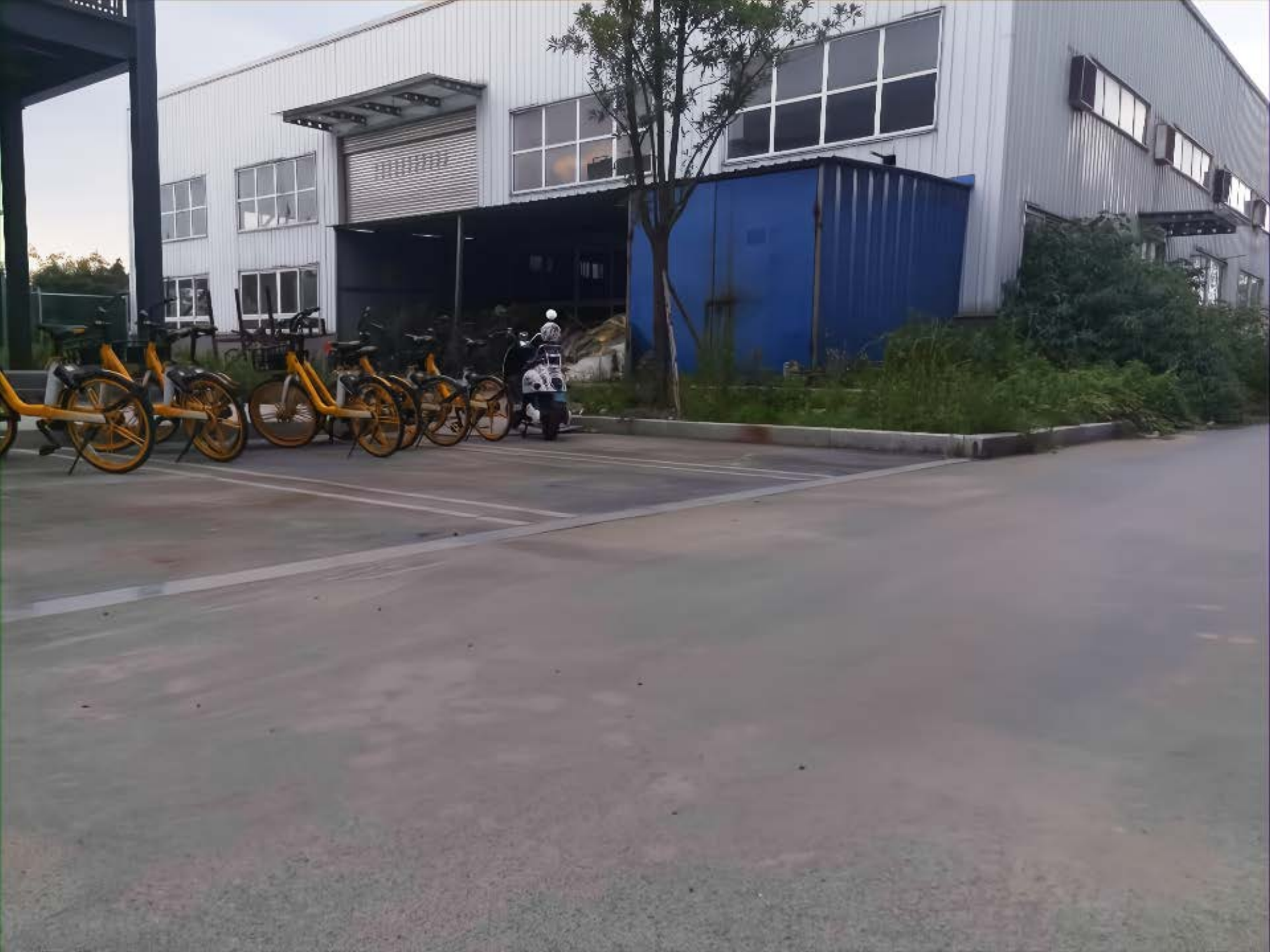}&	
			\includegraphics[width=0.16\linewidth]{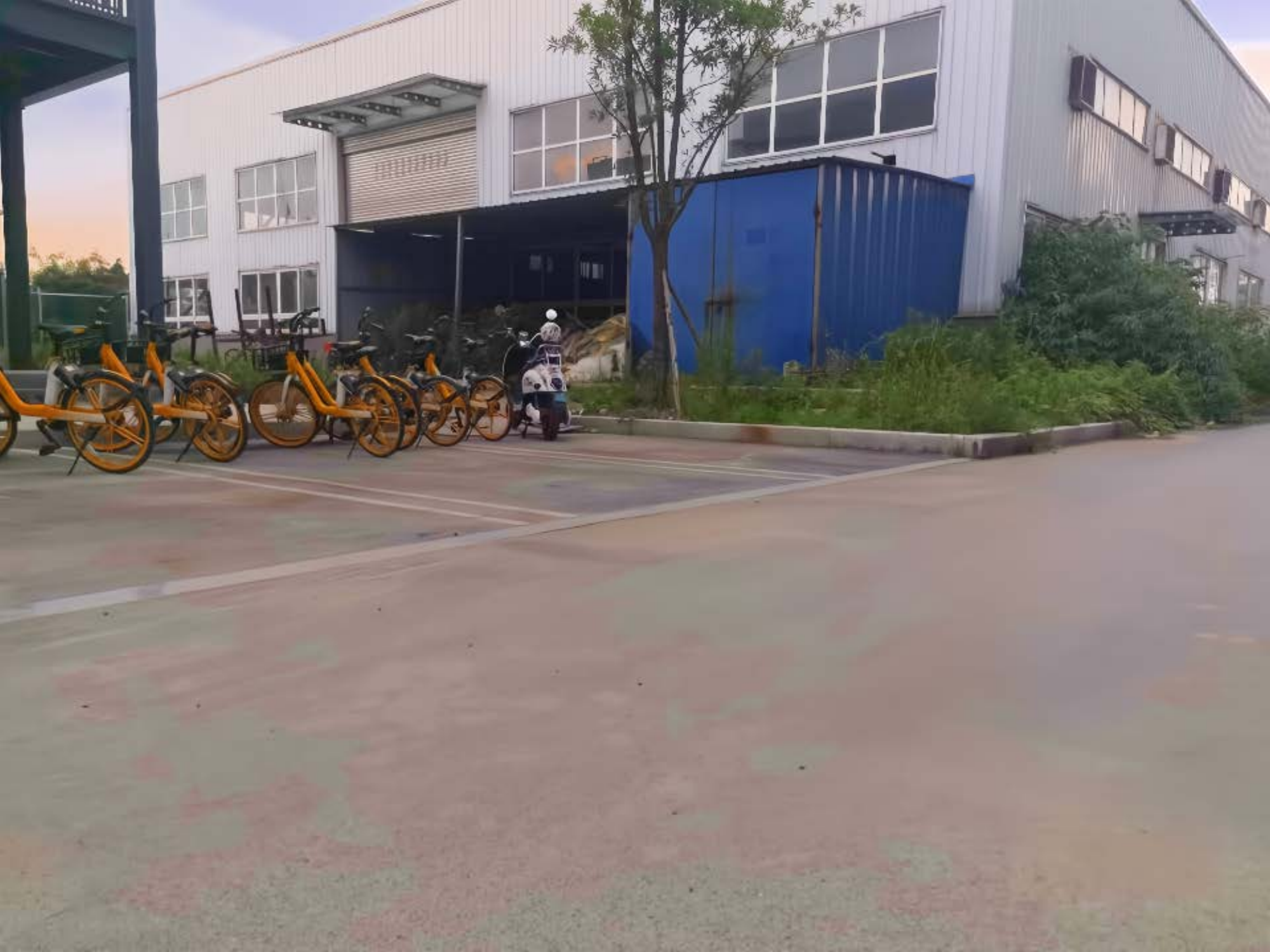}\\
			\vspace{-0.1cm}	
			\includegraphics[width=0.16\linewidth]{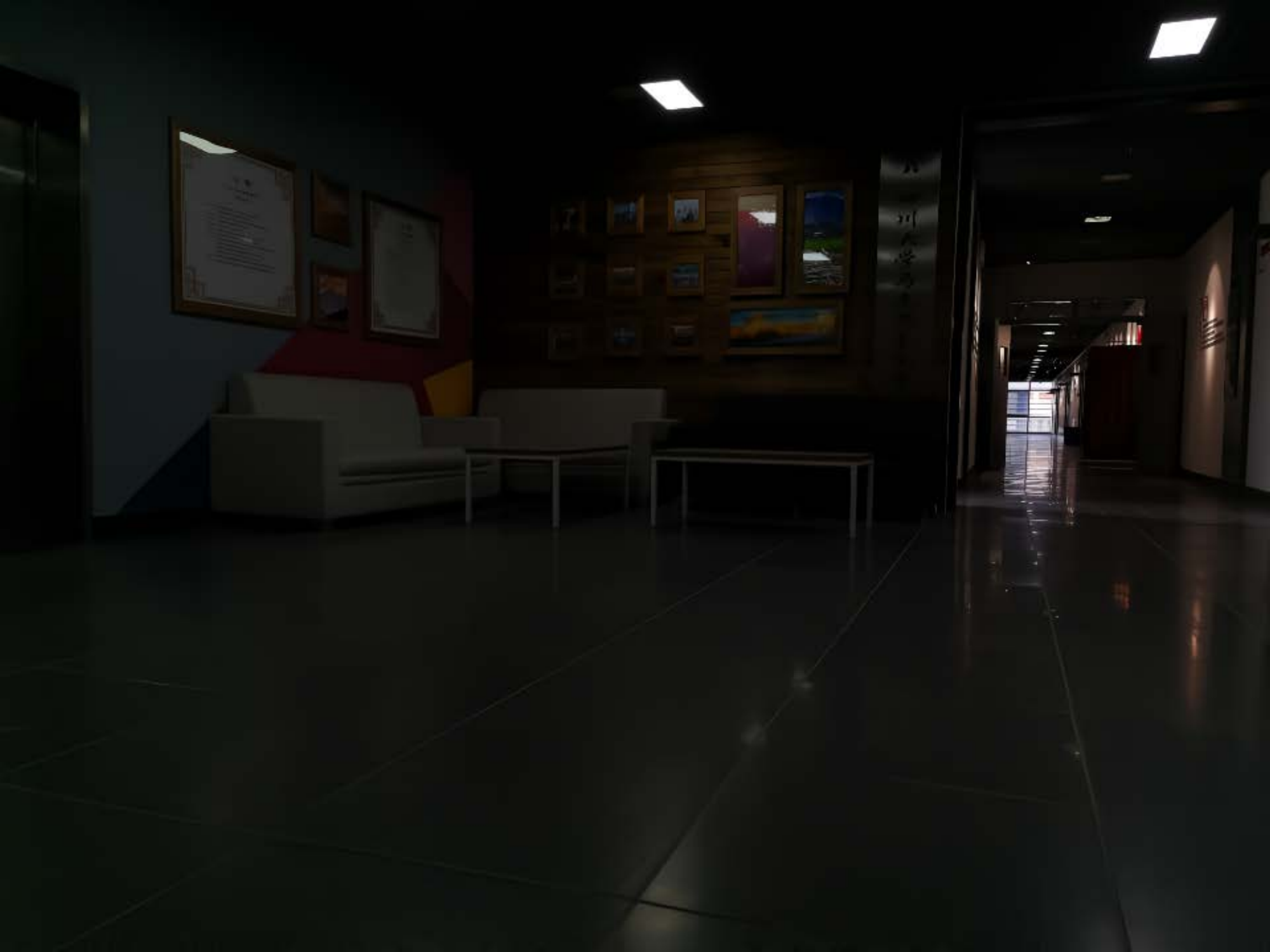}&	
			\includegraphics[width=0.16\linewidth]{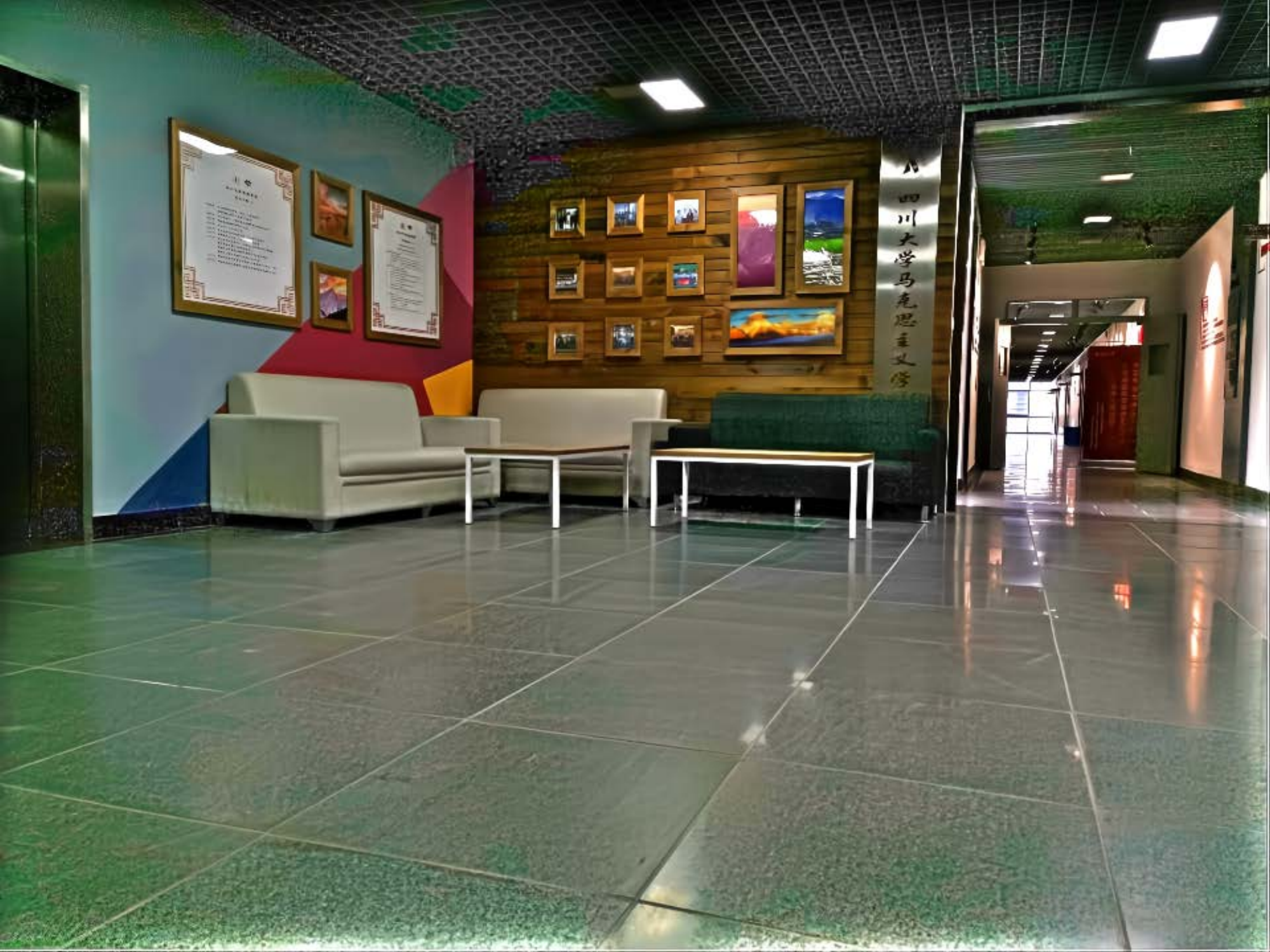}&	
			\includegraphics[width=0.16\linewidth]{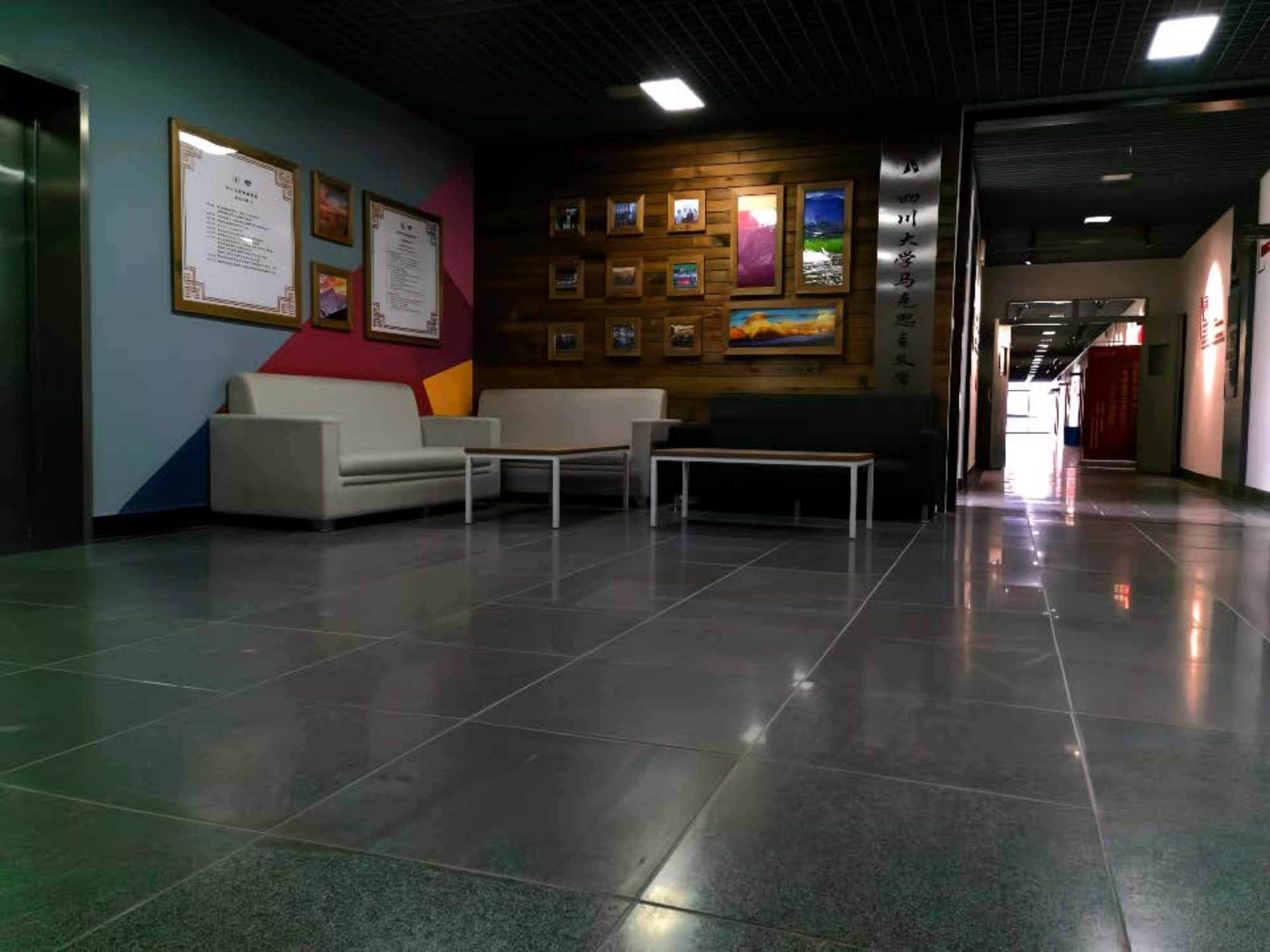}&	
			\includegraphics[width=0.16\linewidth]{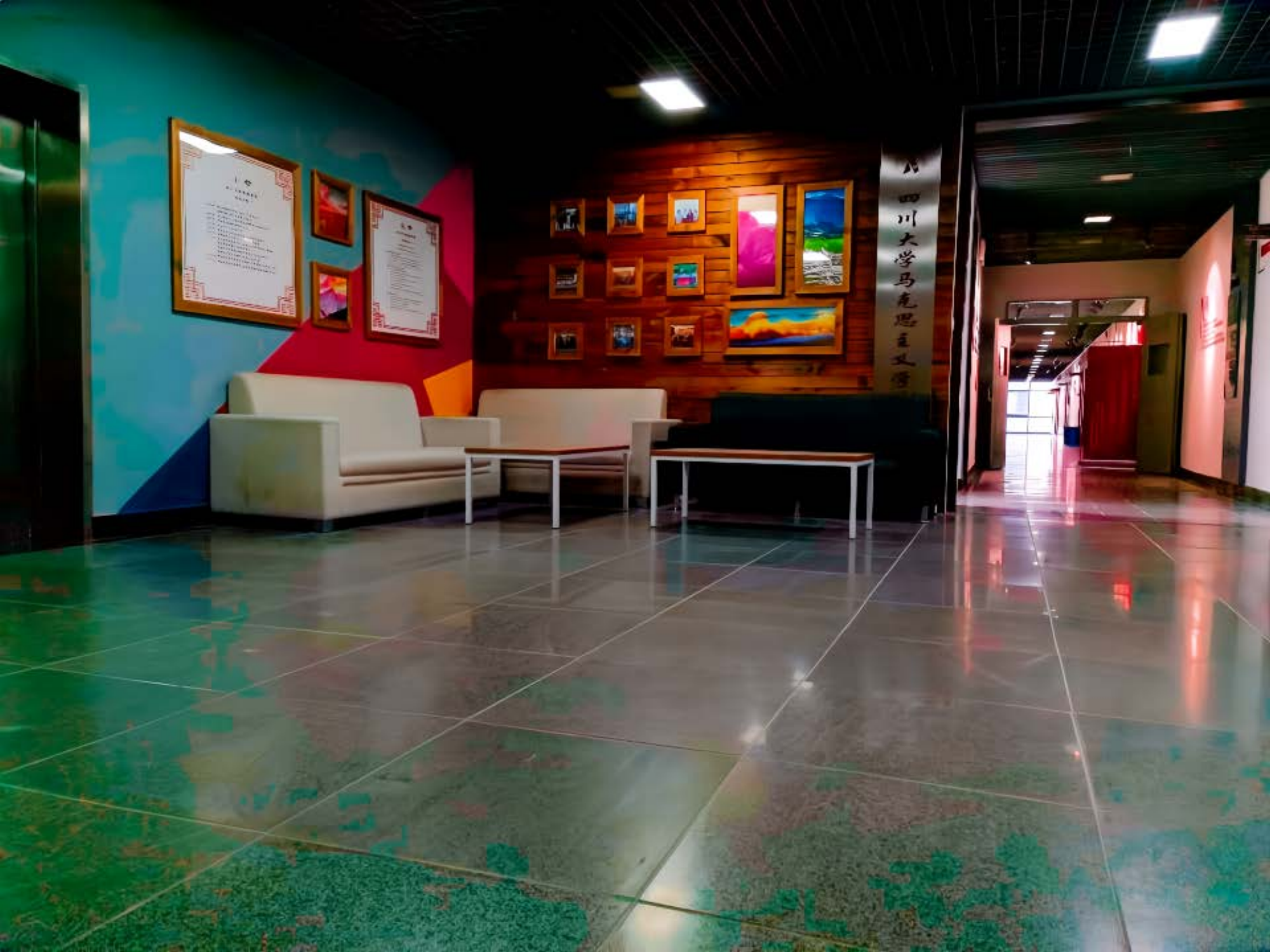}&	
			\includegraphics[width=0.16\linewidth]{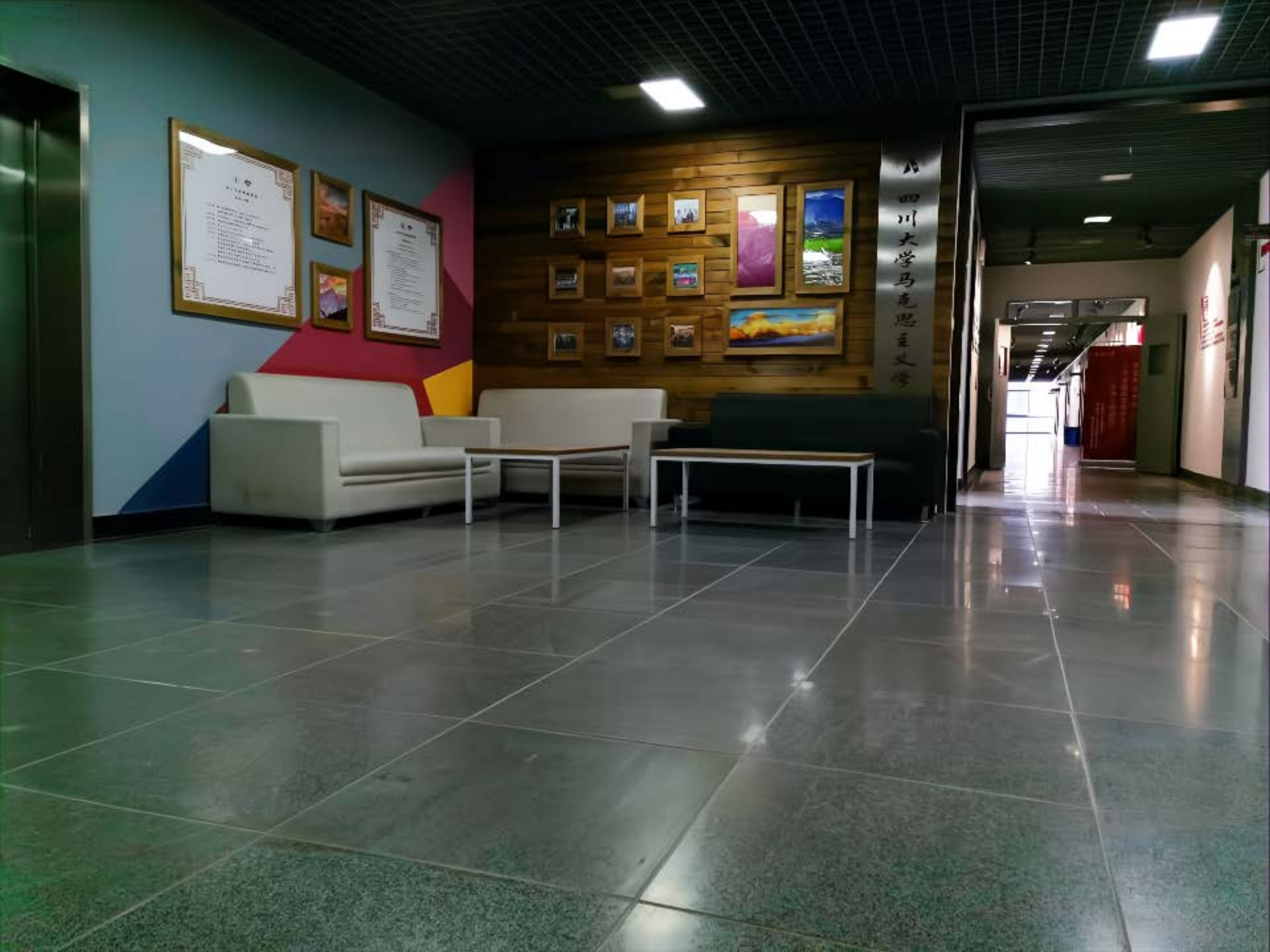}&	
			\includegraphics[width=0.16\linewidth]{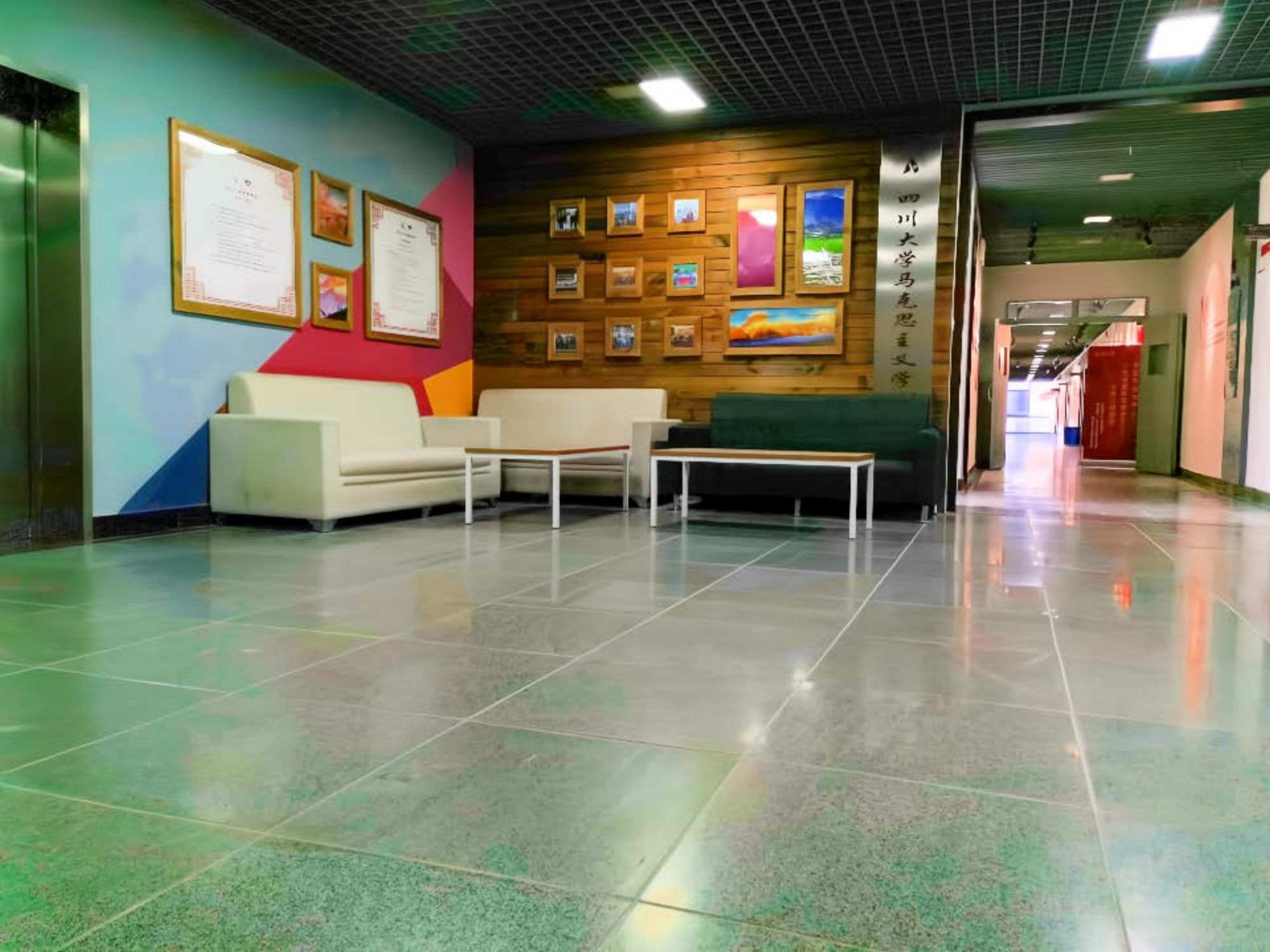}\\
			\vspace{-0.1cm}	
			\includegraphics[width=0.16\linewidth]{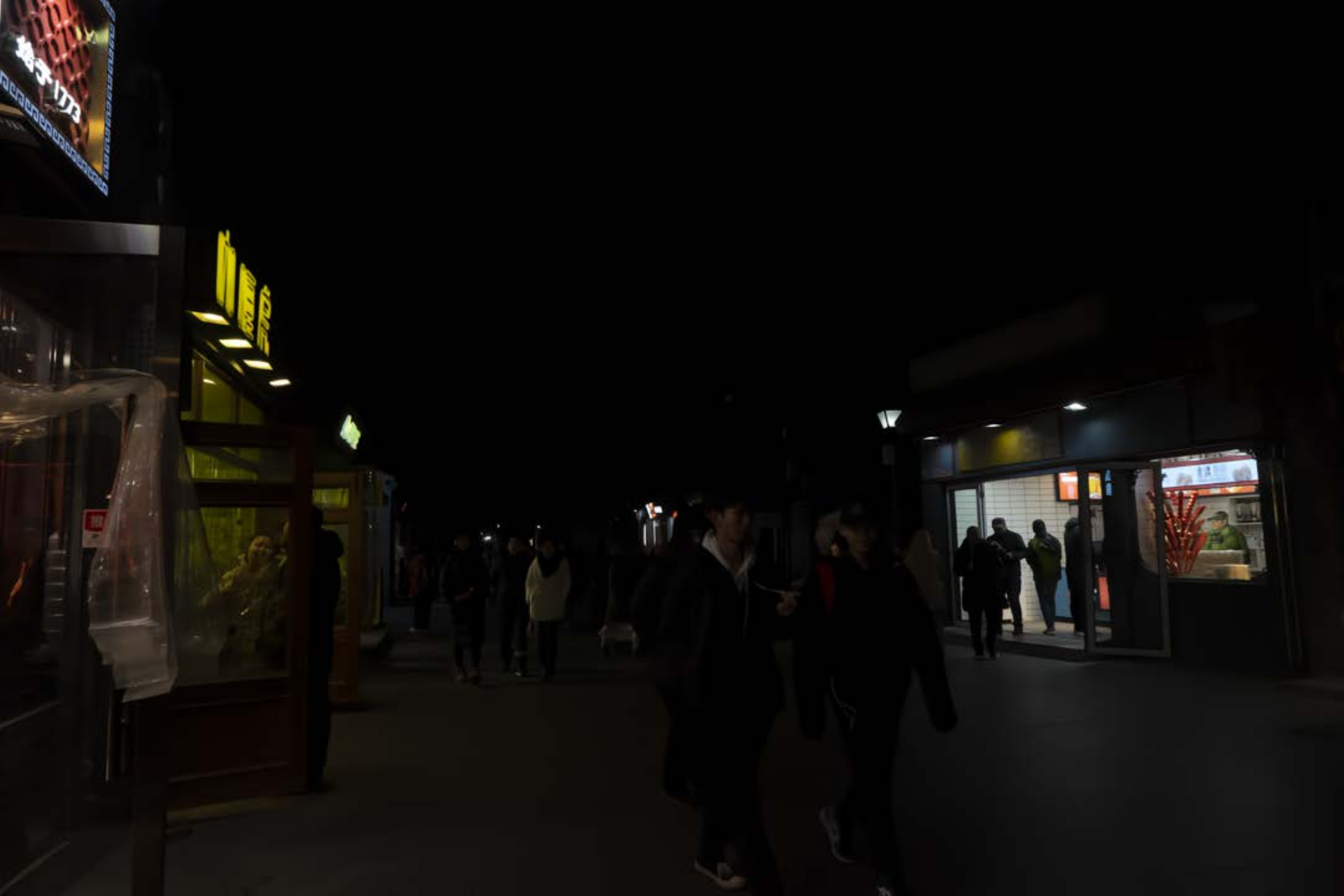}&	
			\includegraphics[width=0.16\linewidth]{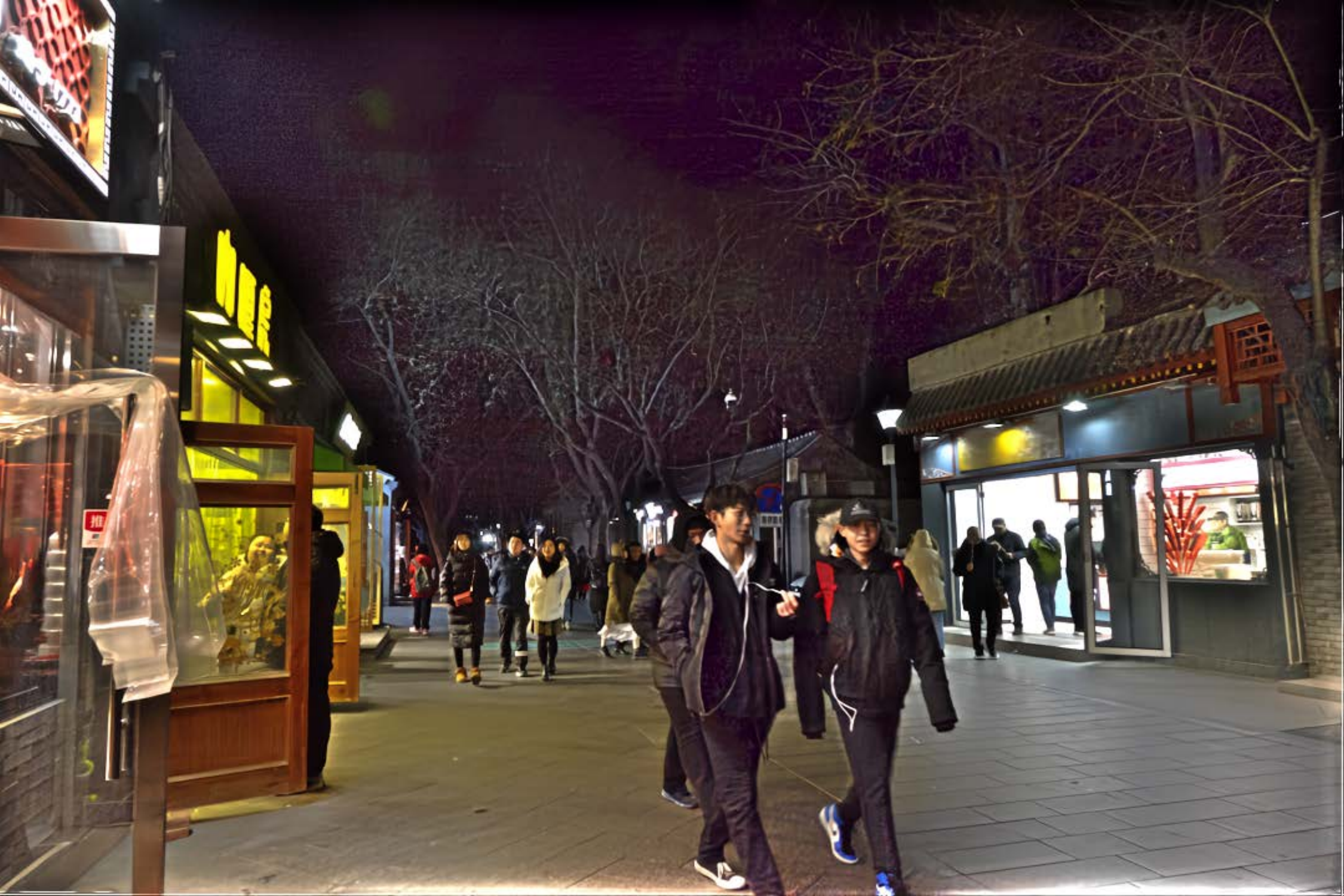}&	
			\includegraphics[width=0.16\linewidth]{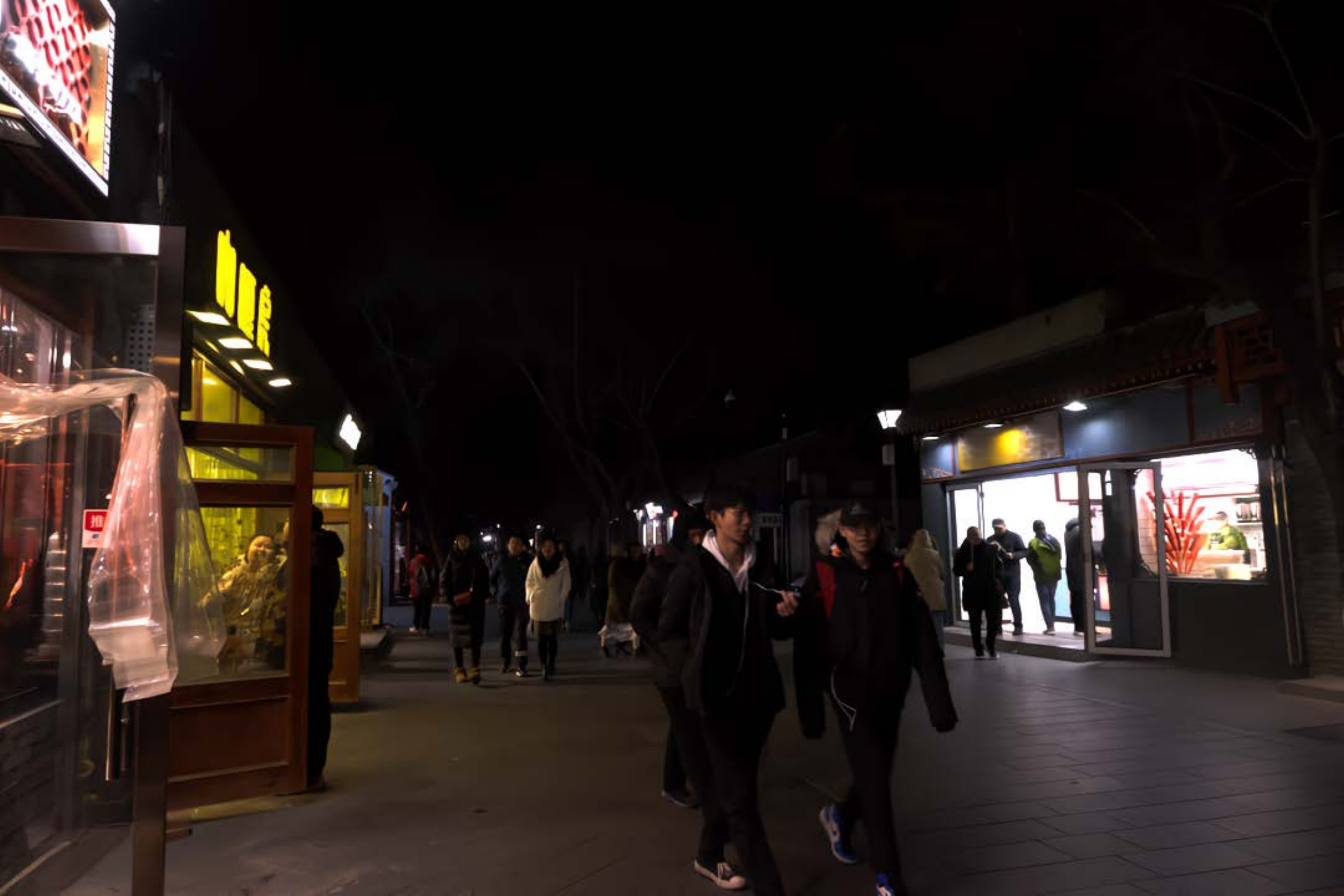}&	
			\includegraphics[width=0.16\linewidth]{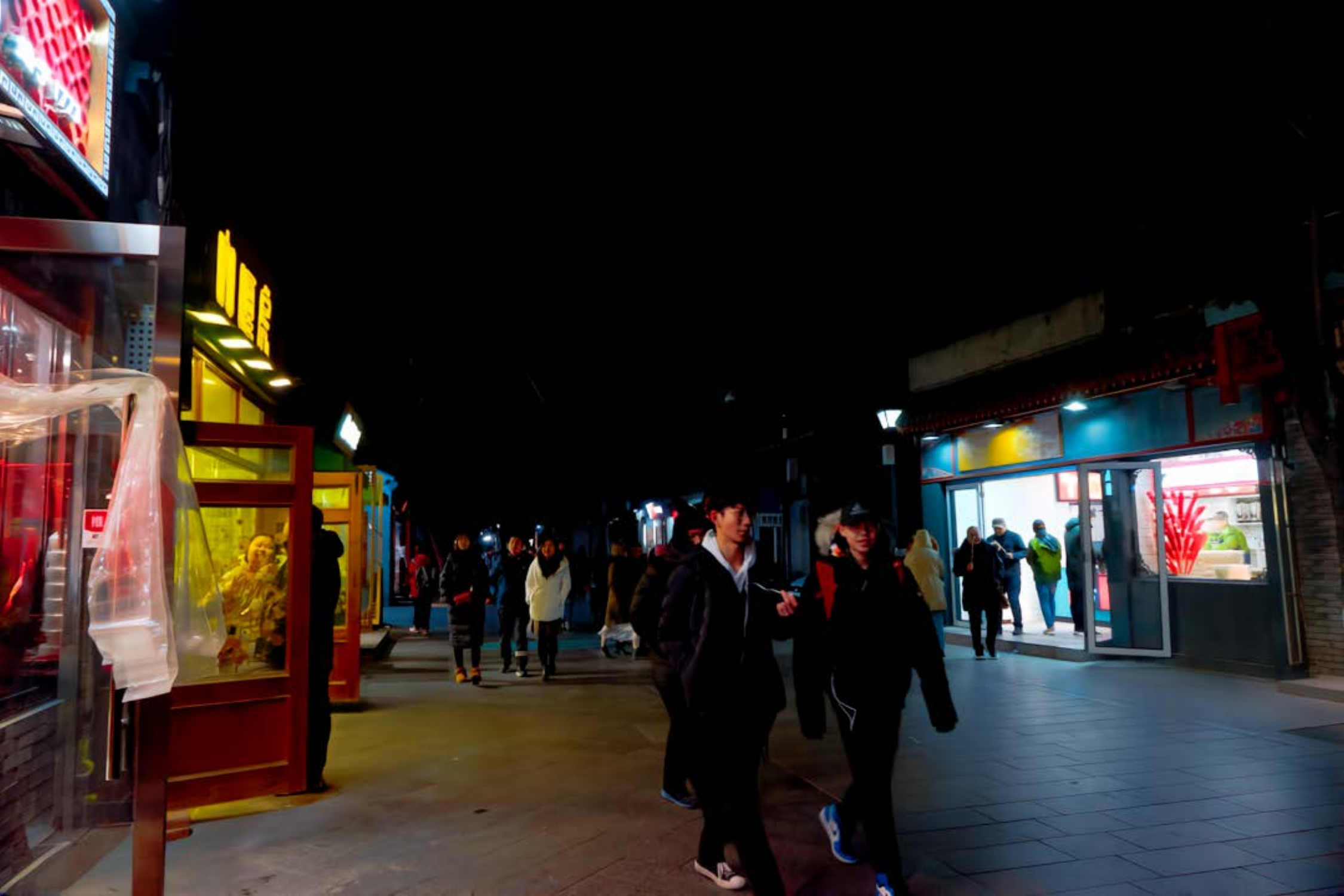}&	
			\includegraphics[width=0.16\linewidth]{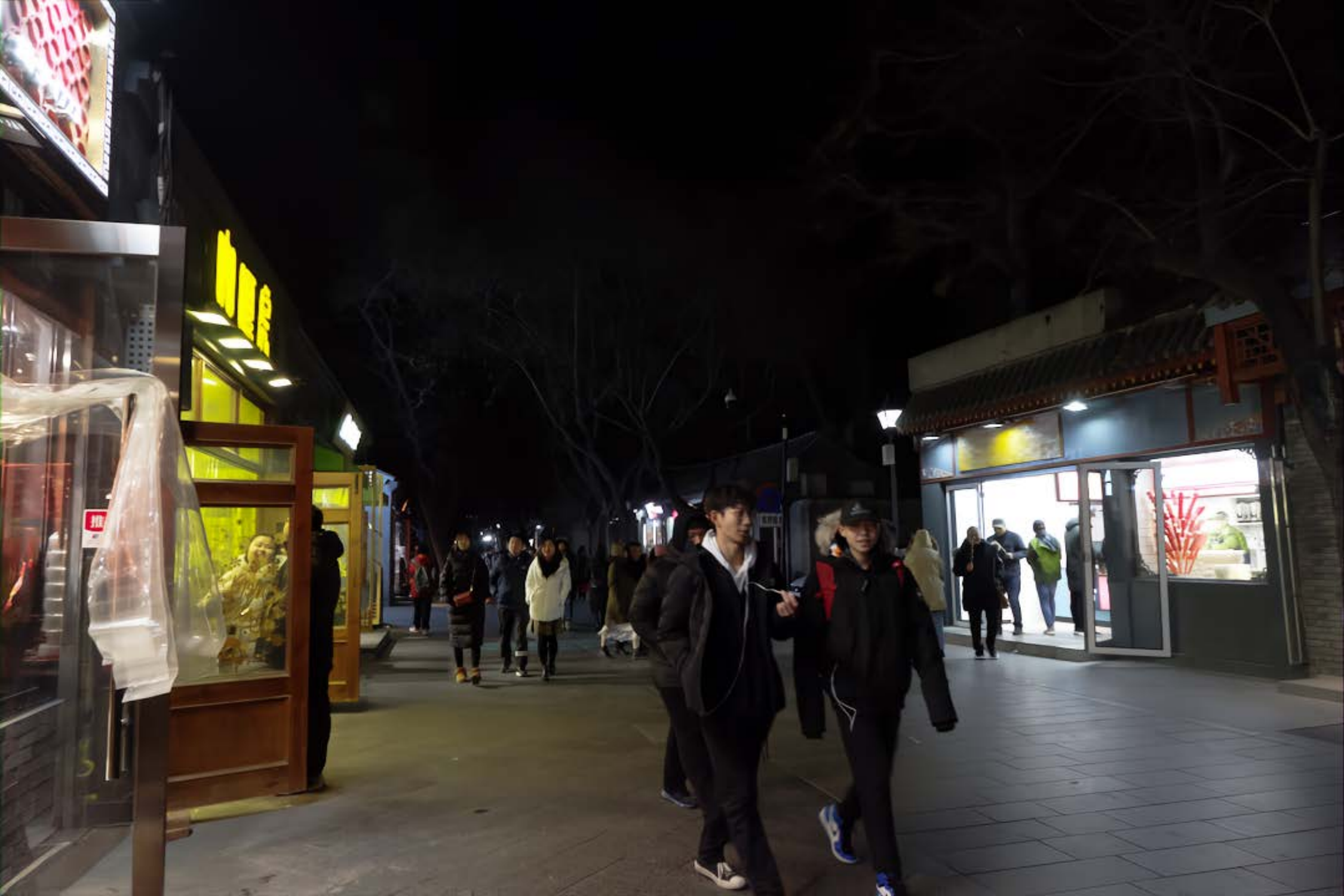}&	
			\includegraphics[width=0.16\linewidth]{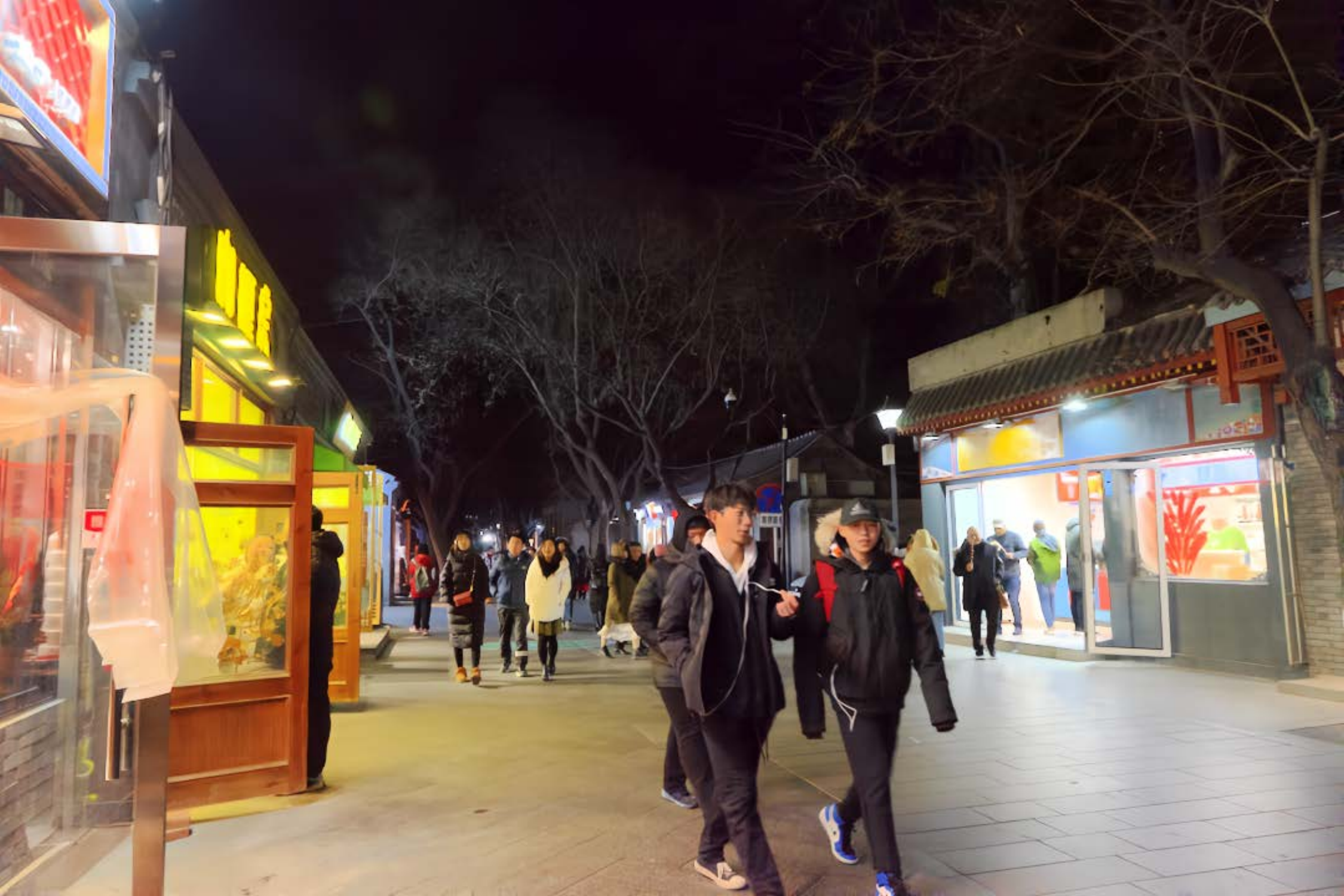}\\
			\vspace{-0.2cm}
			\footnotesize Input&\footnotesize LIME~\cite{guo2017lime}&\footnotesize RUAS~\cite{liu2021retinex}&\footnotesize FEC~\cite{huang2022eccv}&\footnotesize SCI~\cite{ma2022toward}&\footnotesize PEC\\
		\end{tabular}
		\caption{Visual comparison of under-exposure scenarios with noise. Except for the input, all correction results were processed by denoising (SCUNet) for better visualization.}
		\label{fig: Scenes_under1}
	\end{figure*}
	
	\begin{figure*}[t]
		\centering
		\begin{tabular}{c@{\extracolsep{0.2em}}c@{\extracolsep{0.2em}}c@{\extracolsep{0.2em}}c@{\extracolsep{0.2em}}c@{\extracolsep{0.2em}}c}	
			\vspace{-0.1cm}	
			\includegraphics[width=0.16\linewidth]{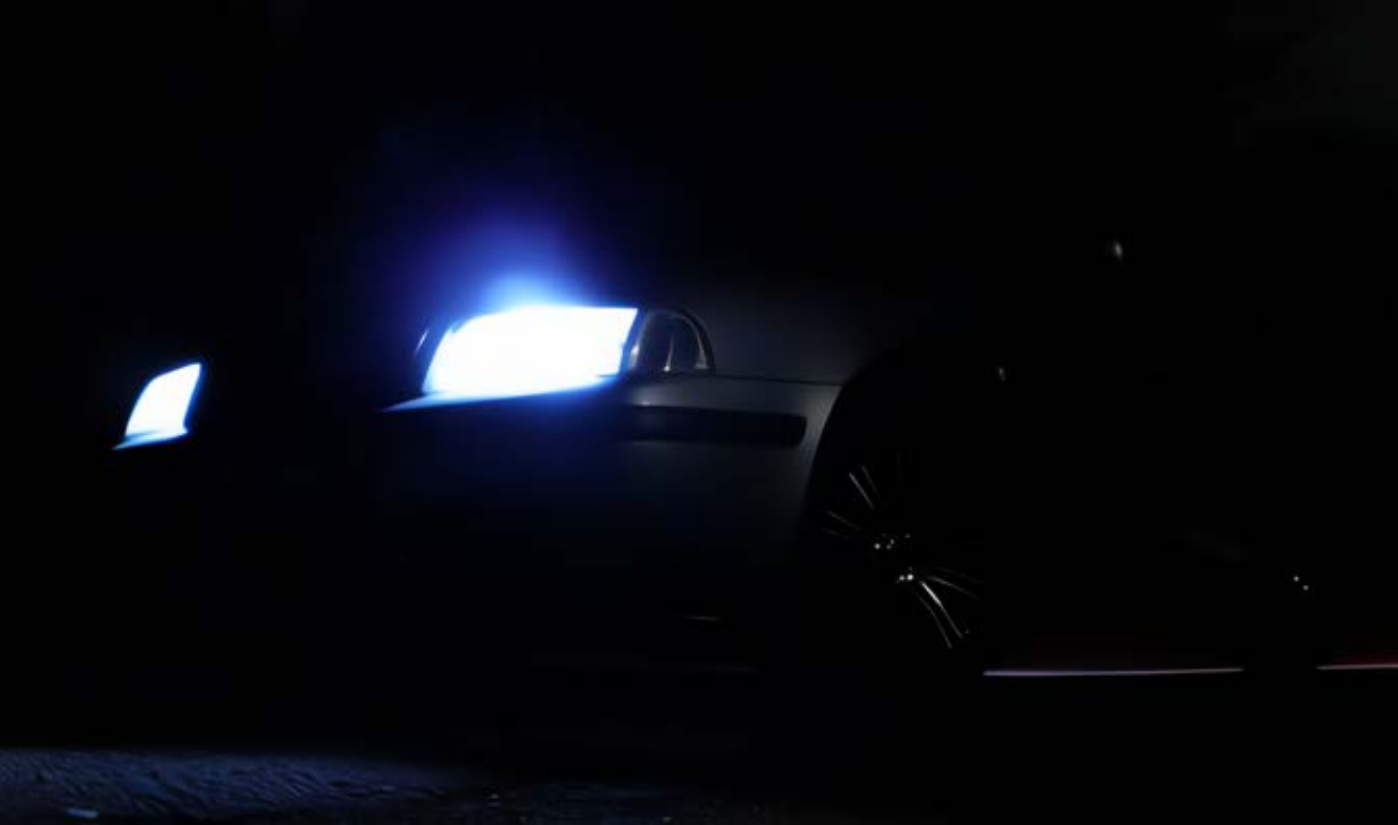}&	
			\includegraphics[width=0.16\linewidth]{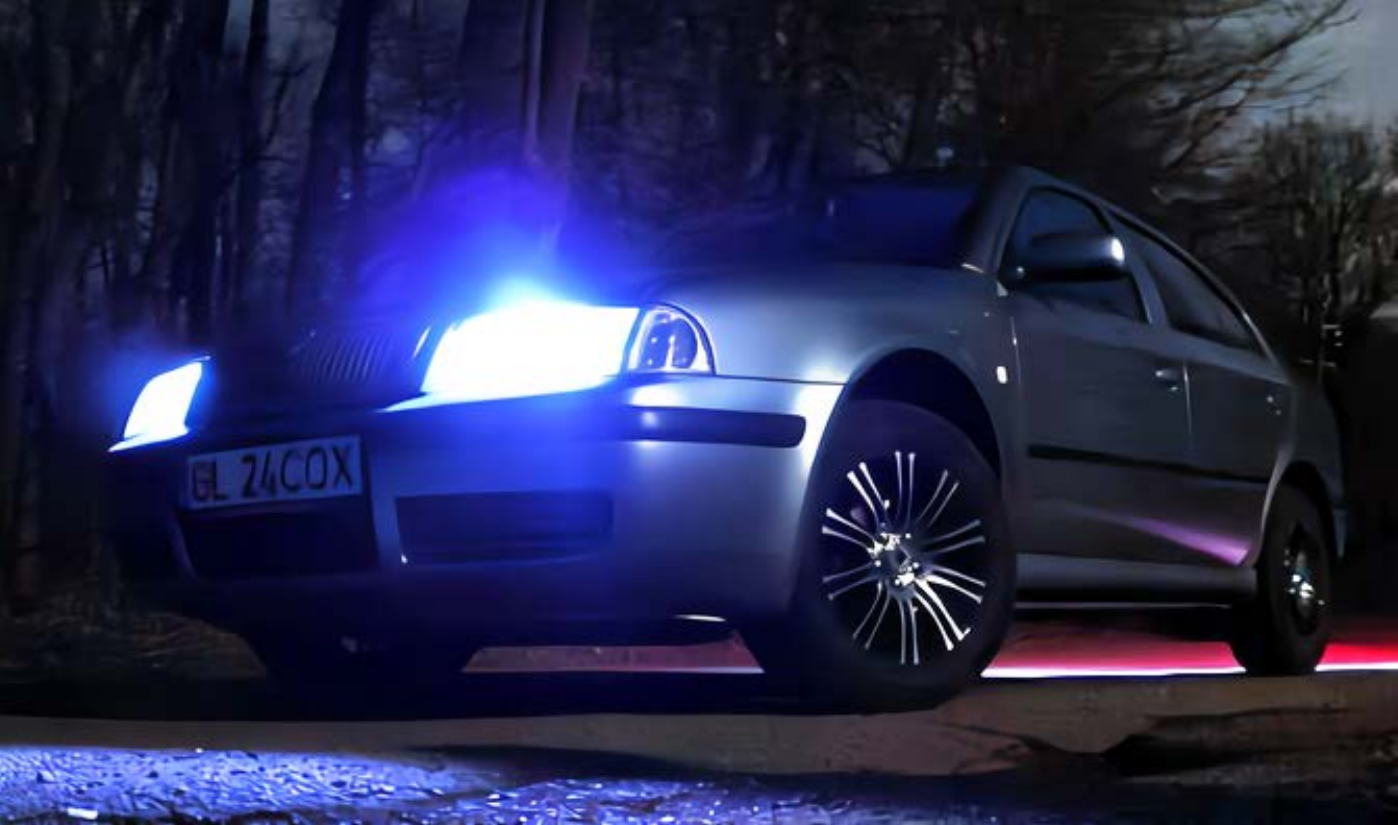}&	
			\includegraphics[width=0.16\linewidth]{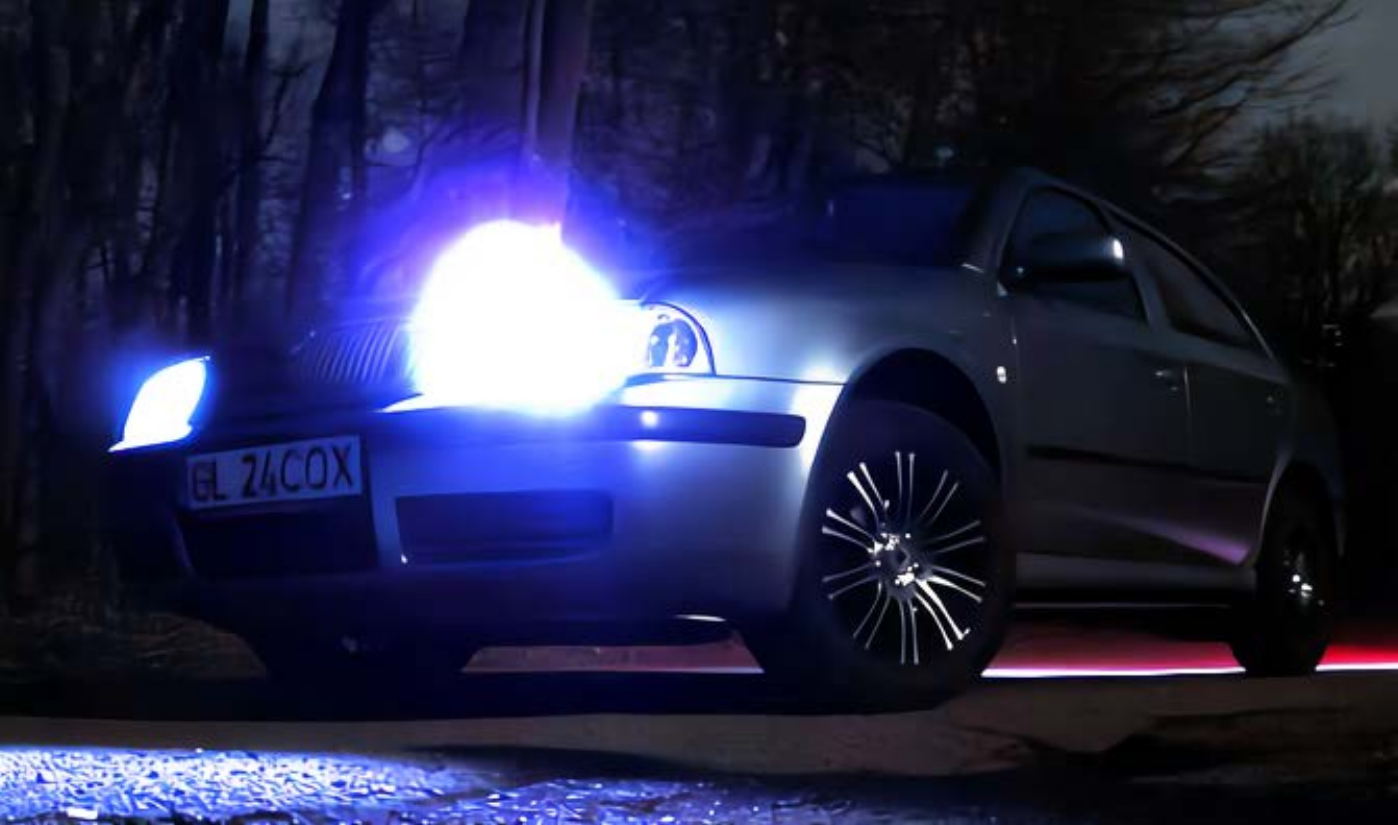}&	
			\includegraphics[width=0.16\linewidth]{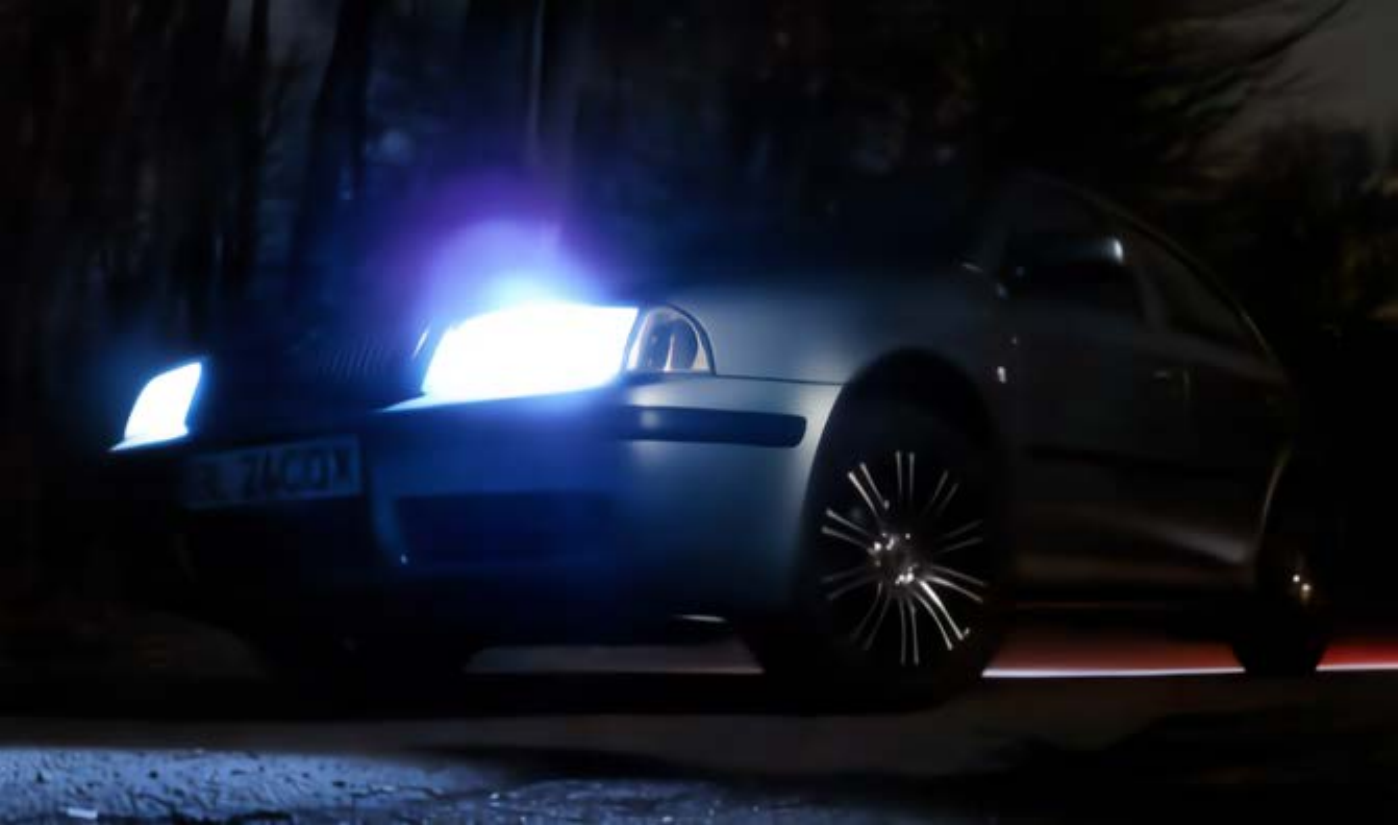}&	
			\includegraphics[width=0.16\linewidth]{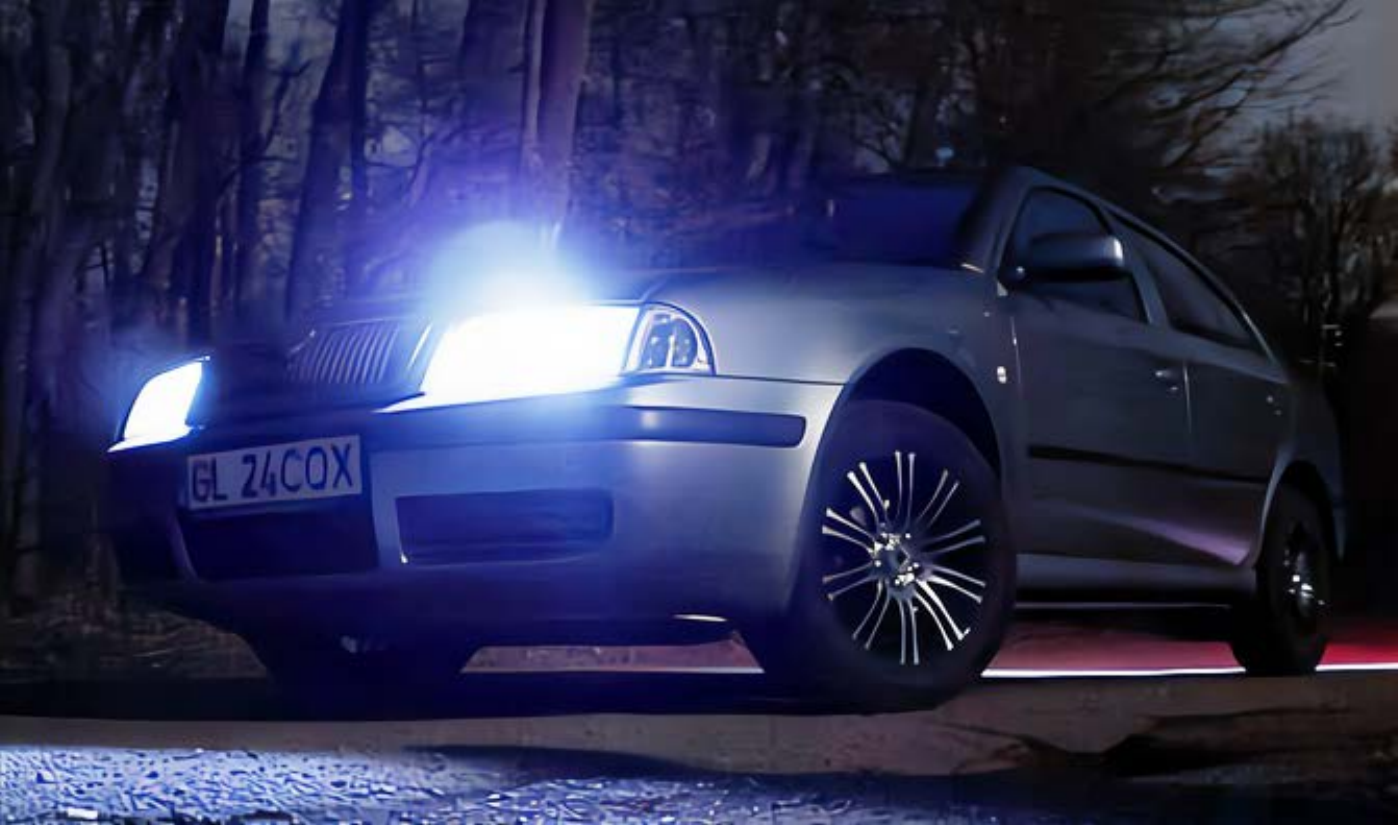}&	
			\includegraphics[width=0.16\linewidth]{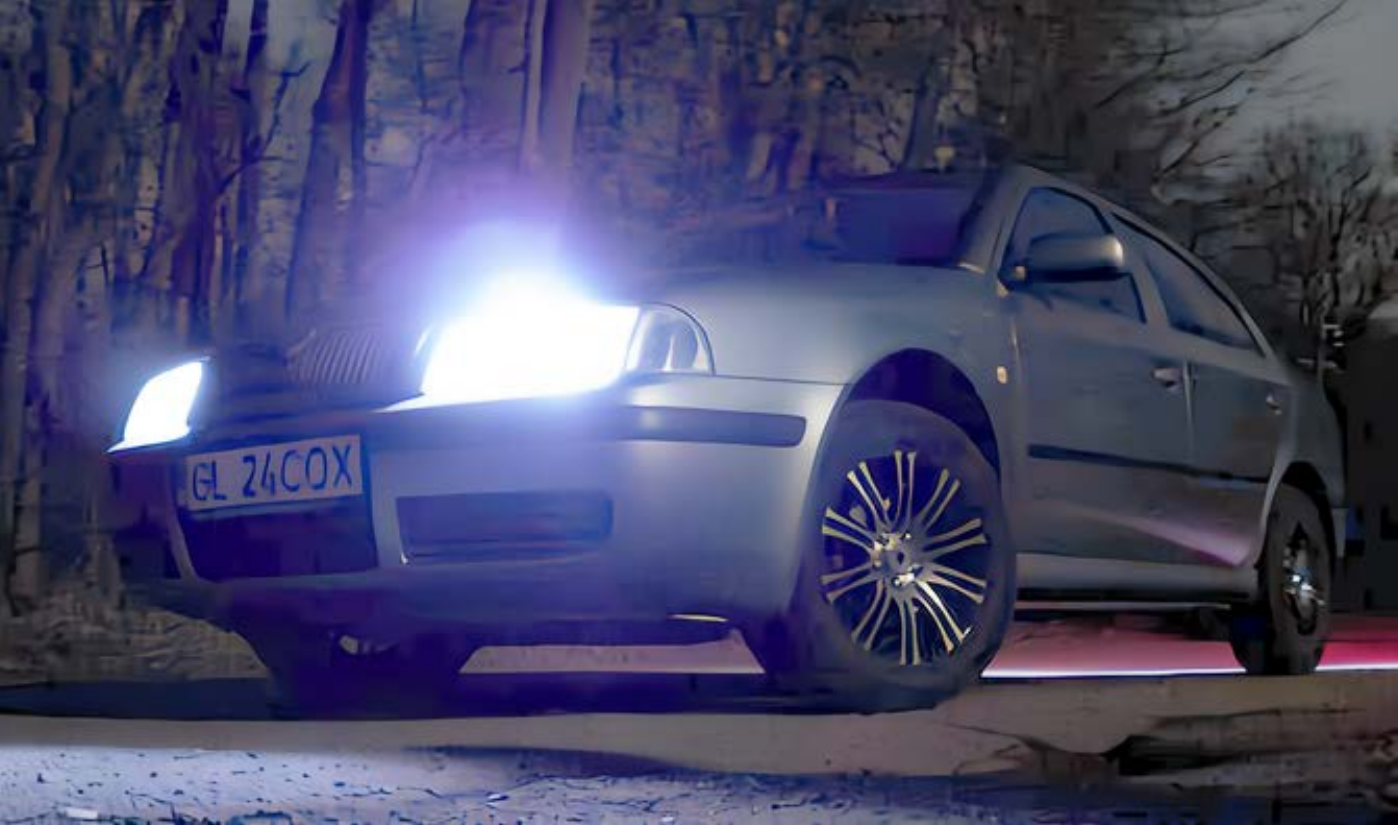}\\
			\vspace{-0.1cm}	
			\includegraphics[width=0.16\linewidth]{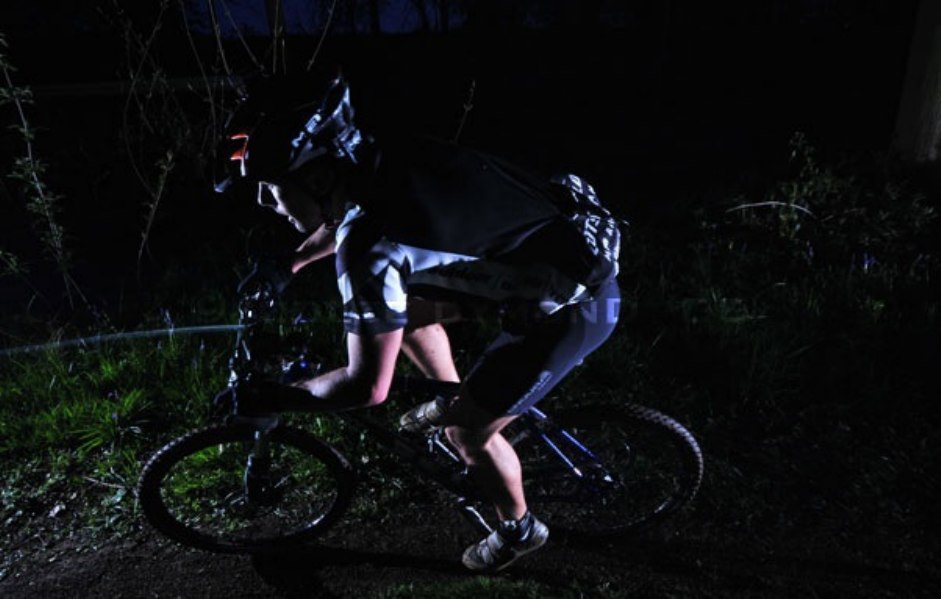}&	
			\includegraphics[width=0.16\linewidth]{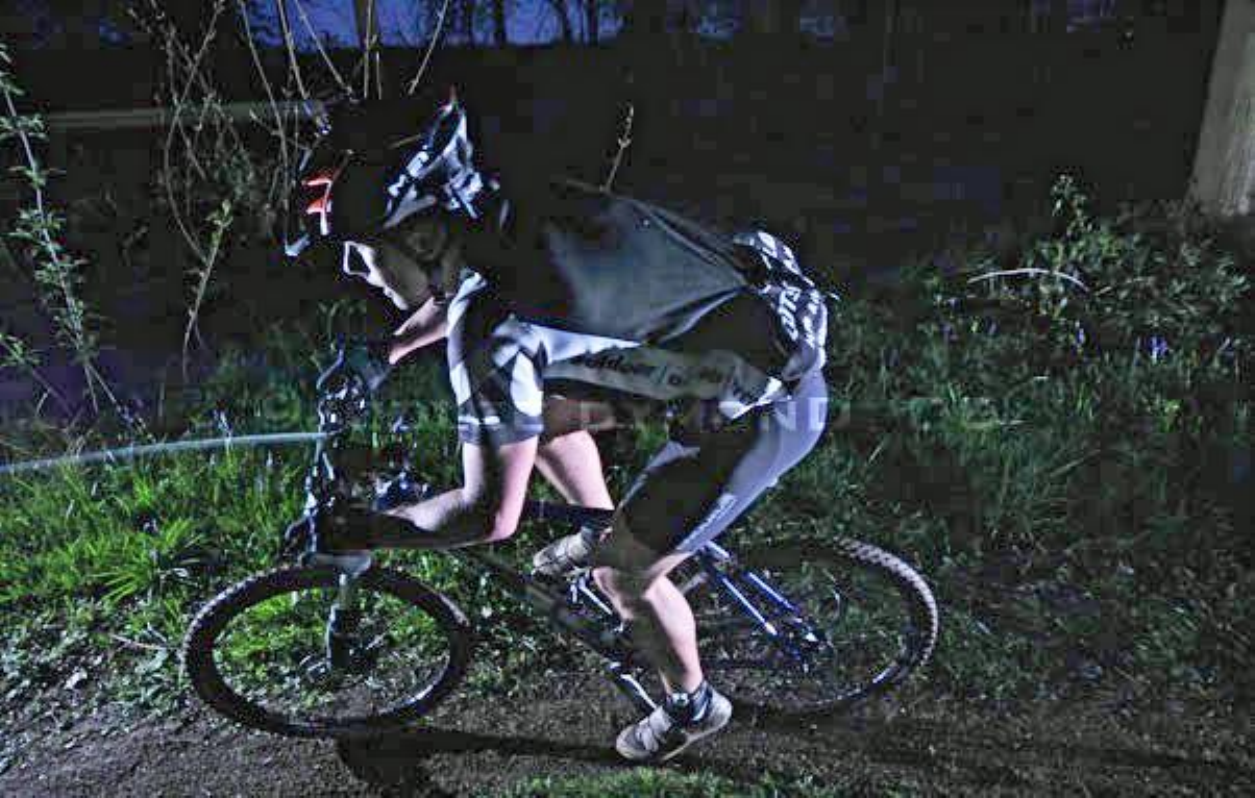}&	
			\includegraphics[width=0.16\linewidth]{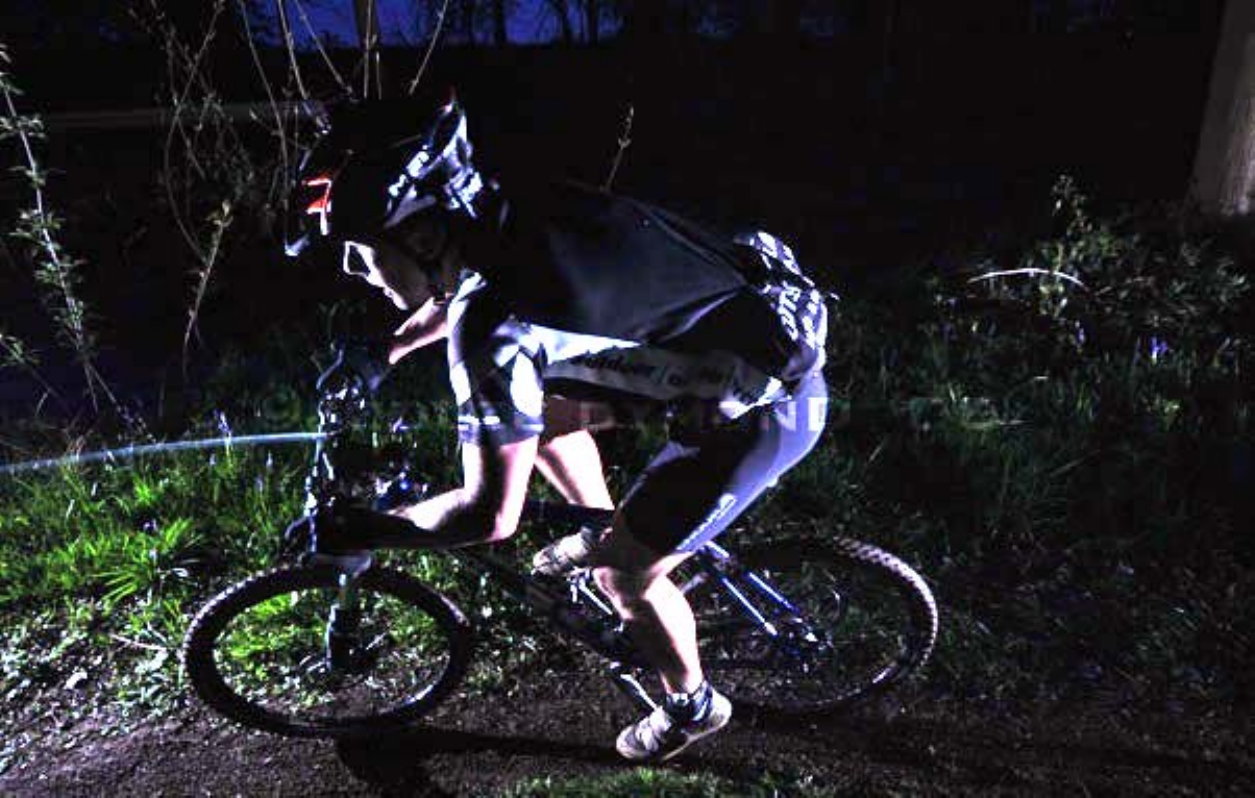}&	
			\includegraphics[width=0.16\linewidth]{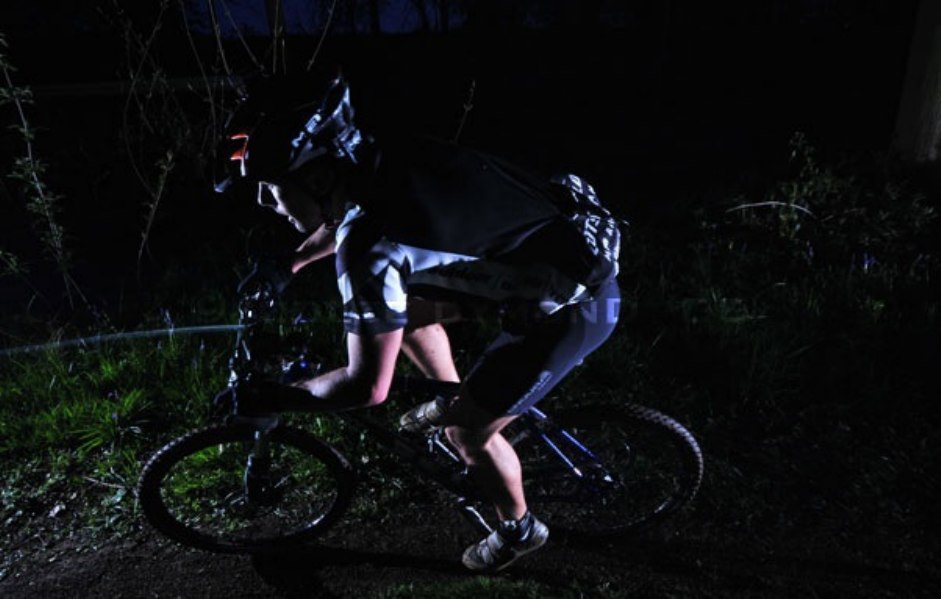}&	
			\includegraphics[width=0.16\linewidth]{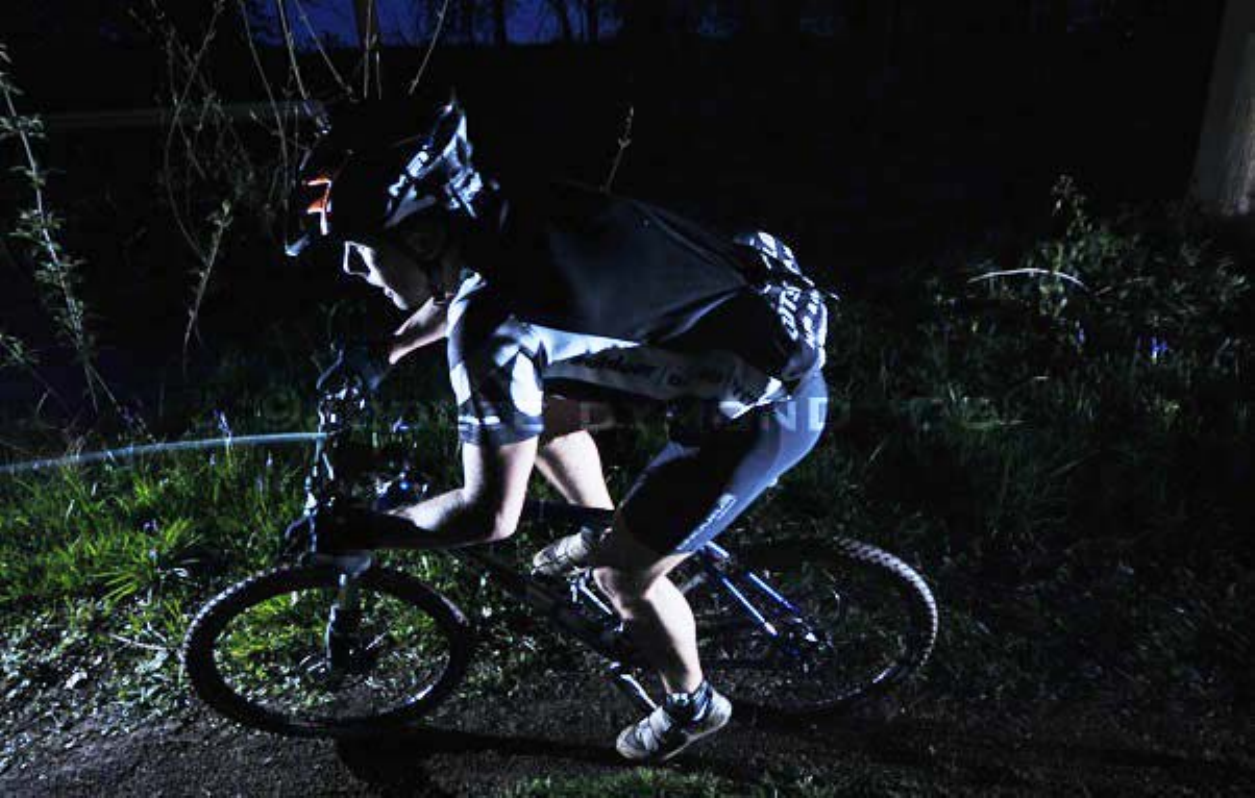}&	
			\includegraphics[width=0.16\linewidth]{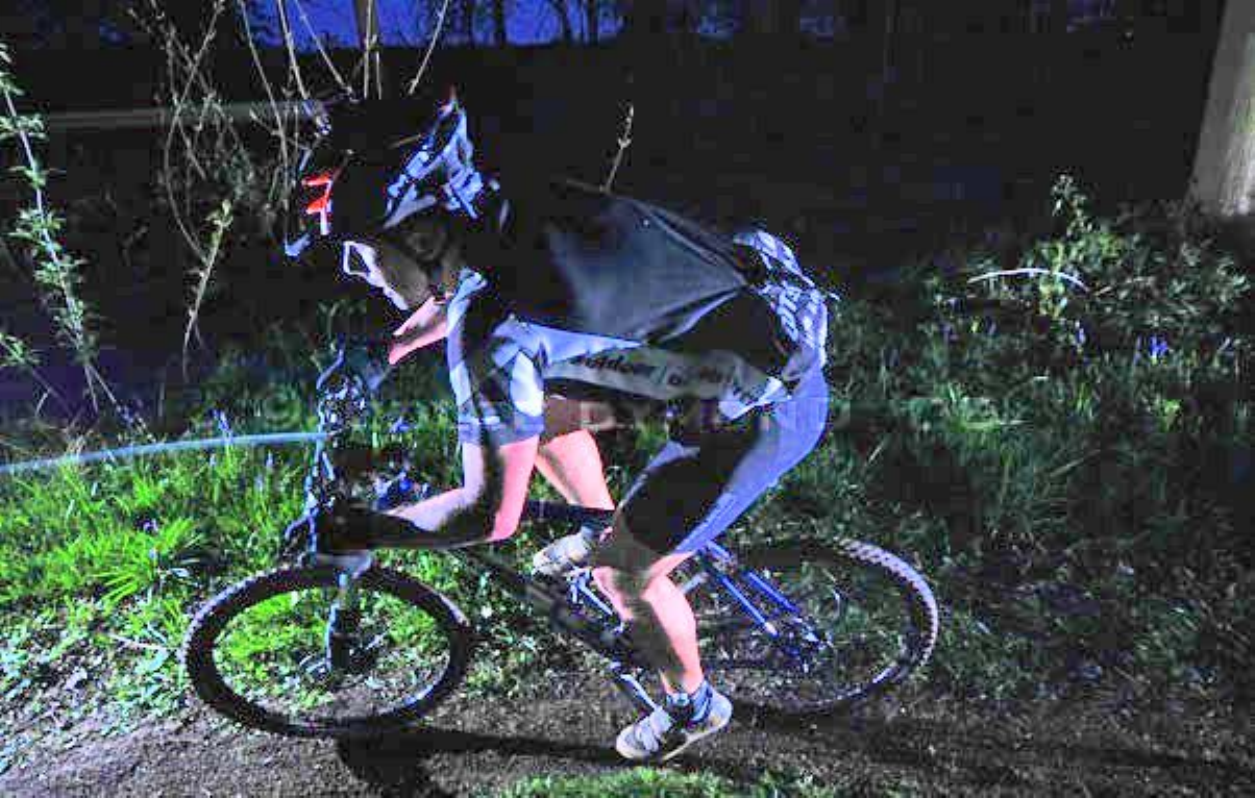}\\
			\vspace{-0.1cm}	
			\includegraphics[width=0.16\linewidth]{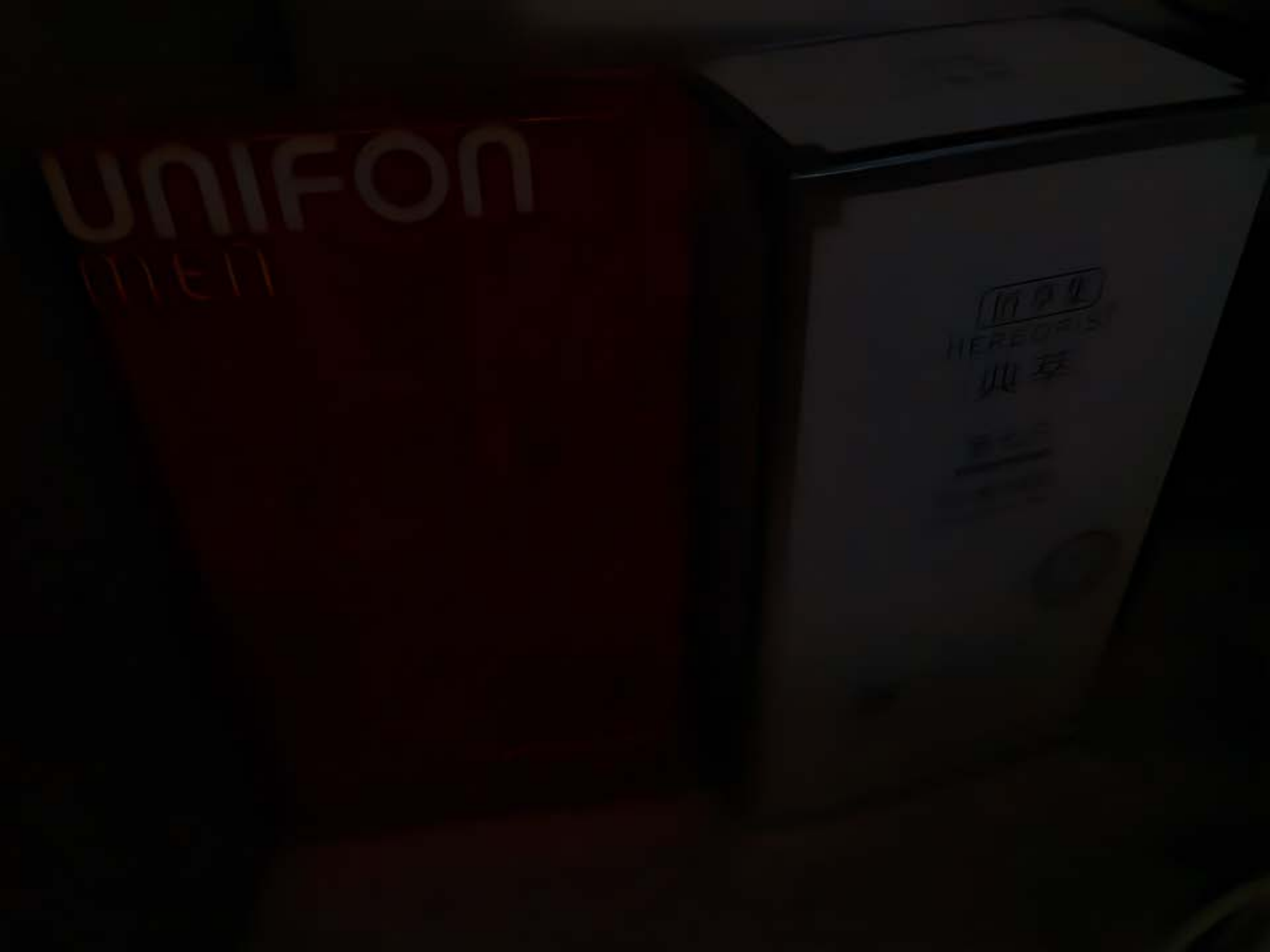}&	
			\includegraphics[width=0.16\linewidth]{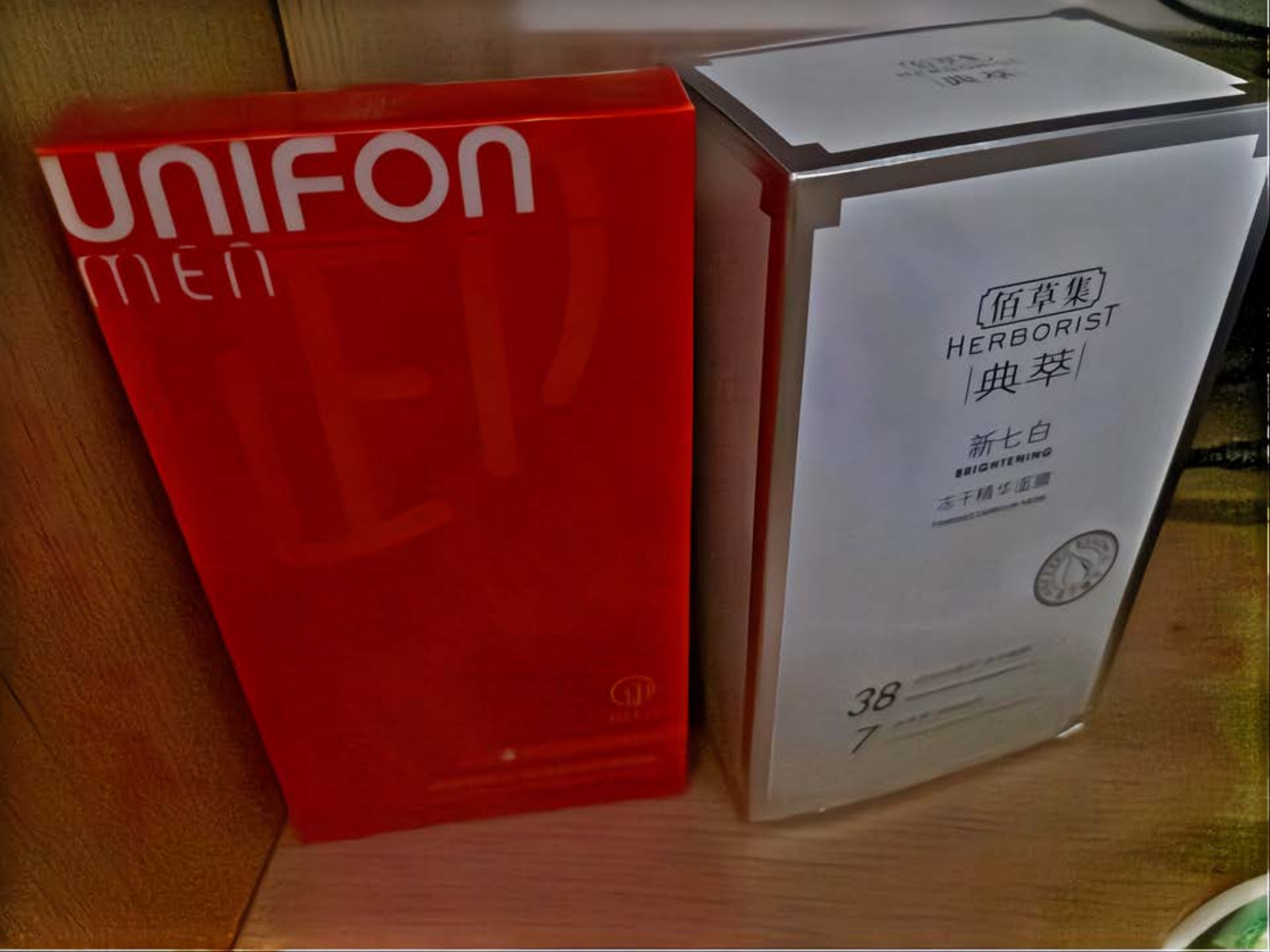}&	
			\includegraphics[width=0.16\linewidth]{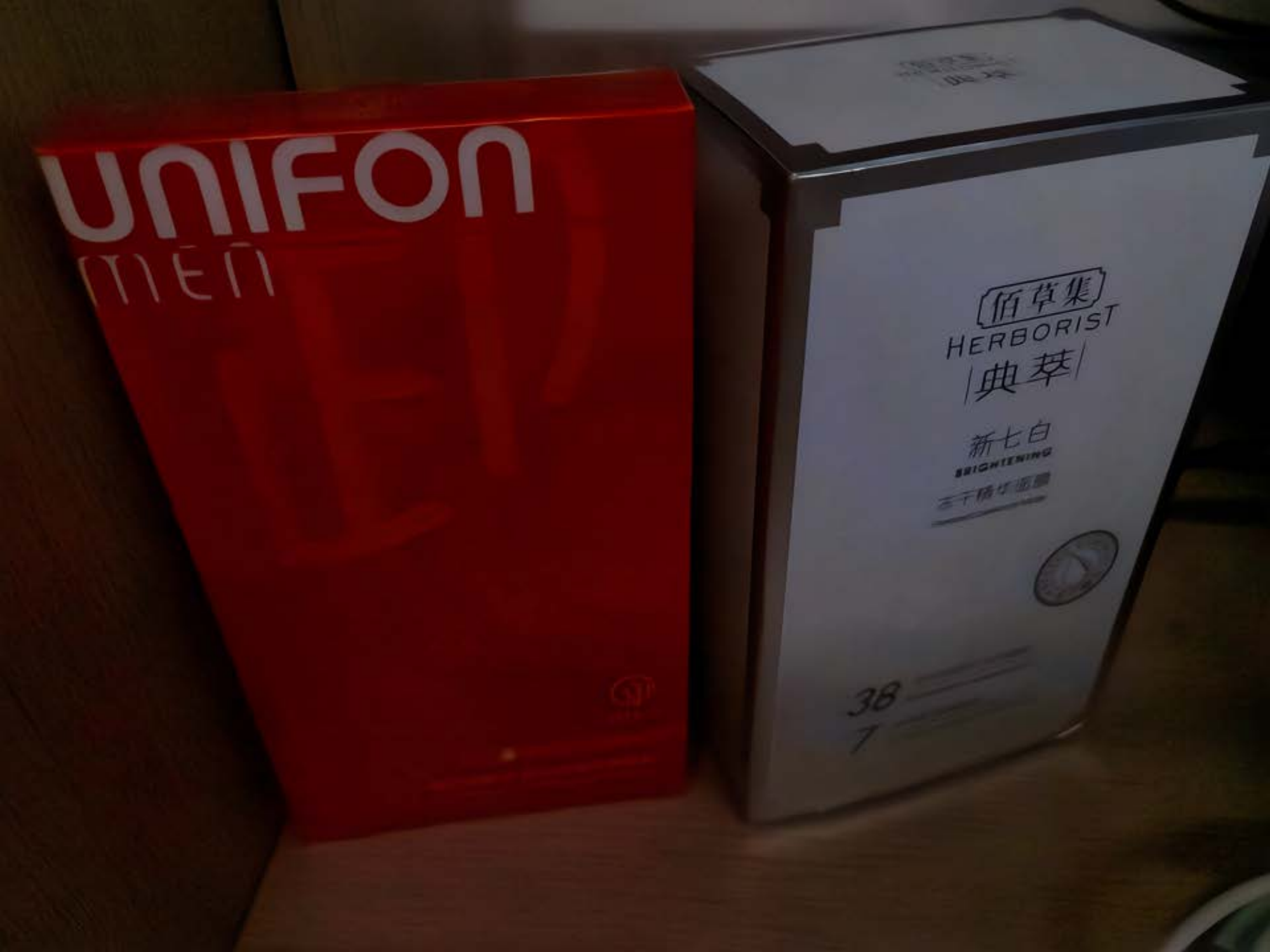}&	
			\includegraphics[width=0.16\linewidth]{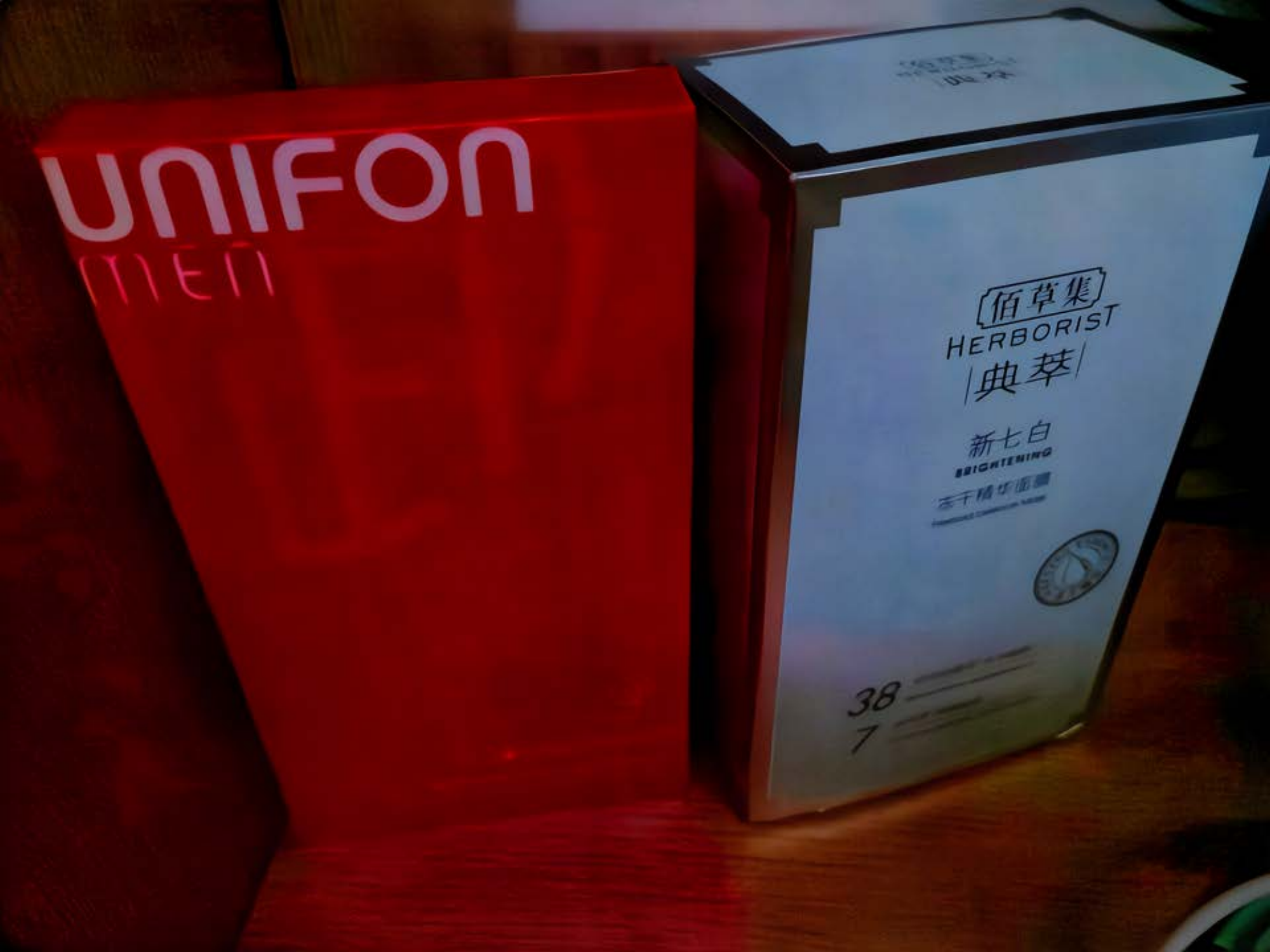}&	
			\includegraphics[width=0.16\linewidth]{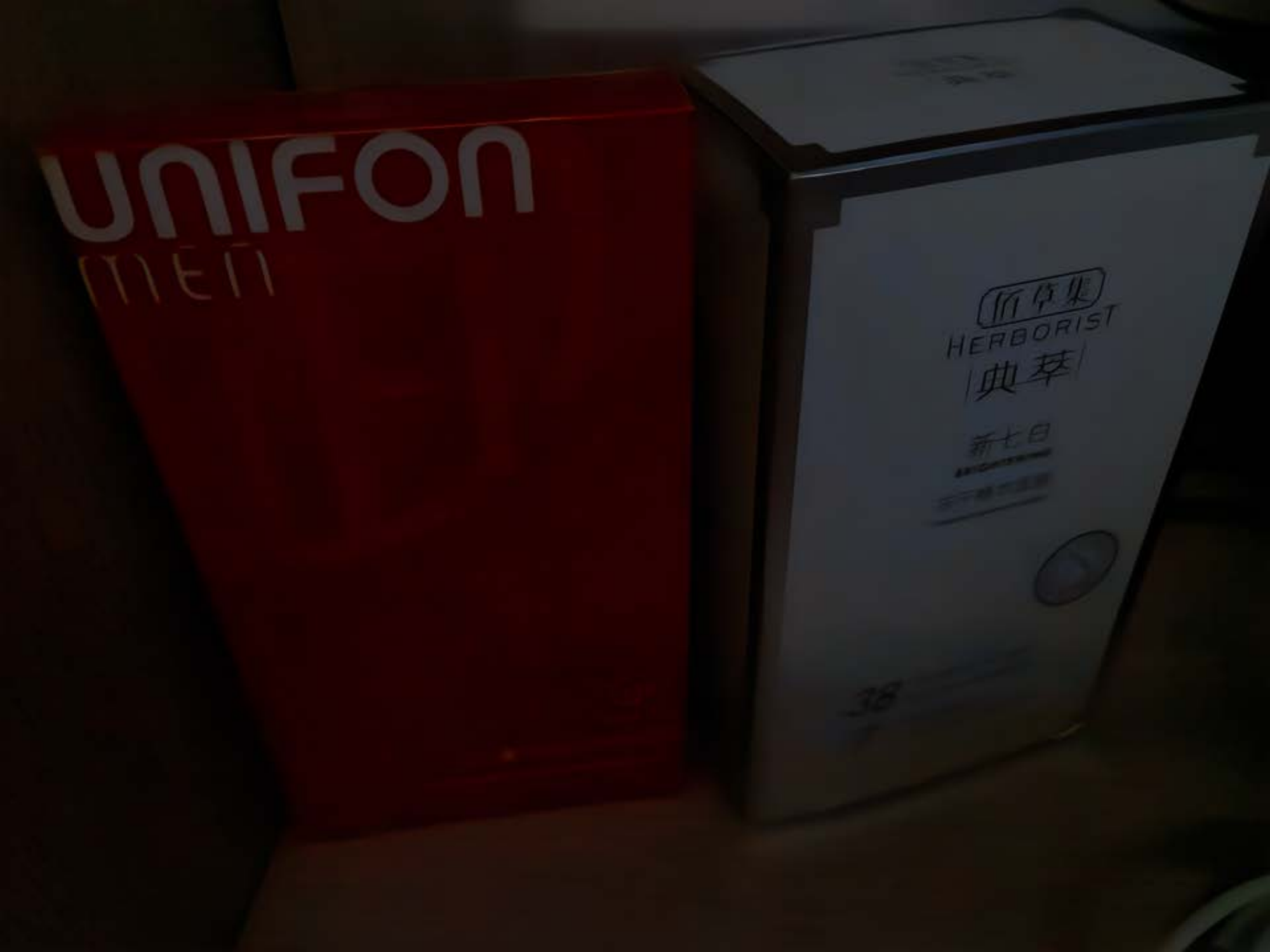}&	
			\includegraphics[width=0.16\linewidth]{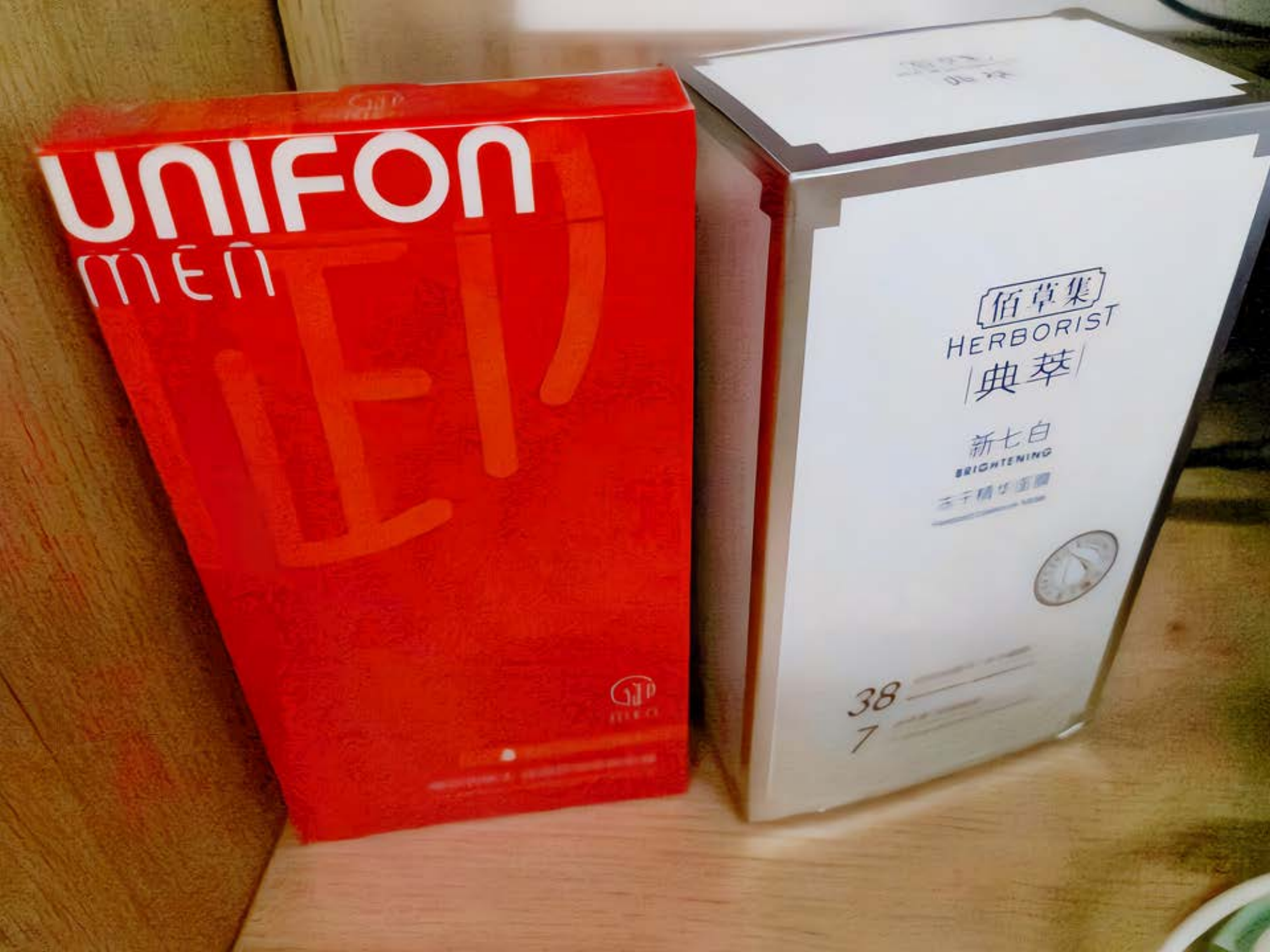}\\
			\vspace{-0.1cm}	
			\includegraphics[width=0.16\linewidth]{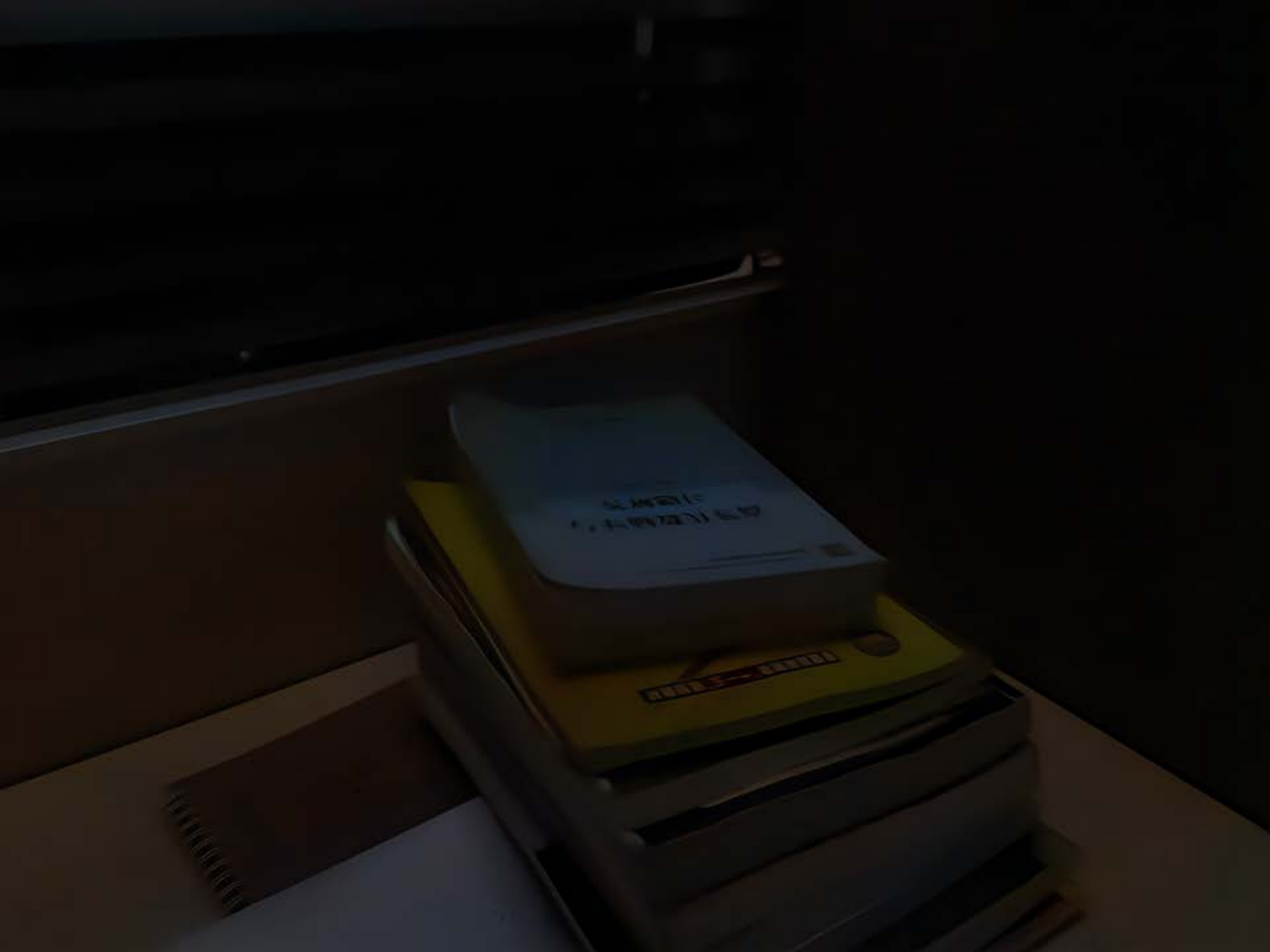}&	
			\includegraphics[width=0.16\linewidth]{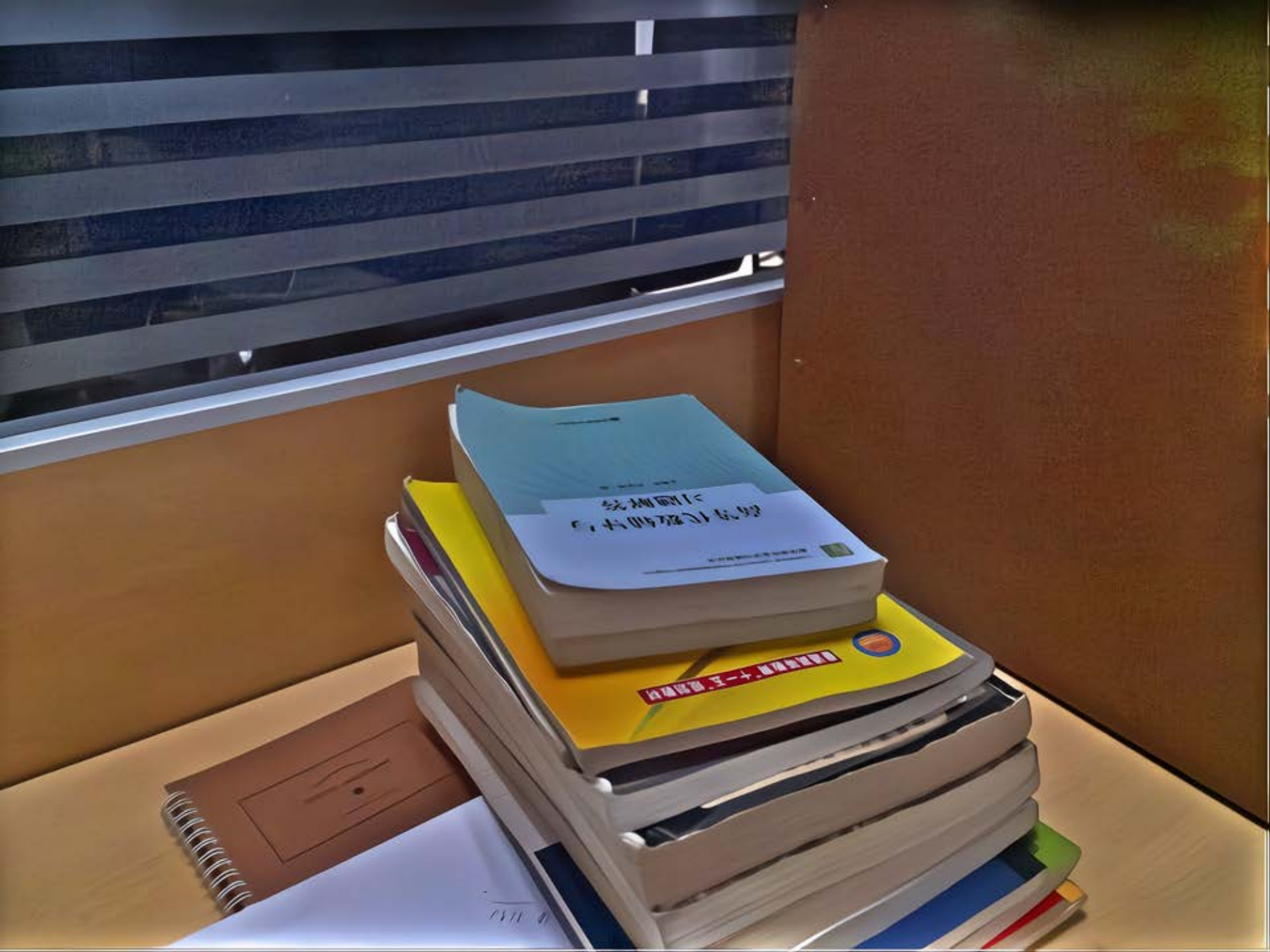}&	
			\includegraphics[width=0.16\linewidth]{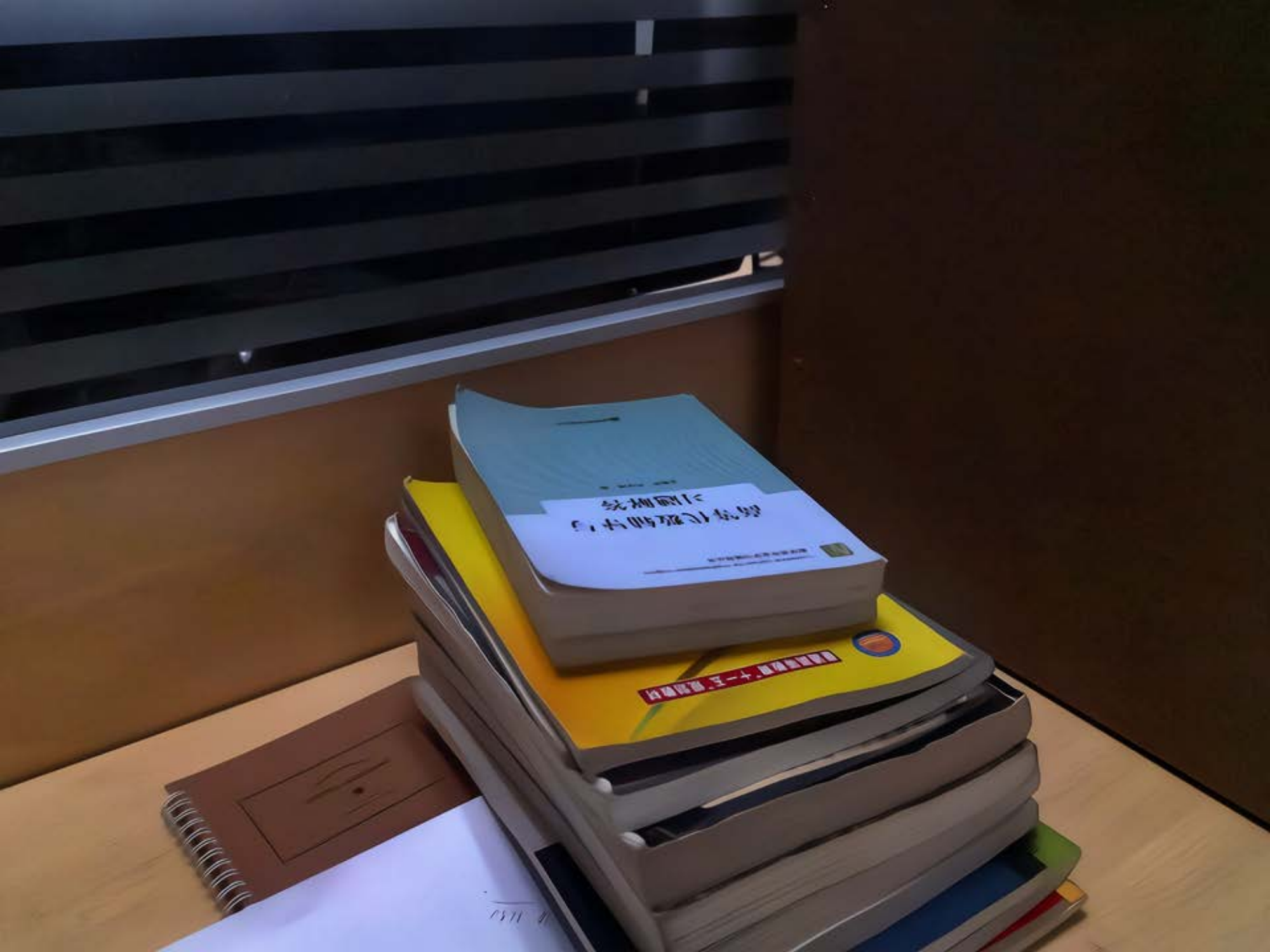}&	
			\includegraphics[width=0.16\linewidth]{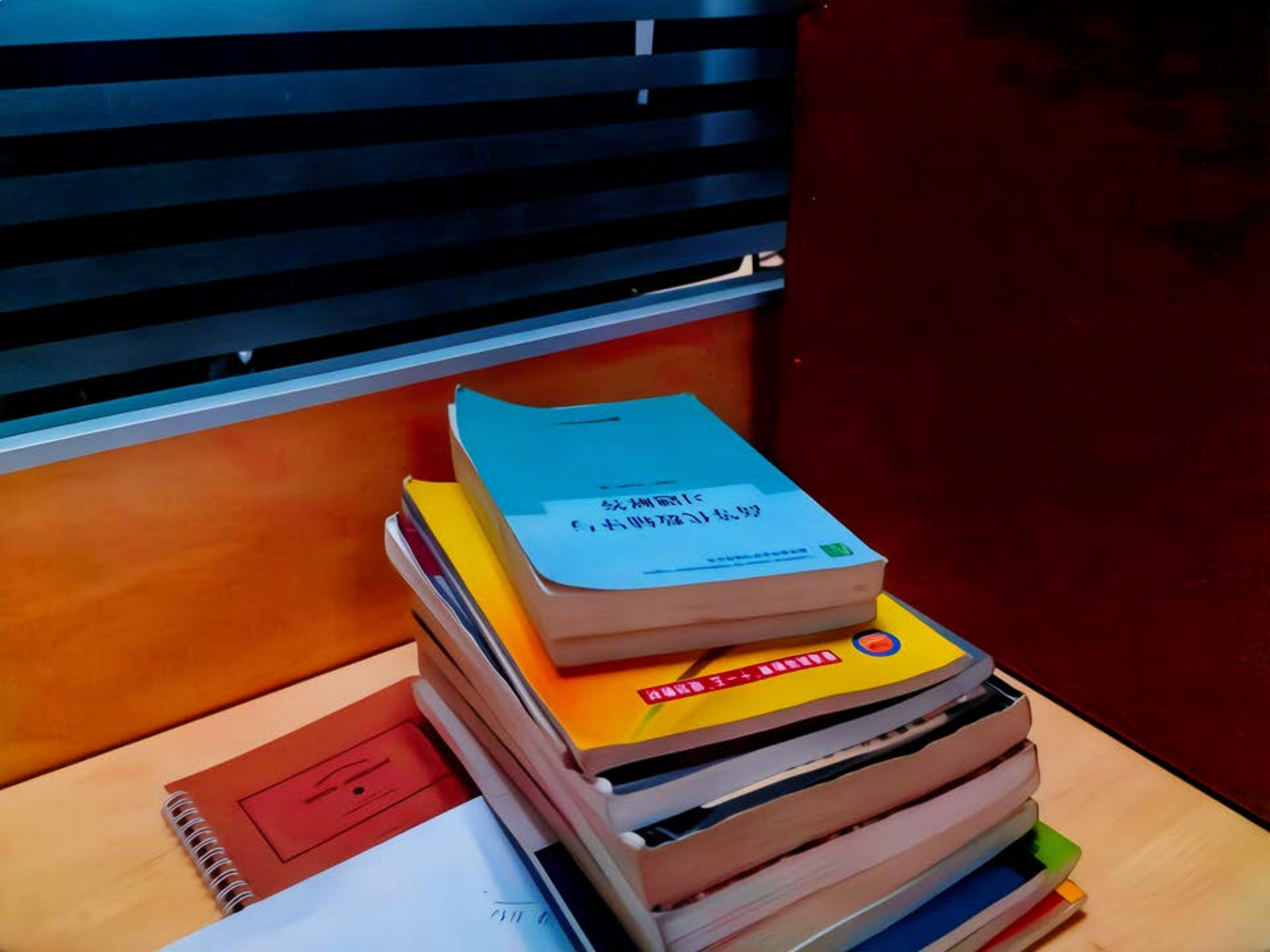}&	
			\includegraphics[width=0.16\linewidth]{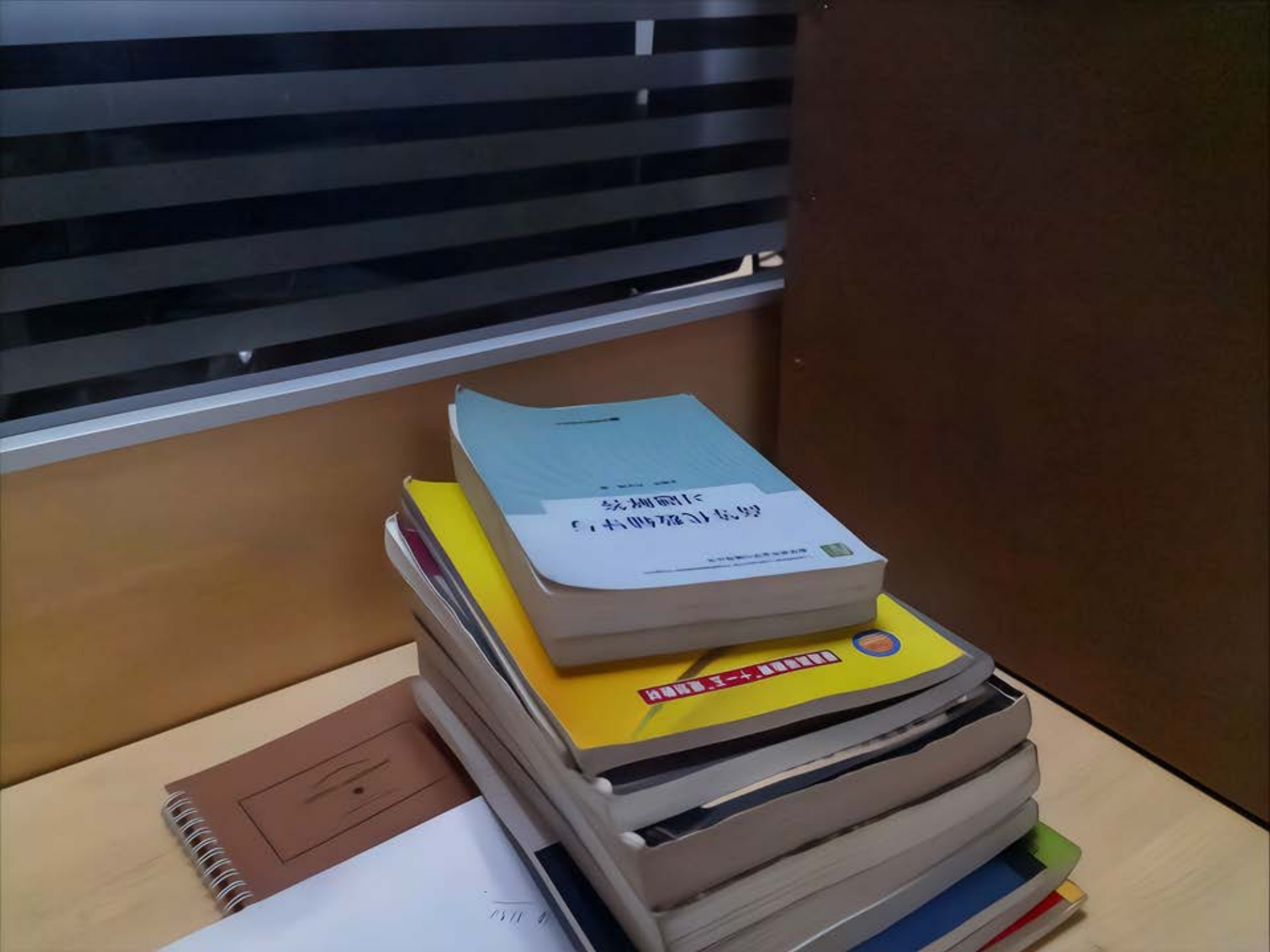}&	
			\includegraphics[width=0.16\linewidth]{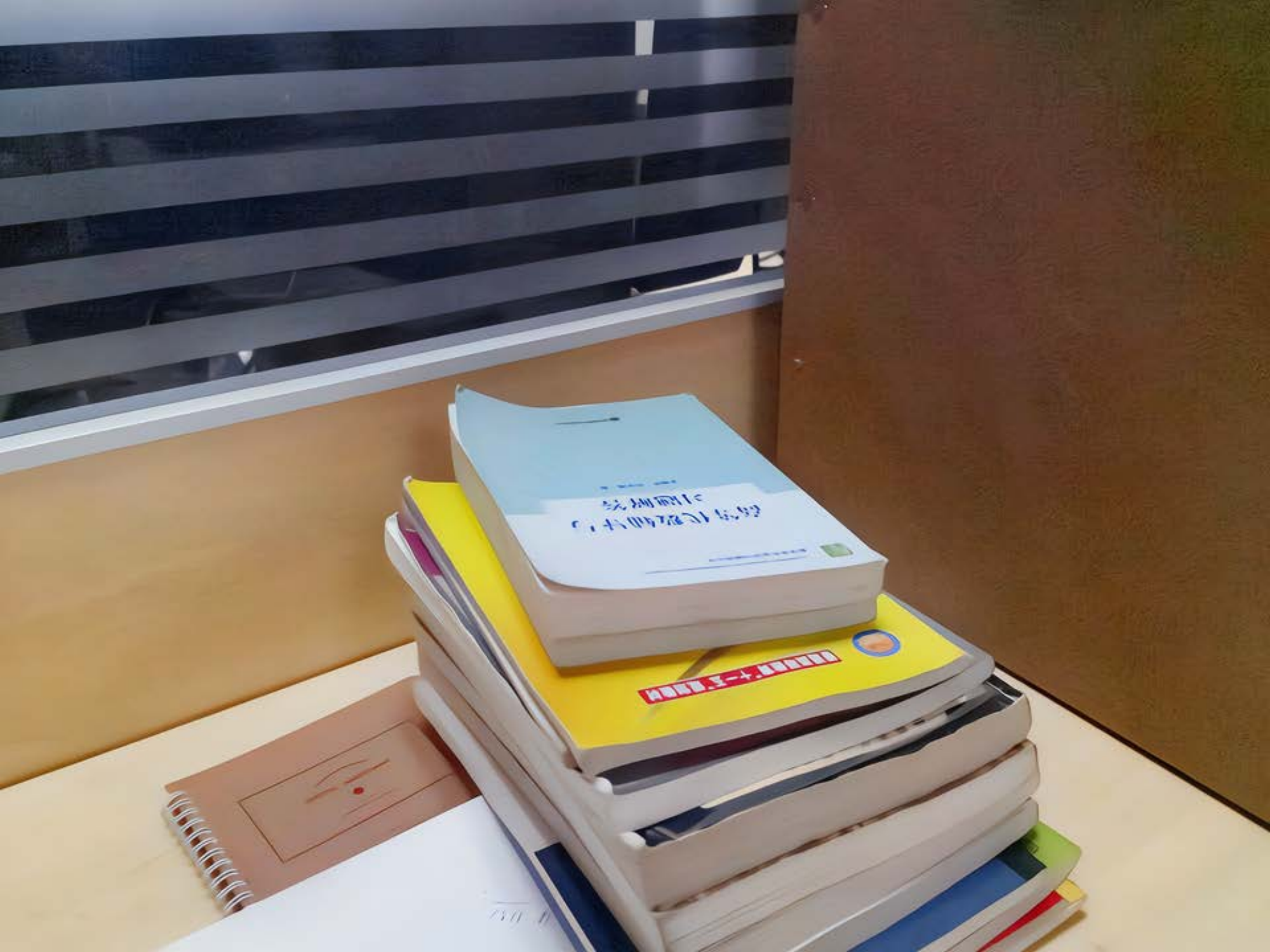}\\
			\vspace{-0.2cm}
			\footnotesize Input&\footnotesize LIME~\cite{guo2017lime}&\footnotesize RUAS~\cite{liu2021retinex}&\footnotesize FEC~\cite{huang2022eccv}&\footnotesize SCI~\cite{ma2022toward}&\footnotesize PEC\\
		\end{tabular}
		\caption{Visual comparison of underexposure scenarios with artifacts. Except for the input, all correction results were processed by FBCNN for artifact removal to improve visualization quality.}
		\label{fig: Scenes_under2}
	\end{figure*}

    \begin{table}[t]
		\footnotesize
		\renewcommand\arraystretch{1.2} 
		\setlength{\tabcolsep}{2mm}
		\centering
		\caption{Comparison of running speed (in seconds) for images with 3 different kinds of resolutions using typical mobile devices.} 
		\begin{tabular}{|c|c|c|c|c|c|c|c|c|}
			\hline
			&\multicolumn{4}{c|}{Snapdragon 865 DSP (\textit{XIAOMI 10})}&\multicolumn{4}{c|}{Kirin 990 NPU (\textit{HUAWEI MATE 30 PRO})}\\
			\hline
			{Resolution}& ZeroDCE~\cite{guo2020zero} &RUAS~\cite{liu2021retinex}&SCI~\cite{ma2022toward}&PEC&  ZeroDCE~\cite{guo2020zero} &RUAS~\cite{liu2021retinex}&SCI~\cite{ma2022toward}&PEC\\
			\hline
			{1280$\times$720}&8.0520&1.8600&\underline{0.0284}&\textbf{0.0101}{\tiny  \textbf{$\uparrow${64\%}}}&1.7040&2.2020&\underline{0.0489}&\textbf{0.0114}{\tiny  \textbf{$\uparrow${77\%}}}\\
			\hline
			{1920$\times$1080}&23.3810&4.0980&\underline{0.0730}&\textbf{0.0268}{\tiny  \textbf{$\uparrow${63\%}}}&4.6780&4.8150&0.0846&\textbf{0.0324}{\tiny  \textbf{$\uparrow${62\%}}}\\
			\hline
			{2560$\times$1440}&43.3570&7.4570&\underline{0.2050}&\textbf{0.0400}{\tiny  \textbf{$\uparrow${80\%}}}&40.2300&9.1220&\underline{0.1970}&\textbf{0.0747}{\tiny  \textbf{$\uparrow${62\%}}} \\
			\hline 
		\end{tabular}	
		\label{table: Mobile Time}
	\end{table} 
    
    \begin{table*}[ht]
		\centering
		\footnotesize
		\renewcommand{\arraystretch}{1.2}
		\setlength{\tabcolsep}{4.6pt}
		\caption{Numerical scores for challenging underexposure scenes with noises/artifacts. Note that all scores were calculated after applying the corresponding post-processing.}
		\vspace{0.2cm}
		\begin{tabular}{|c|ccccc|ccccc|}
			\hline
			\multirow{2}{*}{Metric} & \multicolumn{5}{c|}{Denoising (+SCUNet)} & \multicolumn{5}{c|}{Artifact removal (+FBCNN)} \\ \cline{2-11} 
			& \multicolumn{1}{c|}{LIME~\cite{guo2017lime}}   & \multicolumn{1}{c|}{RUAS~\cite{liu2021retinex}}   & \multicolumn{1}{c|}{FEC~\cite{huang2022eccv}}    & \multicolumn{1}{c|}{SCI~\cite{ma2022toward}}    & {PEC}             	& \multicolumn{1}{c|}{LIME~\cite{guo2017lime}}   & \multicolumn{1}{c|}{RUAS~\cite{liu2021retinex}}   & \multicolumn{1}{c|}{FEC~\cite{huang2022eccv}}    & \multicolumn{1}{c|}{SCI~\cite{ma2022toward}}    & {PEC}             \\ \hline 
			LOE$\downarrow$ & \multicolumn{1}{c|}{725.68} & \multicolumn{1}{c|}{188.01} & \multicolumn{1}{c|}{274.63} & \multicolumn{1}{c|}{\underline{173.47}} & \textbf{168.39} & \multicolumn{1}{c|}{671.01} & \multicolumn{1}{c|}{249.75} & \multicolumn{1}{c|}{270.93} & \multicolumn{1}{c|}{\underline{205.16}} & \textbf{202.59} \\ \hline
			DE$\uparrow$ & \multicolumn{1}{c|}{\underline{6.9582}} & \multicolumn{1}{c|}{5.8401} & \multicolumn{1}{c|}{5.7403} & \multicolumn{1}{c|}{6.4996} & \textbf{7.0464} & \multicolumn{1}{c|}{\underline{7.0834}} & \multicolumn{1}{c|}{6.4410} & \multicolumn{1}{c|}{6.3092} & \multicolumn{1}{c|}{6.9027} & \textbf{7.1038} \\ \hline
			NIQE$\downarrow$ & \multicolumn{1}{c|}{\underline{3.1904}} & \multicolumn{1}{c|}{4.1828} & \multicolumn{1}{c|}{4.0910} & \multicolumn{1}{c|}{3.6972} & \textbf{3.0880} & \multicolumn{1}{c|}{\underline{3.3859}} & \multicolumn{1}{c|}{3.6444} & \multicolumn{1}{c|}{3.7716} & \multicolumn{1}{c|}{3.4903} & \textbf{3.2986} \\ \hline
		\end{tabular}
		\label{tab: B1}
	\end{table*}

	\subsection{Discussion}
	Essentially, the developed PEC belongs to a curve-based scheme, similar to ZeroDCE. Here, we explore the distinctions between them.
	\begin{itemize}
		\item \textit{\textbf{Parameter space}: ZeroDCE is derived from a quadratic curve, and its objective is to learn a pixel-wise curve parameter map. This results in a significant computational burden, particularly for large-scale images, because the size of the map matches that of the input image. In contrast, PEC adheres to the traditional curve computational flow and only requires the pre-definition of three one-dimensional (1-D) parameters, thus reducing the testing burden.} 
		\item \textit{\textbf{Parameter setting}: When encountering unfamiliar scenarios, ZeroDCE requires the definition of three hyperparameters in the loss function and basic parameters (\textit{e.g.,} learning rate) to learn the desired pixel-wise curve parameter map. Distinct from ZeroDCE, PEC only involves three parameters within a limited range, simplifying the adaptation process in unknown environments.} 
	\end{itemize}
	
	Here, we analyze the key compensation by comparing PEC with other representative methods~\cite{ma2022toward, afifi2021learning, guo2017lime, guo2020zero, wang2022local, huang2022eccv}. As shown in Figure~\ref{fig: Compensation}, in the underexposure case, the developed PEC performs better in recovering the details of scenes, with more balanced brightness control (\textit{e.g.,} in the areas of the light tube and ceiling). Conversely, in the overexposure case, the developed method provides higher contrast, and from the perspective of exposure compensation, it highlights structural information more prominently, demonstrating a clear advantage. 
	By applying the same derived compensation process, we can obtain the compensation of the other methods for comparison. These constructed compensations indeed yield similar content. However, as depicted in the right panel, because of the constraints imposed by data-specific learning patterns, undesired artifacts are introduced with the MSEC method, leading to blurring of the structural information (see the bottom row for MSEC). In conclusion, this experiment provides strong evidence that the developed approach is traceable and remarkably effective.

	\section{Experimental results}	
	This section presents a comprehensive evaluation of the developed PEC through comparison with existing state-of-the-art methods across various scenarios. We also investigate the applicability of the developed PEC to other vision tasks. Furthermore, we perform a series of analytical experiments to demonstrate its effectiveness. 
	
	\subsection{Comparison with advanced networks}
	\textbf{Benchmark methods, datasets, and metrics.}
	We begin the evaluation by comparing the developed PEC with several advanced data-driven deep networks, which include three supervised underexposure correctors (KinD~\cite{zhang2019kindling}, UTVNet~\cite{zheng2021adaptive}, and URetinex~\cite{wu2022uretinex}), three unsupervised underexposure correctors (SCI~\cite{ma2022toward}, ZeroDCE~\cite{guo2020zero}, and RUAS~\cite{liu2021retinex}), and three general exposure correctors (MSEC~\cite{afifi2021learning}, LCD~\cite{wang2022local}, and FEC~\cite{huang2022eccv}). 
	To evaluate the objective and subjective performance of the PEC, we randomly sample 500 images from the recently generated Exposure-Errors testing dataset~\cite{afifi2021learning}. This dataset contains images with a broader exposure range, spanning relative EVs from -1.5 to +1.5. 
	For comprehensive statistical assessment, we utilize three full-reference metrics: the peak signal-to-noise ratio (PSNR), structural similarity index (SSIM), and a data-driven measurement called learned perceptual image patch similarity (LPIPS)\cite{zhang2018unreasonable}. 
	We also utilize three no-reference metrics, namely the discrete entropy (DE)~\cite{ye2007discrete}, which measures the color richness and sharpness, the lightness-order error (LOE)~\cite{wang2013naturalness}, which evaluates the lightness distortion, and the naturalness image quality evaluator (NIQE)~\cite{mittal2012making}, which assesses naturalness preservation. \textit{Note that ideal references are challenging to capture. Therefore, no-reference metrics are more meaningful and valuable in most real-world environments.}

	\textbf{Quantitative evaluation.}
	Table~\ref{table: ExposureErrors} summarizes the objective numerical scores for the Exposure-Errors dataset. 
	Clearly, the general correctors (MSEC, LCD, and FEC), which are trained on the same distribution as the testing dataset, achieve the best results in full-reference metrics. The underexposure-oriented supervised methods (KinD, UTVNet, and URetinex) and the unsupervised methods (ZeroDCE, RUAS, and SCI), which did not encounter the testing scenes during training, face a common challenge: they struggle to adapt to new scenarios because of their data-dependent nature.
	In contrast, the proposed learning-free method consistently outperforms the data-driven approaches in no-reference metrics. This demonstrates the potential power of the learning-free paradigm, as it does not rely on large-scale training data and can effectively handle various exposure conditions. The subjective evaluations presented hereinafter further support its effectiveness.
	
	\textbf{Qualitative evaluation.} 
	Figure~\ref{fig: MSEC} presents visual comparisons with state-of-the-art methods~\cite{zhang2019kindling, guo2020zero, afifi2021learning, liu2021retinex, zheng2021adaptive, wang2022local, huang2022eccv, wu2022uretinex, ma2022toward} on scenes from the Exposure-Errors dataset~\cite{afifi2021learning}, with zoomed-in regions for better visualization. The comparison reveals that all learning-based methods struggle to provide structure-prominent, exposure-consistent visual quality for the underexposure case. Although KinD, ZeroDCE, and URetinex achieve competitive exposure, they fail to preserve the hierarchical structures, leading to a degradation in the overall naturalness. RUAS leads to overexposure in the sky region, and UTVNet introduces distinct color distortion. For the overexposure case, MSEC introduces local artifacts, and LCD and FEC result in unfriendly color rendition. In contrast, the PEC generates visually appealing results with vivid colors and prominent structures, for both underexposure and overexposure cases.

	\begin{figure*}[t]
		\centering
		\begin{tabular}{c@{\extracolsep{0.2em}}c@{\extracolsep{0.2em}}c@{\extracolsep{0.2em}}c@{\extracolsep{0.2em}}c}	
			\vspace{-0.1cm}	
			\includegraphics[width=0.19\linewidth]{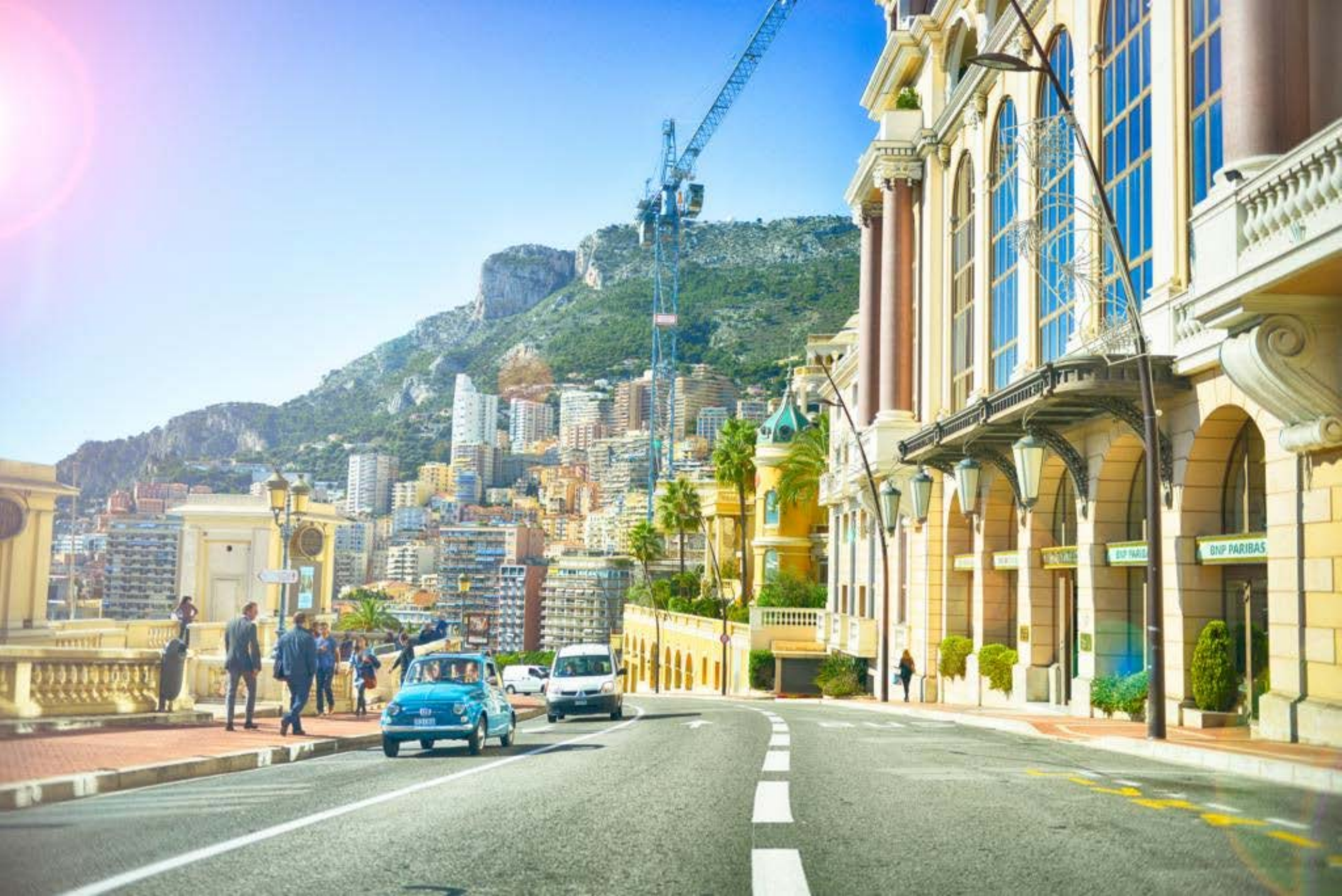}&	
			\includegraphics[width=0.19\linewidth]{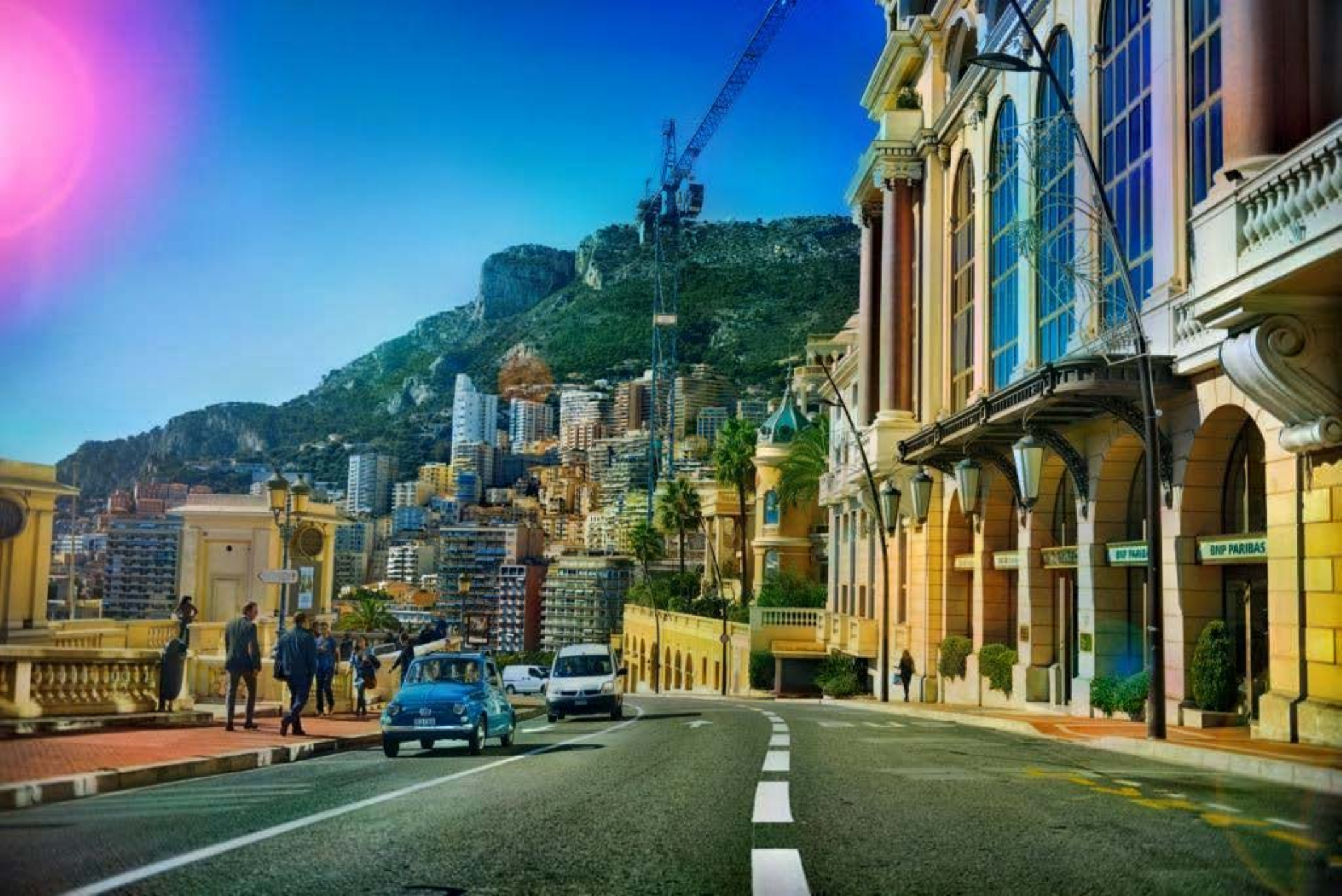}&	
			\includegraphics[width=0.19\linewidth]{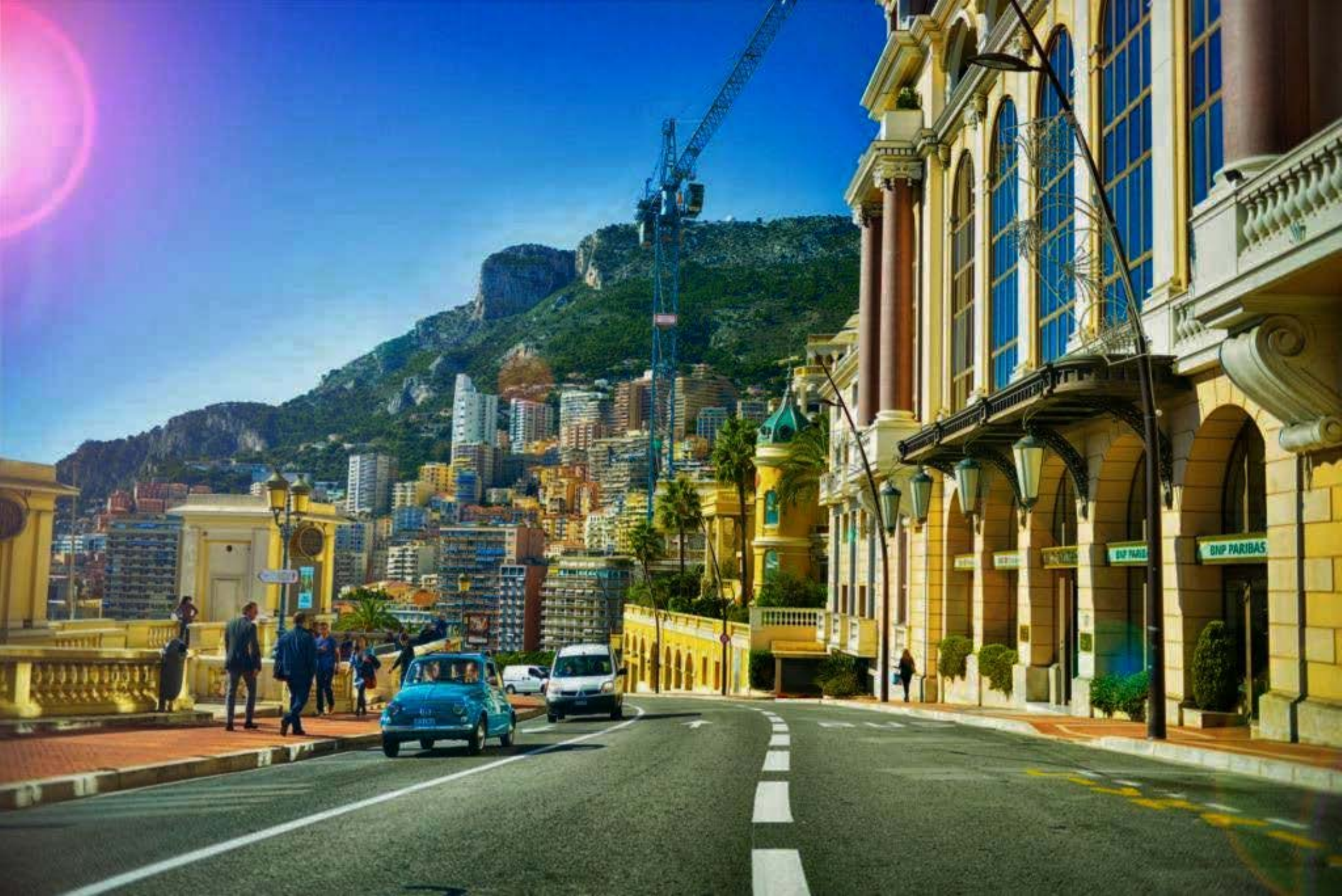}&	
			\includegraphics[width=0.19\linewidth]{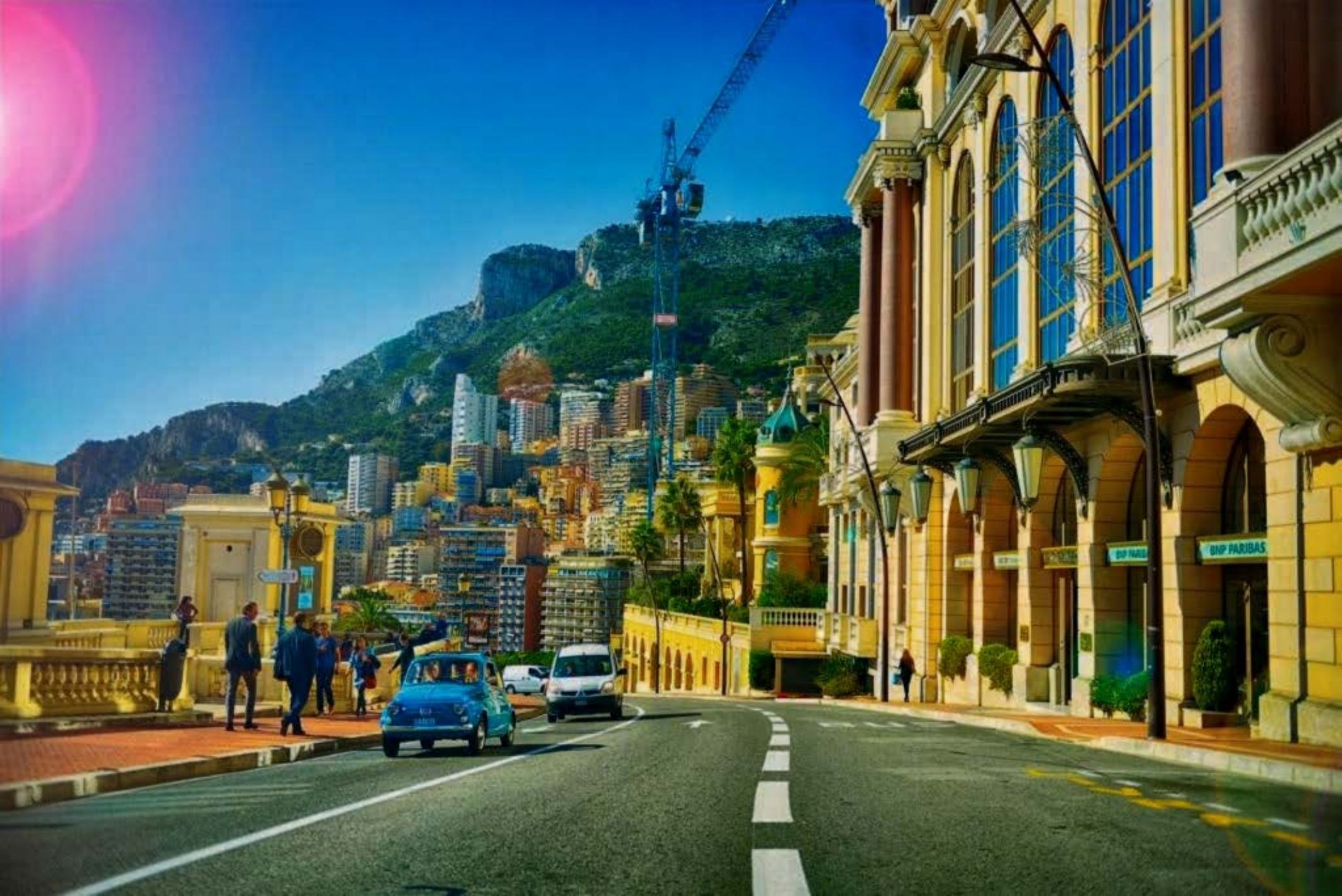}&	
			\includegraphics[width=0.19\linewidth]{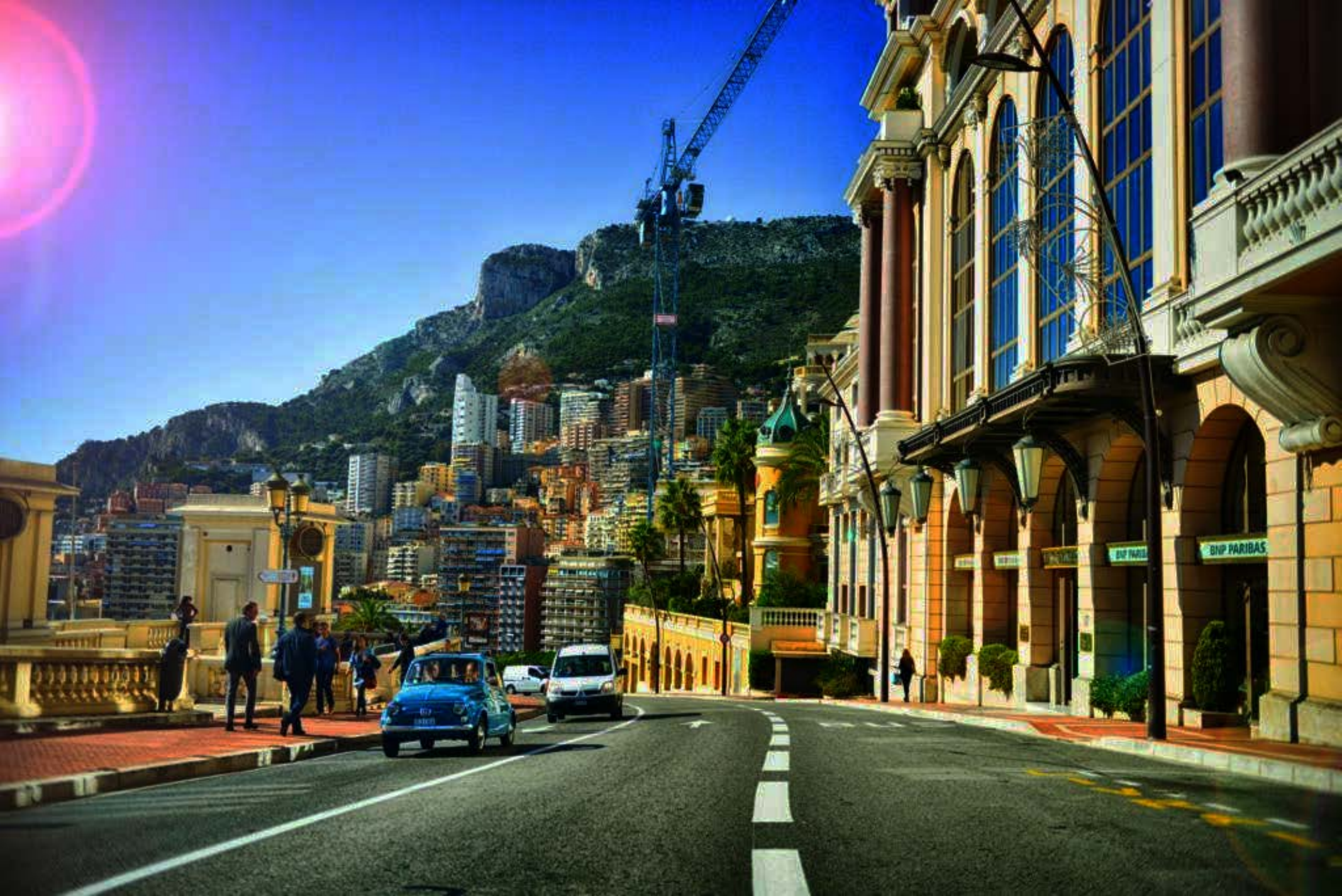}\\
			\vspace{-0.1cm}
			\includegraphics[width=0.19\linewidth]{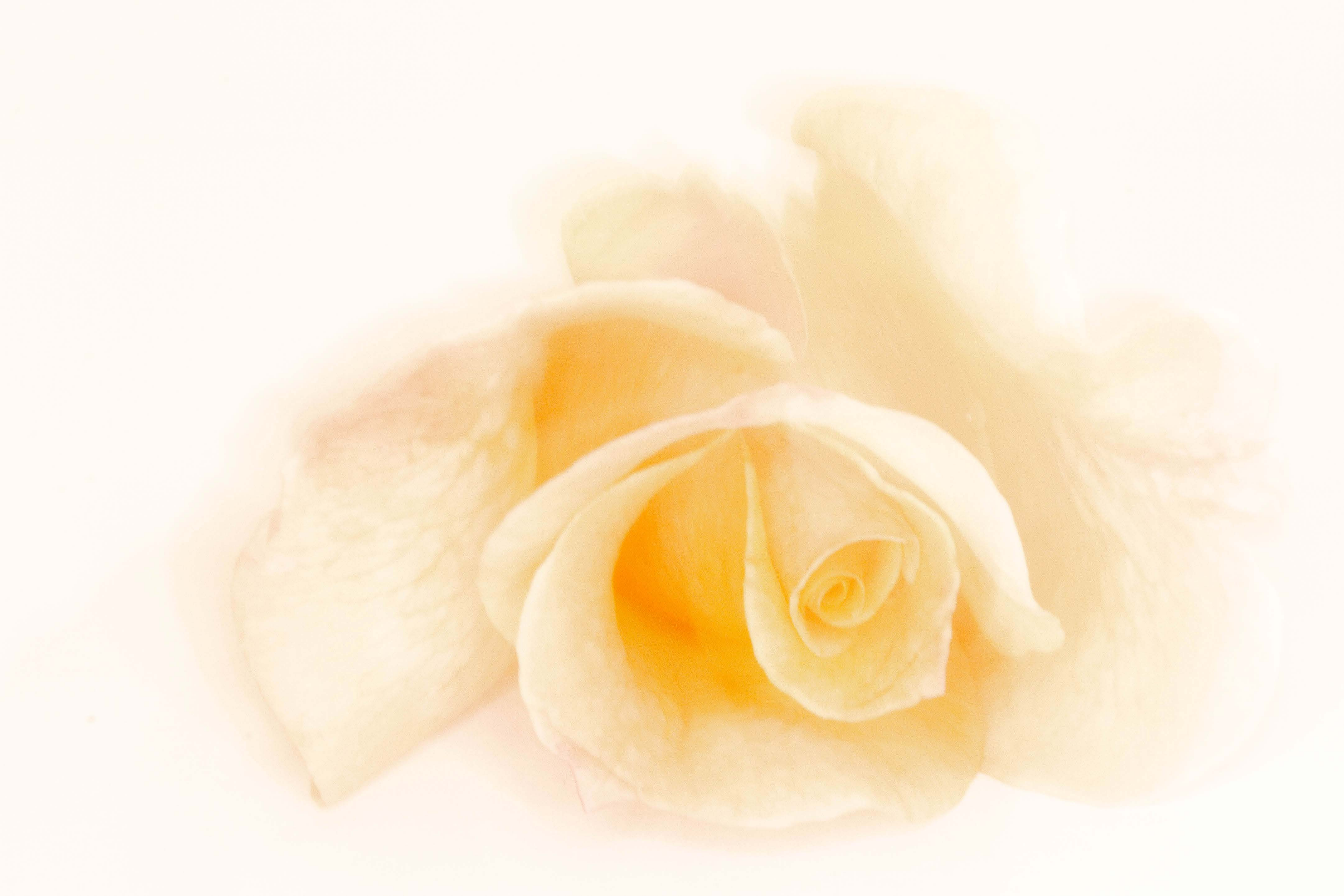}&	
			\includegraphics[width=0.19\linewidth]{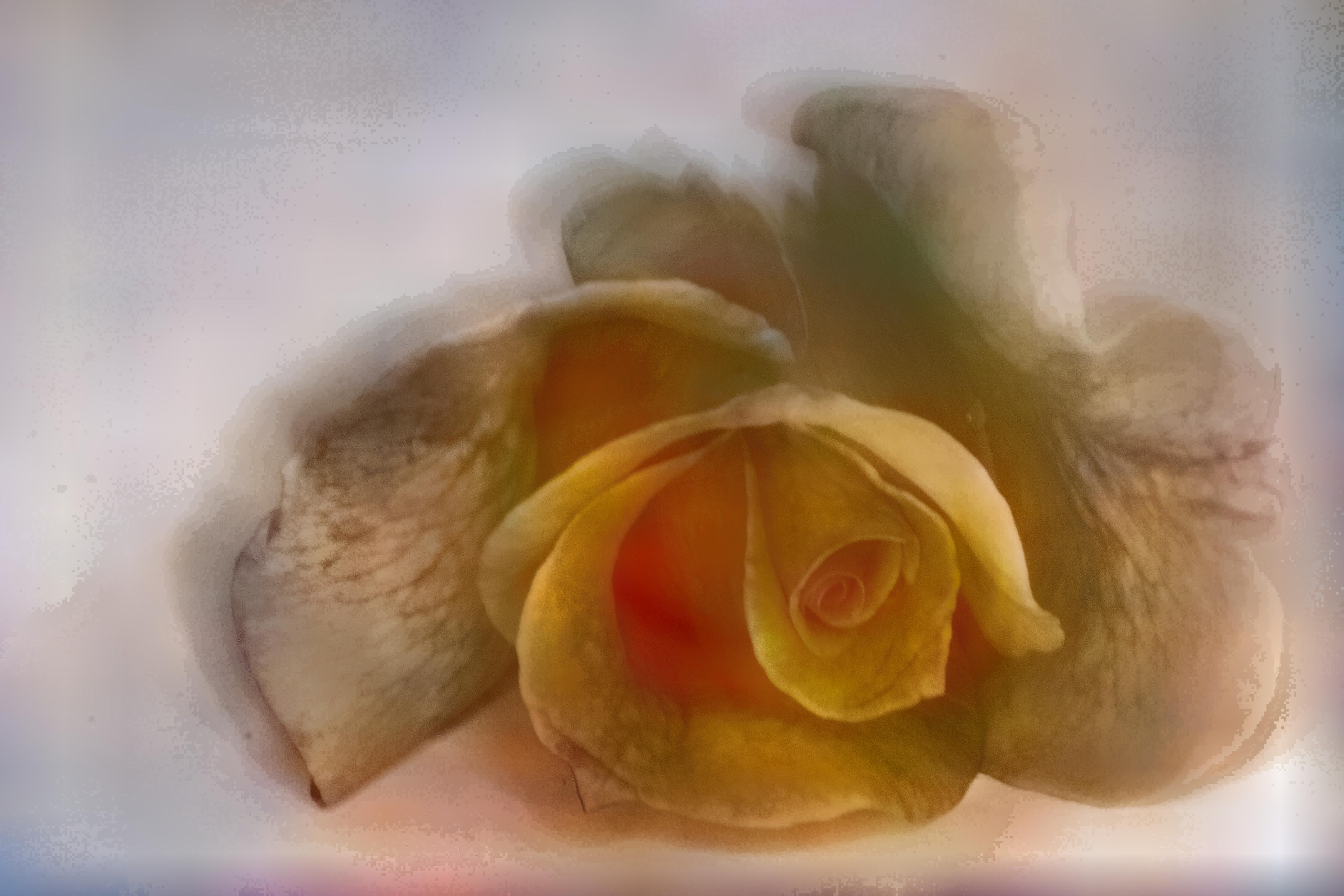}&	
			\includegraphics[width=0.19\linewidth]{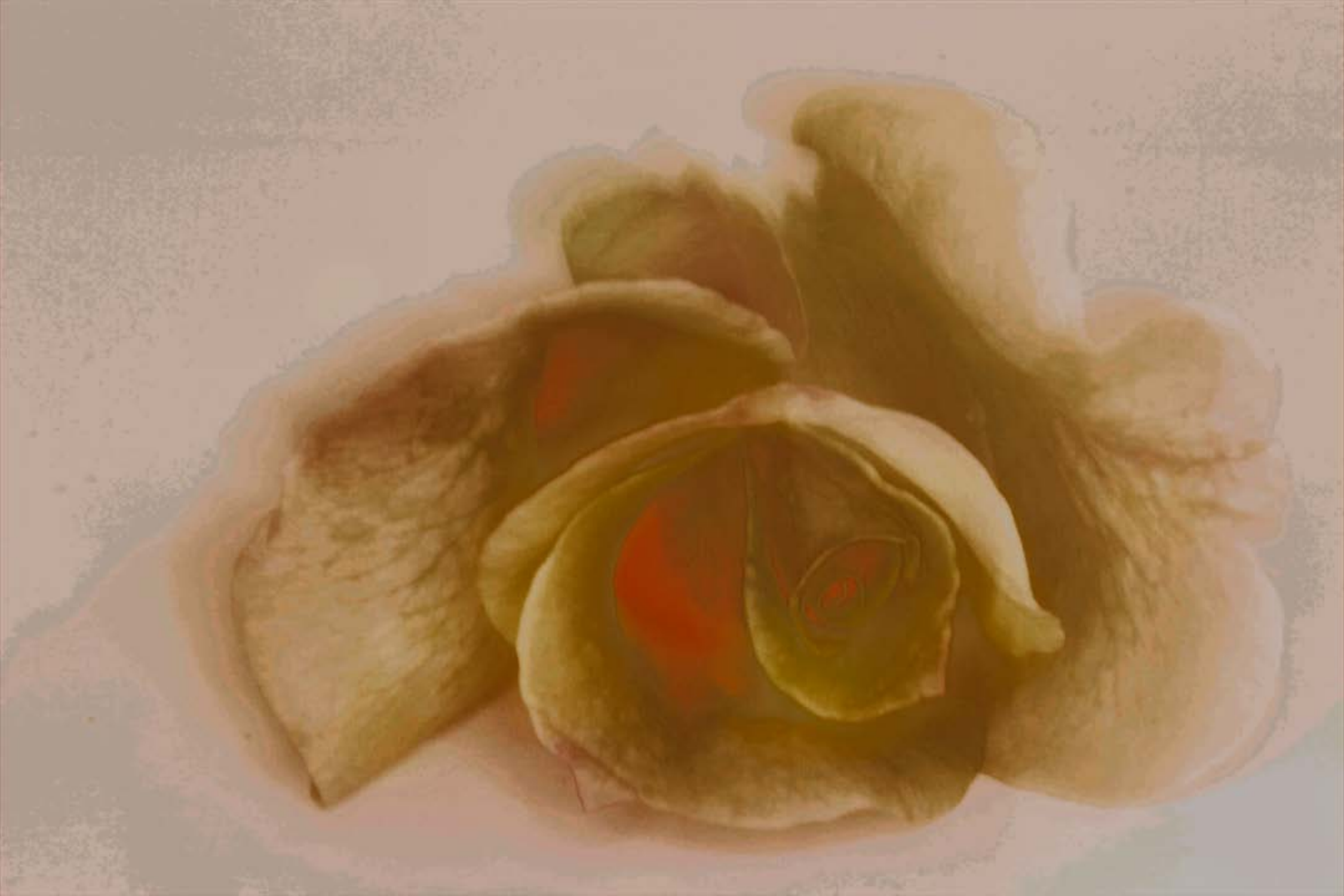}&	
			\includegraphics[width=0.19\linewidth]{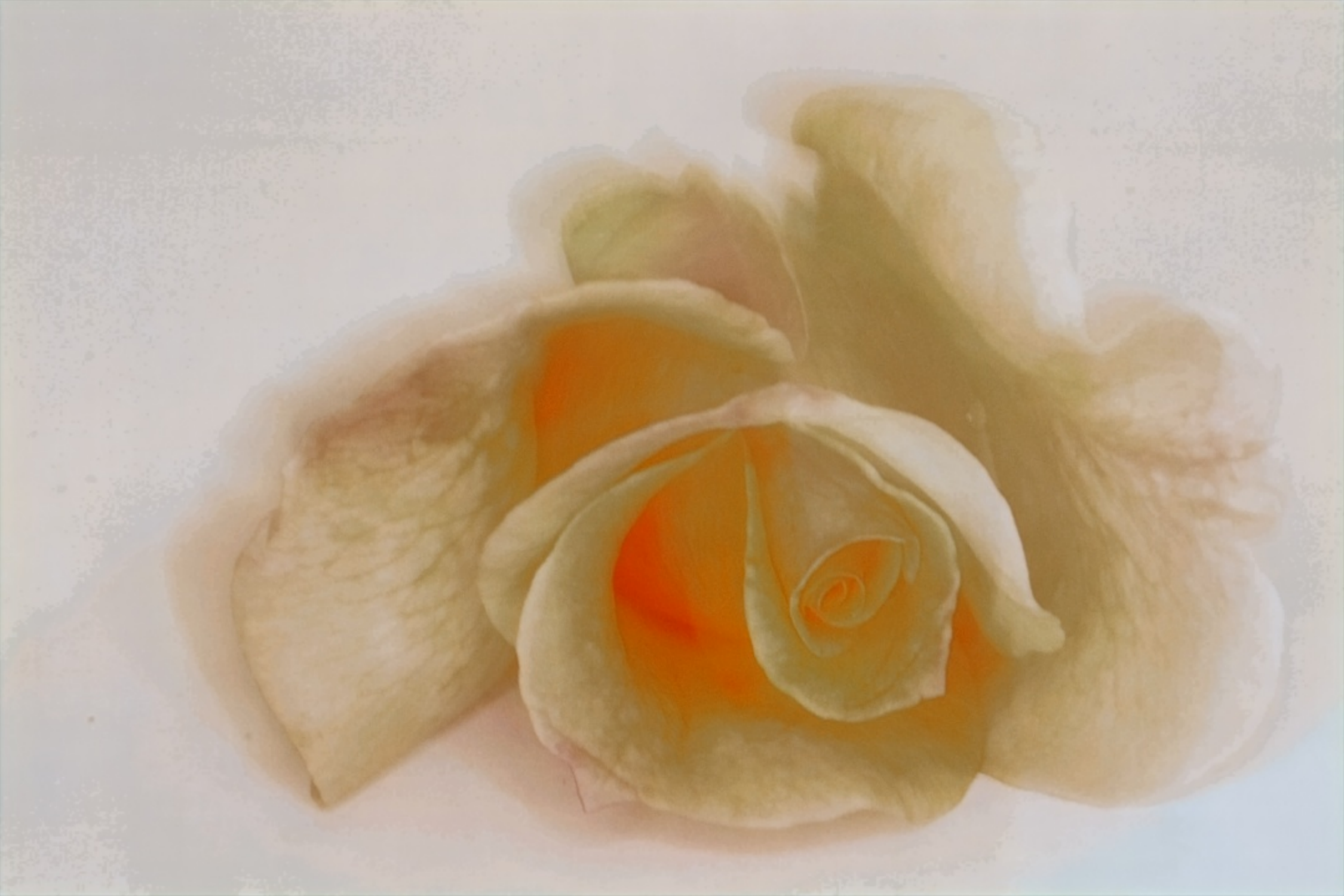}&	
			\includegraphics[width=0.19\linewidth]{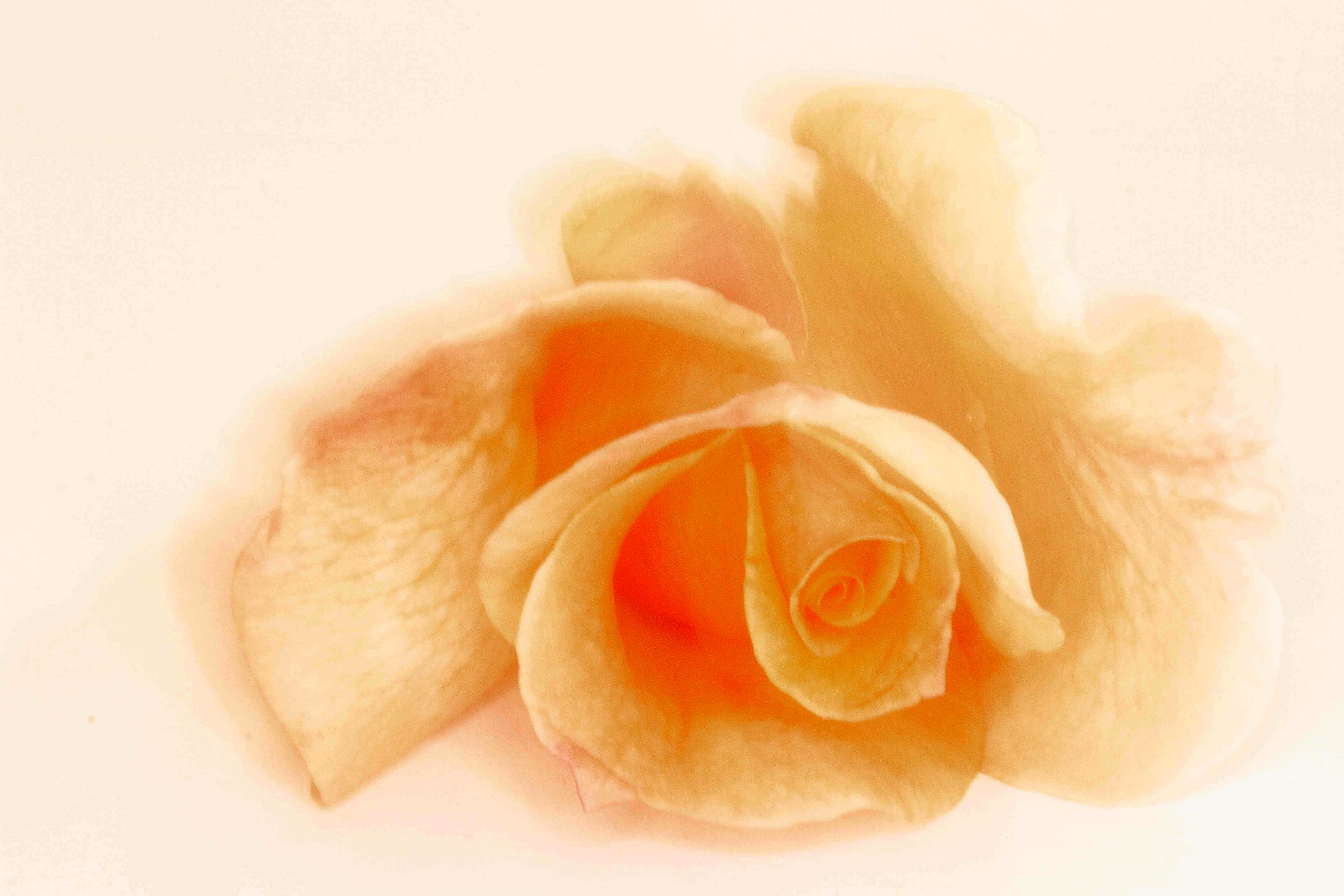}\\
			\vspace{-0.1cm}
			\includegraphics[width=0.19\linewidth]{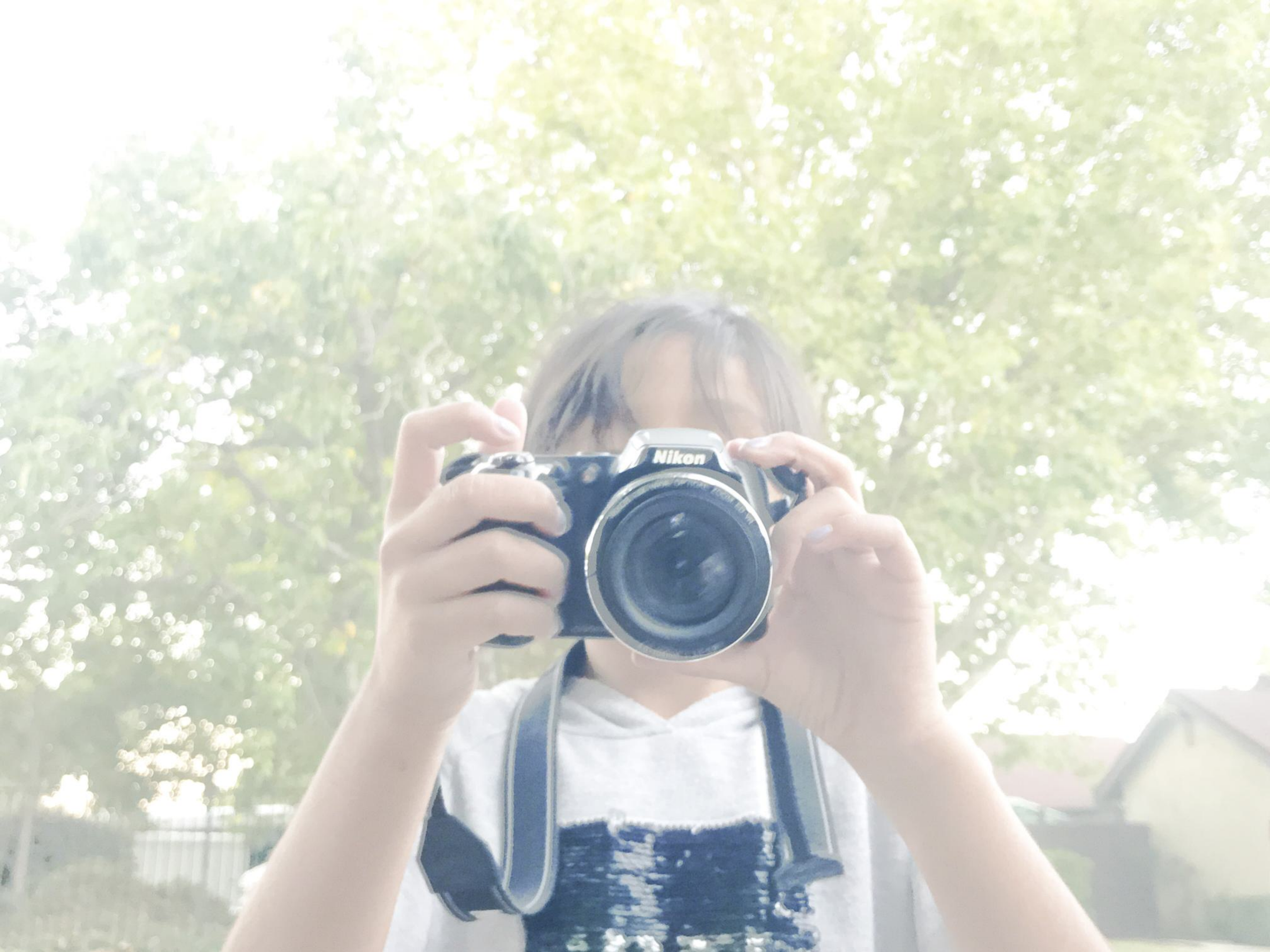}&	
			\includegraphics[width=0.19\linewidth]{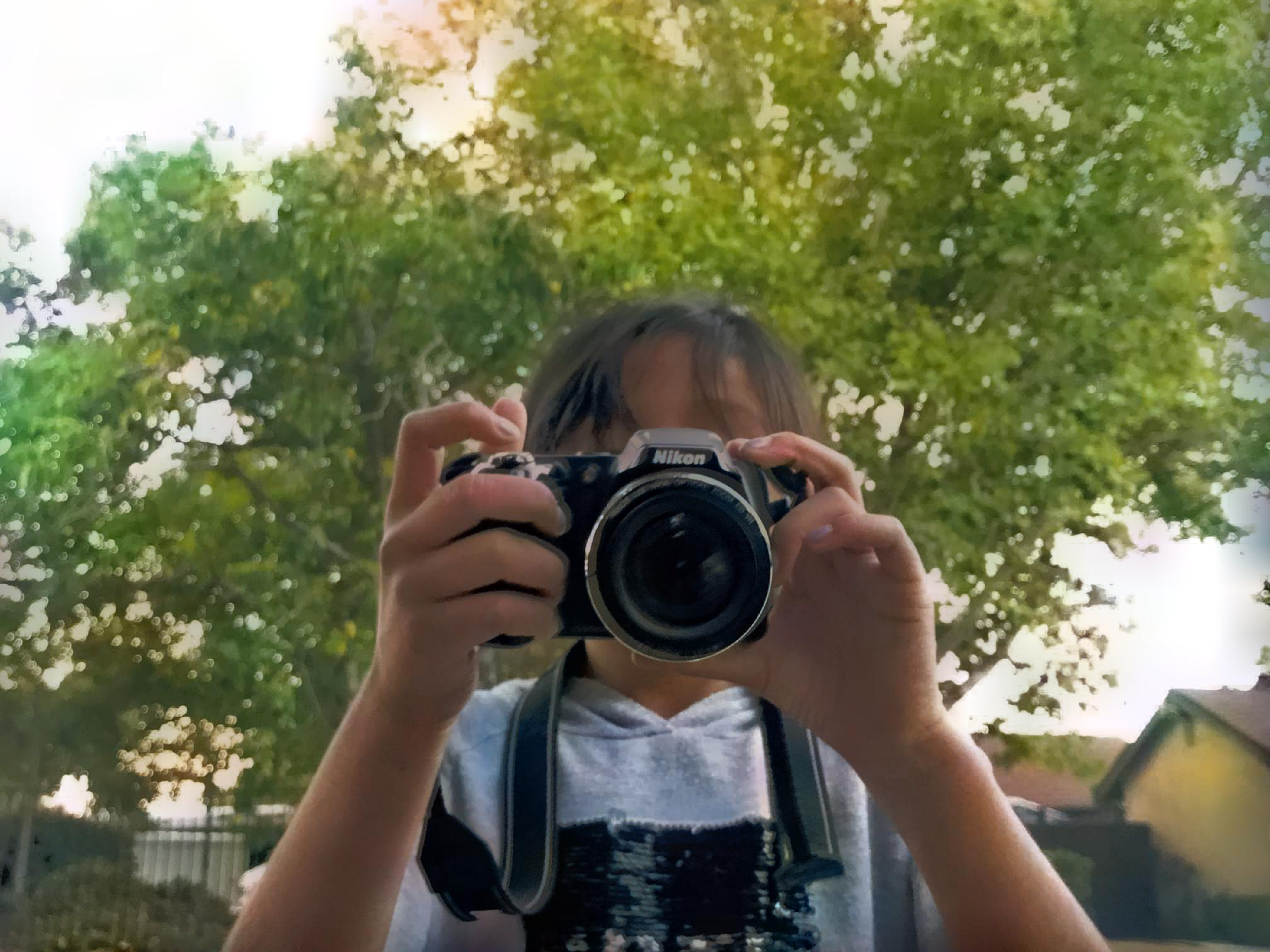}&				\includegraphics[width=0.19\linewidth]{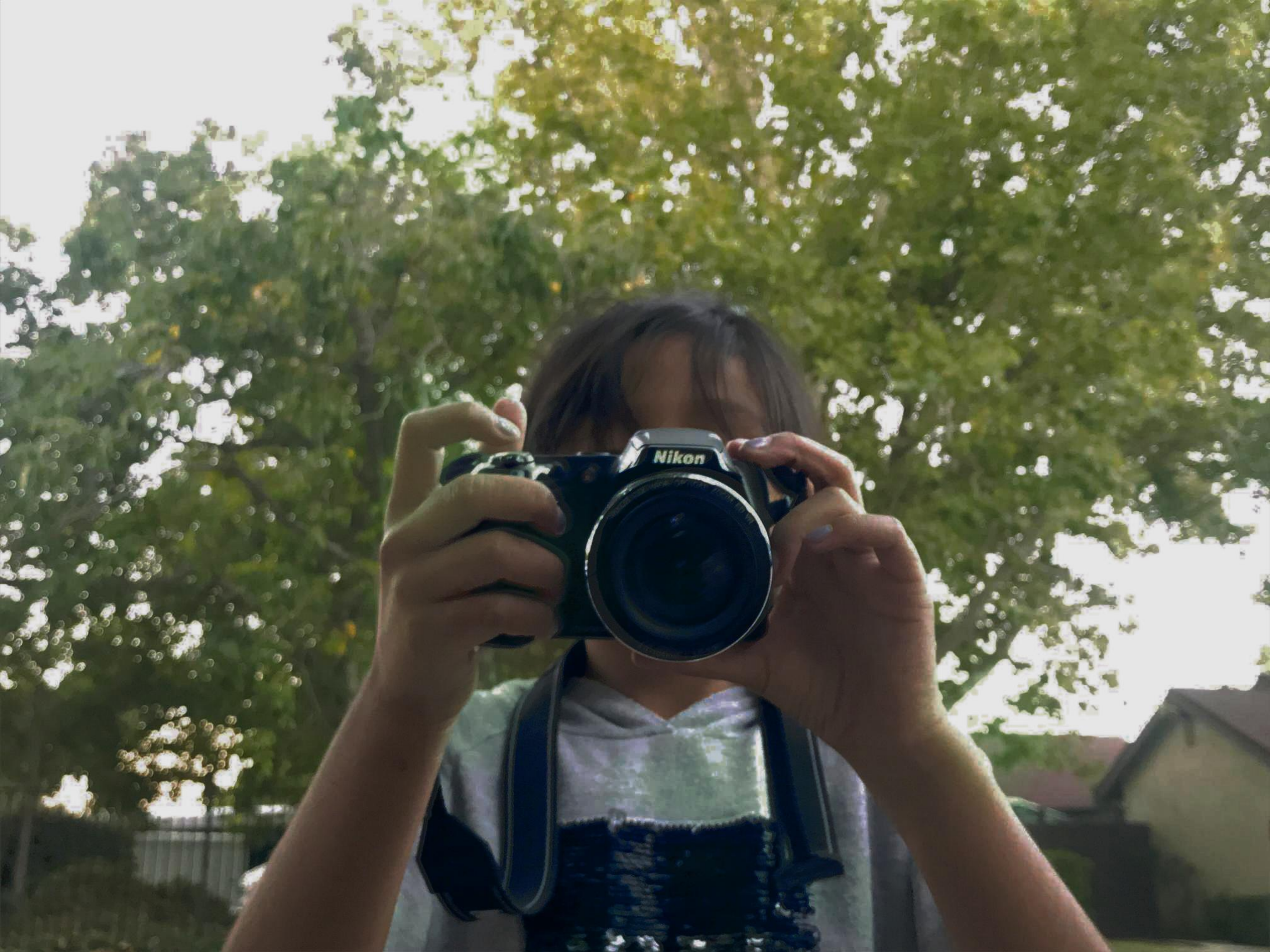}&	
			\includegraphics[width=0.19\linewidth]{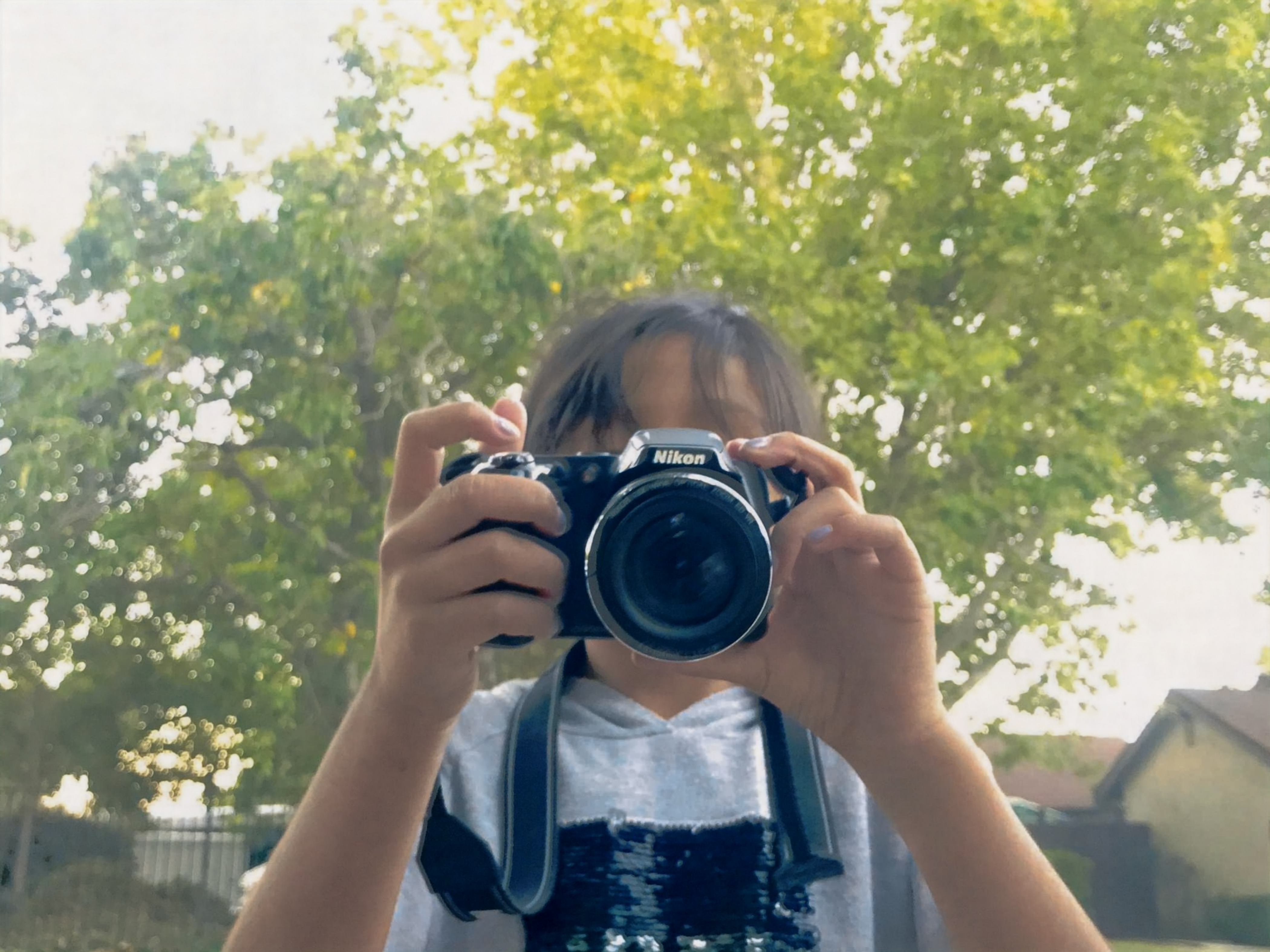}&	
			\includegraphics[width=0.19\linewidth]{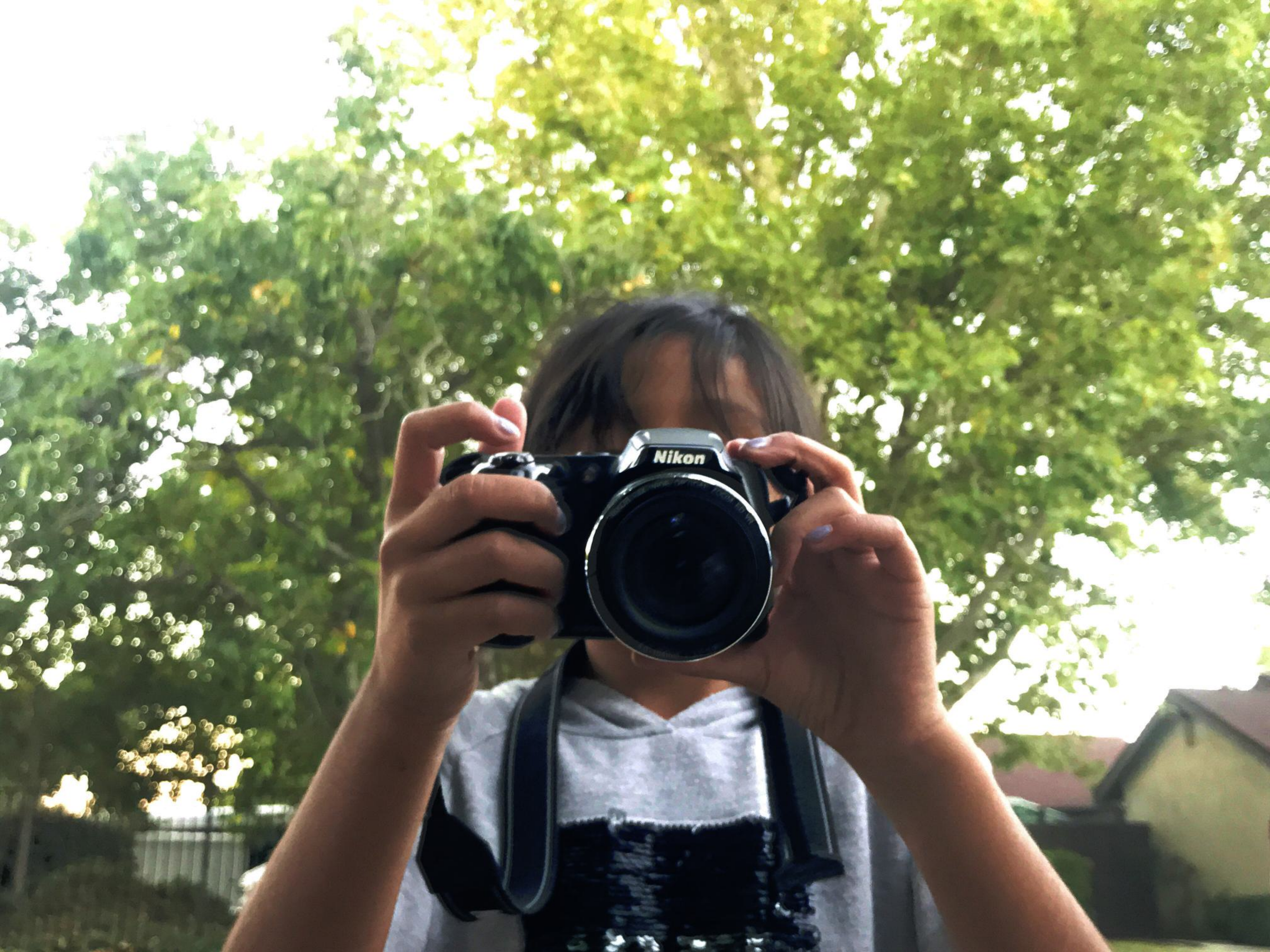}\\
			\vspace{-0.2cm}
			\includegraphics[width=0.19\linewidth]{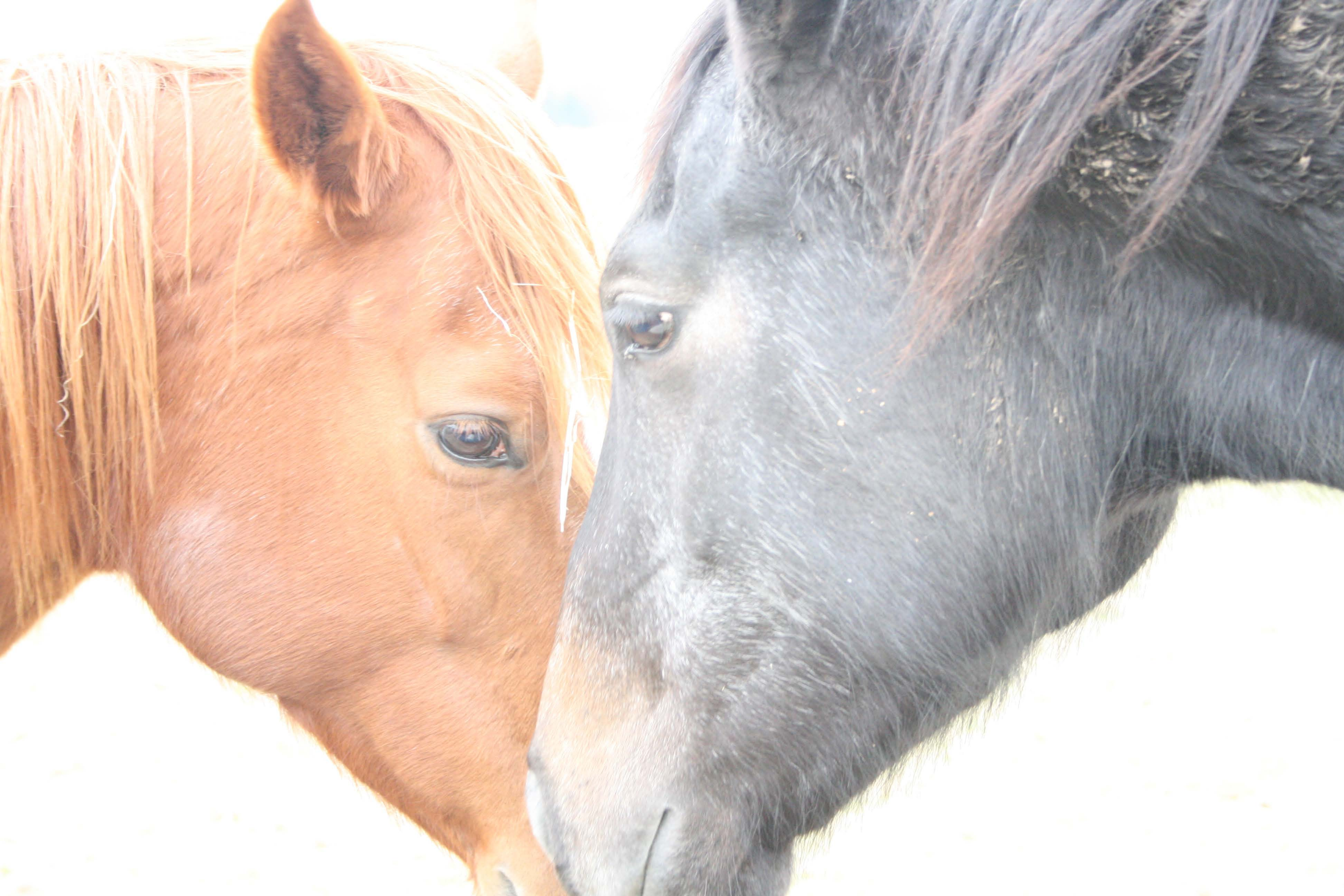}&	
			\includegraphics[width=0.19\linewidth]{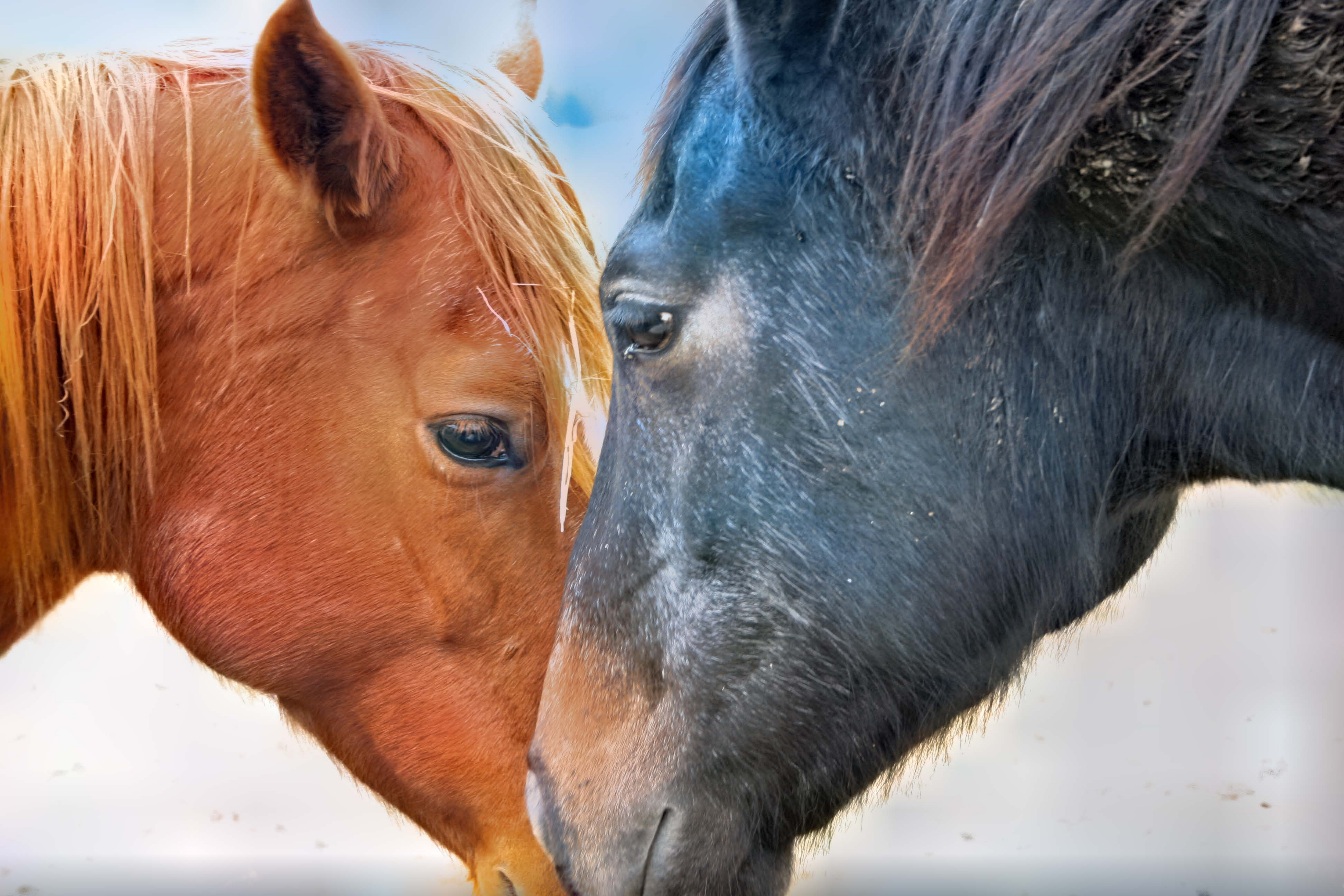}&	
			\includegraphics[width=0.19\linewidth]{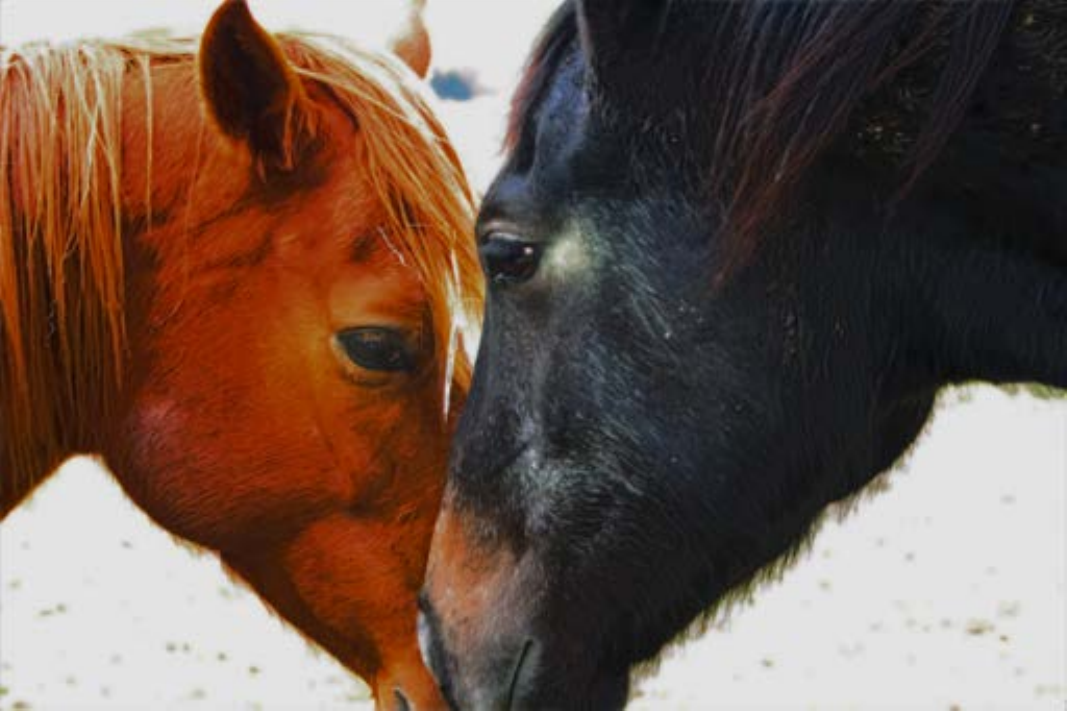}&	
			\includegraphics[width=0.19\linewidth]{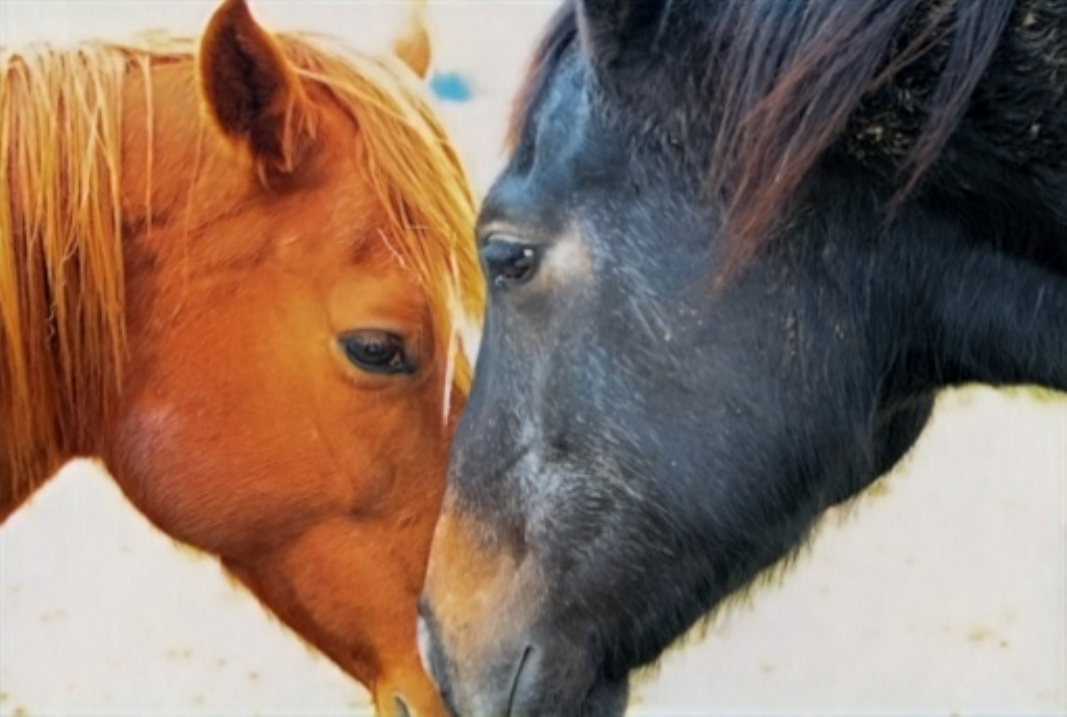}&	
			\includegraphics[width=0.19\linewidth]{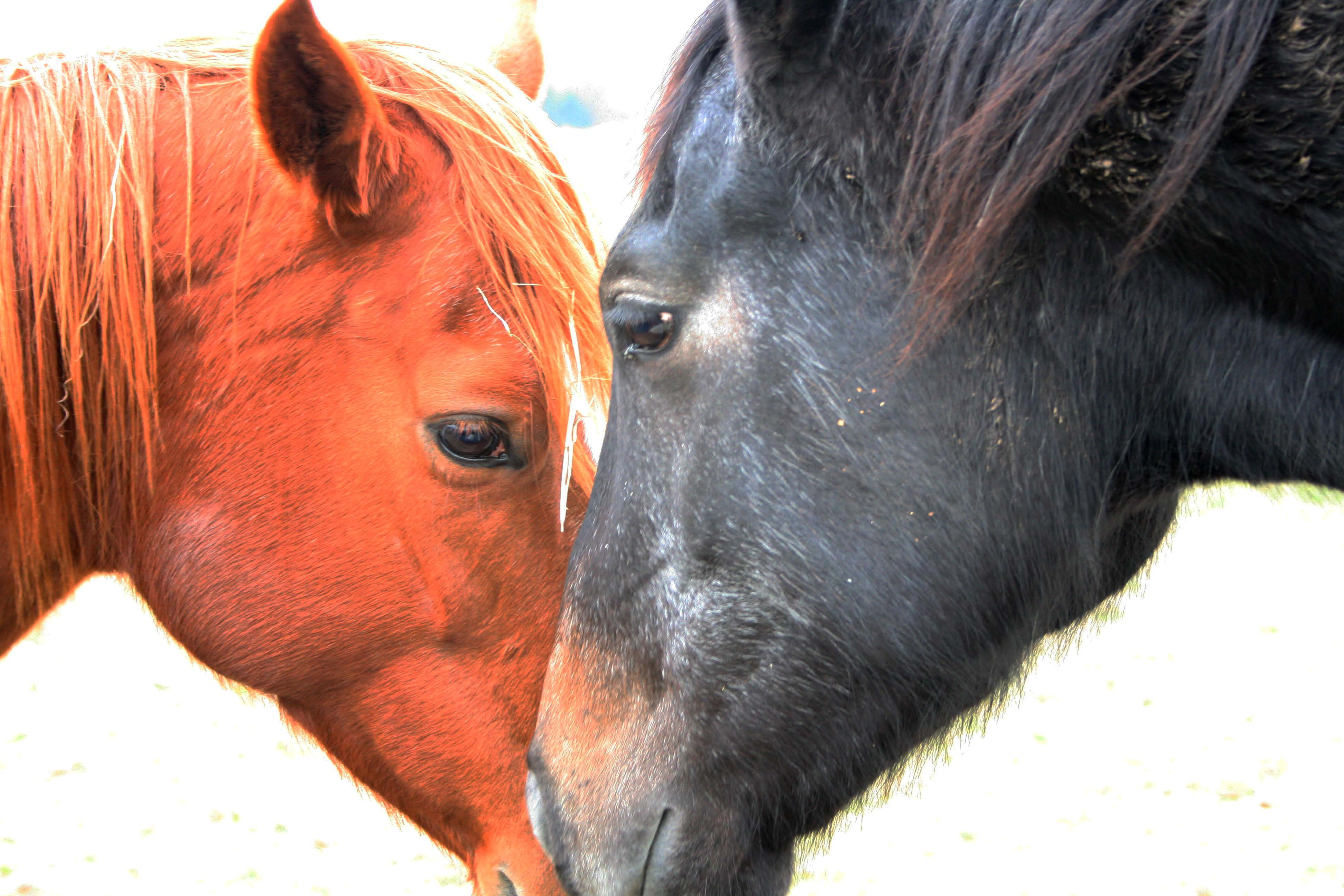}\\
			\footnotesize Input&\footnotesize MSEC~\cite{afifi2021learning}&\footnotesize LCD~\cite{wang2022local}&\footnotesize FEC~\cite{huang2022eccv}&\footnotesize PEC\\
		\end{tabular}
		\caption{Visual comparison of overexposure correction. \textit{Best viewed full screen to see details.}}
		\label{fig: Scenes_over}
	\end{figure*}

	\subsection{Comparison with traditional schemes}
	\textbf{Comparative methods, datasets, and metrics.} Model-based traditional methods are another prominent approach for underexposure correction. These methods are also learning-free, making them comparable to the PEC from the macro perspective. Therefore, it is essential to compare our PEC with these traditional methods. 
	Here, we evaluate the proposed PEC alongside nine benchmark approaches, namely WVM~\cite{fu2016weighted}, LIME~\cite{guo2017lime}, STAR~\cite{xu2020star}, NPEA~\cite{wang2013naturalness}, SRIE~\cite{fu2015probabilistic}, JIEP~\cite{cai2017joint}, RRM~\cite{li2018structure}, LR3M~\cite{ren2020lr3m}, and SDD~\cite{hao2020low}. The evaluation is performed using two familiar no-reference datasets: LIME~\cite{guo2017lime}, containing 10 degraded, low-light images captured at nighttime, and MEF~\cite{ma2015perceptual}, consisting of 17 low-light images captured in indoor and outdoor environments. To assess the quantitative scores, we utilize three no-reference metrics (DE, LOE, and NIQE).
	
	\textbf{Quantitative evaluation.}
	The results of all methods~\cite{fu2016weighted, guo2017lime, xu2020star,wang2013naturalness,fu2015probabilistic,cai2017joint,li2018structure,ren2020lr3m,hao2020low} in terms of three no-reference metrics are provided in Table~\ref{table: Score of traditional}, our method ranks within the top two in most metrics on both the MEF and LIME datasets. While achieving competitive results, PEC requires significantly less computation time compared to the existing fastest LIME, running almost 100 times faster (see Figure~\ref{fig:FirstFigure}~(a)), making it highly practical compared to the other state-of-the-art methods. 
	
	\textbf{Qualitative evaluation.} Figure~\ref{fig: Others} displays the visual results along with the difference maps for the corrected outputs generated using representative methods~\cite{fu2016weighted, guo2017lime, xu2020star,wang2013naturalness,fu2015probabilistic,cai2017joint,hao2020low} and the original input. The benchmark methods do not provide satisfying visual quality, \textit{e.g.,} see the yellow wall in the right part. Only LIME and PEC provide relatively clear structures, but LIME introduces obvious artifacts into the sky, where the difference map further verifies the performance. Simply, the developed PEC is significantly superior to other model-based traditional methods.

	\subsection{Running time}
	In the foregoing sections, we evaluated the running time and compared it with that of model-based traditional methods, showcasing the fast inference speed achieved herein. 
	Here, we perform an in-depth experimental assessment of the running time on various computational platforms composed of PC (CPU and GPU) and mobile (DSP/NPU) resources. To comprehensively evaluate the performance of the method, we consider three different image resolutions: 1280$\times$720, 1920$\times$1080, and 2560$\times$1440.
	
		\textbf{PC devices.} Table~\ref{table: PC Time} summarizes the running times of the compared methods~\cite{fu2016weighted,guo2017lime,xu2020star,wang2013naturalness,fu2015probabilistic,cai2017joint,li2018structure,ren2020lr3m,hao2020low} and our PEC on a PC equipped with a GeForce RTX 2080Ti GPU and an Intel Xeon W-2135 processor. The developed PEC is significantly and consistently faster than all the advanced networks at all resolutions. Notably, the PEC realizes a 37-FPS running speed for the image with the size of 1280~$\times$~720 on the CPU platform, which satisfies the real-time requirement (more than 20 FPS), whereas the other methods fail to accomplish this. Moreover, the PEC requires only 0.0009 s, which is equal to 1111 FPS for a video with 2K resolution, and far surpasses the frame rate of the human eye.

	\textbf{Mobile devices.} We further evaluated the running time for the top-three fastest methods (including SCI~\cite{ma2022toward}, ZeroDCE~\cite{guo2020zero}, and RUAS~\cite{liu2021retinex}) in Table~\ref{table: PC Time} and the PEC using two representative mobiles (XIAOMI 10 with Snapdragon 865 DSP and HUAWEI MATE 30 PRO with Kirin 990 NPU). AI Benchmark version 5.0.1 was utilized for evaluating the algorithms on the mobile devices. 
	As shown in Table~\ref{table: Mobile Time}, the PEC obtains an average improvement of 62\% over that of the second-best method, SCI. The results further verify the practicality of this approach for terminal mobiles.

	\begin{figure*}[t]
		\centering
		\footnotesize
		\begin{tabular}{c}		
			\vspace{-0.1cm}		
			\includegraphics[width=0.95\linewidth]{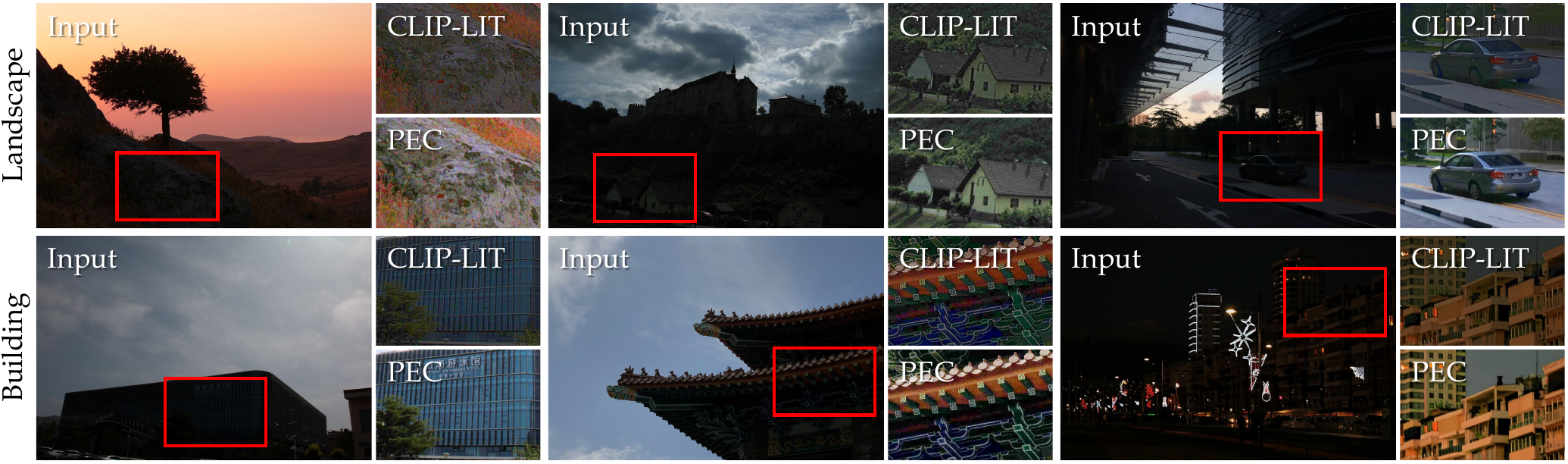}\\		
		\end{tabular}
		\caption{Evaluation of enhancement effects in different backlit images by comparing PEC with the recent CLIP-LIT~\cite{liang2023iterative}.}
		\label{fig: backlit}
	\end{figure*}
	\begin{figure*}[t]
		\centering
		\begin{tabular}{c@{\extracolsep{0.2em}}c@{\extracolsep{0.2em}}c@{\extracolsep{0.2em}}c@{\extracolsep{0.2em}}c}	
			\vspace{-0.1cm}	
			\includegraphics[width=0.185\linewidth]{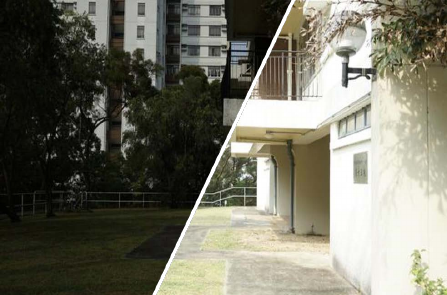}&		
			\includegraphics[width=0.185\linewidth]{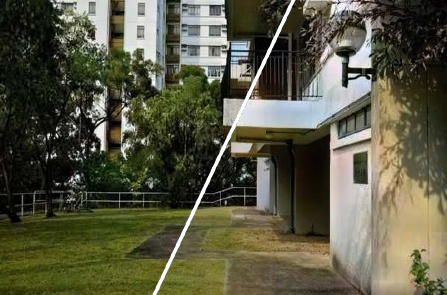}&
			\includegraphics[width=0.185\linewidth]{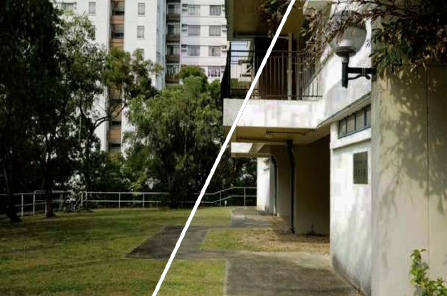}&
			\includegraphics[width=0.185\linewidth]{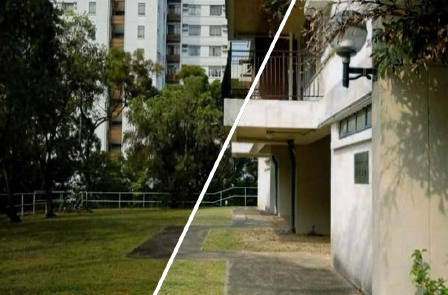}&
			\includegraphics[width=0.185\linewidth]{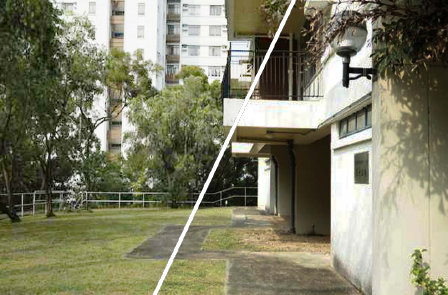}\\
			\vspace{-0.2cm}
			\includegraphics[width=0.185\linewidth]{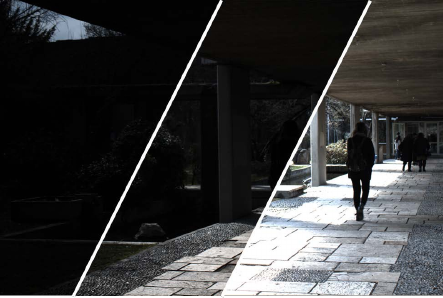}&		
			\includegraphics[width=0.185\linewidth]{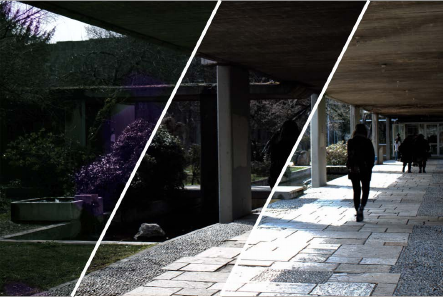}&
			\includegraphics[width=0.185\linewidth]{Figures/VisualComparisons/Consistency/2/LCD}&
			\includegraphics[width=0.185\linewidth]{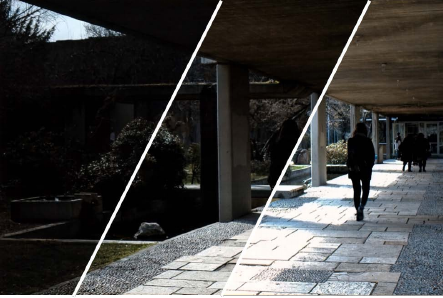}&
			\includegraphics[width=0.185\linewidth]{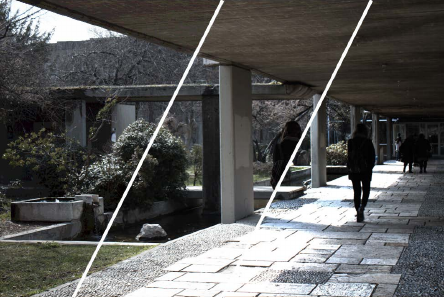}\\
			\footnotesize Input&\footnotesize MSEC~\cite{afifi2021learning}&\footnotesize LCD~\cite{wang2022local}&\footnotesize FEC~\cite{huang2022eccv}&\footnotesize PEC\\
		\end{tabular}
		\caption{Comparison of exposure consistency for scenes with different exposure levels using different exposure correctors.}
		\label{fig: Consistency}
	\end{figure*}

	\subsection{Scene adaptability}
	To fully evaluate the scene adaptability, we consider several challenging examples from different real-world datasets, including underexposure, overexposure, and backlit cases.
	
	\textbf{Underexposure scenes.}
	Here, we conduct an in-depth analysis of challenging low-exposure scenarios, followed by a discussion of the necessity of introducing post-processing as a solution. Finally, we present a series of quantitative and qualitative comparison results for performance evaluation. In real-world environments with insufficient exposure, unwanted degradations often arise, in addition to improper lighting. In fact, such challenging underexposure scenarios can generally be categorized into two types: those with noise and those with artifacts. The former is primarily caused by the limitations of imaging sensors or the amplification of inherent noise signals due to increased ISO to compensate for under-exposure. The latter mainly arises from image compression algorithms (\textit{e.g.,} JPEG), in-camera processing algorithms, and lens imperfections. These two types of scenarios present significant challenges in the field.
	
	Figure~\ref{fig: Analysis} presents the results obtained with the developed method and a series of representative approaches (\textit{i.e.,} SCI~\cite{ma2022toward}, LIME~\cite{guo2017lime}, FEC~\cite{huang2022eccv}, and RUAS\cite{liu2021retinex}) under these two challenging scenarios. All methods struggle to effectively address the inherent degradations of the images on their own, underscoring the necessity of incorporating post-processing operations (\textit{i.e.,} denoising and artifact removal). Importantly, the core contribution of this work lies in addressing degradation caused by improper brightness, rather than designing additional image quality enhancement solutions for specific scenarios. Instead, we directly employ widely-used methods in the field for post-processing to optimize the visual presentation. Based on the above analysis, we conducted a more systematic performance evaluation in these two challenging scenarios by employing SCUNet~\cite{zhang2023practical} for denoising and FBCNN~\cite{jiang2021towards} for artifact removal~\cite{lin2020learning}.
	
	\begin{table*}[t]
		\footnotesize
		\renewcommand\arraystretch{1.2} 
		\setlength{\tabcolsep}{0.2mm}
		\centering
		\caption{Comparison of performance in other vision tasks. We retrained the detector/segmentator containing the enhancer.} 
		\begin{tabular}{|c|c|c|c|c|c|c|c|c|c|c|c|}
			\hline
			&\multicolumn{3}{c|}{\textit{Dark Face Detector}}&\multicolumn{8}{c|}{\textit{Enhancer + Detector (Finetune)}}\\
			\hline 
			Method&HLA~\cite{wang2022unsupervised}&REG~\cite{liang2021recurrent}&MAET~\cite{cui2021multitask}&LIME~\cite{guo2017lime}&ZeroDCE~\cite{guo2020zero}&MSEC~\cite{afifi2021learning}&RUAS~\cite{liu2021retinex}&LCD~\cite{wang2022local}&FEC~\cite{huang2022eccv}&SCI~\cite{ma2022toward}&PEC\\
			\hline 
			mAP&0.607&0.514&0.526&0.644&{\underline{0.665}}&0.659&0.642&0.654&0.603&0.663&{\textbf{0.677}}\\
			\hline
			\hline
			&\multicolumn{3}{c|}{\textit{Nighttime Semantic Segmentator}}&\multicolumn{8}{c|}{\textit{Enhancer + Segmentator (Finetune)}}\\
			\hline 
			Method&DANNet~\cite{wu2021one}&CIC~\cite{lengyel2021zero}&GPS-GLASS~\cite{lee2022gps}&LIME~\cite{guo2017lime}&ZeroDCE~\cite{guo2020zero}&MSEC~\cite{afifi2021learning}&RUAS~\cite{liu2021retinex}&LCD~\cite{wang2022local}&FEC~\cite{huang2022eccv}&SCI~\cite{ma2022toward}&PEC\\
			\hline
			mIoU&0.398&0.264&0.380&0.397&{\underline{0.399}}&0.377&0.363&0.381&0.375&0.393&{\textbf{0.416}}\\
			\hline
		\end{tabular}
		\label{table: Finetune}
	\end{table*}
	
	\begin{figure*}[t]
		\centering
		\begin{tabular}{c@{\extracolsep{0.15em}}c@{\extracolsep{0.15em}}c@{\extracolsep{0.15em}}c@{\extracolsep{0.15em}}c}
			\vspace{-0.2cm}
			\includegraphics[width=0.19\linewidth]{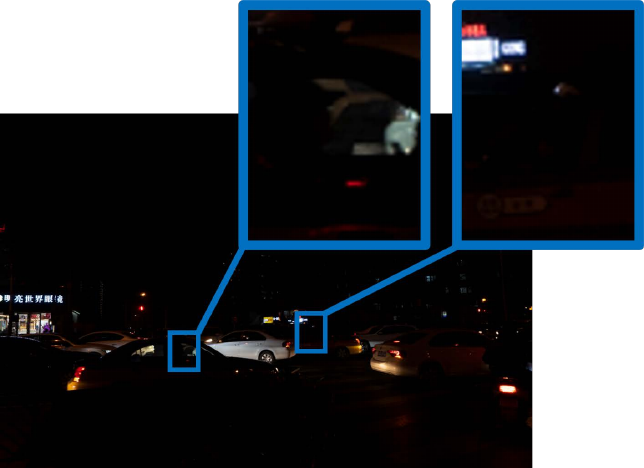}&
			\includegraphics[width=0.19\linewidth]{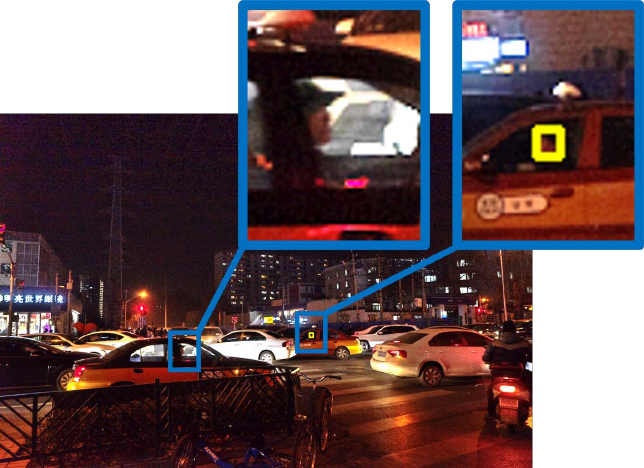}&
			\includegraphics[width=0.19\linewidth]{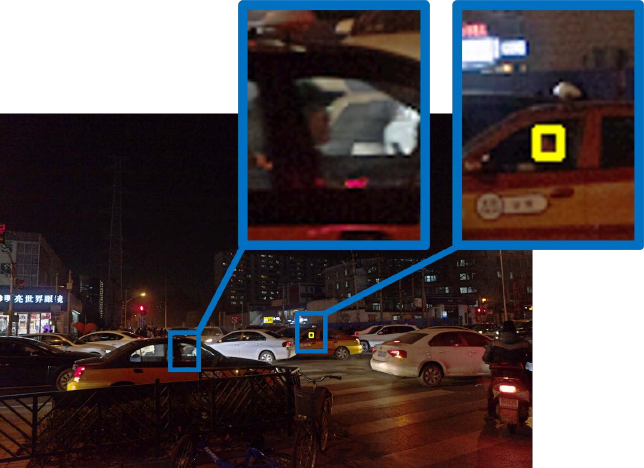}&
			\includegraphics[width=0.19\linewidth]{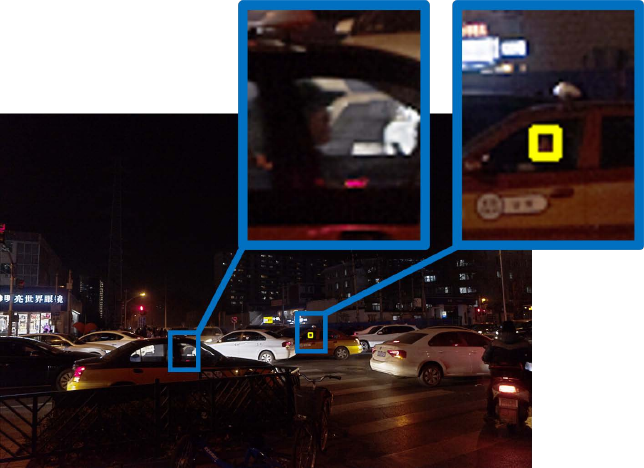}&
			\includegraphics[width=0.19\linewidth]{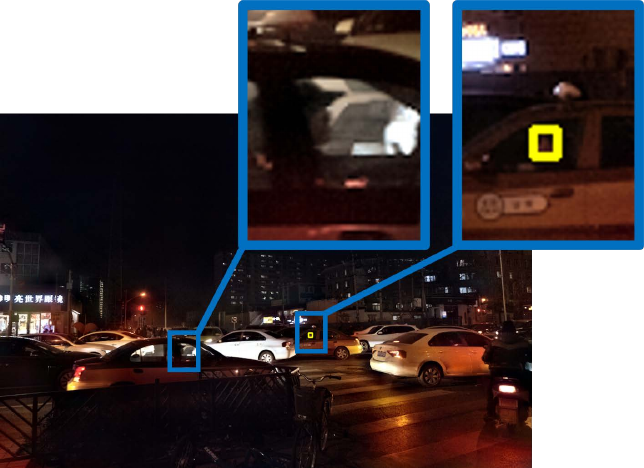}\\
			\footnotesize Input &	\footnotesize LIME~\cite{guo2017lime} &\footnotesize ZeroDCE~\cite{guo2020zero} &\footnotesize SCI~\cite{ma2022toward} &\footnotesize MSEC~\cite{afifi2021learning}\\
			\vspace{-0.2cm}
			\includegraphics[width=0.19\linewidth]{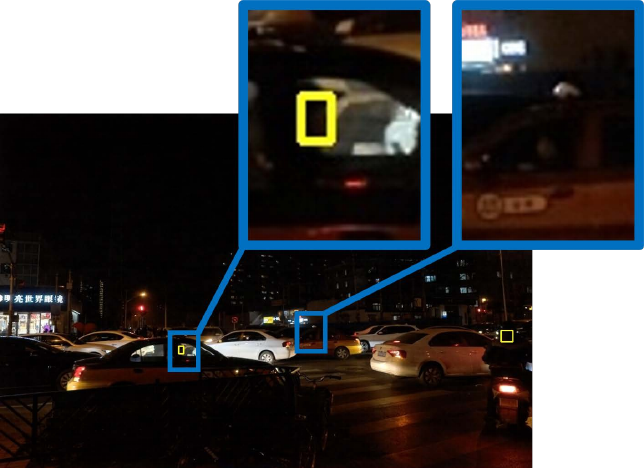}&
			\includegraphics[width=0.19\linewidth]{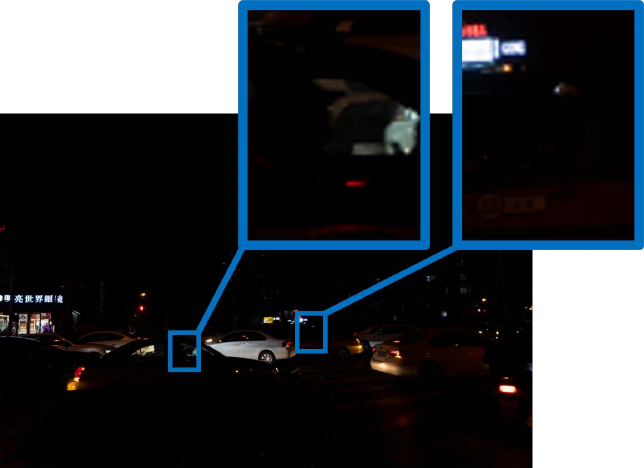}&
			\includegraphics[width=0.19\linewidth]{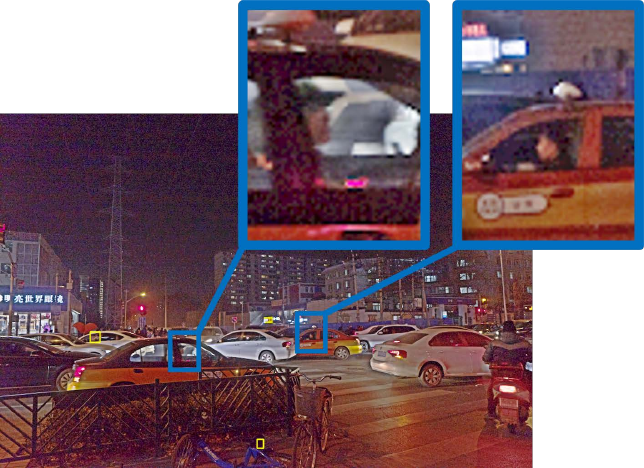}&
			\includegraphics[width=0.19\linewidth]{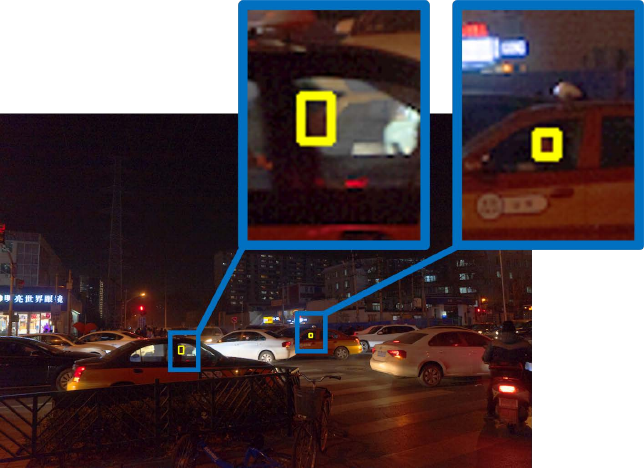}&
			\includegraphics[width=0.19\linewidth]{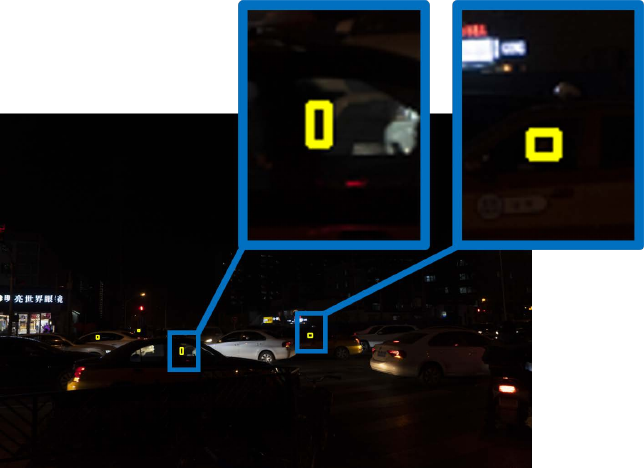}\\
			\footnotesize FEC~\cite{huang2022eccv} &	\footnotesize MAET~\cite{cui2021multitask} &\footnotesize HLA~\cite{wang2022unsupervised} &\footnotesize Ours &\footnotesize Label\\
		\end{tabular}
		\caption{Visual comparison for dark face detection.}
		\label{fig: detection}
	\end{figure*}
	
	\begin{figure*}[!htb]
		\centering
		\begin{tabular}{c@{\extracolsep{0.15em}}c@{\extracolsep{0.15em}}c@{\extracolsep{0.15em}}c@{\extracolsep{0.15em}}c}
			\vspace{-0.2cm}
			\includegraphics[width=0.19\linewidth]{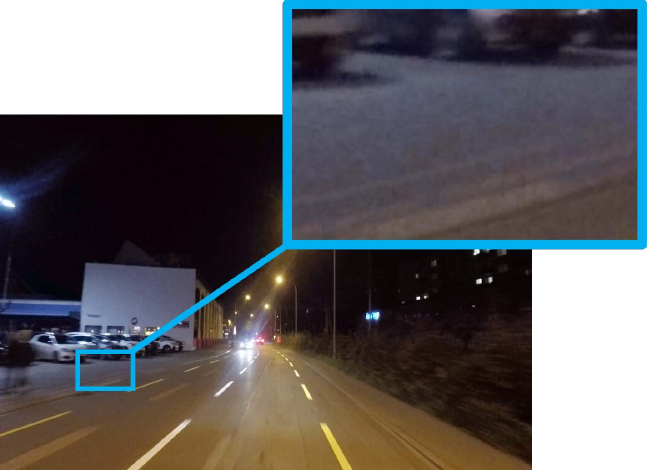}&
			\includegraphics[width=0.19\linewidth]{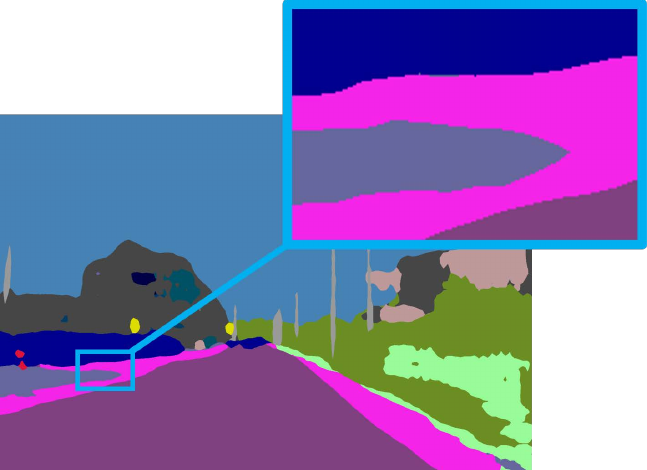}&
			\includegraphics[width=0.19\linewidth]{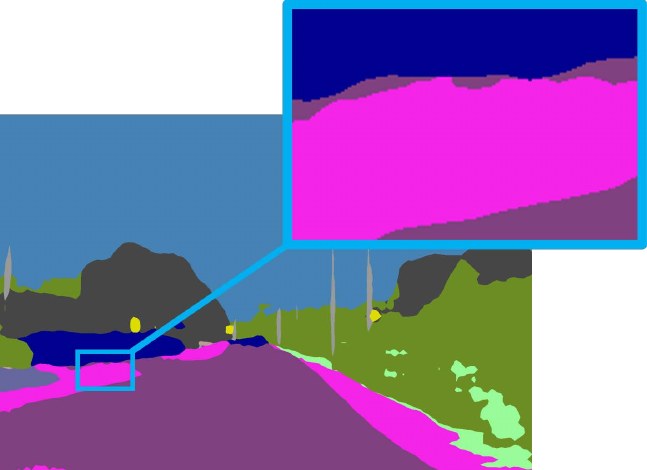}&
			\includegraphics[width=0.19\linewidth]{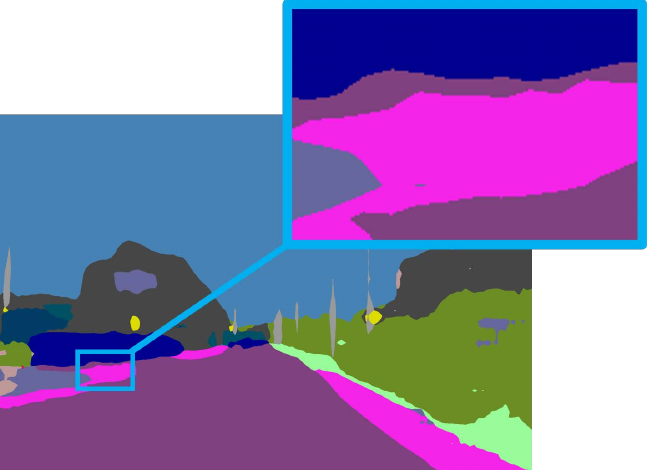}&
			\includegraphics[width=0.19\linewidth]{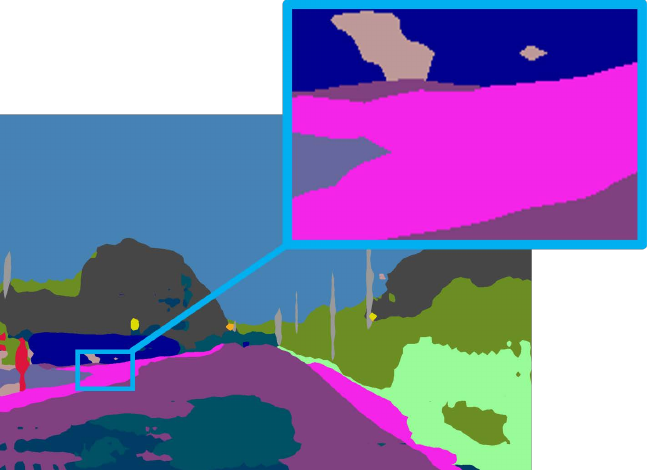}\\
			\footnotesize Input &	\footnotesize LIME~\cite{guo2017lime} &\footnotesize ZeroDCE~\cite{guo2020zero} &\footnotesize SCI~\cite{ma2022toward} &\footnotesize MSEC~\cite{afifi2021learning}\\
			\vspace{-0.2cm}
			\includegraphics[width=0.19\linewidth]{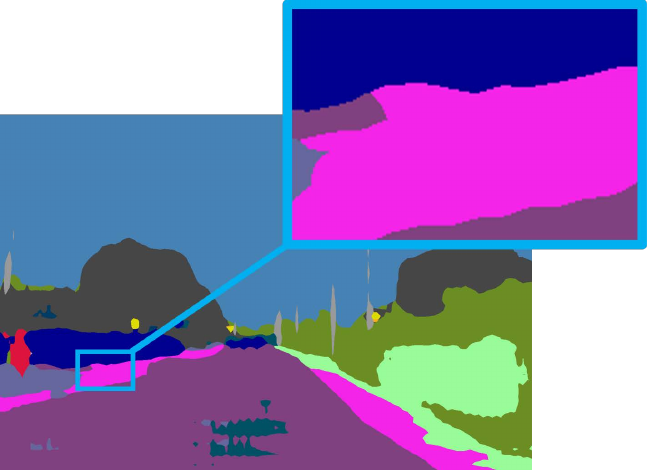}&
			\includegraphics[width=0.19\linewidth]{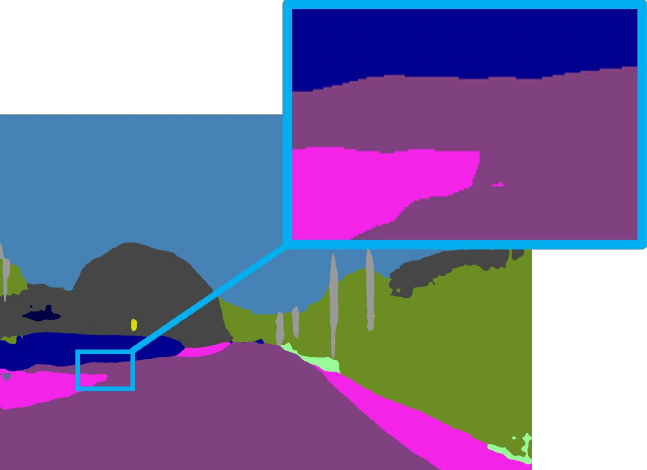}&
			\includegraphics[width=0.19\linewidth]{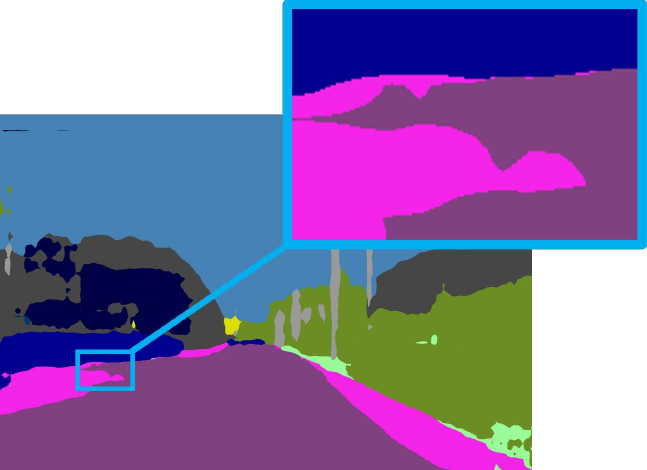}&
			\includegraphics[width=0.19\linewidth]{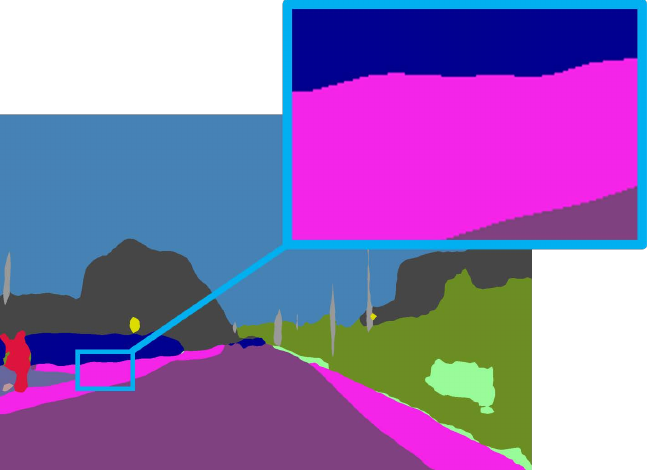}&
			\includegraphics[width=0.19\linewidth]{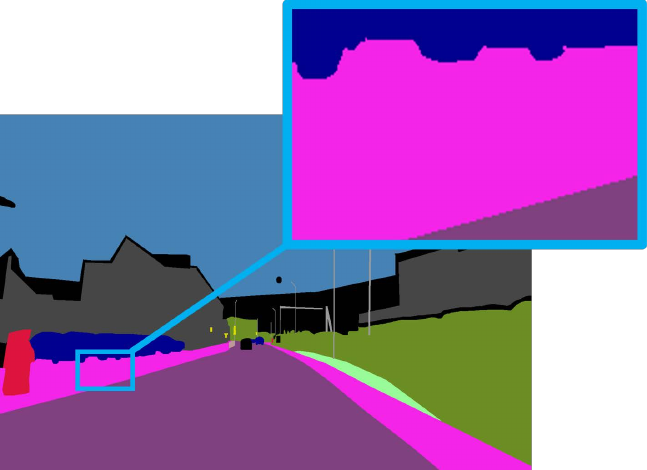}\\
			\footnotesize FEC~\cite{huang2022eccv} &	\footnotesize GPS-GLASS~\cite{lee2022gps} &\footnotesize DANNet~\cite{wu2021one} &\footnotesize Ours &\footnotesize Label\\
		\end{tabular}
		\caption{Visual comparisons for nighttime semantic segmentation.}
		\label{fig: segmentation}
	\end{figure*}
	
	Table~\ref{tab: B1} presents a comparison of the performance with different methods ~\cite{ma2022toward,guo2017lime,huang2022eccv, liu2021retinex} across two challenging scenarios using the LOE~\cite{wang2013naturalness}, DE~\cite{ye2007discrete}, and NIQE~\cite{mittal2012making} as evaluation metrics. For this evaluation, by collecting and reorganizing existing datasets (\textit{i.e.,} LSRW~\cite{hai2023r2rnet} and ExDARK~\cite{loh2019getting}), we constructed two small-scale datasets\footnote{We have uploaded them to our project page.} for evaluating the performance in the aforementioned challenging scenarios: one consisting of 40 underexposure images with noise and the other containing 40 underexposure images with artifacts. 
	Comparison of the numerical results in Table~\ref{tab: B1} clearly shows that for both challenging scenarios, the developed method outperforms the benchmark methods across all metrics. We present the relevant visual comparison results in Figures~\ref{fig: Scenes_under1} and~\ref{fig: Scenes_under2}, demonstrating that the PEC performs better in terms of the exposure control and color restoration across all challenging scenarios. Compared to the other methods, the PEC achieves more uniform brightness in the correction results, effectively avoiding issues such as overexposure or underexposure in localized regions. Additionally, the PEC achieves more natural color restoration, accurately recovering the true colors of objects and avoiding color distortion.

	\textbf{Overexposure scenes.}
	Figure~\ref{fig: Scenes_over} shows the visual results with three state-of-the-art methods~\cite{afifi2021learning, wang2022local, huang2022eccv} for the overexposure cases.  The observations in the top two rows are from the UFDD dataset\cite{nada2018pushing}, while the observations in the bottom two rows are sourced from the free stock photo website Pexels\footnote{\url{https://www.pexels.com/}.}.
	Because most of the methods mentioned above are designed for underexposure cases, here, we focus on comparing the PEC with three general correctors, namely MSEC, LCD, and FEC. All methods lead to unknown artifacts and color distortion to varying degrees. However, the PEC achieves the best visual appearance, particularly in cases where it provides appropriate correction for some regions that are originally overexposed.
	
	\textbf{Backlit scenes.} Gratifyingly, the PEC can effectively recover objects in backlit scenes, as demonstrated in Figure~\ref{fig: Scenes_under2}. Despite the challenging lighting conditions, the developed method performs well and manages to restore details and objects in the images. Importantly, the task of handling backlit scenes has been independently studied by other researchers in recent years~\cite{lv2022backlitnet,liang2023iterative}. In this context, we aim to further evaluate the effectiveness of the PEC by comparison with the latest development in this area, namely CLIP-LIT~\cite{liang2023iterative}. Figure~\ref{fig: backlit} shows visualizations of different examples from the constructed dataset presented in~\cite{liang2023iterative}. Notably, the proposed PEC is capable of recovering more details and producing vivid colors compared to CLIP-LIT.

	\begin{figure*}[t]
		\centering
		\footnotesize
		\begin{tabular}{c@{\extracolsep{0.3em}}c@{\extracolsep{0.3em}}c@{\extracolsep{0.3em}}c}			
			\vspace{-0.1cm}
			\includegraphics[height=0.21\linewidth]{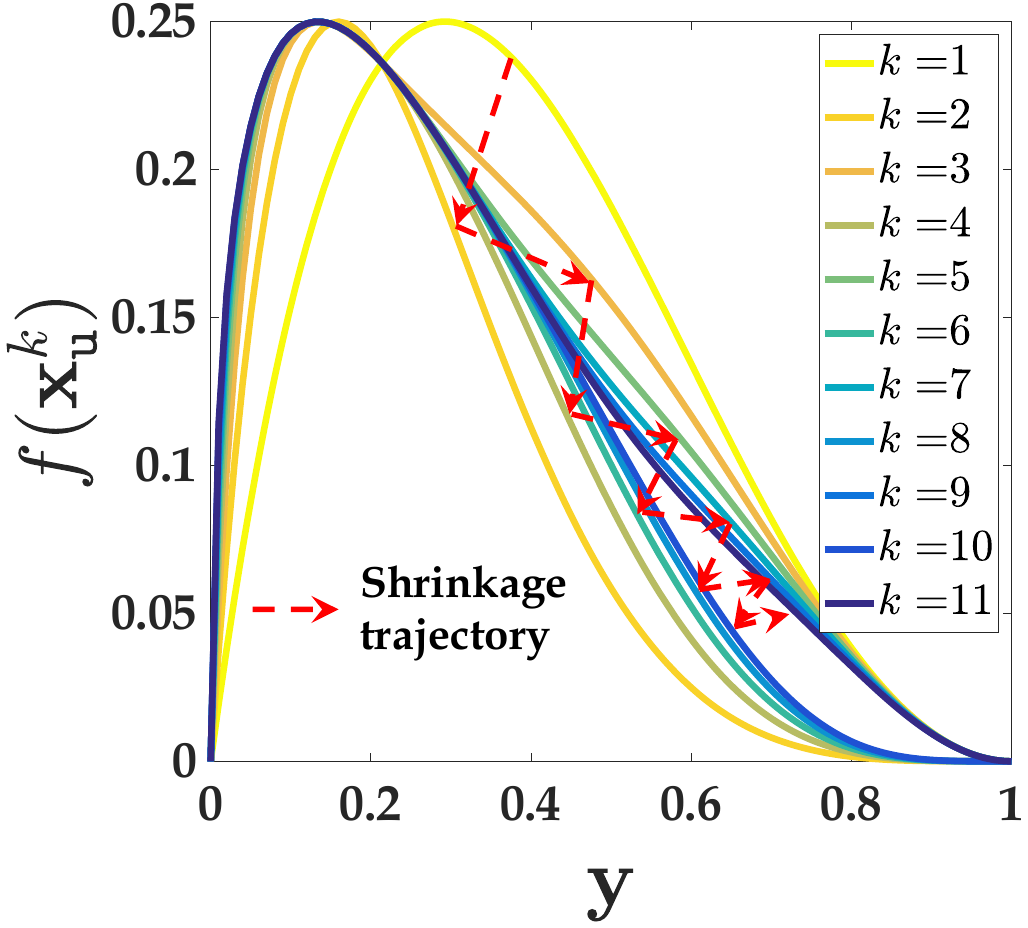}&
			\includegraphics[height=0.21\linewidth]{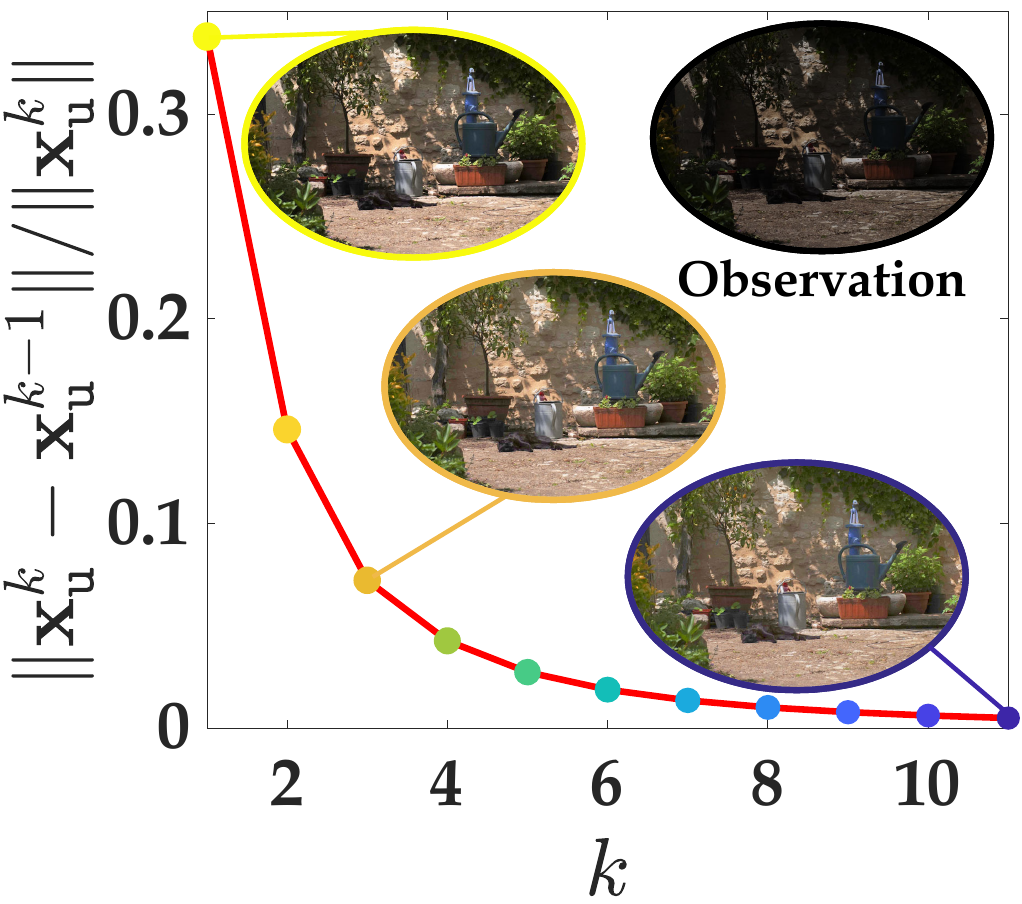}&
			\includegraphics[height=0.21\linewidth]{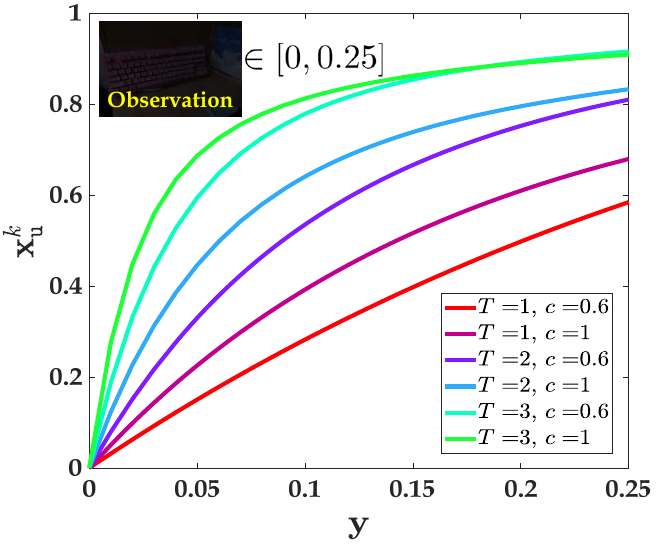}&
			\includegraphics[height=0.21\linewidth]{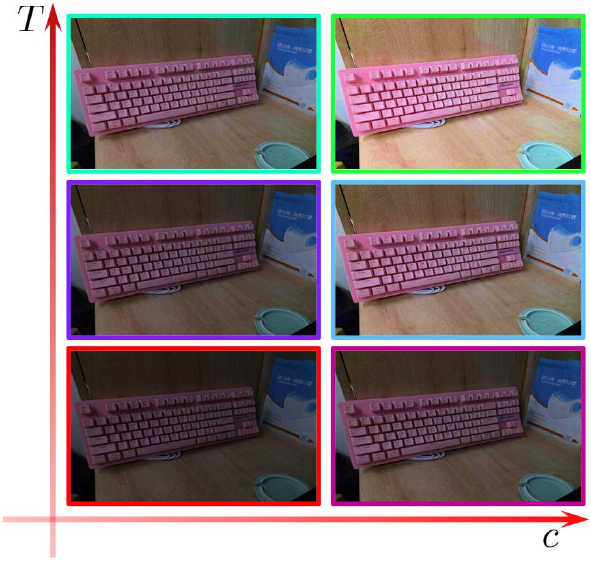}\\
			(a) \textit{Toy}: $f(\mathbf{x}_\mathtt{u}^{k})$ v.s. $\mathbf{y}$&(b) \textit{Practical}: $\frac{\|\mathbf{x}_\mathtt{u}^{k}-\mathbf{x}_\mathtt{u}^{k-1}\|}{\|\mathbf{x}_\mathtt{u}^{k}\|}$ v.s. $k$ & (c) Curves of different settings& (d) Outputs for cases in (c)\\
		\end{tabular}
		\caption{Algorithmic analyses. (a)–(b) Plot of convergence parameters for proposed PEC. (Left) Plot of convergence parameters for a toy example; (Right) plot of relative errors and intermediate results, where the underexposed observation $\mathbf{y}\in\mathbb{U}$ is shown in the right top corner of (b). (c)–(d) Analysis for iteration $T$ and coefficient $c$ parameters.}
		\label{fig: Convergence}
	\end{figure*}	
	
	\subsection{Exposure consistency}\label{sec: exposureconsistency}
	To further explore the stability and consistency of the PEC when faced with different exposure levels, here, we provide visual comparisons with methods~\cite{afifi2021learning,wang2022local,huang2022eccv} for various examples. As shown in Figure~\ref{fig: Consistency}, the top two observations at two exposure levels are from the SICE dataset~\cite{cai2018Learning} and the bottom two observations, each with three exposure levels, are from Tursun’s dataset~\cite{tursun2016objective}. 
	The PEC consistently outputs harmonious content with consistent exposure across various scenes, setting it apart from other state-of-the-art approaches.

	\subsection{Other vision applications}
	Image quality is typically assessed~\cite{ma2022toward,guo2020zero} by utilizing the enhanced results to serve downstream vision tasks. Thus, we evaluate the perception performance in two tasks (dark face detection and nighttime semantic segmentation).  
	We use the DARKFACE~\cite{yang2020advancing} and ACDC~\cite{sakaridis2021acdc} datasets for evaluating the detection and segmentation, respectively, and we follow the data partition presented in~\cite{ma2022toward}. To ensure better adaptation, we finetune all the vision models using the corrected outputs. The number of epochs for detection and segmentation is set to 10 and 30, respectively. 
	We employ the SGD optimizer for these two tasks. The momentum, weight decay, and learning rate are set as \{0.9, 3$\times 10^{-4}$, 5$\times 10^{-4}$\}, and \{0.9, 5$\times 10^{-4}$, 2.5$\times 10^{-4}$\} for detection and segmentation, respectively. 
	
	To fully evaluate the performance, we compare a part of the representative correction methods~\cite{guo2017lime,guo2020zero,afifi2021learning,liu2021retinex,wang2022local,huang2022eccv,ma2022toward} and also consider methods specifically designed for detection (HLA~\cite{wang2022unsupervised}, REG~\cite{liang2021recurrent}, and MAET~\cite{cui2021multitask}) and segmentation (DANNet~\cite{wu2021one}, CIC~\cite{lengyel2021zero}, GPS-GLASS~\cite{lee2022gps}). 	
	As shown in Table~\ref{table: Finetune}, the PEC achieves the highest objective scores compared to the other methods Figure~\ref{fig: detection} and Figure~\ref{fig: segmentation} present visual comparisons of the detection and segmentation outcomes, respectively. The PEC can detect more small objects and provide a structure-aware segmentation map. These experiments further highlight the practicality of this approach in downstream vision tasks.

	\begin{figure*}[t]
		\centering
		\footnotesize
		\begin{tabular}{c@{\extracolsep{0.2em}}c@{\extracolsep{0.2em}}c@{\extracolsep{0.4em}}c@{\extracolsep{0.2em}}c@{\extracolsep{0.2em}}c}
			\vspace{-0.1cm}
			\includegraphics[width=0.155\linewidth]{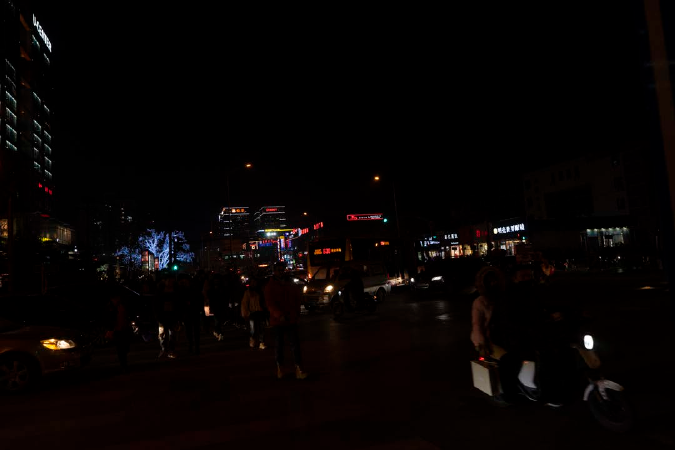}&	
			\includegraphics[width=0.155\linewidth]{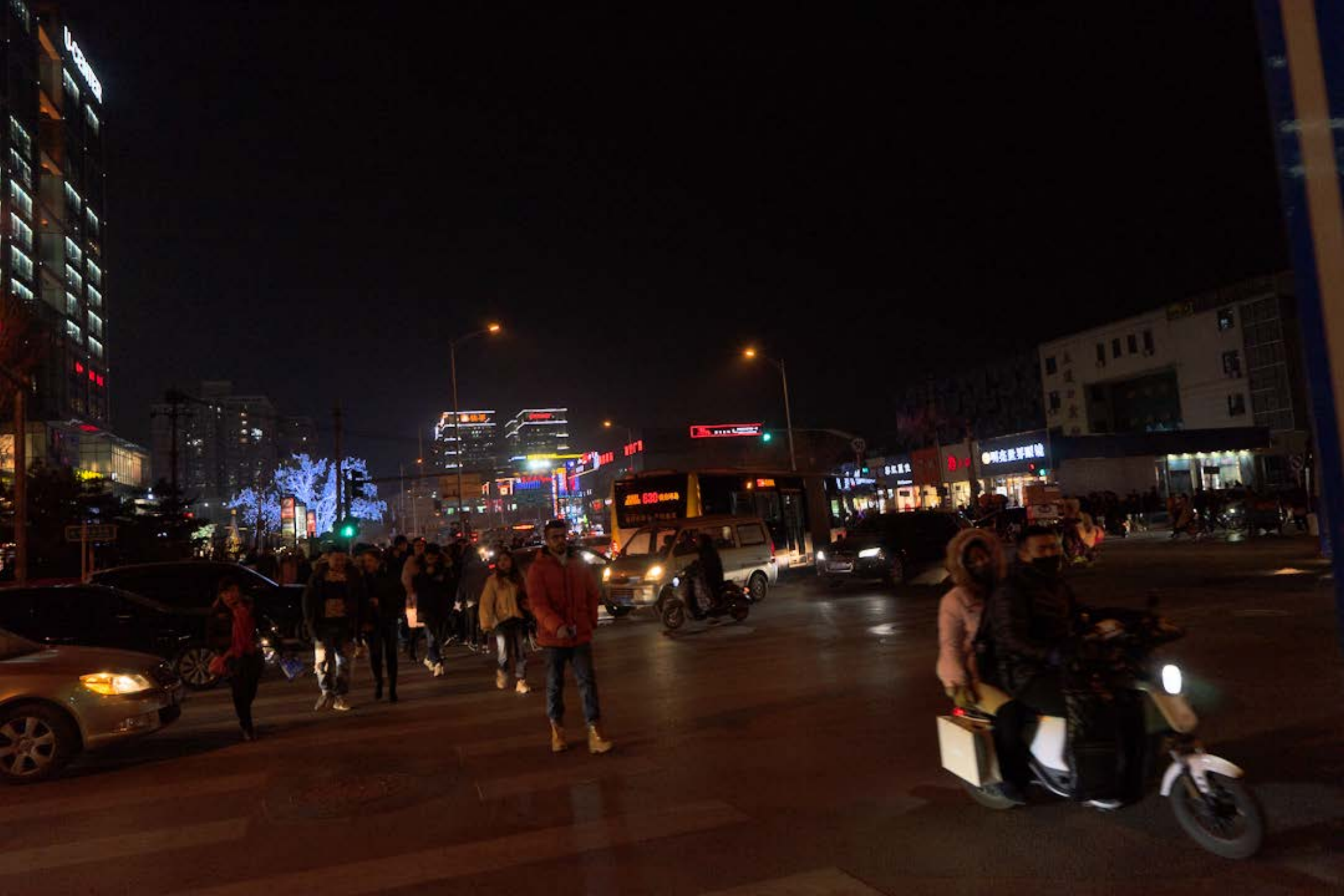}&		
			\includegraphics[width=0.155\linewidth]{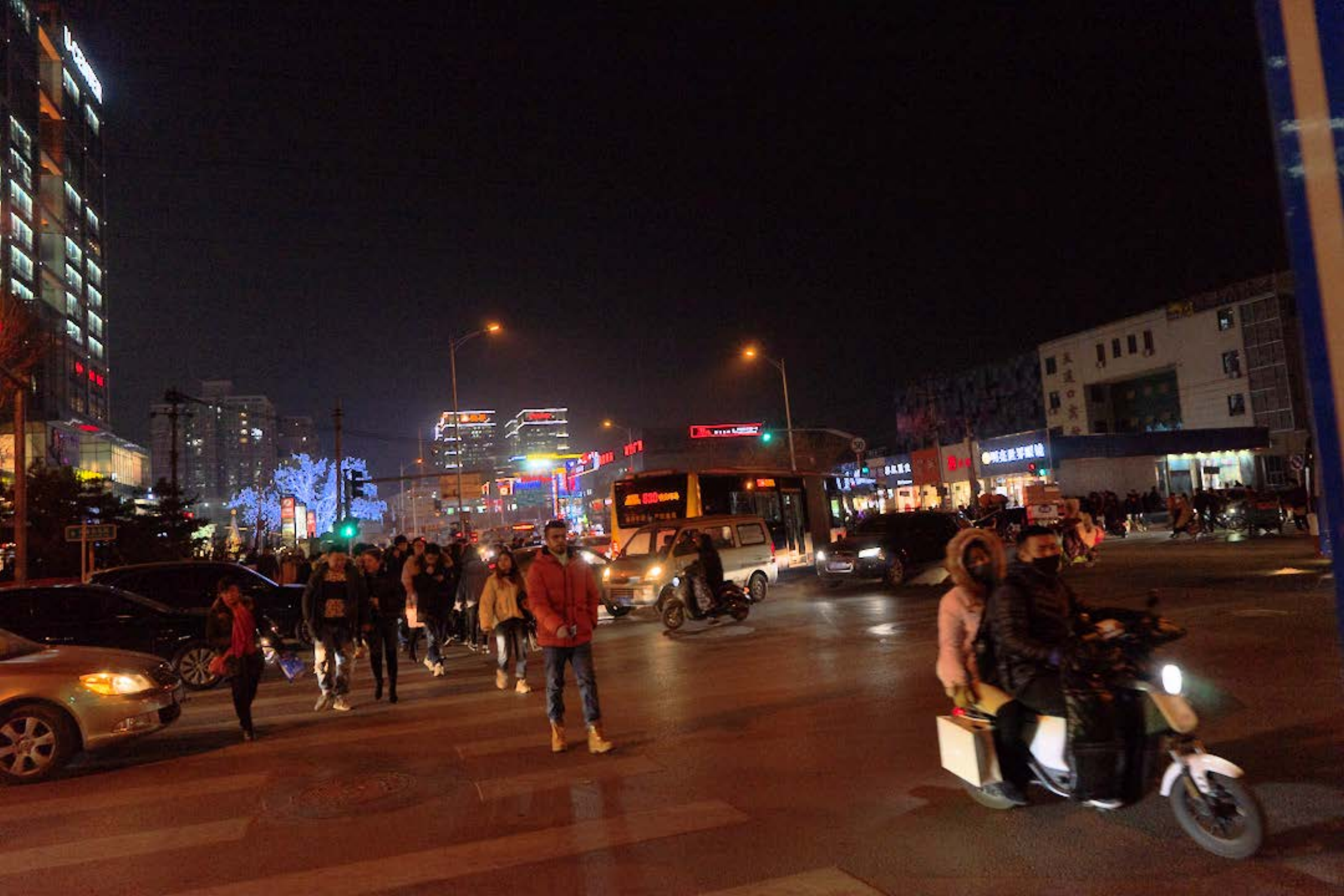}&
			\includegraphics[width=0.155\linewidth]{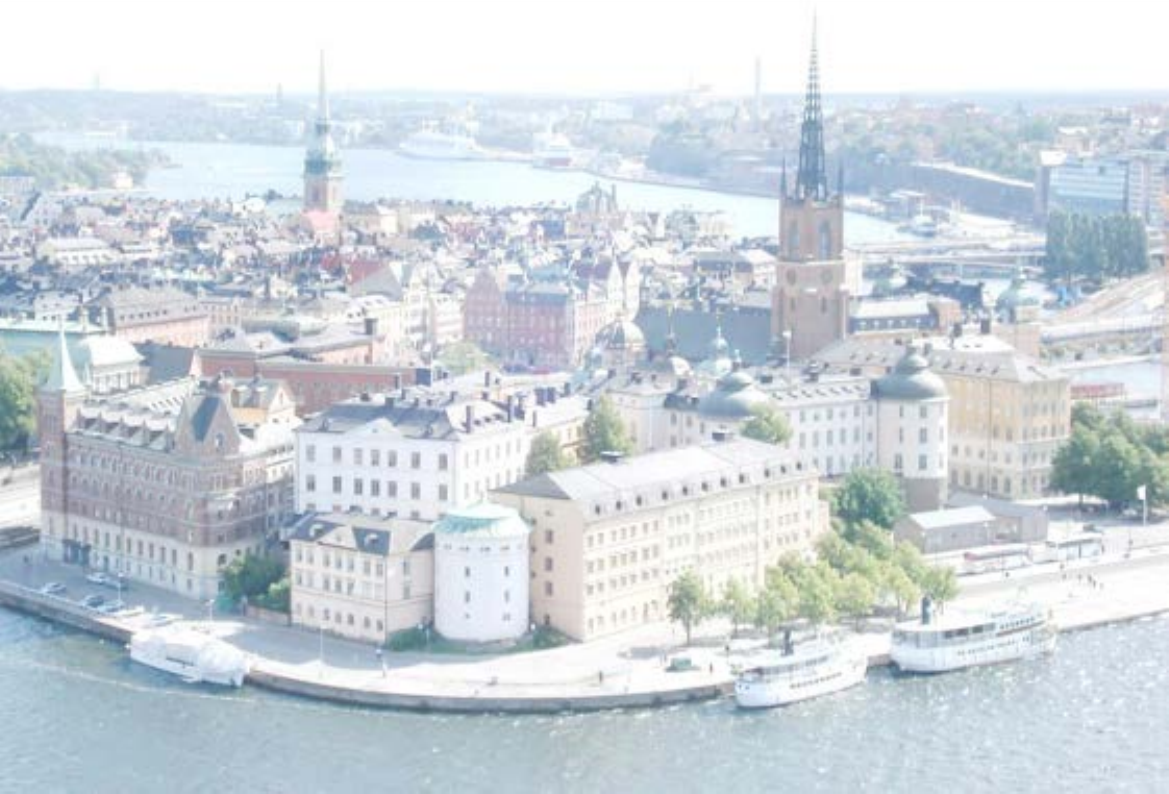}&	
			\includegraphics[width=0.155\linewidth]{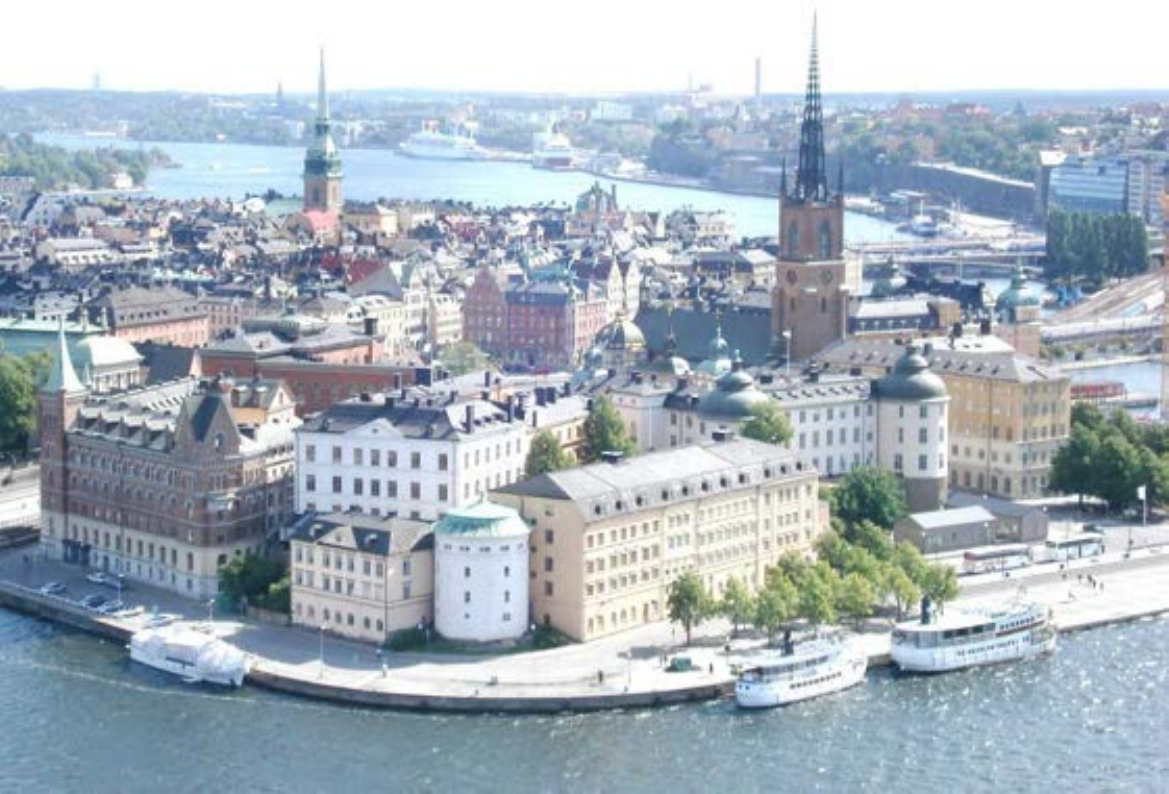}&
			\includegraphics[width=0.155\linewidth]{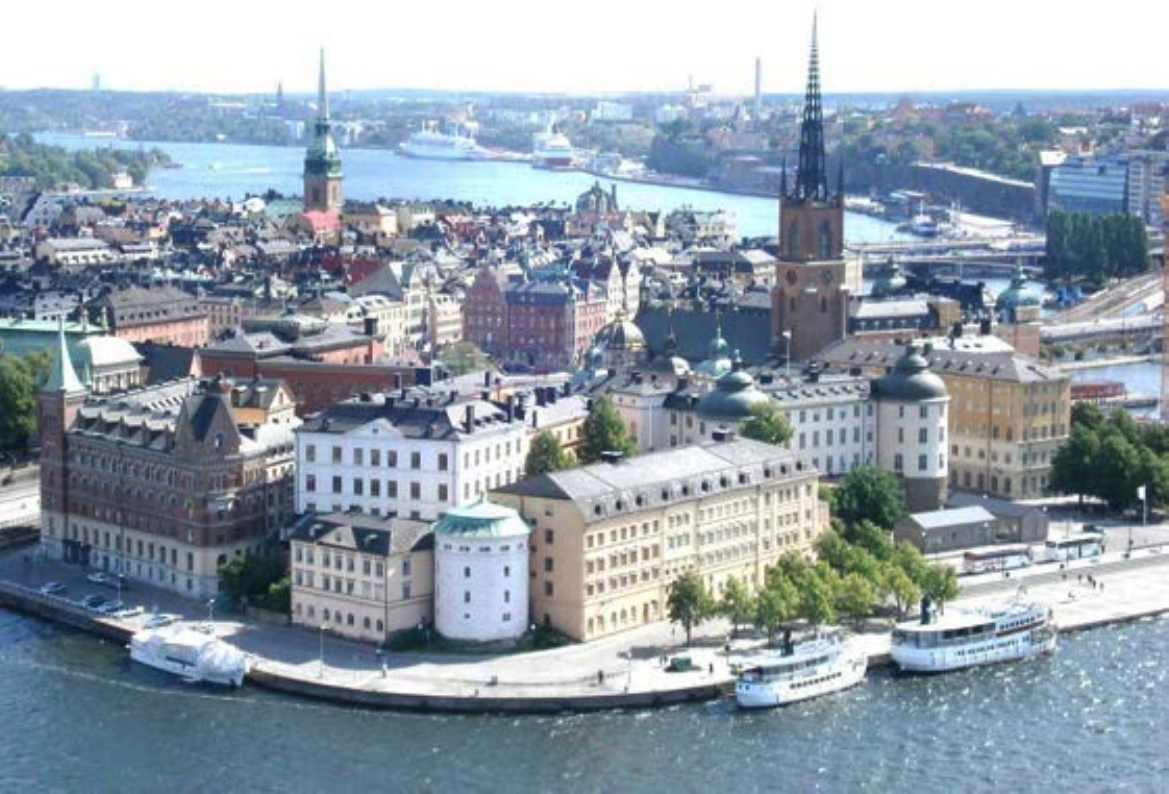}\\
			Time (S)$\downarrow$/DE$\uparrow$& 0.00018/5.8018 & 0.00026/6.3611 &
			Time (S)$\downarrow$/DE$\uparrow$& 0.00007/7.1429& 0.00009/7.3219\\
			Input & w/o Warm Start&w/ Warm Start&Input & w/o Warm Start&w/ Warm Start\\
		\end{tabular}
		\caption{Ablation study of the effect of the warm start. We report the visual results after removing the warm start in both underexposure and overexposure scenarios. The corresponding running times and performance evaluation metrics are also provided below each image.}
		\label{fig: A3}
	\end{figure*}
	
	\begin{figure*}[t]
		\centering
		\begin{tabular}{c}
			\includegraphics[width=0.98\linewidth]{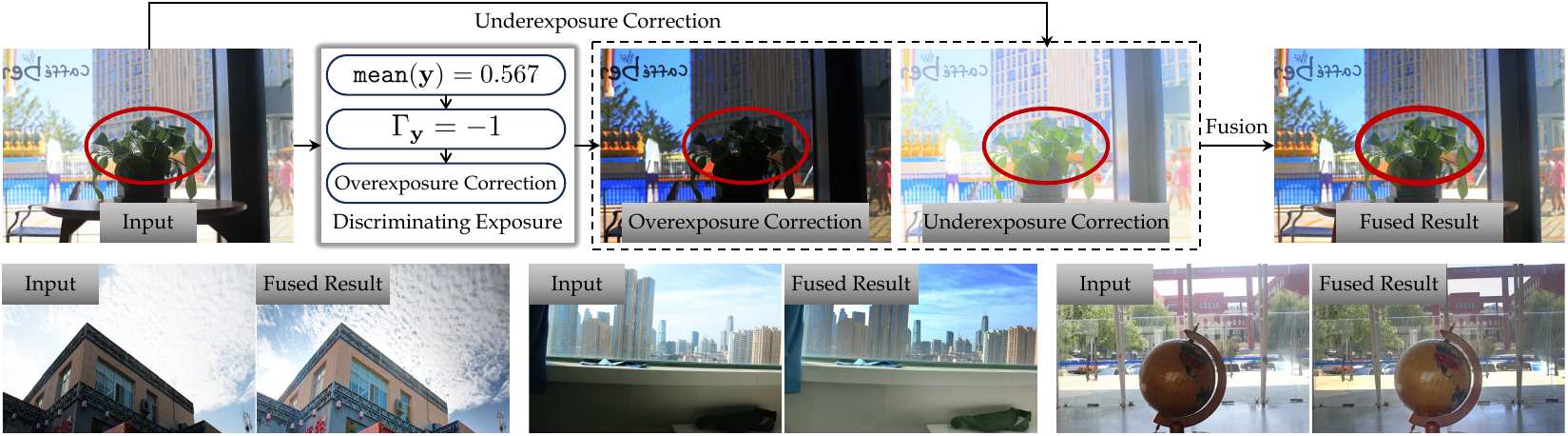}\\
		\end{tabular}
		\caption{Discrimination of exposure level. The top row presents the results of overexposure correction using the proposed exposure level discrimination tool, with the results when underexposure correction is forcibly applied. Because of the particularity of such HDR scenes, it is highly likely that the input image contains excessive overexposed areas, which may affect the exposure level discrimination process. Note that PEC can only achieve unidirectional improvement in the exposure control. To this end, we propose the fusion of these outputs using an existing technique~\cite{mertens2009exposure} to ensure better visual quality for these scenes. The bottom row presents additional examples.}
		\label{fig: A5}
	\end{figure*}

	\begin{figure}[t]
		\centering
		\footnotesize
		\begin{tabular}{c}			
			\includegraphics[width=0.8\linewidth]{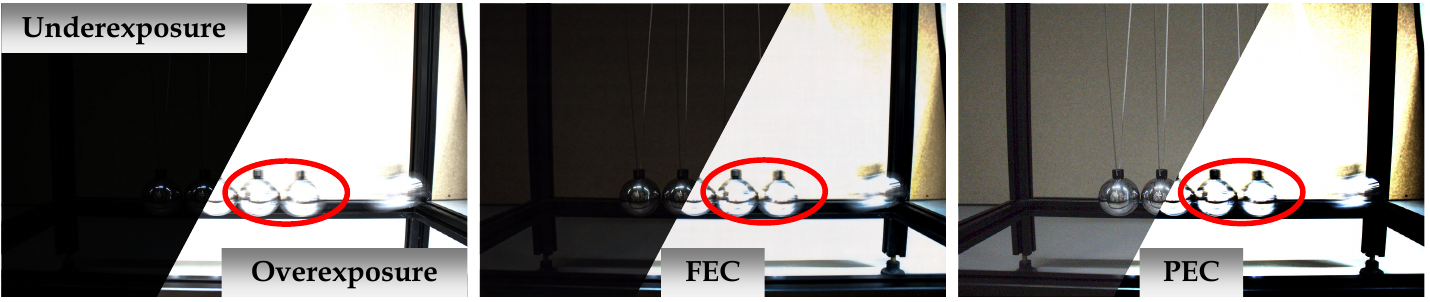}\\
		\end{tabular}
		\caption{Limitations under extreme imaging conditions.}
		\label{fig: Limitations}
	\end{figure}

	\subsection{Algorithmic analyses}\label{sec:analyses}
	
	\textbf{Analysis of convergence behavior.}
	We further explore the convergence behavior using the toy and practical examples. As shown in Figure~\ref{fig: Convergence} (a)–(b), the observed phenomenon of a shrinkage trajectory in the toy examples reflects the convergent nature of the PEC. Moreover, from the relative error curve and intermediate results in panel (b), it is apparent that the corrected results consistently maintain stable exposure, and the local textures and details are progressively highlighted with increasing iterations. Overall, we can argue that the PEC indeed possesses excellent convergence ability. The corrected result in the third iteration is the most satisfying. Thus, we empirically define $1\leq K\leq 3$. 
	
	\textbf{Parameter analysis.}
	In the PEC (see Algorithm~\ref{alg:PEC}), three groups of parameters must be manually defined, \textit{i.e.,} the iteration numbers $T$ and $K$, and the exposure control coefficient $c$. The results in the previous subsection proved that the structural information is progressively compensated as the iteration number $k$ increases. Figure~\ref{fig: Convergence} (c)–(d) presents the effects of $T$ and $c$ on the corrected output. Panel (c) presents the mapping in the range of a given underexposed observation. Panel (d) shows the corrected results among the cases appearing in (c). $T$ and $c$ clearly decide the exposure level, where $T$ has more perceptible brightening effects than the coefficient $c$. Moreover, we set $c=1$ for most underexposed scenes, whereas for the overexposed cases, we set $c<1$ because the levels of overexposure are not sufficiently extreme.
	
	\textbf{Effects of warm start.}
	After analyzing the core parameters of the PEC, we further investigated the role of the crucial warm start component. The relevant visualizations and numerical results are presented in Figure~\ref{fig: A3}. Removal of the warm start leads to varying degrees of performance degradation, both in underexposure and overexposure scenarios. Although removing the warm start can slightly improve the computational efficiency, numerical analysis of the computation time indicates that this improvement is negligible. Sacrificing part of the visual quality for such a minor gain is not worthwhile. This further validates the effectiveness of the proposed warm start component.

	\textbf{Process for discriminating exposure level.}
	This section presents a more comprehensive analysis of the effectiveness of the designed exposure level discrimination process. The results presented in Figure~\ref{fig: exposurelevel} (a) partially illustrate the original intention and rationality behind the developed exposure level discrimination tool. However, in some extreme cases (\textit{e.g.,} those containing both significantly overexposed and underexposed areas), the developed method may exhibit certain limitations. As shown in the top row of Figure~\ref{fig: A5}, when the PEC handles such examples, the overexposed areas (\textit{i.e.,} the background) dominate the image, leading the discrimination tool to classify the image as overexposed and to apply overexposure correction. Because the PEC can achieve only unidirectional improvement in exposure control, although the visual quality of the background area is enhanced, the exposure level of areas such as the foreground vase is not improved and may even be negatively affected. Even if low exposure correction is forcibly applied, the issue of not being able to balance both regions persists. To address this issue, inspired by the work in~\cite{ma2020joint,zhang2019dual}, we propose a fusion strategy to better handle such situations, ensuring higher visual quality. The upper right corner of Figure~\ref{fig: A5} shows the results after applying the fusion strategy, where the developed method achieves good results in both overexposure correction for the background and underexposure enhancement for the foreground, effectively balancing multiple exposure level areas. Additional related results are presented in the bottom row of Figure~\ref{fig: A5} to further validate the effectiveness of the proposed fusion strategy.

	\subsection{Limitation}
	Despite the evident advantages of the PEC demonstrated above, certain limitations still persist. As shown in Figure~\ref{fig: Limitations}, the PEC outperforms the FEC~\cite{huang2022eccv} in underexposure images but struggles to recover structures for overexposure images. This difficulty arises from the constraints imposed by imaging devices and the storage capacity they allow.
	A multi-exposure fusion approach built on the developed PEC is a potential solution to this issue.

	\section{Concluding remarks}
	A novel, practical, and learning-free exposure correction algorithm that effectively combines the benefits of remarkable computational efficiency and visually pleasing image quality was developed. The algorithm of the PEC also exhibits excellent stability and robustness. Extensive evaluations demonstrated the superior performance of the PEC in exposure correction.
	
	The PEC introduces a new perspective that leverages the intrinsic demand of the task and valuable information extracted from observations. It efficiently generates the corrected outputs based on a simple solution. Owing to its real-time processing capability and the properties of the algorithm, the PEC can be easily implemented in hardware chips (\textit{e.g.,} FPGA) to meet practical demands.

	\Acknowledgements{This work was partially supported by the National Key R\&D Program of China (Grant 2022YFA1004101), the National Natural Science Foundation of China (Grant Nos. 62506060, U22B2052 and 62027826),  the Distinguished Youth Funds of the Liaoning Natural Science Foundation (No.2025JH6/101100001), the Distinguished Young Scholars Funds of Dalian (No. 2024RJ002), and the Fundamental Research Funds for the Central Universities.}

\end{document}